\documentclass[10pt,journal,compsoc]{IEEEtran}

\usepackage{cite}
\usepackage{url}
\usepackage{ragged2e}
\usepackage{epsfig}
\usepackage{graphicx}
\usepackage{amsmath,amssymb} 
\usepackage{subfigure}
\usepackage{algorithm}
\usepackage{algorithmicx}
\usepackage{algpseudocode}
\usepackage{graphics}
\usepackage{threeparttable}
\usepackage{color}
\usepackage[normalem]{ulem}
\usepackage{multirow}
\usepackage{float}
\usepackage{amsfonts}
\usepackage{bm}
\usepackage{array}
\usepackage[table]{xcolor}
\usepackage{colortbl}
\usepackage{pifont}
\usepackage{diagbox}
\usepackage{rotating}
\usepackage{booktabs}
\usepackage{overpic}
\usepackage{textcomp}
\usepackage{contour}
\usepackage{enumitem}
\usepackage[pagebackref=false,breaklinks=true,colorlinks=true, bookmarks=false]{hyperref}
\usepackage{geometry}
\usepackage{caption}
\usepackage{subcaption}
\def\ie{\emph{i.e.}}
\def\eg{\emph{e.g.}}
\def\etc{\emph{etc}}
\def\etal{{\em et al.~}}

\definecolor{bblue}{rgb}{0,150,230}
\definecolor{mygray}{gray}{.92}

\newcommand{\figref}[1]{Fig.~\ref{#1}}
\newcommand{\tabref}[1]{Tab.~\ref{#1}}

\graphicspath{{./Imgs/}}
\DeclareGraphicsExtensions{.jpg,.pdf,.png}
\begin{document}
\newgeometry{top=0.7in, bottom=0.7in, left=0.6in, right=0.6in}

\newcommand{\pjl}[1]{{\textcolor{red}{#1}}}

\def\ie{\emph{i.e.}}
\def\eg{\emph{e.g.}}
\def\etc{\emph{etc}}
\def\etal{{\em et al.}}

\title{Evaluation Study on SAM 2 for Class-agnostic
Instance-level Segmentation}

\author{Jialun Pei, Zhangjun Zhou,
        Tiantian Zhang$^{\dagger}$
\IEEEcompsocitemizethanks{
\IEEEcompsocthanksitem Jialun Pei is with the Department of Computer Science and Engineering, The Chinese University of Hong Kong, HKSAR, China.
\IEEEcompsocthanksitem Tiantian Zhang is with the Department of Computer Science and Engineering, The Chinese University of Hong Kong, HKSAR, China.
\IEEEcompsocthanksitem Zhangjun Zhou is with the School of Software Engineering, Huazhong University of Science and Technology, Wuhan, China.
\IEEEcompsocthanksitem  
$^{\dagger}$ Corresponding author: Jialun Pei (Email: peijialun@gmail.com).}
}

\markboth{}%
{Pei \MakeLowercase{\textit{et al.}}}

\IEEEtitleabstractindextext{%
\begin{abstract}
\justifying
Segment Anything Model (SAM) has demonstrated powerful zero-shot segmentation performance in natural scenes. The recently released Segment Anything Model 2 (SAM2) has further heightened researchers' expectations towards image segmentation capabilities. To evaluate the performance of SAM2 on class-agnostic instance-level segmentation tasks, we adopt different prompt strategies for SAM2 to cope with instance-level tasks for three relevant scenarios: Salient Instance Segmentation (SIS), Camouflaged Instance Segmentation (CIS), and Shadow Instance Detection (SID). In addition, to further explore the effectiveness of SAM2 in segmenting granular object structures, we also conduct detailed tests on the high-resolution Dichotomous Image Segmentation (DIS) benchmark to assess the fine-grained segmentation capability. Qualitative and quantitative experimental results indicate that the performance of SAM2 varies significantly across different scenarios. 
Besides, SAM2 is not particularly sensitive to segmenting high-resolution fine details.
We hope this technique report can drive the emergence of SAM2-based adapters, aiming to enhance the performance ceiling of large vision models on class-agnostic instance segmentation tasks. 
Project link: \url{https://github.com/PJLallen/InstanceSAM2Eval}.
\end{abstract}

\begin{IEEEkeywords}
Foundation Model, Large Vision Model, SAM 2, Instance-level Segmentation, Dichotomous Image Segmentation.
\end{IEEEkeywords}}

\maketitle

\IEEEdisplaynontitleabstractindextext

\IEEEpeerreviewmaketitle

\IEEEraisesectionheading{\section{Introduction}\label{sec:introduction}}

\IEEEPARstart{T}{h}e advent of large foundation models, including ChatGPT, GPT-4, and LLaMA, has revolutionized the artificial intelligence (AI) landscape. Powered by extensive datasets, these models excel in multi-modal processing, \eg, language, image, video, and audio, showcasing substantial progress in AI capabilities. Building on these developments, the Segment Anything Model (SAM)~\cite{kirillov2023segment} stands out breakthrough in scene segmentation with large vision models.
The generality and Adaptability of SAM highlight its potential for understanding complex scenarios and targets, further expanding the frontiers of image segmentation tasks.

SAM allows users to input custom prompts, such as points or bounding boxes, resulting in highly accurate segmentation masks. This adaptability enables SAM to perform a wide range of image segmentation tasks. More recently, the release of SAM2~\cite{ravi2024sam} further overcomes the limitation that SAM does not handle video content well. In the field of image segmentation, SAM2 has shown improvement in segmentation accuracy and inference efficiency~\footnote{\url{https://sam2.metademolab.com/demo}}. A variety of evaluations have recently emerged to examine the segmentation performance of SAM2 in different scenarios~\cite{lian2024evaluation, yan2024biomedical,zhu2024medical,ma2024segment,tang2024evaluating,lou2024zero,yu2024sam}. For instance, Lian \etal~\cite{lian2024evaluation} assessed its instance segmentation performance in underwater environments, while Yan \etal~\cite{yan2024biomedical} explored its effectiveness in endoscopic and microscopic images. Additionally, Ma \etal~\cite{ma2024segment} conducted a comprehensive benchmark of SAM2 across 11 medical image modalities and videos, highlighting its strengths and weaknesses compared to SAM and MedSAM. Furthermore, Tang \etal~\cite{tang2024evaluating} compared SAM2 and SAM on the camouflage object detection benchmark. This research found that SAM2 significantly degrades the performance of SAM2 compared to SAM for detecting camouflaged objects when no prompts are provided, and significantly improves its performance over SAM when segmentation prompts are available. These interesting findings raise curiosity about SAM2's performance in class-agnostic instance-level segmentation tasks.

In this paper, we evaluate the performance of SAM2 in class-agnostic instance-level segmentation tasks, focusing on three distinct scenarios: Salient Instance Segmentation (SIS)~\cite{li2017instance}, Camouflaged Instance Segmentation (CIS)~\cite{pei2022osformer}, and Shadow Instance Detection (SID)~\cite{wang2022instance}. Moreover, We also thoroughly evaluate SAM2 on the high-resolution dichotomous image segmentation (DIS) benchmark~\cite{qin2022highly} to analyze its ability to segment granular target structures. We compare SAM2 with SAM and well-known specific models on multiple benchmarks. Based on extensive experimental results, we summarize the following conclusions:
\begin{itemize}
    \item SAM2 outperforms task-specific methods for CIS and SIS when using bounding boxes as prompt inputs. However, the performance of SAM2 drops remarkably without box prompts, especially for camouflaged instances.
    \item SAM2 performs poorly on the DIS task, whether or not it uses bounding boxes as prompts. It indicates that SAM2 is not well suited for fine-grained segmentation of complex object structures.
    \item For the SID task, while SAM2 performs well in segmenting instances, it struggles with shadow matching.
    \item SAM2 with fewer parameters achieves superior results compared to SAM across four tasks when using bounding boxes as prompts. In contrast, SAM2 without box prompts performs inferior to SAM for SIS, CIS, and SID.
\end{itemize}

\begin{table*}[t!]
\begin{center}

\footnotesize
\renewcommand{\arraystretch}{1.2}
\renewcommand{\tabcolsep}{2.0mm}
\caption{\textbf{Quantitative comparisons of SAM2, SAM, and SIS-specific methods on ILSO, SIS10K, SOC, and SIP.} SAM-B, SAM-L, and SAM-H represent ViT-B, ViT-L, and ViT-H model types of SAM, respectively. SAM2-T, SAM2-B+, and SAM2-L represent Hiera-Tiny, Hiera-Base+, and Hiera-Large model types of SAM2, respectively. The best and second-best results are bolded and underlined.}
\label{table:SIS}

\begin{tabular}{r||r|r|ccc|ccc|ccc|ccc}
\toprule
\multirow{2}{*}{Methods} & \multirow{2}{*}{Pub. \& Year}& \multirow{2}{*}{Settings} 
& \multicolumn{3}{c|}{ILSO~\cite{li2017instance}}   
& \multicolumn{3}{c|}{SIS10K~\cite{pei2022transformer}} 
& \multicolumn{3}{c|}{SOC~\cite{fan2018salient}}   
& \multicolumn{3}{c}{SIP~\cite{fan2020rethinking}} 
\\
& & & AP & AP$^{50}$ & AP$^{70}$ 
& AP & AP$^{50}$ & AP$^{70}$ 
& AP & AP$^{50}$ & AP$^{70}$ 
& AP & AP$^{50}$ & AP$^{70}$ 
\\ \hline
S4Net~\cite{fan2019s4net} & PR 2019 &ResNet-50 & 52.3 & 86.7 & 63.6 & 55.5 & 83.3 & 67.0 & 35.4 & 64.6 & 42.0 & 49.6 & 76.0 & 63.7 \\
SCNet~\cite{pei2020salient}&Neuro 2020 & DenseNet-201 & 56.8 & 84.6 & 67.4 & 52.9 & 82.5 & 69.2 & 32.6 & 60.9 & 41.4 & -- & -- & -- \\
RDPNet~\cite{wu2021regularized} & TIP 2021 &ResNet-50 & 58.4 & 88.5 & 73.4 & 56.5 & 82.0 & 69.4 & 37.8 & 60.6 & 48.2 & 59.0 & 80.1 & 74.1 \\
SCG~\cite{liu2021scg} &TIP 2021 &ResNet-50 &  64.8 & 88.8 & 78.6 & -- & -- & -- & --  & -- & -- & -- & -- & -- \\
OQTR~\cite{pei2022transformer}&TMM 2022 & ResNet-50 & 63.7 & 90.4 & 79.9 & 67.2 & 88.1 & 81.7 & 60.9 & 79.5 & 72.5 & 59.9 & 83.1 & 76.3 \\
\hline
SAM-B~\cite{kirillov2023segment}& ICCV 2023 &Automatic & 30.0 & 41.0 & 34.2 & 26.3 & 34.7 & 30.3 & 23.5 & 33.5 & 25.9 & 41.9 & 59.5 & 42.6 \\
SAM-L~\cite{kirillov2023segment}& ICCV 2023 &Automatic & 69.9 & 89.2 & 83.6 & 66.1 & 84.5 & 78.6 & 66.0 & 82.0 & 74.9 & 84.4 & 93.3 & 90.4 \\
SAM-H~\cite{kirillov2023segment}& ICCV 2023&Automatic & 72.2 & 92.0 & 86.9 & 68.4 & 87.1 & 81.7 & 69.2 & 85.6 & 78.5 & 80.8 & 94.3 & 89.7 \\
\hline
SAM-B~\cite{kirillov2023segment}& ICCV 2023&GT-Bbox & 73.0 & 95.9 & 90.2 & 73.8 & 95.9 & 89.4 & 71.5 & 92.1 & 83.1 & 86.9 & \textbf{97.0} & 93.9 \\
SAM-L~\cite{kirillov2023segment}& ICCV 2023&GT-Bbox & 77.8 & 98.0 & 93.6 & 78.5 & 97.9 & 93.4 & 78.8 & 95.1 & 90.1 & 91.8 & 98.0 & 97.0 \\
SAM-H~\cite{kirillov2023segment}& ICCV 2023&GT-Bbox & 79.2 & 98.0 & 94.8 & 79.5 & 97.9 & 93.6 & 80.6 & 95.4 & 90.8 & 92.6 & 98.0 & 96.9 \\
\hline
\rowcolor[RGB]{216,241,244}
\textbf{SAM2-T}~\cite{ravi2024sam} & -- &Automatic & 30.2 & 36.6 & 34.0 & 24.7 & 29.5 & 27.7 & 21.7 & 26.2 & 23.7 & 64.1 & 68.7 & 65.5 \\
\rowcolor[RGB]{216,241,244}
\textbf{SAM2-B+}~\cite{ravi2024sam} & -- &Automatic & 51.8 & 62.5 & 59.1 & 41.8 & 50.0 & 47.9 & 43.7 & 54.0 & 47.9 & 77.7 & 82.6 & 79.8 \\
\rowcolor[RGB]{216,241,244}
\textbf{SAM2-L}~\cite{ravi2024sam} & -- &Automatic & 49.1 & 57.8 & 55.9 & 45.2 & 53.0 & 51.1 & 47.9 & 55.1 & 52.4 & 79.3 & 83.1 & 81.4 \\
\hline
\rowcolor[RGB]{216,241,244}
\textbf{SAM2-T}~\cite{ravi2024sam} &--  &GT-Bbox & 81.0 & \underline{98.0} & \textbf{96.8} & 81.0 & 98.0 & \underline{94.7} & 80.7 & 96.7 & 91.5 & 91.9 & \textbf{98.0} & \underline{97.0} \\
\rowcolor[RGB]{216,241,244}
\textbf{SAM2-B+}~\cite{ravi2024sam} & --&GT-Bbox & \underline{81.8} & \underline{98.0} & \textbf{96.8} & \underline{81.7} & \underline{97.9} & 94.5 & \underline{82.6} & \textbf{96.8} & \underline{92.5} & \underline{93.2} & \textbf{98.0} & 98.0 \\
\rowcolor[RGB]{216,241,244}
\textbf{SAM2-L}~\cite{ravi2024sam} & --&GT-Bbox & \textbf{82.2} & \textbf{99.0} & \underline{96.7} & \textbf{82.3} & \textbf{98.0} & \textbf{94.8} & \textbf{83.1} & \underline{96.7} & \textbf{93.5} & \textbf{93.4} & \textbf{98.0} & \textbf{98.0} \\

\bottomrule
\end{tabular}

\end{center}
\end{table*}

\section{Experiments}\label{sec:exp}
This section provides the guidelines and details of our basic and extensive experiments, \ie, datasets, the evaluation protocol, implementation settings, and the quantitative and qualitative results of SAM2 on four tasks.

\subsection{Datasets}\label{sec:datasets}
In line with \cite{pei2022transformer,pei2024calibnet}, we utilize the ILSO~\cite{li2017instance}, SOCK~\cite{fan2018salient}, SIS10K~\cite{pei2022transformer}, and SIP~\cite{fan2020rethinking} datasets for the SIS task. For the CIS task, we employ COD10K~\cite{fan2020camouflaged} and NC4K~\cite{lv2021simultaneously} to evaluate the performance. For the SID task, we use the SOBA-challenge and SOBA-test datasets~\cite{wang2022instance}. For DIS, We conduct experiments on DIS5K~\cite{qin2022highly}, including DIS-VD and DIS-TE. DIS-TE is further divided
into four subsets \ie, DIS-TE1, DIS-TE2, DIS-TE3, and DIS-TE4, representing four levels of testing difficulty.
The number of test samples for datasets in each task is summarised below:
\begin{itemize}
    \item \textbf{SIS}: \emph{ILSO}: 300; \emph{SOC}: 600; \emph{SIS10K}: 1,170; \emph{SIP}: 929.
    \item \textbf{CIS}: \emph{COD10K}: 2,026; \emph{NC4K}: 4,121.
    \item \textbf{SID}: \emph{SOBA-challenge}: 100; \emph{SOBA-test}: 160.
    \item \textbf{DIS}: \emph{DIS-VD}: 470; \emph{DIS-TE1}: 500; \emph{DIS-TE2}: 500; \emph{DIS-TE3}: 500; \emph{DIS-TE4}: 500; \emph{Overall DIS-TE}: 2,000.
\end{itemize}


\begin{figure*}[!t]
    \centering
    \renewcommand{\arraystretch}{0.8}
    \begin{tabular}{@{}c@{\hskip 5pt}c@{\hskip 1pt}c@{\hskip 1pt}c@{\hskip 1pt}c@{\hskip 1pt}c@{\hskip 1pt}c@{}}
        \raisebox{0.5cm}{\makebox[0pt][c]{\rotatebox{90}{\small Image}}} & 
        \includegraphics[height=2.1cm]{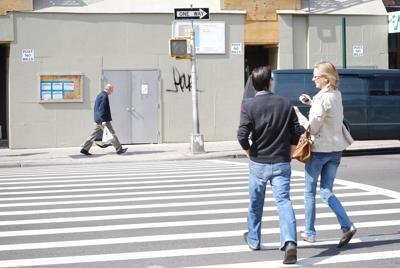} &
        \includegraphics[height=2.1cm]{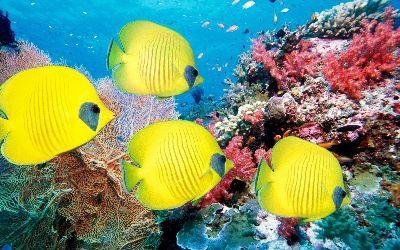} &
        \includegraphics[height=2.1cm]{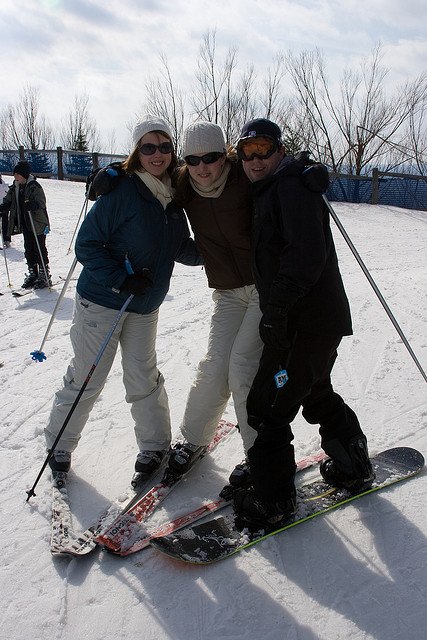} &
        \includegraphics[height=2.1cm]{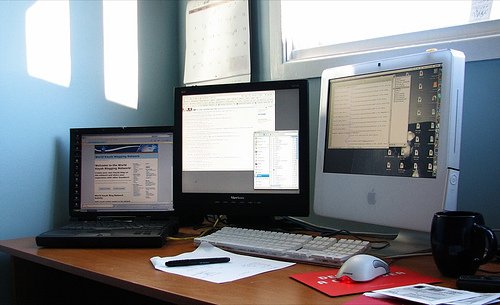} &
        \includegraphics[height=2.1cm]{SIS/soc/image/COCO_train2014_000000036303.jpg} &
        \includegraphics[height=2.1cm]{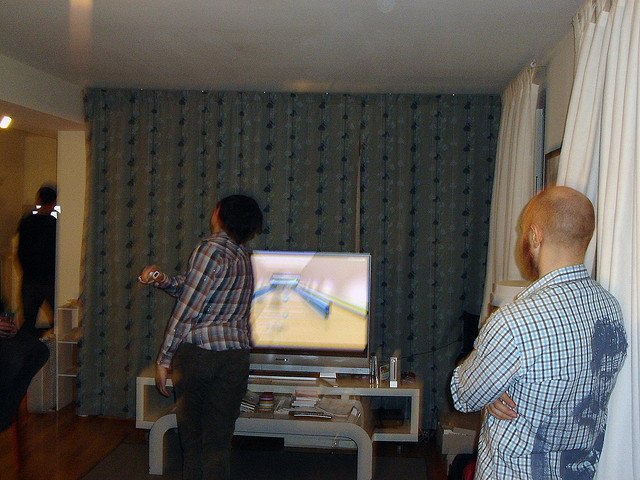}\\ 
        \raisebox{0.05cm}{\makebox[0pt][c]{\rotatebox{90}{\small Ground Truth}}} & 
        \includegraphics[height=2.1cm]{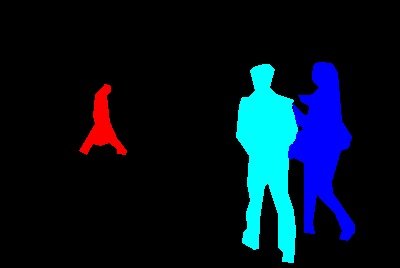} &
        \includegraphics[height=2.1cm]{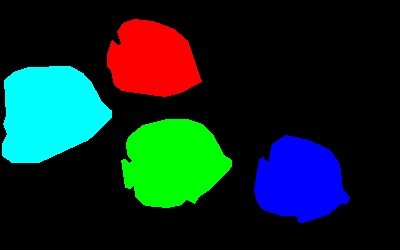} &
        \includegraphics[height=2.1cm]{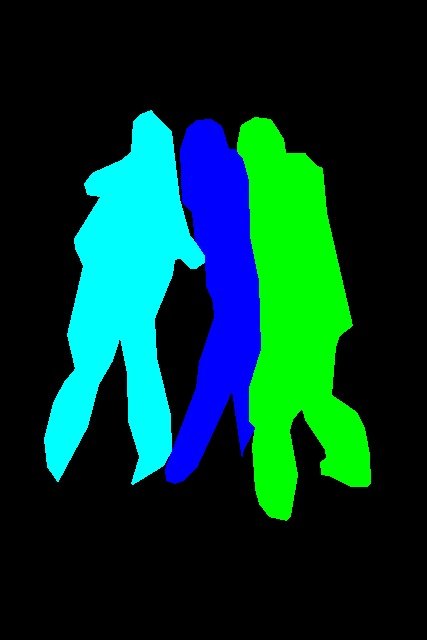} &
        \includegraphics[height=2.1cm]{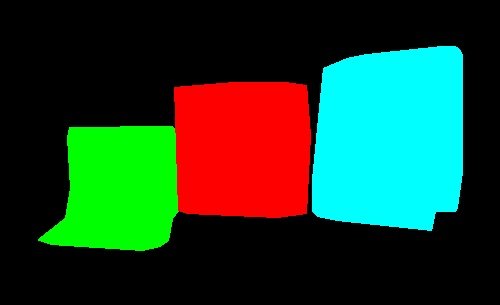} &
        \includegraphics[height=2.1cm]{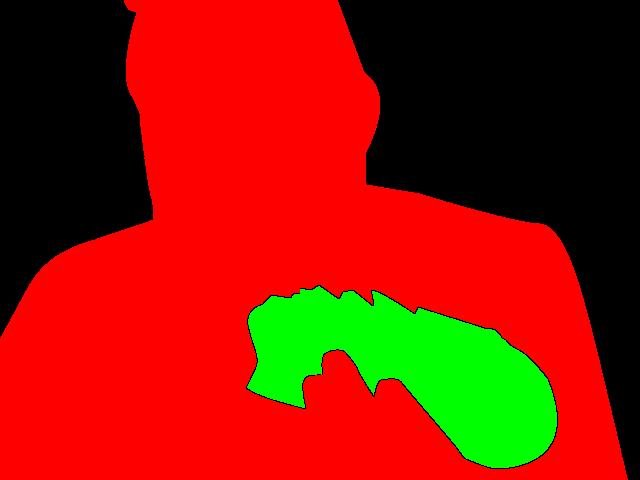} &
        \includegraphics[height=2.1cm]{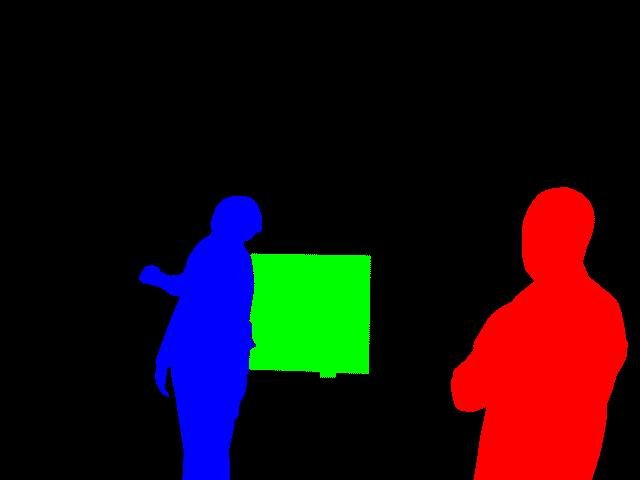}\\ 
        \raisebox{0.55cm}{\makebox[0pt][c]{\rotatebox{90}{\small S4Net}}} & 
        \includegraphics[height=2.1cm]{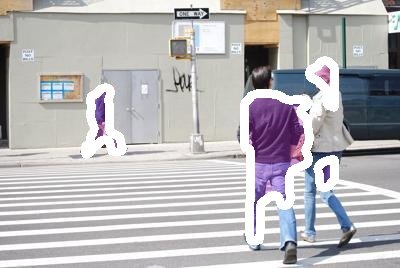} &
        \includegraphics[height=2.1cm]{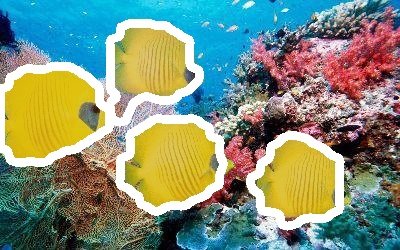} &
        \includegraphics[height=2.1cm]{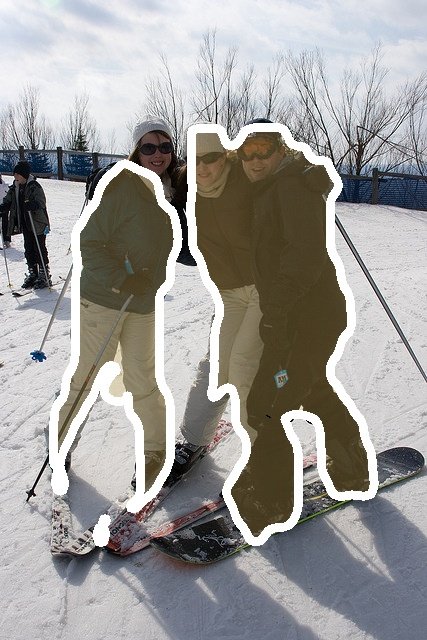} &
        \includegraphics[height=2.1cm]{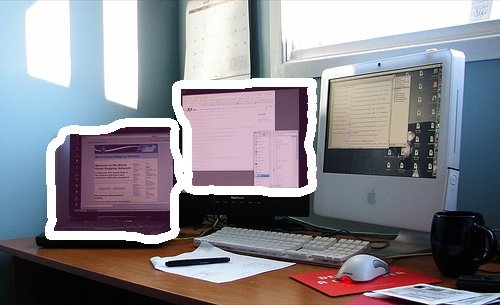} &
        \includegraphics[height=2.1cm]{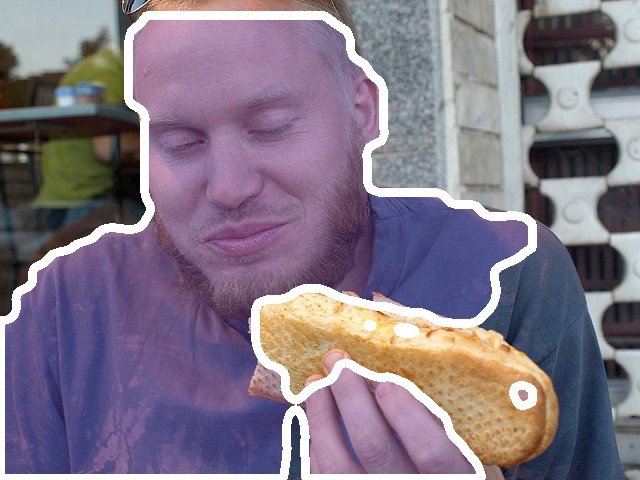} &
        \includegraphics[height=2.1cm]{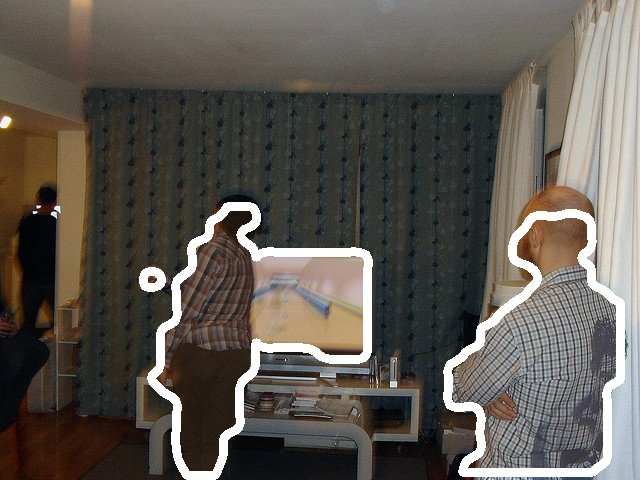}\\ 
        \raisebox{0.45cm}{\makebox[0pt][c]{\rotatebox{90}{\small RDPNet}}} & 
        \includegraphics[height=2.1cm]{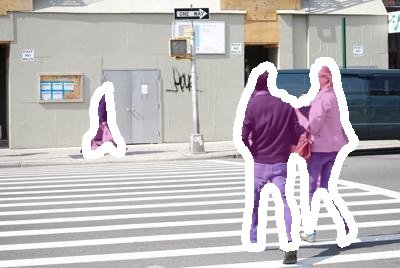} &
        \includegraphics[height=2.1cm]{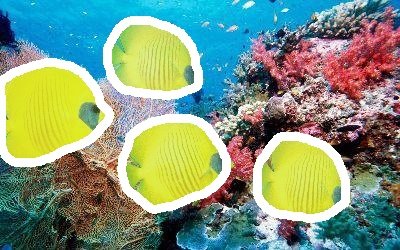} &
        \includegraphics[height=2.1cm]{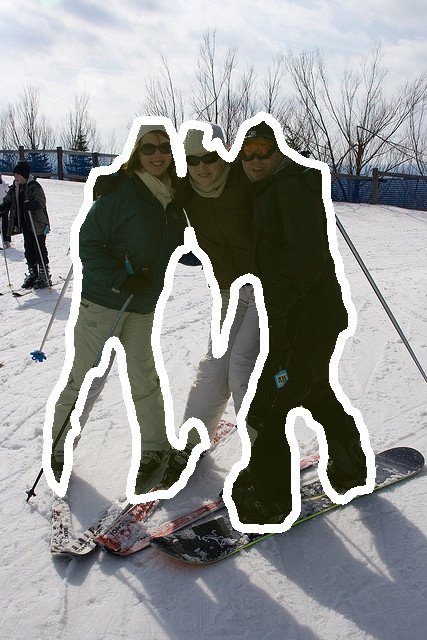} &
        \includegraphics[height=2.1cm]{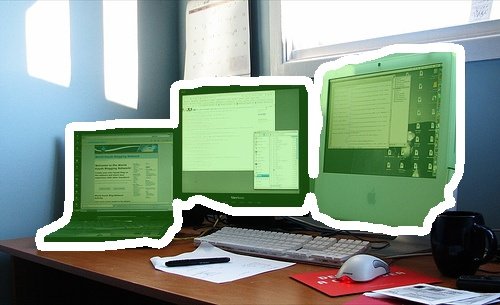} &
        \includegraphics[height=2.1cm]{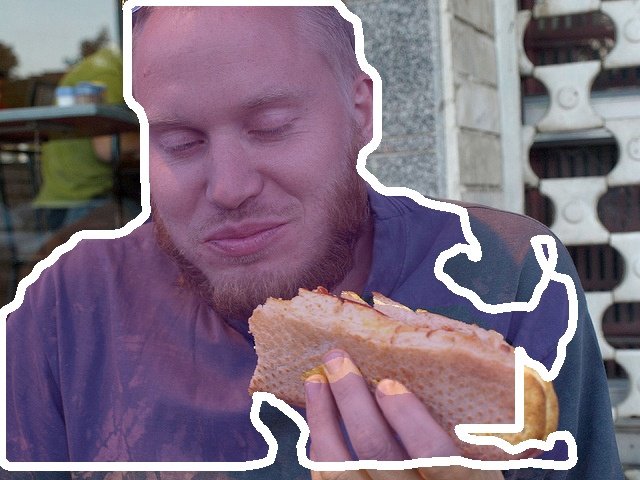} &
        \includegraphics[height=2.1cm]{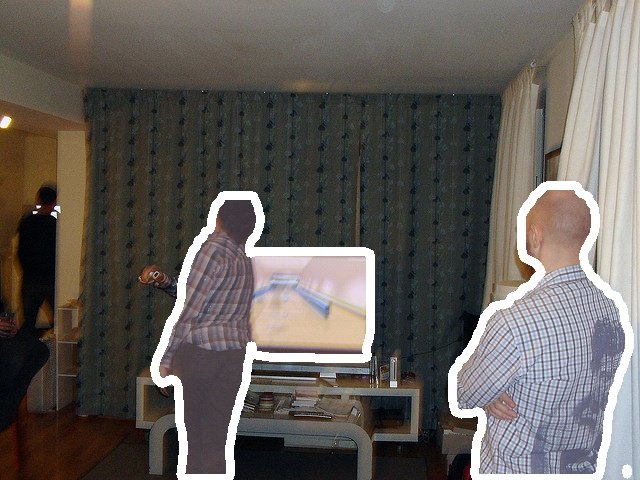}\\ 
        \raisebox{0.5cm}{\makebox[0pt][c]{\rotatebox{90}{\small OQTR}}} & 
        \includegraphics[height=2.1cm]{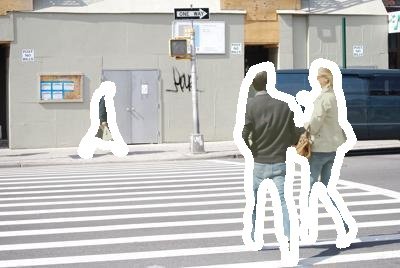} &
        \includegraphics[height=2.1cm]{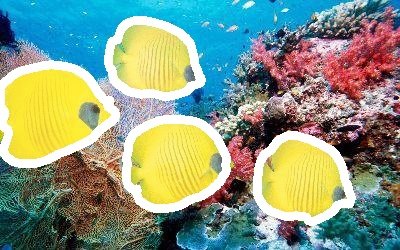} &
        \includegraphics[height=2.1cm]{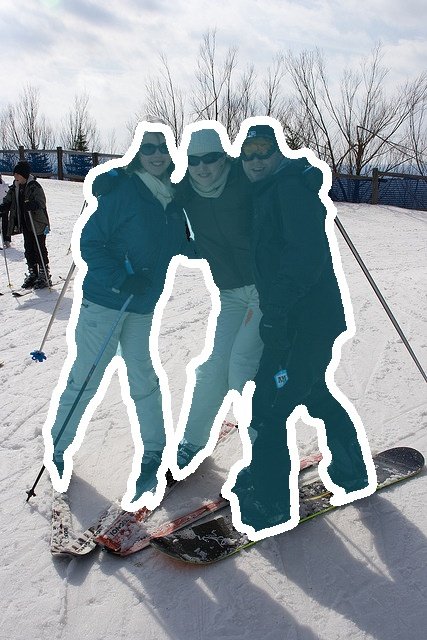} &
        \includegraphics[height=2.1cm]{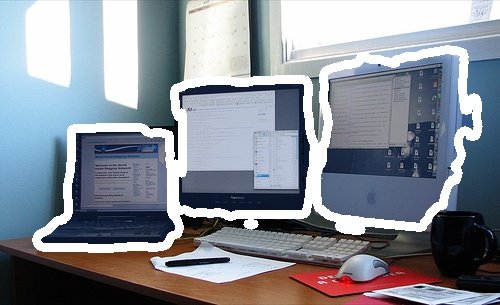} &
        \includegraphics[height=2.1cm]{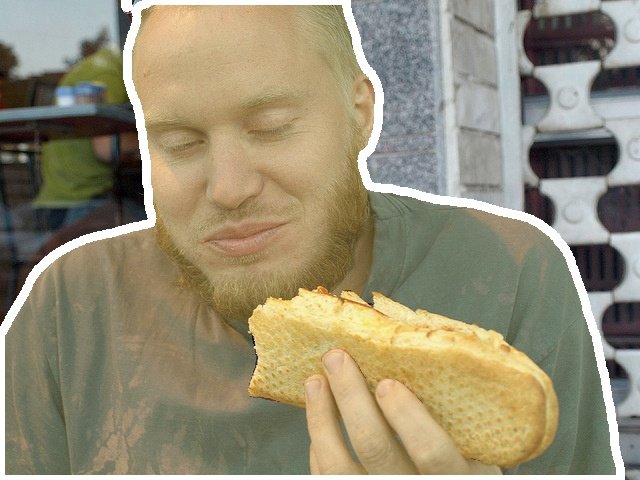} &
        \includegraphics[height=2.1cm]{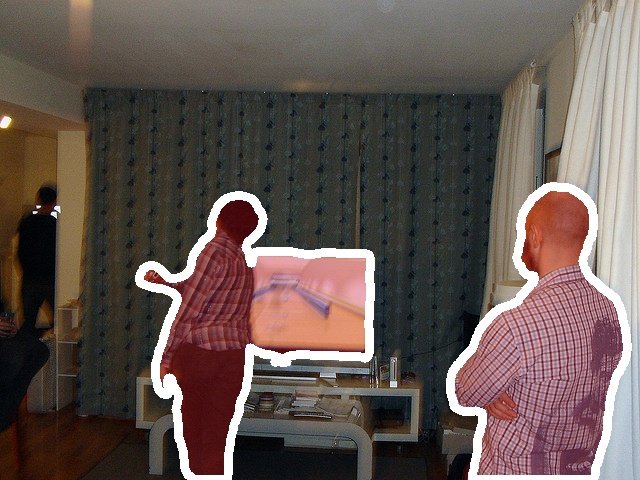}\\ 
        \raisebox{0.2cm}{\makebox[0pt][c]{\rotatebox{90}{\small SAM-Auto}}} & 
        \includegraphics[height=2.1cm]{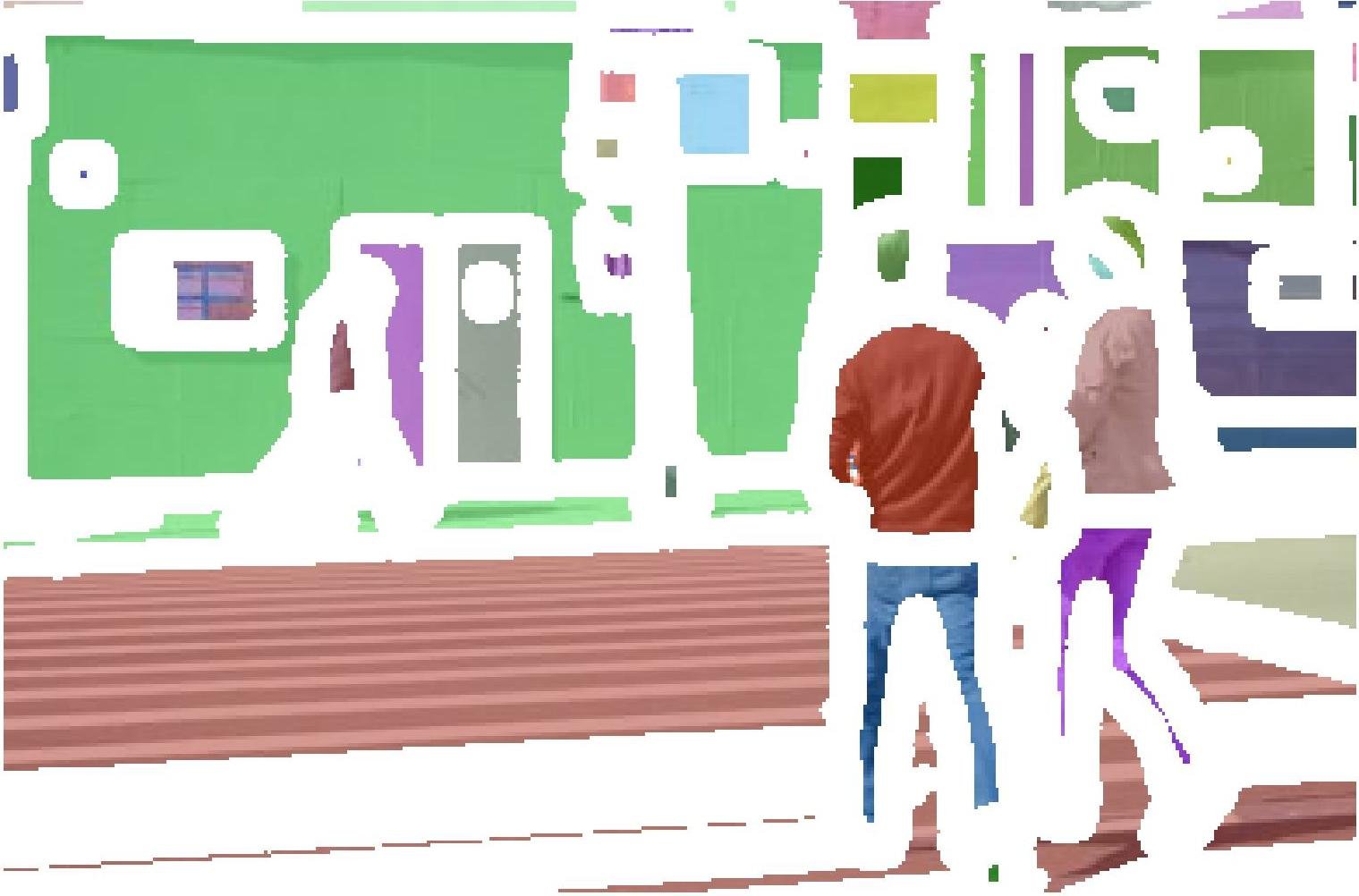} &
        \includegraphics[height=2.1cm]{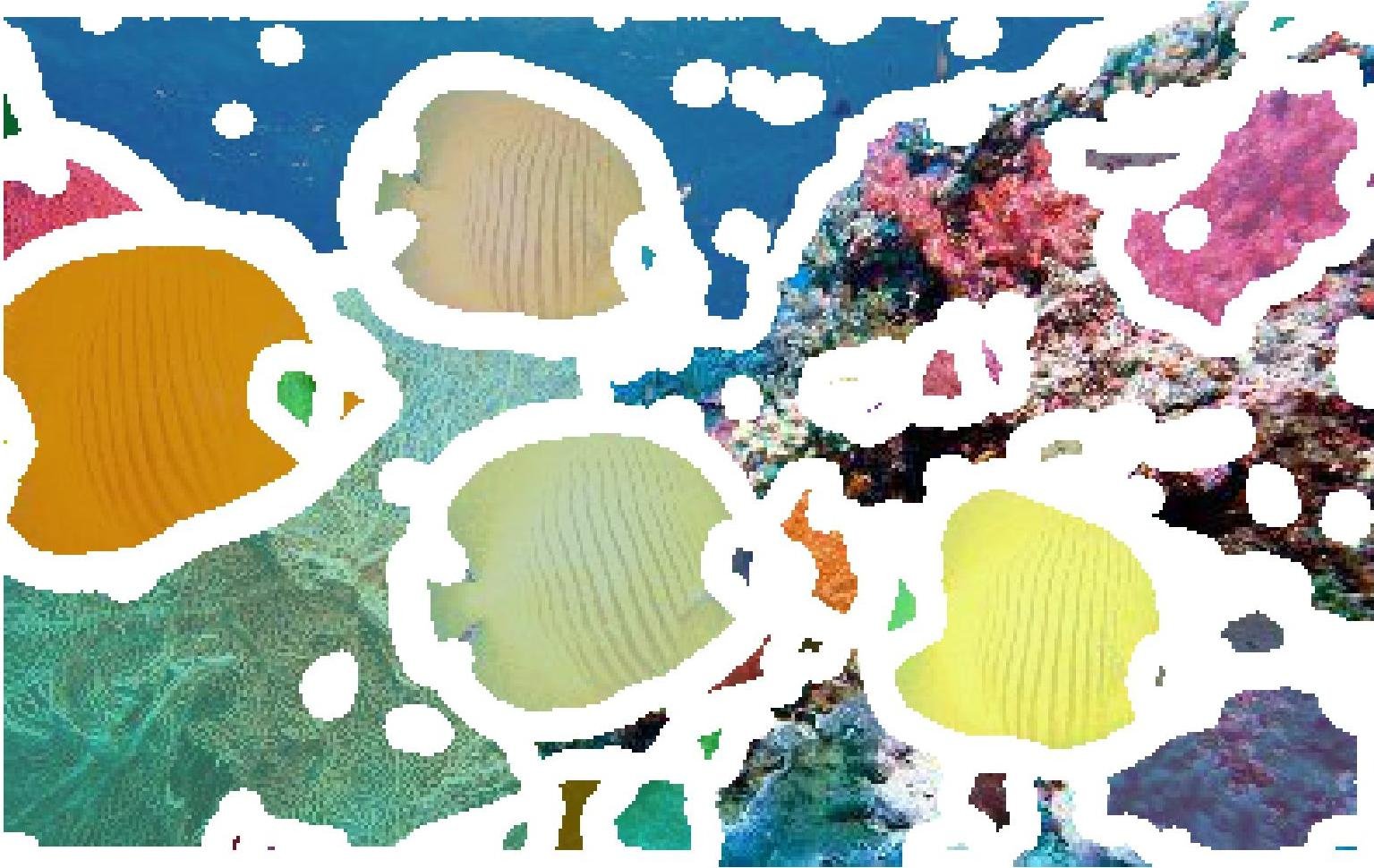} &
        \includegraphics[height=2.1cm]{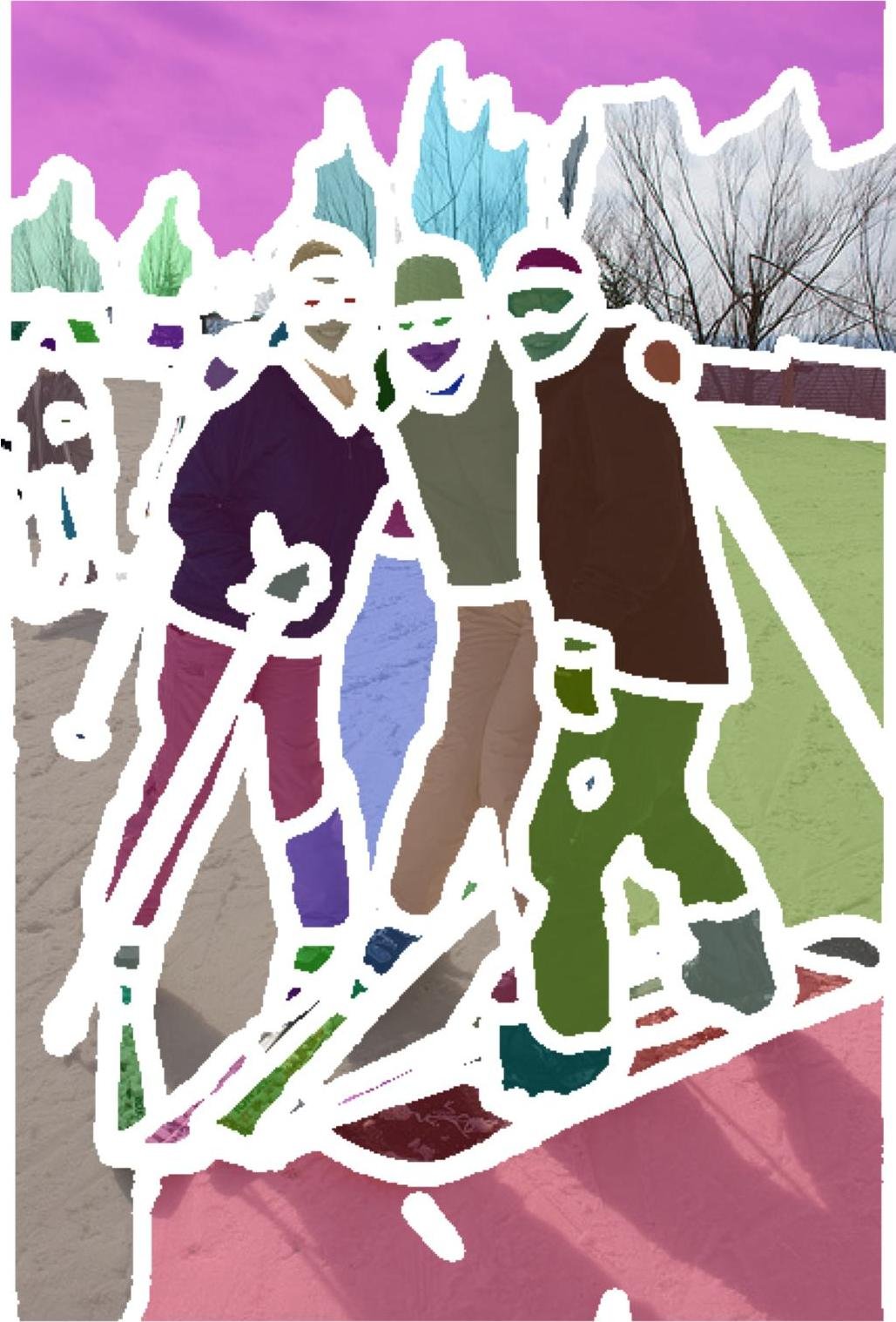} &
        \includegraphics[height=2.1cm]{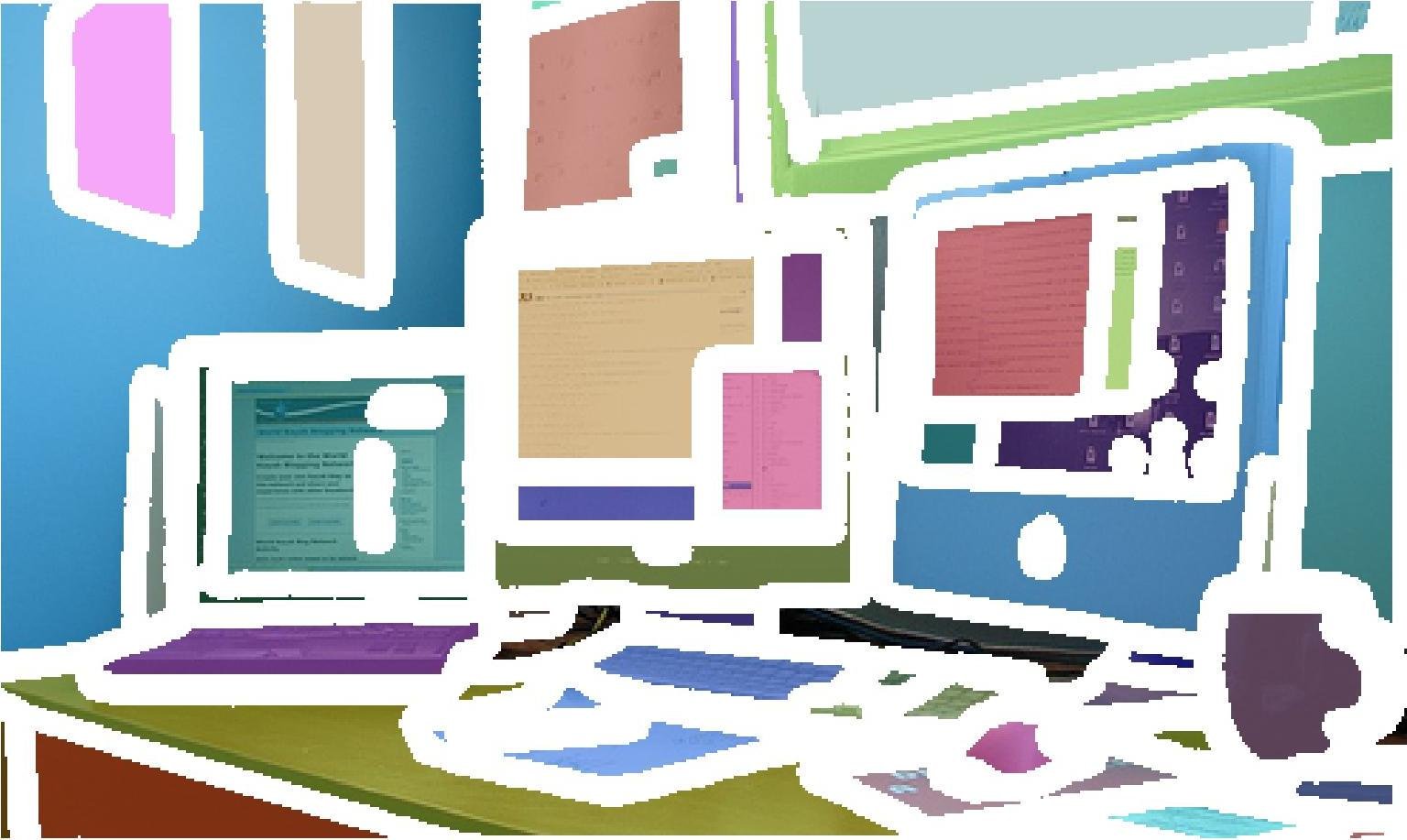} &
        \includegraphics[height=2.1cm]{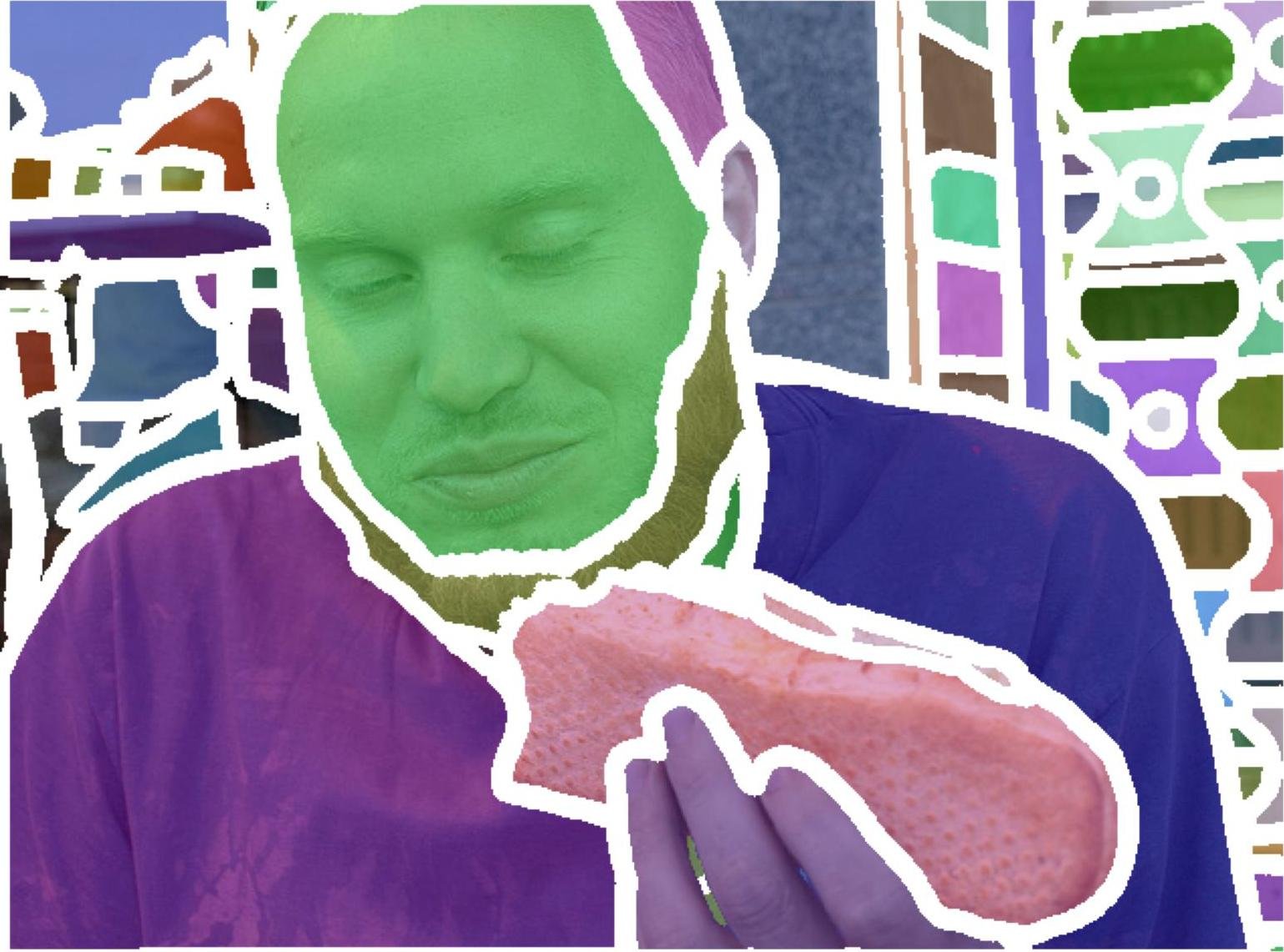} &
        \includegraphics[height=2.1cm]{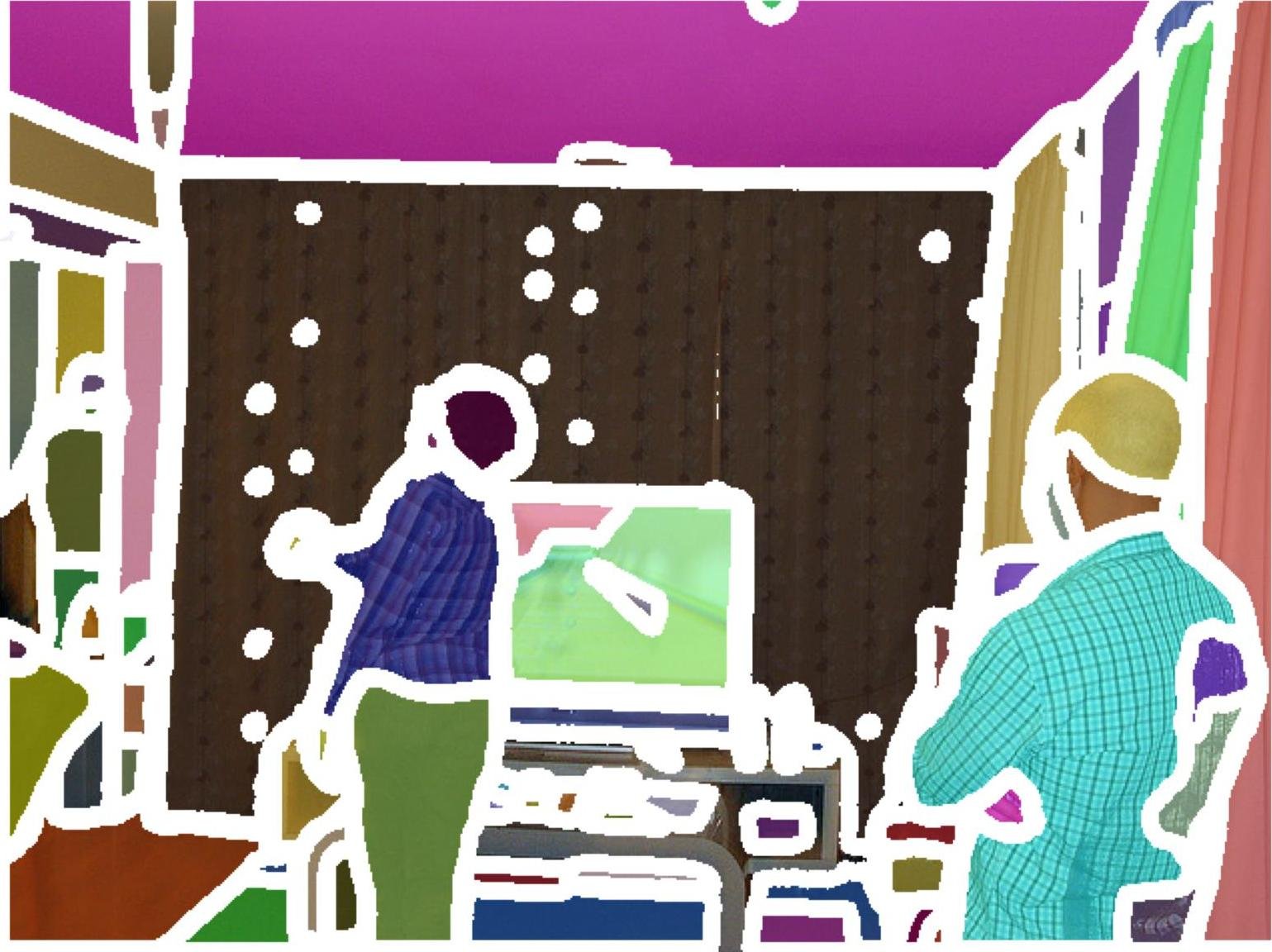}\\ 
        \raisebox{0.2cm}{\makebox[0pt][c]{\rotatebox{90}{\small SAM-Bbox}}} & 
        \includegraphics[height=2.1cm]{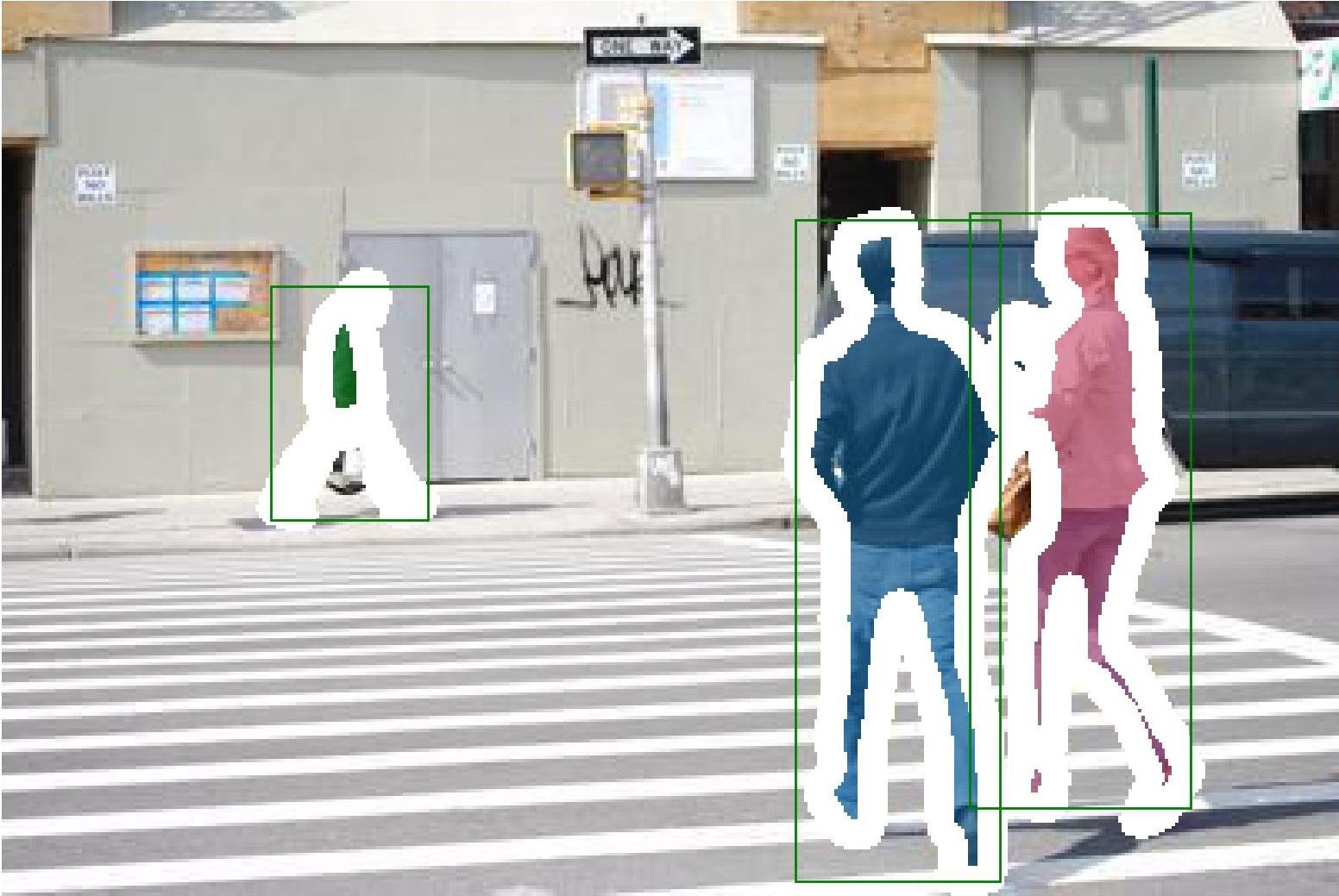} &
        \includegraphics[height=2.1cm]{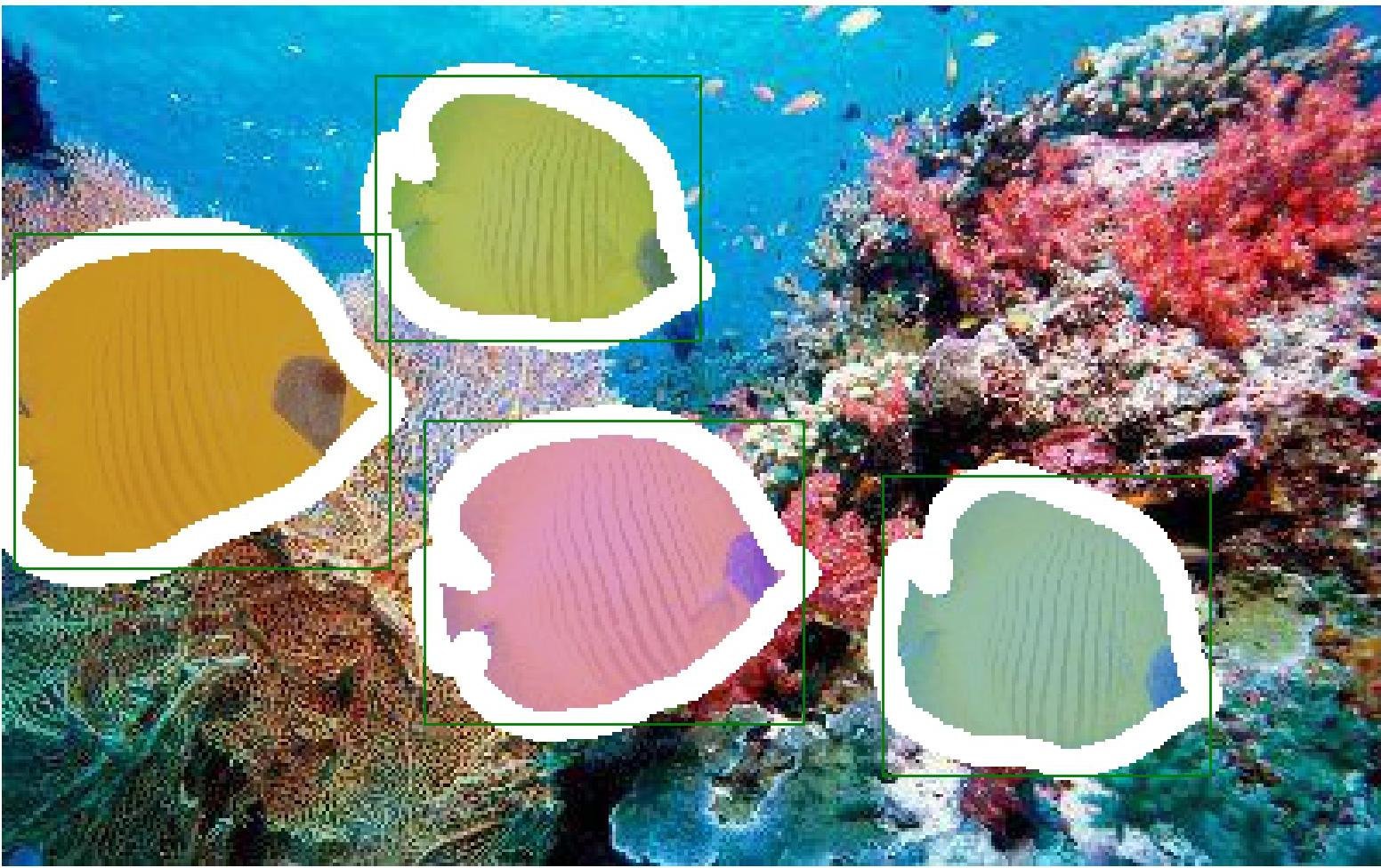} &
        \includegraphics[height=2.1cm]{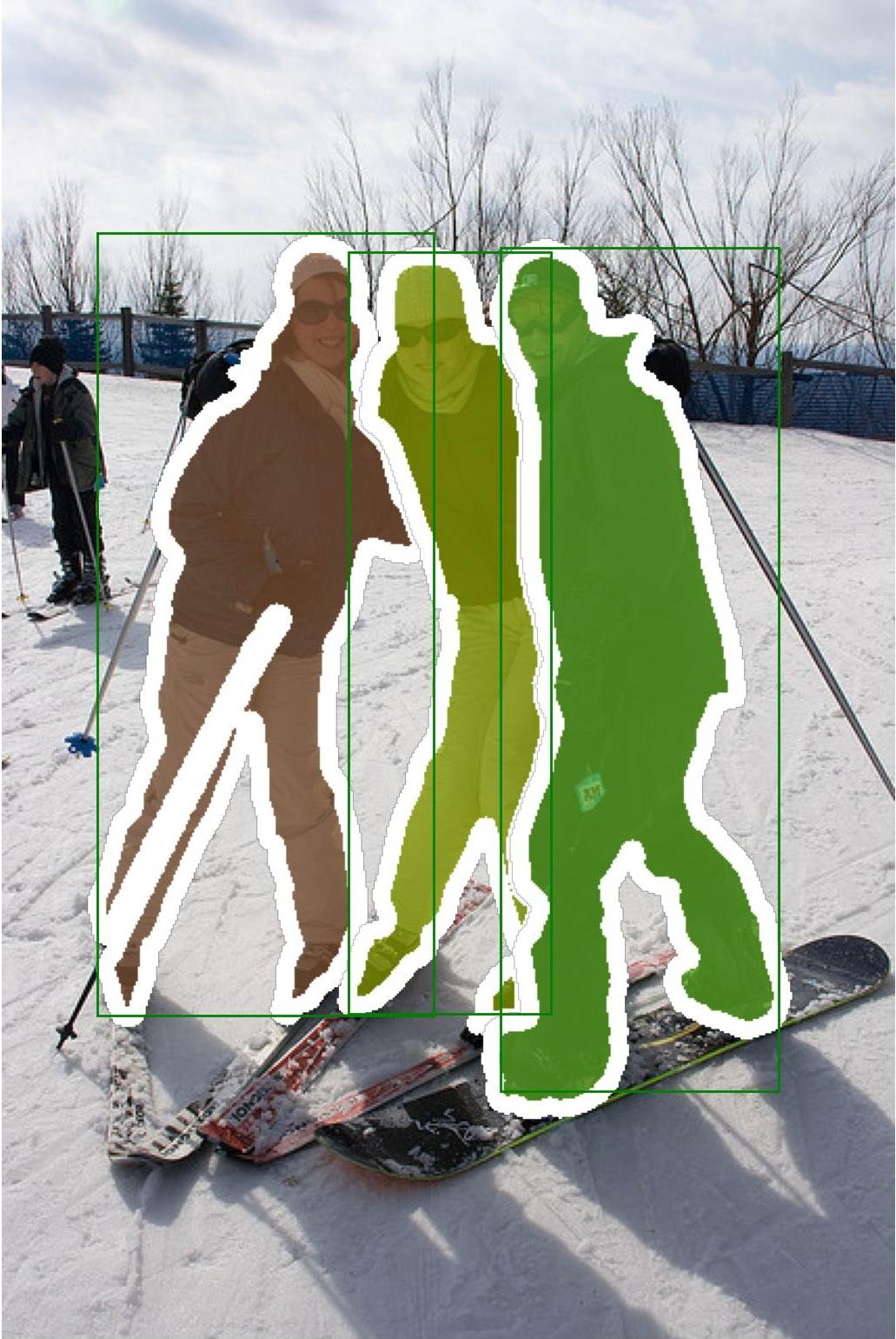} &
        \includegraphics[height=2.1cm]{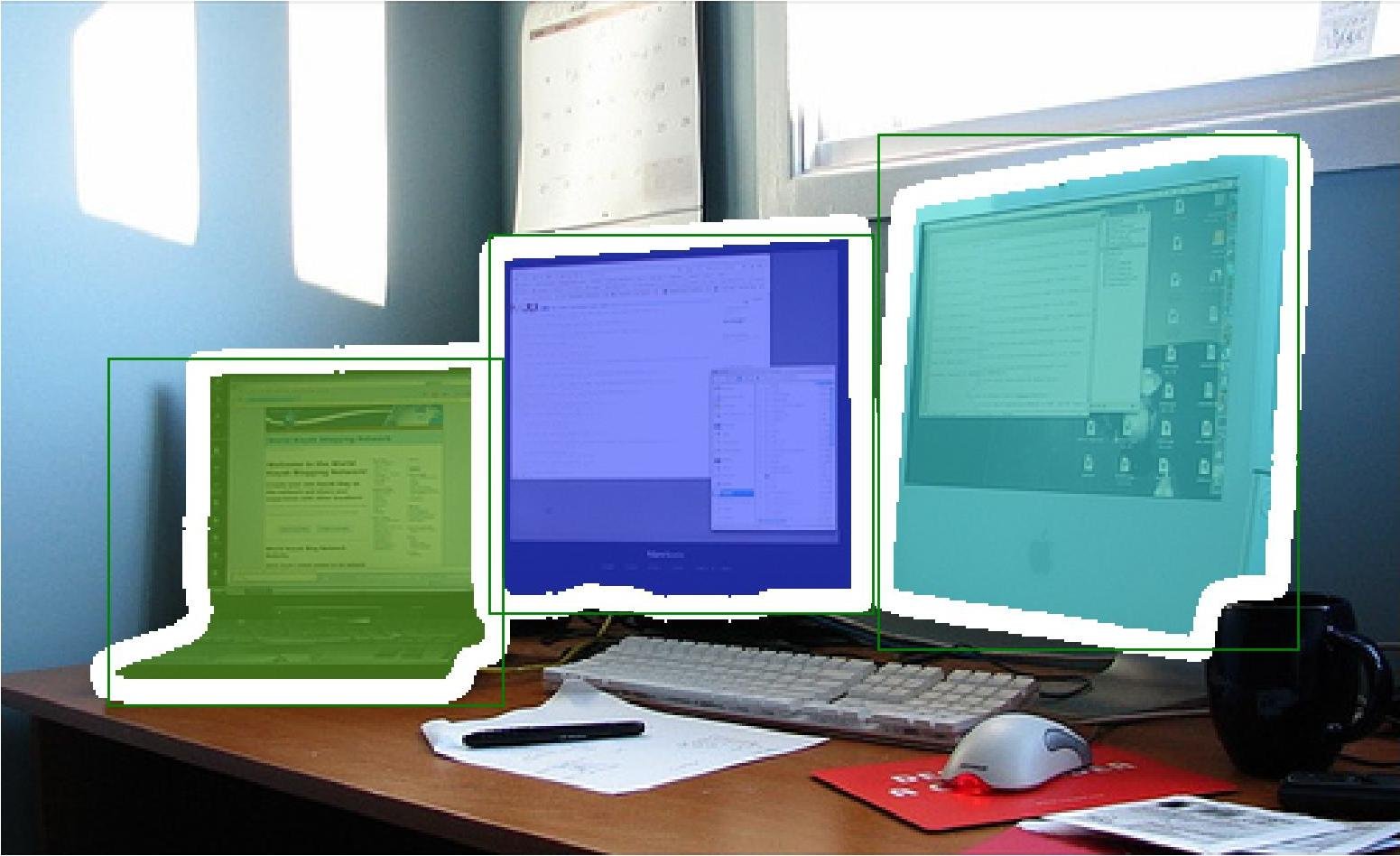} &
        \includegraphics[height=2.1cm]{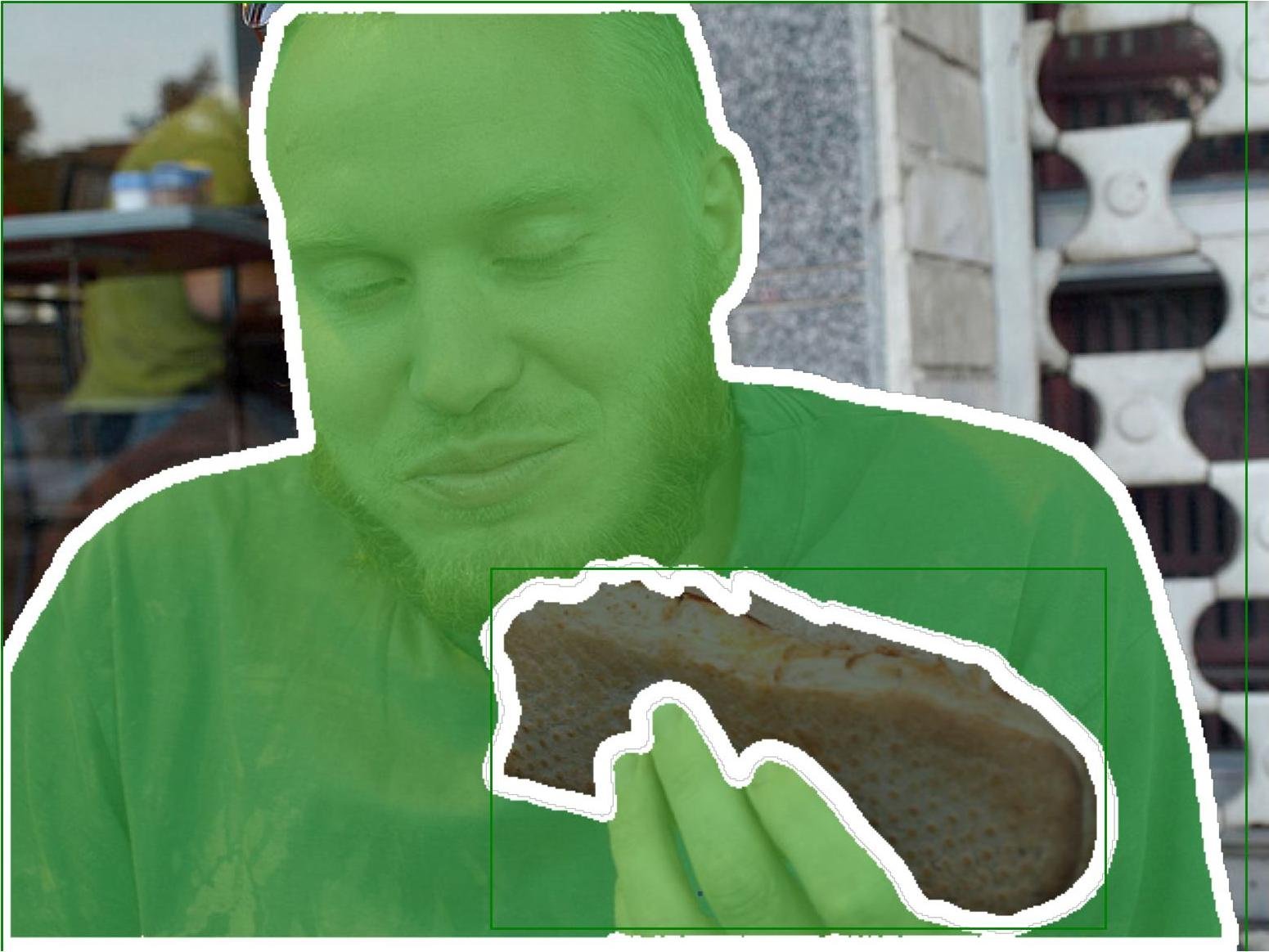} &
        \includegraphics[height=2.1cm]{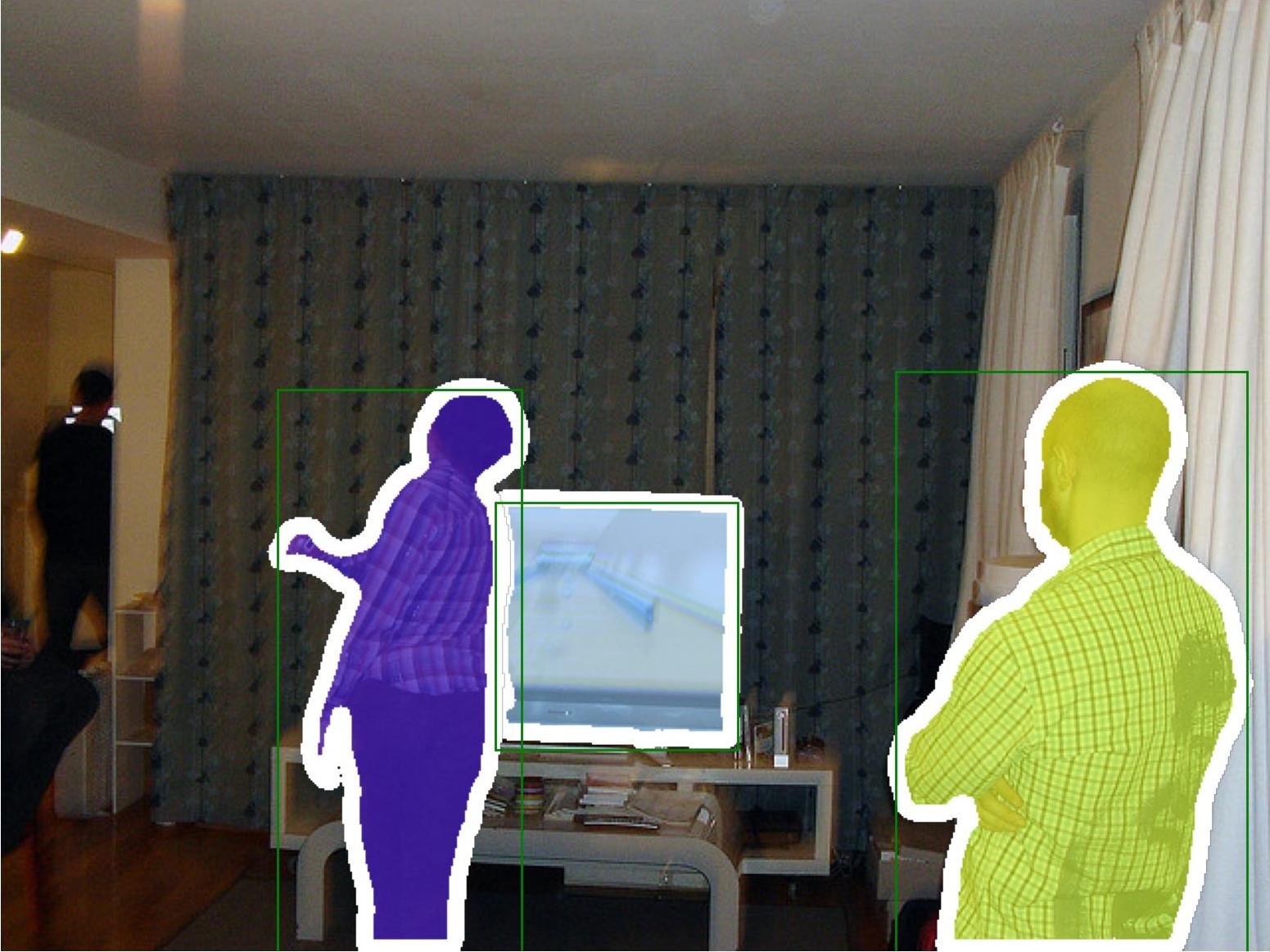}\\ 
        \raisebox{0.1cm}{\makebox[0pt][c]{\rotatebox{90}{\small \textbf{SAM2-Auto}}}} & 
        \includegraphics[height=2.1cm]{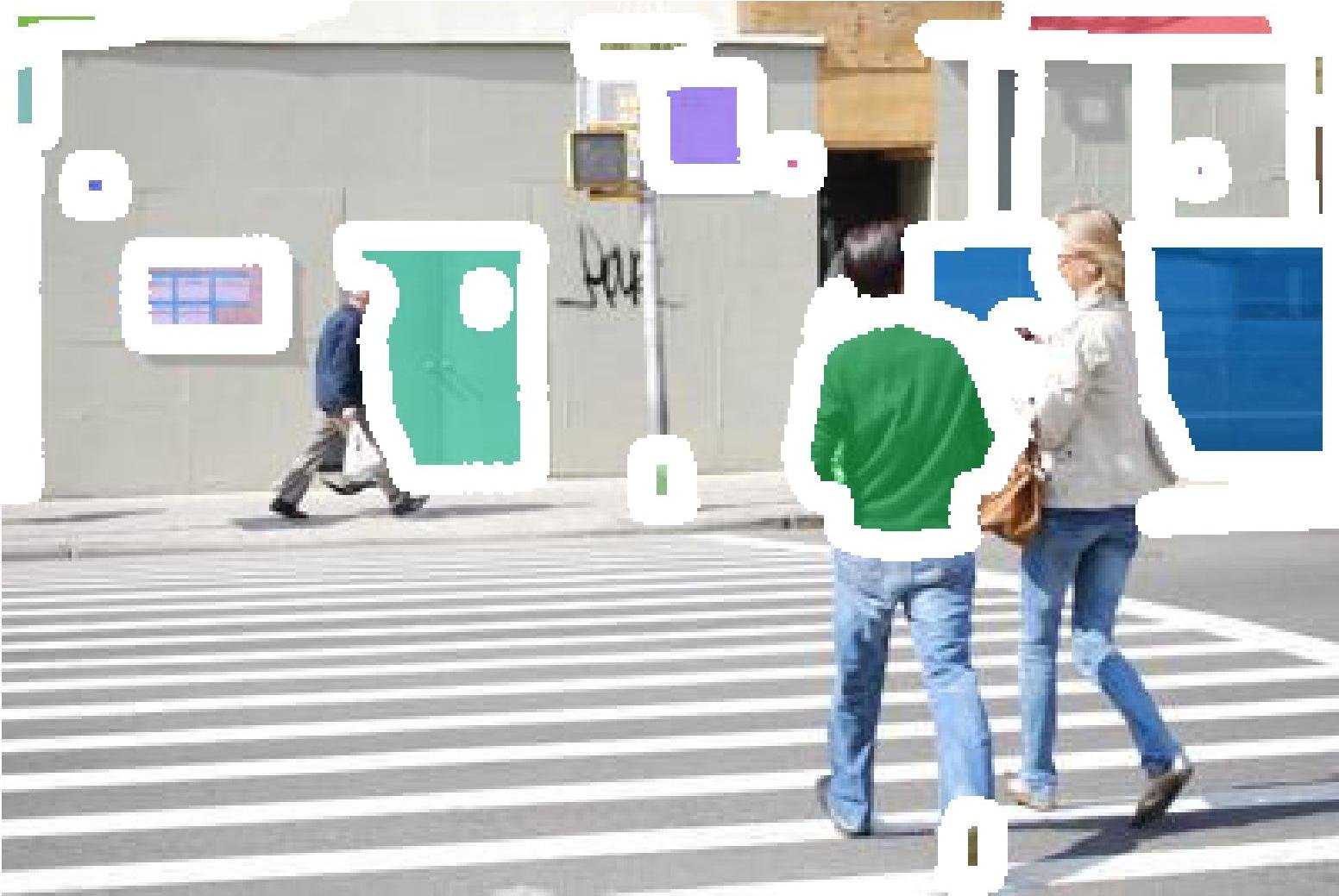} &
        \includegraphics[height=2.1cm]{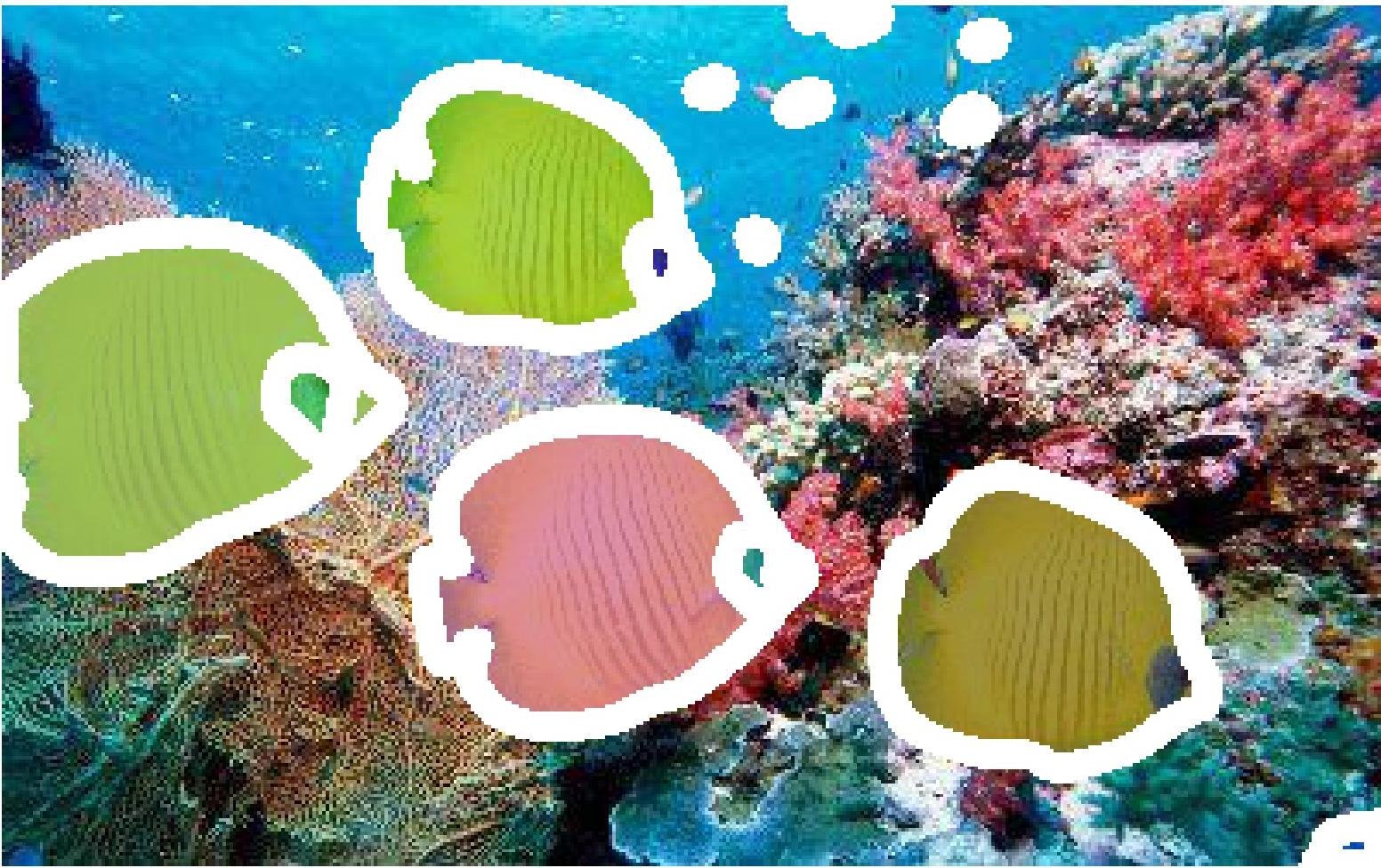} &
        \includegraphics[height=2.1cm]{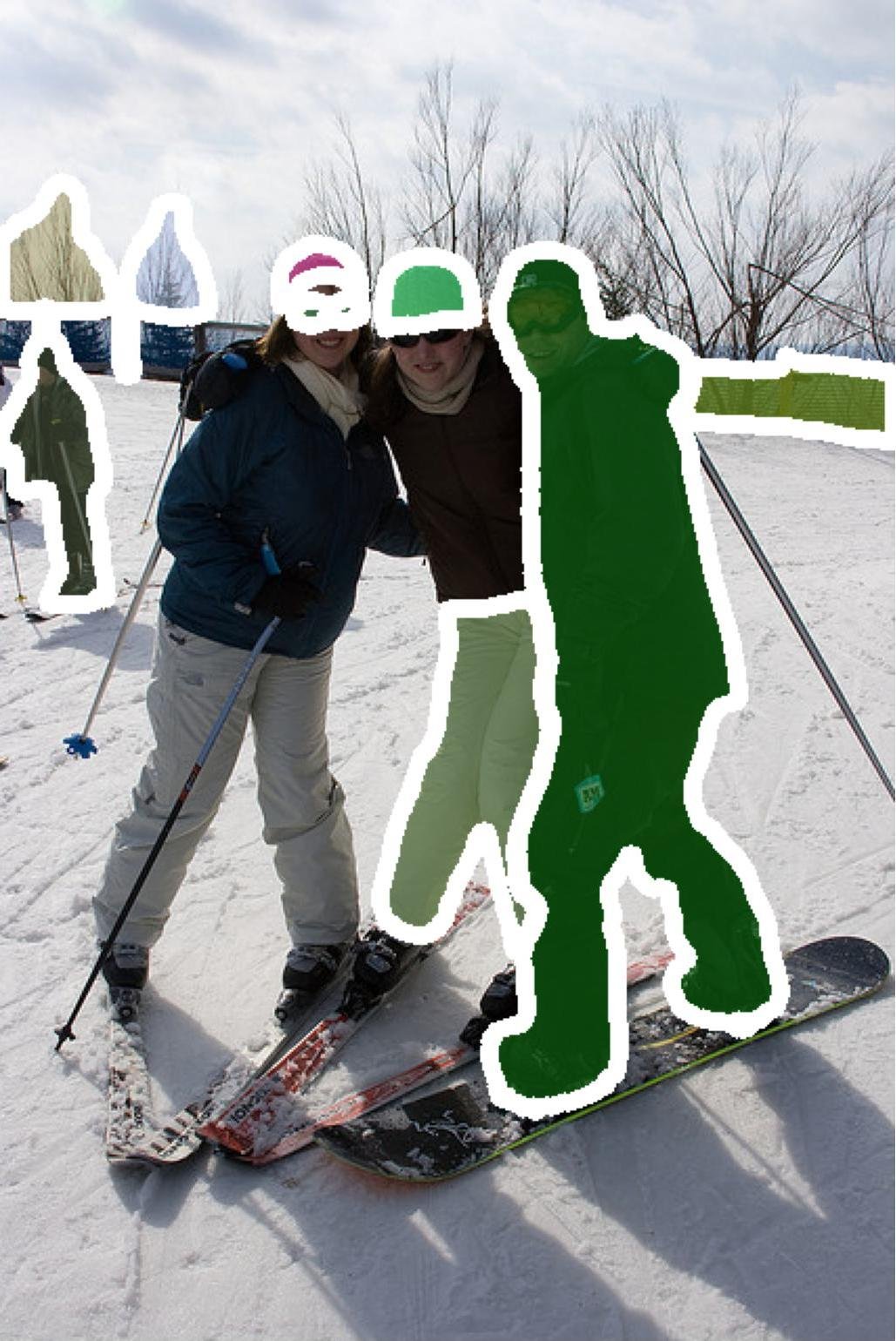} &
        \includegraphics[height=2.1cm]{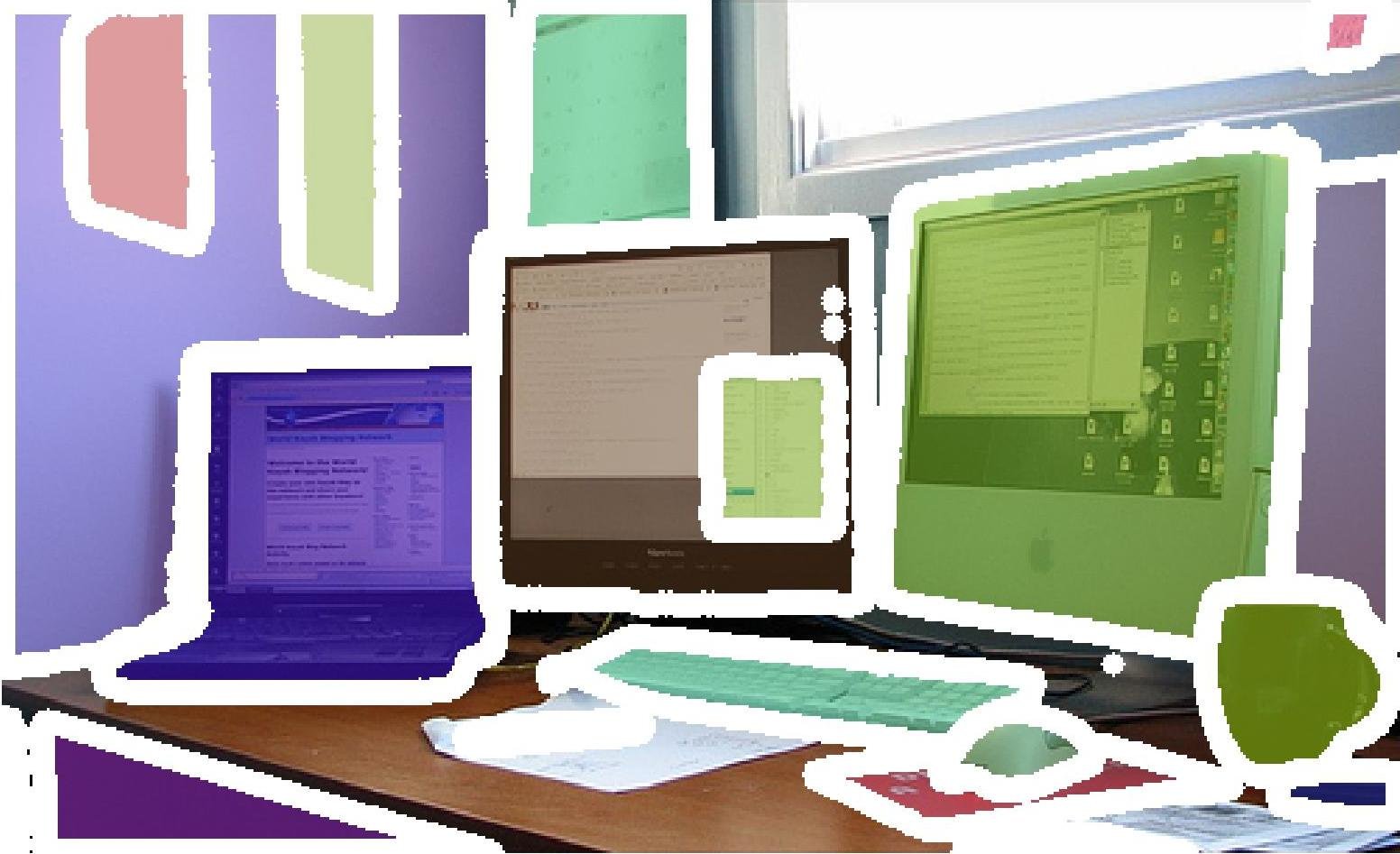} &
        \includegraphics[height=2.1cm]{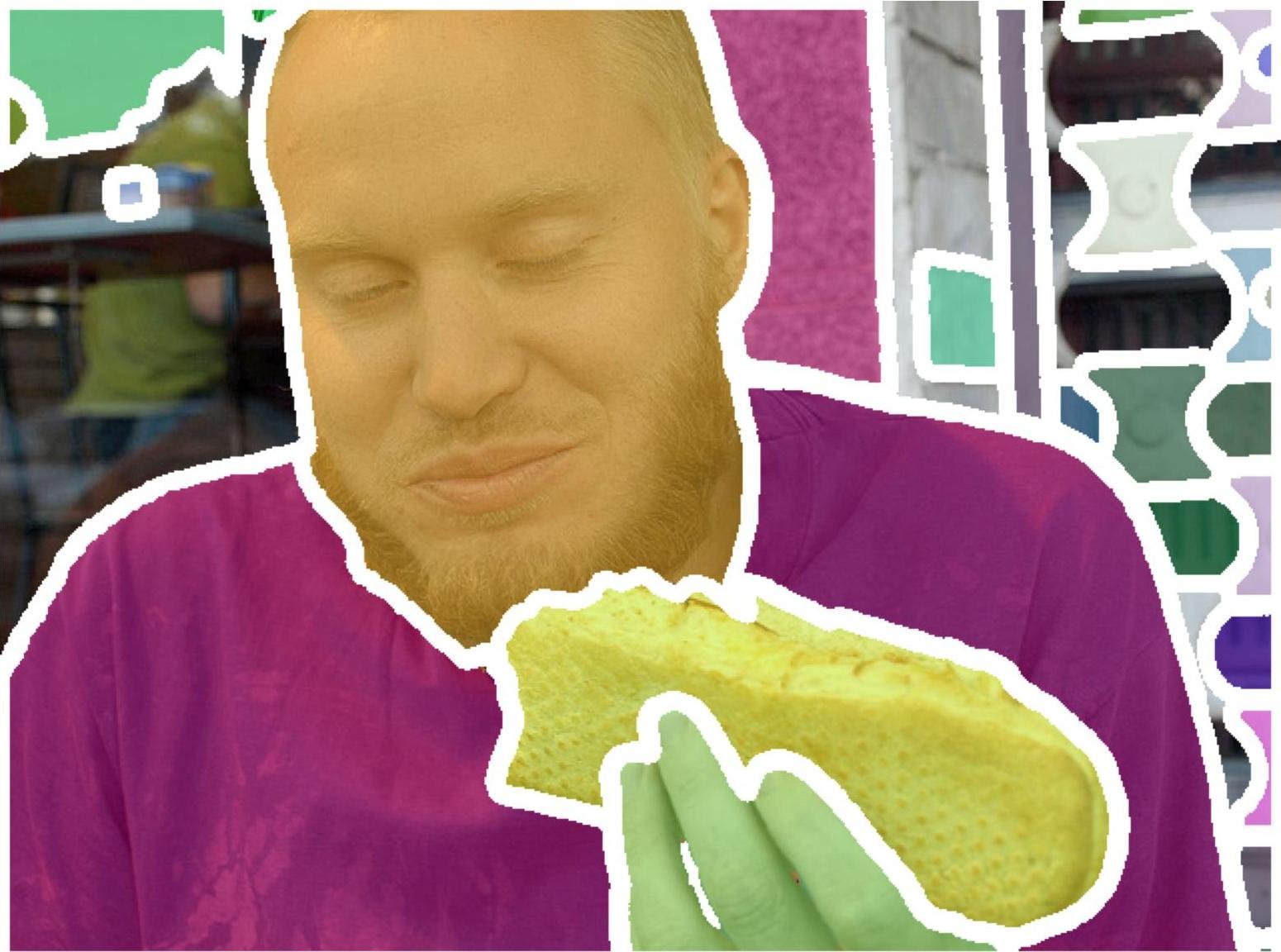} &
        \includegraphics[height=2.1cm]{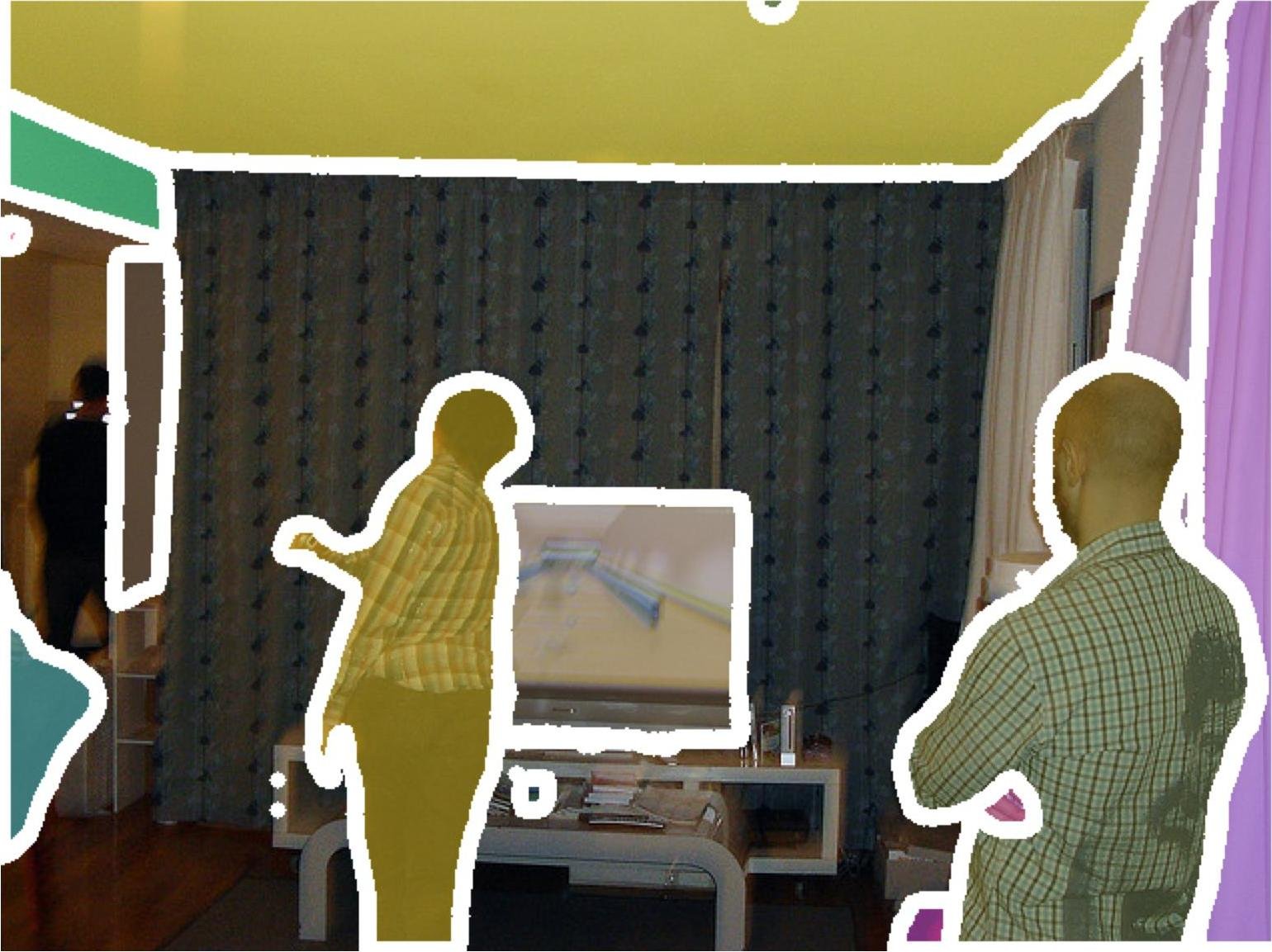}\\ 
        \raisebox{0.1cm}{\makebox[0pt][c]{\rotatebox{90}{\small \textbf{SAM2-Bbox}}}} & 
        \includegraphics[height=2.1cm]{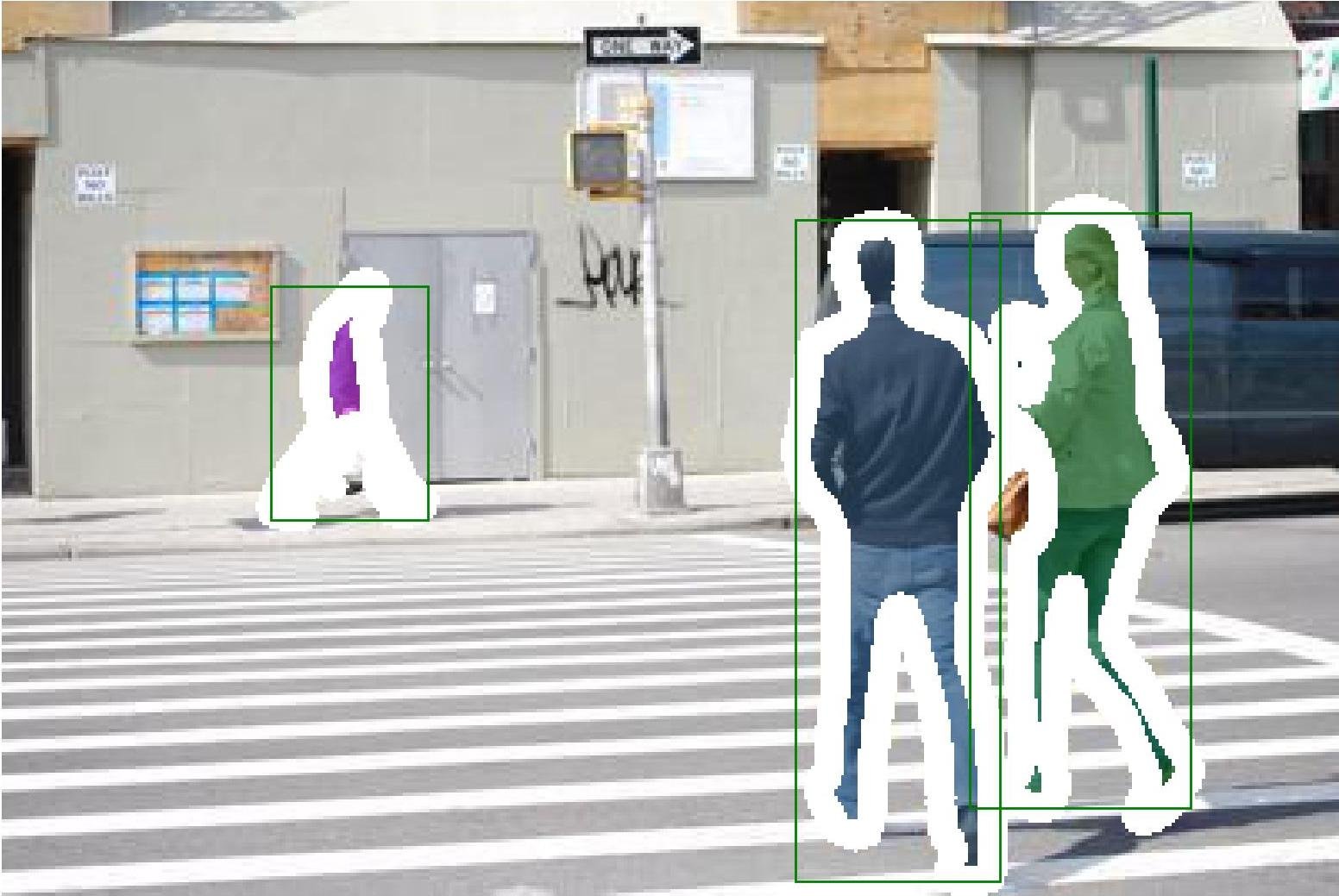} &
        \includegraphics[height=2.1cm]{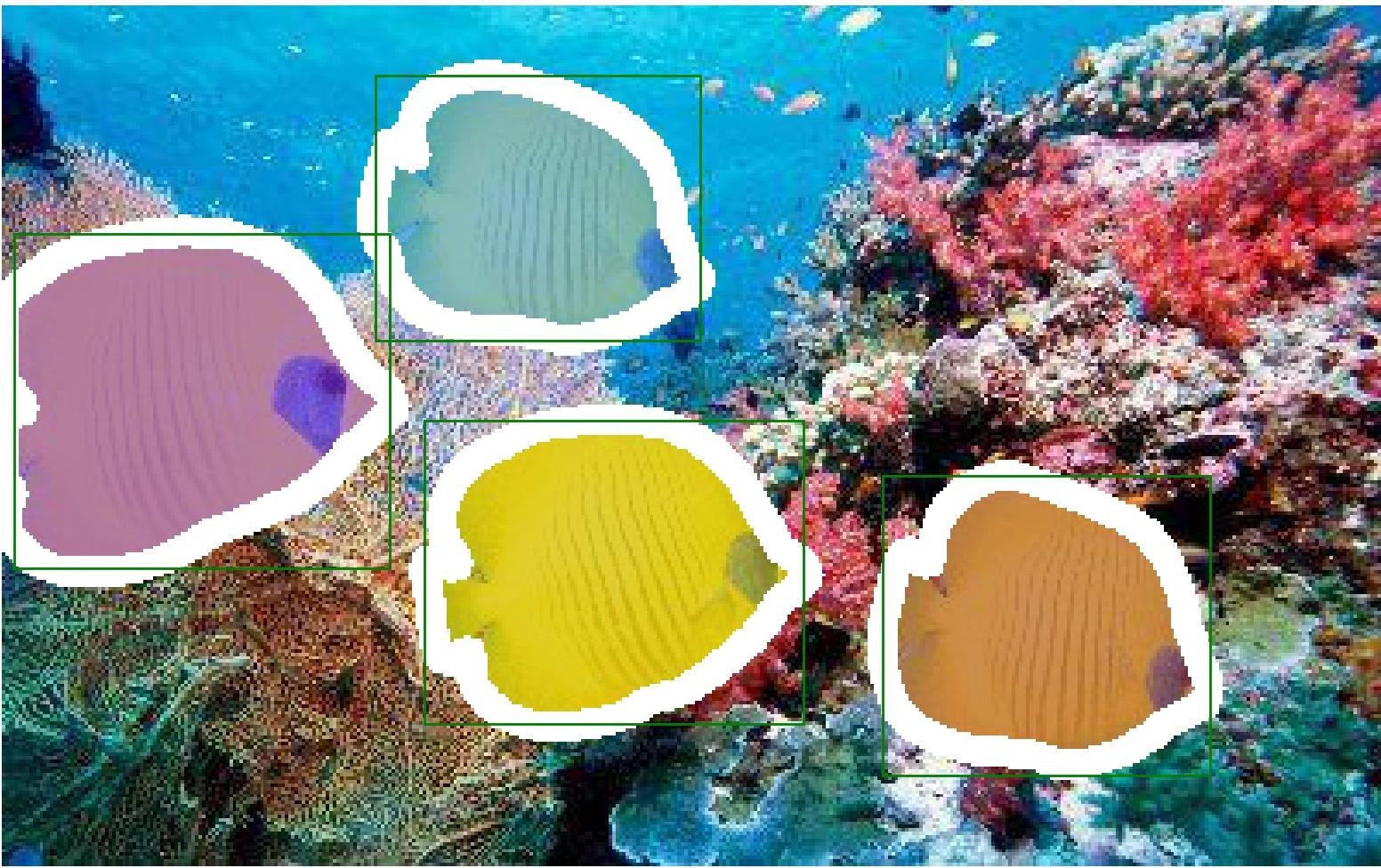} &
        \includegraphics[height=2.1cm]{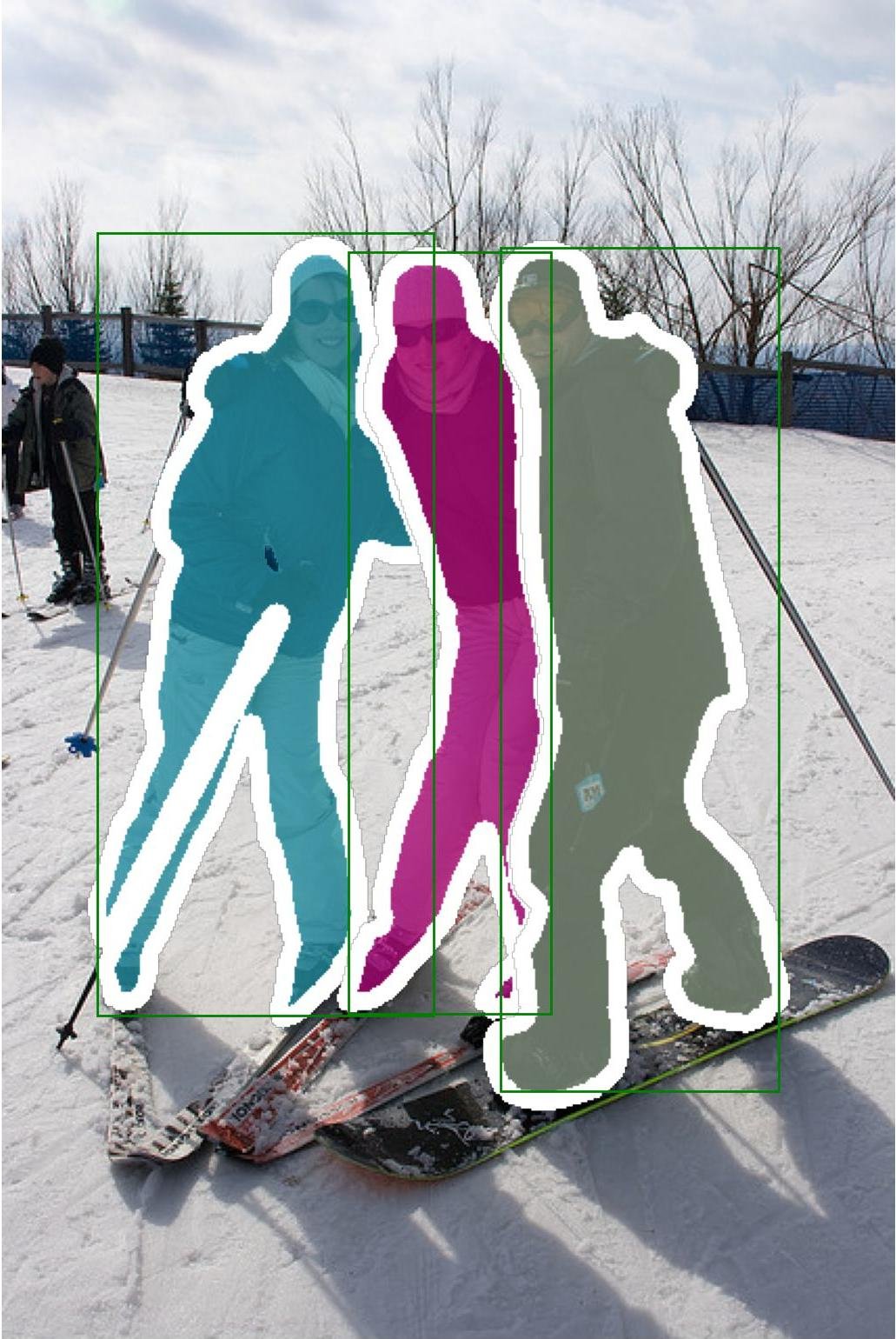} &
        \includegraphics[height=2.1cm]{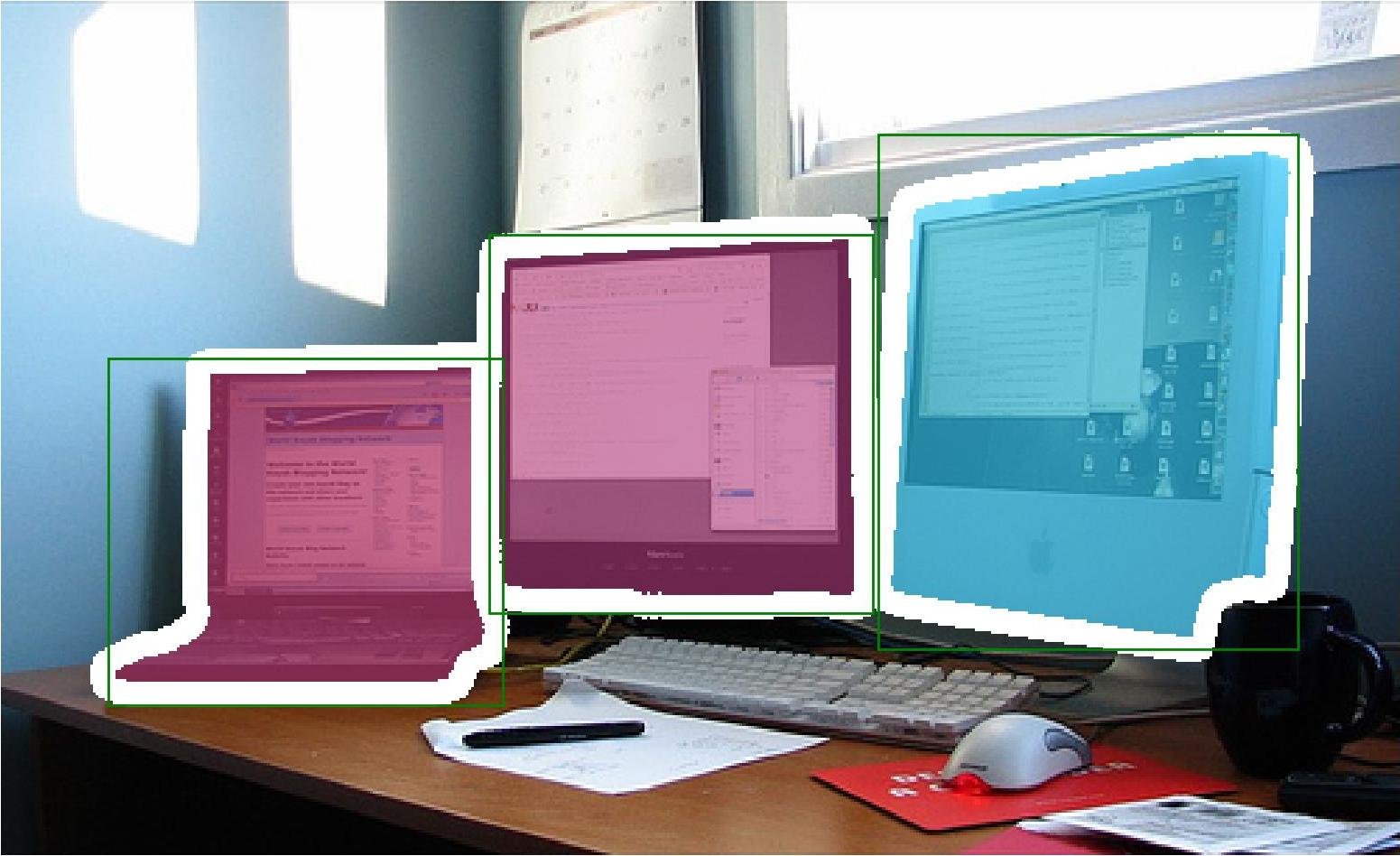} &
        \includegraphics[height=2.1cm]{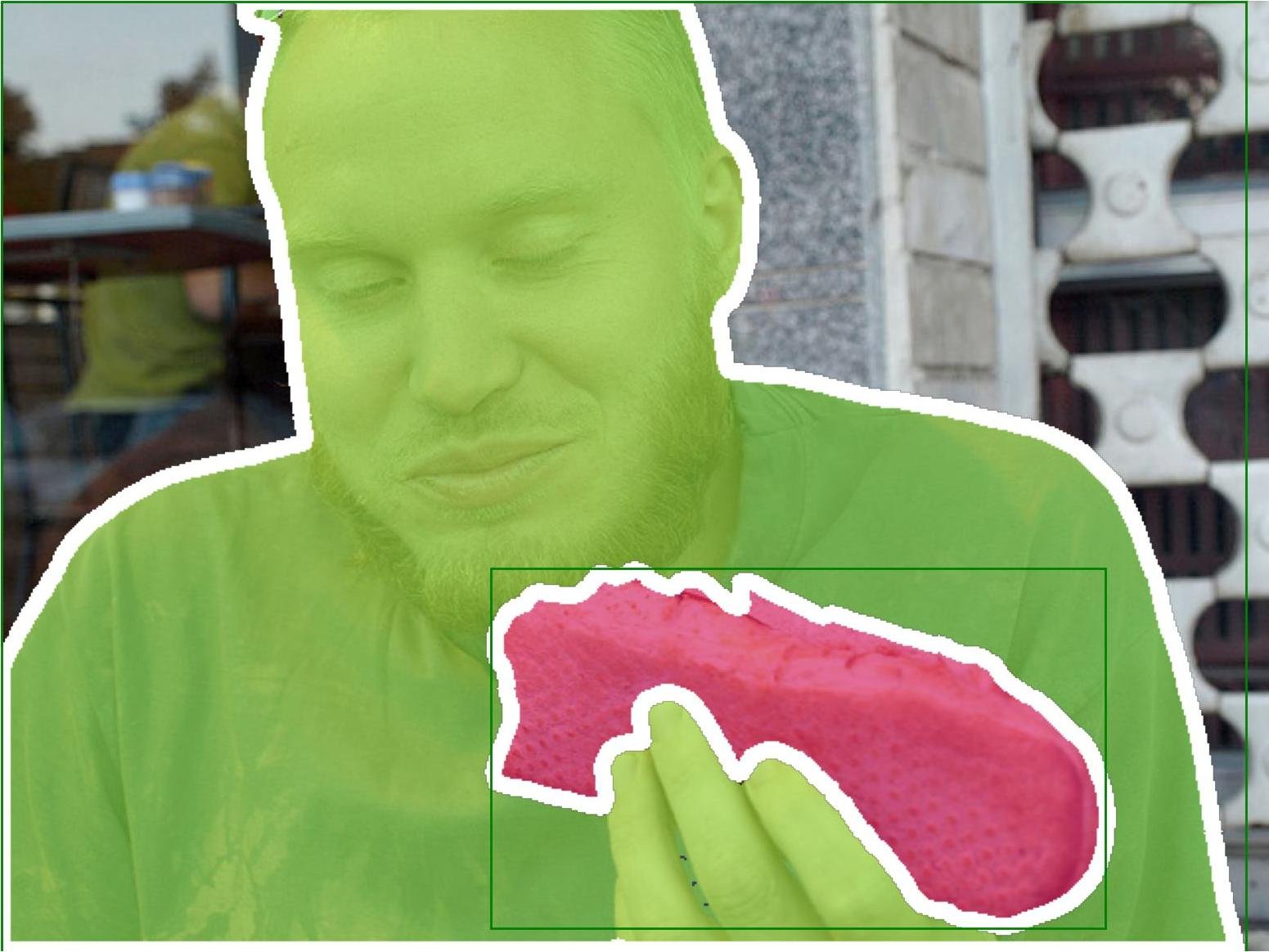} &
        \includegraphics[height=2.1cm]{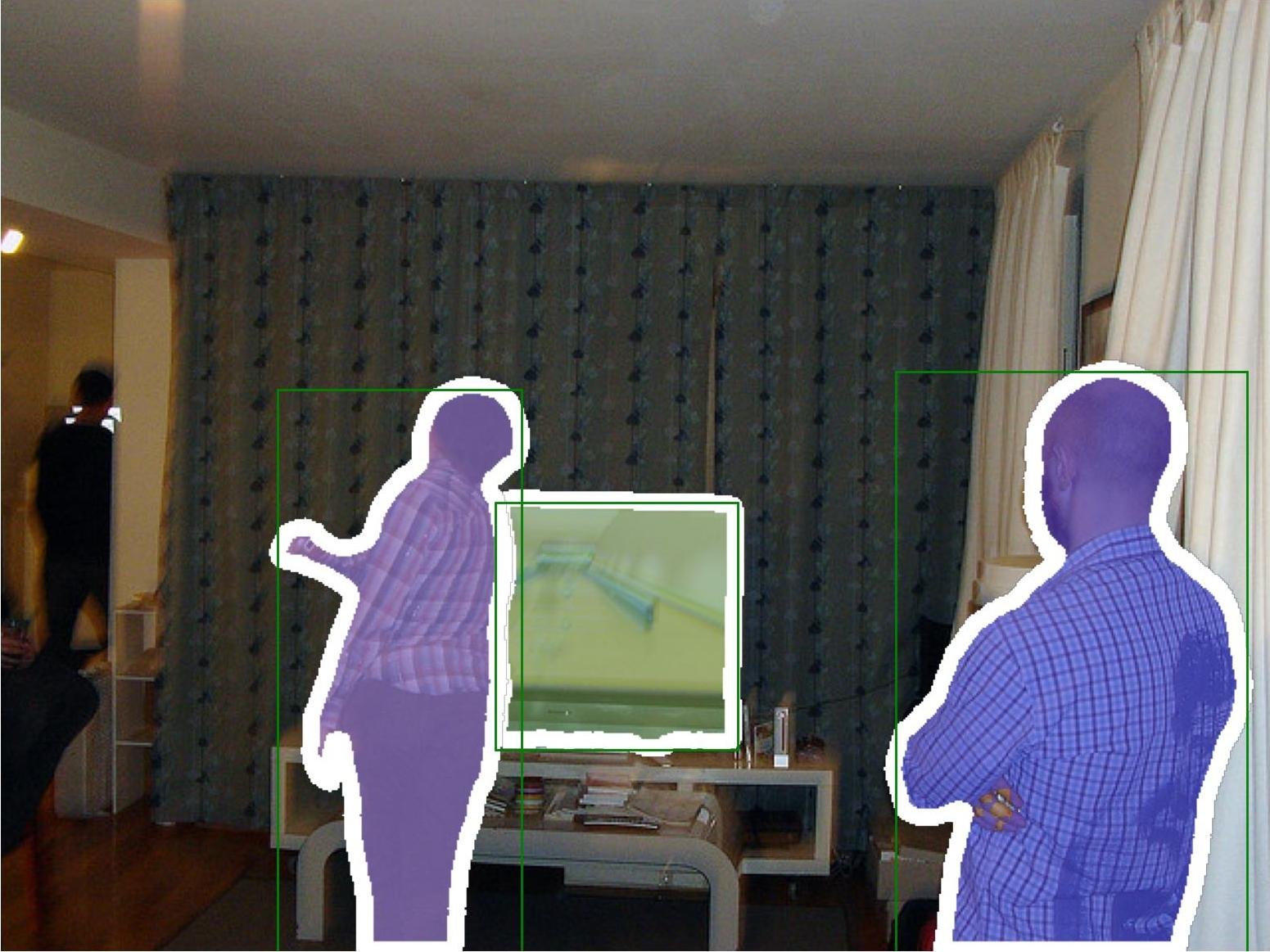}\\ 
    \end{tabular}
    \caption{Qualitative comparisons on ILSO, SIS10K, SOC datasets for salient instance segmentation.}
    \vspace{-5pt}
    \label{fig:SIS}
\end{figure*}

\subsection{Evaluation Protocol}
To evaluate camouflaged instance segmentation, we employ COCO-style evaluation metrics, including AP$_{50}$, AP$_{75}$, and AP values. For salient instance segmentation, we adopt the AP$_{70}$ metric, which is commonly used in related literatures~\cite{fan2019s4net,wu2021regularized}, instead of the AP$_{75}$ metric. In shadow instance segmentation, while task-specific methods employ the SOAP metric to assess object and shadow matching, SAM2 do not involve this matching mechanism. In this regard, we focus only on performance with the instance AP metric.

To assess high-accuracy DIS, we employ six evaluation metrics to evaluate SAM2, SAM, and DIS-specific models, including maximal F-measure ($F_\beta^\text{max}\uparrow$)~\cite{achanta2009frequency}, weighted F-measure ($F_\beta^\omega\uparrow$)~\cite{margolin2014evaluate}, Mean Absolute Error (MAE, $M\downarrow$)~\cite{perazzi2012saliency}, Structural measure (S-measure, $S_\alpha\uparrow$)~\cite{fan2017structure}, mean Enhanced alignment measure (E-measure, $E_{\phi}^\text{m}\uparrow$)~\cite{fan2018enhanced}, and Human Correction Efforts (HCE$_{\gamma}\downarrow$)~\cite{qin2022highly}.

\subsection{Implementation Details}

To ensure a fair comparison, we use the original official code of SAM2 and SAM to test on different datasets. Both SAM2 and SAM are evaluated under two settings: automatic mode and ground truth bounding box~(GT-Bbox) mode. In automatic mode, we use the default setting of a 32$\times$32 point prompt for both. In GT-Bbox mode, the ground truth bounding box serves as the box prompt input. All parameters remain at their default settings, and the input images are resized to 1024$\times$1024. Furthermore, we use different backbones for SAM and SAM2. For SAM, we use ViT-Base, ViT-Large, and ViT-Huge. For SAM2, we use Hiera-Tiny, Hiera-Base+, and Hiera-Large.
All experiments are implemented with a single Tesla A40 GPU.

\subsection{Results}\label{sec:result}

\begin{table*}[t!]
\begin{center}

\footnotesize
\renewcommand{\arraystretch}{1.0}
\renewcommand{\tabcolsep}{3.5mm}
\caption{\textbf{Quantitative comparisons  of SAM2, SAM, and CIS-specific methods on COD10K and NC4K.} SAM-B, SAM-L, and SAM-H represent ViT-B, ViT-L, and ViT-H model types of SAM, respectively. SAM2-T, SAM2-B+, and SAM2-L represent Hiera-Tiny, Hiera-Base+, and Hiera-Large model types of SAM2, respectively. The best and second-best results are bolded and underlined.}
\label{table:CIS}
\begin{tabular}{r||r|r|ccc|ccc}
\toprule
\multirow{2}{*}{Methods} & \multirow{2}{*}{Pub. \& Year} & \multirow{2}{*}{Settings} 
& \multicolumn{3}{c|}{COD10K}   
& \multicolumn{3}{c}{NC4K} 
\\
& & & AP & AP$^{50}$ & AP$^{75}$ & AP & AP$^{50}$ & AP$^{75}$ 
\\ \hline
Mask R-CNN~\cite{he2017mask}& ICCV 2017 & Fully-supervised & 28.7&60.1&25.7&36.1&68.9&33.5\\
Mask2Former~\cite{cheng2022masked}&CVPR 2022 & Fully-supervised & 44.3 & 70.5 & 46.0 & 49.2 & 71.6 & 51.4\\
OSFormer~\cite{pei2022osformer}&ECCV 2022 & Fully-supervised & 41.0 & 71.1 & 40.8 & 42.5 & 72.5 & 42.3\\
DCNet~\cite{luo2023camouflaged}&CVPR 2023 & Fully-supervised & 45.3 & 70.7 & 47.5 & 52.8 & 77.1 & 56.5 \\
UQFormer~\cite{dong2023unified}&MM 2023 & Fully-supervised & 45.4 & 71.8 & 47.9 & 50.1 & 76.8 & 52.8 \\
PointSup~\cite{cheng2022pointly}&CVPR 2022 & Point-supervised & 17.9 & 44.1 & 11.9 & 19.1 & 47.6 & 11.6 \\
Tokencut~\cite{wang2023tokencut}&TPAMI 2023 & Unsupervised & 2.6 & 6.5 & 2.0 & 3.5 & 8.3 & 2.5 \\
Cutler~\cite{wang2023cut}&CVPR 2023 & Unsupervised & 11.7 & 29.1 & 7.3 & 15.5 & 37.9 & 10.5 \\
TPNet~\cite{xia2024text}&MM 2024 & Text-prompt & 18.3 & 41.8 & 14.3 & 21.4 & 48.3 & 16.6 \\
\hline
SAM-B~\cite{kirillov2023segment}& ICCV 2023 & Automatic & 7.6 & 12.3 & 8.2 & 5.7 & 8.8 & 6.3 \\
SAM-L~\cite{kirillov2023segment}& ICCV 2023& Automatic & 29.5 & 45.3 & 32.3 & 26.2 & 38.8 & 29.5 \\
SAM-H~\cite{kirillov2023segment}& ICCV 2023& Automatic & 33.7 & 51.2 & 37.7 & 33.1 & 47.9 & 37.6 \\
\hline
SAM-B~\cite{kirillov2023segment}& ICCV 2023& GT-Bbox & 50.2 & 81.5 & 54.3 & 52.9 & 84.2 & 57.7 \\
SAM-L~\cite{kirillov2023segment}& ICCV 2023& GT-Bbox & 58.9 & 88.4 & 65.5 & 62.0 & 90.3 & 69.9 \\
SAM-H~\cite{kirillov2023segment}& ICCV 2023& GT-Bbox & 59.6 & 87.2 & 67.0 & 63.4 & 89.3 & 72.0 \\
\hline
\rowcolor[RGB]{216,241,244}
\textbf{SAM2-T}~\cite{ravi2024sam}& -- & Automatic & 3.1 & 4.0 & 3.4 & 3.4 & 4.2 & 3.8 \\
\rowcolor[RGB]{216,241,244}
\textbf{SAM2-B+}~\cite{ravi2024sam}& -- & Automatic & 11.7 & 15.7 & 13.1 & 8.9 & 11.0 & 9.8 \\
\rowcolor[RGB]{216,241,244}
\textbf{SAM2-L}~\cite{ravi2024sam}& -- & Automatic & 10.6 & 13.2 & 12.1 & 8.8 & 10.3 & 9.6 \\
\hline
\rowcolor[RGB]{216,241,244}
\textbf{SAM2-T}~\cite{ravi2024sam}& -- & GT-Bbox & 59.3 & 88.8 & 65.2 & 66.4 & 92.1 & 74.9 \\
\rowcolor[RGB]{216,241,244}
\textbf{SAM2-B+}~\cite{ravi2024sam}& -- & GT-Bbox & \underline{63.0} & \underline{90.0} & \underline{70.5} & \underline{69.6} & \underline{93.3} & \underline{79.2}  \\
\rowcolor[RGB]{216,241,244}
\textbf{SAM2-L}~\cite{ravi2024sam}& -- & GT-Bbox & \textbf{68.8} & \textbf{94.6} & \textbf{78.3} & \textbf{73.5} & \textbf{95.5} & \textbf{83.7} \\
\bottomrule

\end{tabular}
\end{center}
\end{table*}

\begin{table*}[t!]
\begin{center}

\footnotesize
\renewcommand{\arraystretch}{1.2}
\renewcommand{\tabcolsep}{1.8mm}
\caption{\textbf{Quantitative comparisons of SAM2, SAM, and SID-specific methods on SOBA-challenge and SOBA-test.} SAM-B, SAM-L, and SAM-H represent ViT-B, ViT-L, and ViT-H model types of SAM, respectively. SAM2-T, SAM2-B+, and SAM2-L represent Hiera-Tiny, Hiera-Base+, and Hiera-Large model types of SAM2, respectively. The best and second-best results are bolded and underlined.}
\label{table:shadow}

\begin{tabular}{r||r|r|cccccc|cccccc}
\toprule
\multirow{2}{*}{Methods}& \multirow{2}{*}{Pub. \& Year} & \multirow{2}{*}{Settings} 
& \multicolumn{6}{c|}{SOBA-challenge}   
& \multicolumn{6}{c}{SOBA-test} 
\\
& & & AP & AP$^{50}$ & AP$^{75}$ & AR$^s$ & AR$^m$ & AR$^l$ 
& AP & AP$^{50}$ & AP$^{75}$ & AR$^s$ & AR$^m$ & AR$^l$  
\\ \hline
LISA~\cite{wang2020instance}& CVPR 2020&Fully-supervised & 23.8 & -- & -- & -- & -- & -- & 39.2 & -- & -- & -- & -- & -- \\
SSIS~\cite{wang2021single} & CVPR 2021&Fully-supervised & 25.6 & -- & -- & -- & -- & -- & 43.4 & -- & -- & -- & -- & -- \\
SSISv2~\cite{wang2022instance} &TPAMI 2022 &Fully-supervised & 31.0 & -- & -- & -- & -- & -- & 50.2 & -- & -- & -- & -- & -- \\
\hline
SAM-B~\cite{kirillov2023segment}& ICCV 2023&Automatic & 12.0 & 16.8 & 13.6 & 17.7 & 19.9 & 23.8 & 18.7 & 26.5 & 21.1 & 22.4 & 31.8 & 35.7 \\
SAM-L~\cite{kirillov2023segment}& ICCV 2023&Automatic & 28.6 & 38.5 & 33.0 & 23.1 & 34.0 & 40.1 & 35.6 & 47.8 & 40.8 & 26.6 & 42.5 & 52.6 \\
SAM-H~\cite{kirillov2023segment}& ICCV 2023&Automatic & 29.7 & 40.2 & 34.1 & 23.0 & 35.2 & 40.8 & 35.1 & 47.1 & 40.0 & 24.8 & 41.3 & 53.0 \\
\hline
SAM-B~\cite{kirillov2023segment}& ICCV 2023&GT-Bbox & 46.2 & 74.0 & 48.7 & 40.7 & 52.2 & 52.6 & 53.8 & 86.3 & 56.2 & 45.9 & 57.9 & 65.5 \\
SAM-L~\cite{kirillov2023segment}& ICCV 2023&GT-Bbox & 46.1 & 70.6 & 48.8 & 39.6 & 51.8 & 53.7 & 53.3 & 83.2 & 56.6 & 44.1 & 57.6 & 66.0 \\
SAM-H~\cite{kirillov2023segment}& ICCV 2023&GT-Bbox & 45.2 & 69.2 & 48.5 & 39.0 & 50.4 & 52.2 & 51.7 & 80.6 & 53.4 & 43.2 & 55.8 & 63.5 \\
\hline
\rowcolor[RGB]{216,241,244}
\textbf{SAM2-T}~\cite{ravi2024sam} & --&Automatic & 5.4 & 6.5 & 6.0 & 6.4 & 8.6 & 16.5 & 13.6 & 16.3 & 15.0 & 10.1 & 17.0 & 24.5 \\
\rowcolor[RGB]{216,241,244}
\textbf{SAM2-B+}~\cite{ravi2024sam} & --&Automatic & 13.2 & 15.7 & 14.5 & 7.4 & 15.1 & 26.2 & 22.2 & 26.8 & 24.7 & 16.9 & 24.7 & 34.7 \\
\rowcolor[RGB]{216,241,244}
\textbf{SAM2-L}~\cite{ravi2024sam} & --&Automatic & 13.6 & 16.2 & 15.2 & 10.7 & 14.0 & 27.0 & 22.9 & 26.7 & 25.4 & 16.3 & 22.8 & 34.8 \\
\hline
\rowcolor[RGB]{216,241,244}
\textbf{SAM2-T}~\cite{ravi2024sam} & --&GT-Bbox & \textbf{51.9} & \textbf{80.3} & \textbf{54.7} & \textbf{42.0} & \textbf{54.9} & \textbf{60.8} & \textbf{58.9} & \textbf{86.9} & \underline{62.0} & \textbf{51.7} & \textbf{62.1} & \underline{70.4} \\
\rowcolor[RGB]{216,241,244}
\textbf{SAM2-B+}~\cite{ravi2024sam} & --&GT-Bbox & 48.9 & 72.0 & 52.0 & \underline{39.8} & 53.4 & 57.4 & 56.5 & 81.4 & 60.5 & 47.9 & 60.0 & 68.1 \\
\rowcolor[RGB]{216,241,244}
\textbf{SAM2-L}~\cite{ravi2024sam} & --&GT-Bbox & \underline{49.7} & \underline{73.8} & \underline{53.1} & \underline{39.8} & \underline{54.2} & \underline{59.3} & \underline{58.2} & \underline{82.7} & \textbf{62.6} & \underline{49.5} & \underline{61.1} & \textbf{71.2} \\
\bottomrule
\end{tabular}
\vspace{-5pt}
\end{center}
\end{table*}

\subsubsection{Salient Instance Segmentation}

\noindent\textbf{Quantitative Results.}
The quantitative results of salient instance segmentation are presented in \tabref{table:SIS}. On ILSO and SIS10K datasets, SAM2 models generally outperform in the GT-Bbox setting. For example, SAM2-L achieves an AP score of 82.2 on ILSO, slightly higher than 79.2 of SAM-H. 

However, in the automatic setting, SAM2-L scores lower, with an AP of 49.1 versus SAM-H's 72.2. 
A similar trend is observed on SIS10K, with SAM2-L reaching 45.2 compared to SAM-H's 68.4.
On SOC and SIP datasets, SAM2 models also excel in the GT-Bbox setting, with SAM2-L scoring 83.1 on SOC and 93.4 on SIP. 
Compared to specific methods like SCNet, S4Net, RDPNet, and OQTR, SAM2 models often achieve higher AP scores in the GT-Bbox setting. 
This indicates that SAM2 models show significant improvements with ground truth boxes, outperforming traditional methods in certain scenarios. 
However, it should be noted that SAM2's bounding box mode relies on inputting the ground truth instance locations, which might introduce a slight unfair compared to other frameworks.

\noindent\textbf{Qualitative Results.}
As shown in~\figref{fig:SIS}, in the qualitative analysis of salient instance segmentation, both SAM-Auto and SAM2-Auto perform global segmentation because they do not specify particular objects to segment. 
The segmentation quality of SAM is slightly better, likely attributed to SAM using a larger version (huge) compared to large version of SAM2. 
This difference in model size may account for the finer details in segmentation masks of SAM, which, though they still appear somewhat fragmented. 
Nonetheless, when bounding box prompts are adopted, both SAM-bbox and SAM2-bbox achieve significantly improved and precise segmentation, highlighting the value of guided segmentation.

\begin{figure*}[!t]
    \centering
    \renewcommand{\arraystretch}{0.7}
    \begin{tabular}{@{}c@{\hskip 5pt}c@{\hskip 1pt}c@{\hskip 1pt}c@{\hskip 1pt}c@{\hskip 1pt}c@{\hskip 1pt}c@{}}
        \raisebox{0.5cm}{\makebox[0pt][c]{\rotatebox{90}{\small Image}}} & 
        \includegraphics[height=2cm]{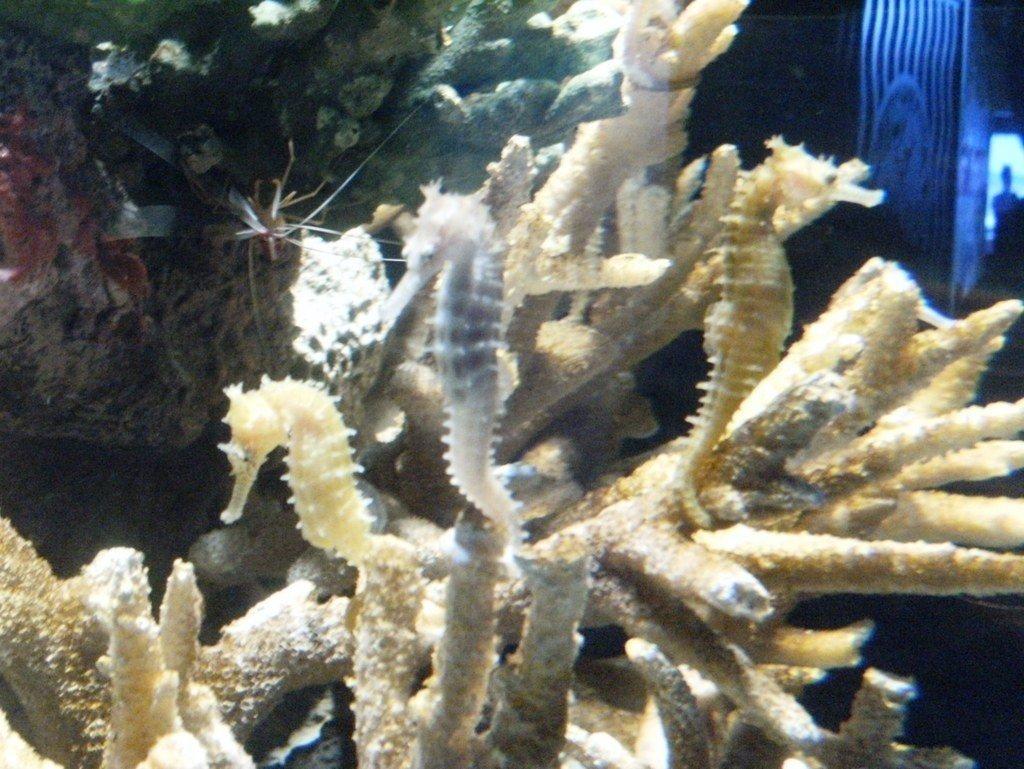} &
        \includegraphics[height=2cm]{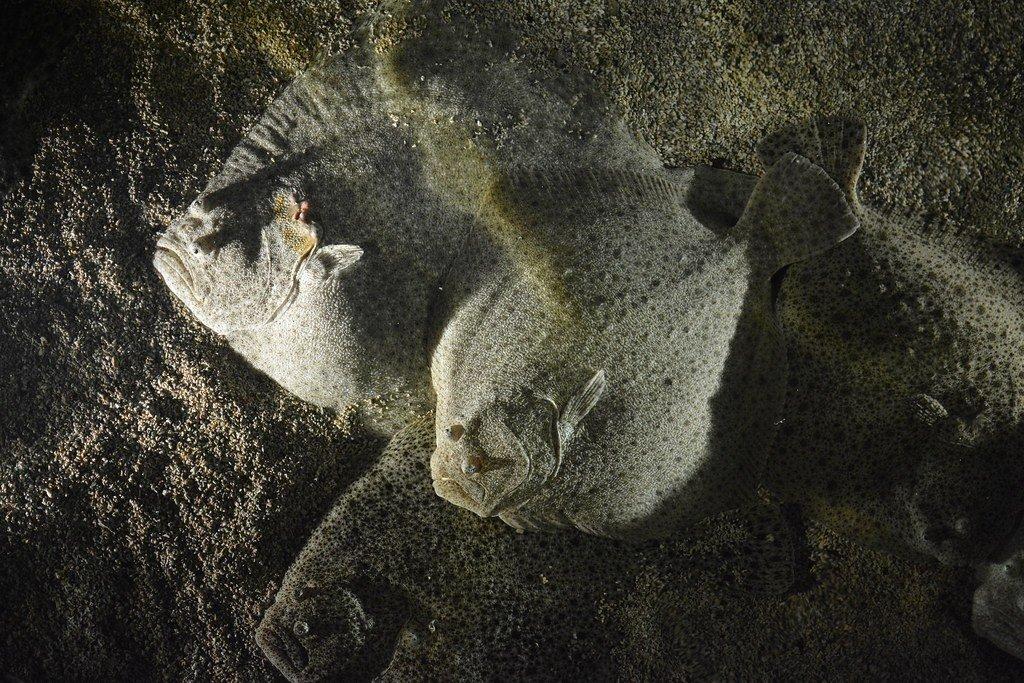} &
        \includegraphics[height=2cm]{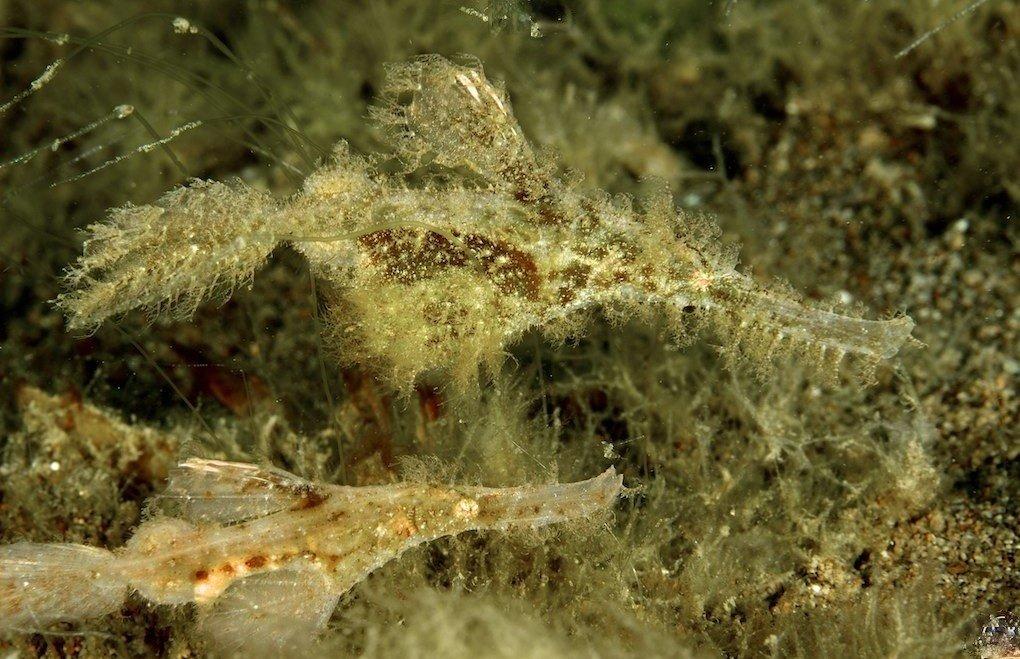} &
        \includegraphics[height=2cm]{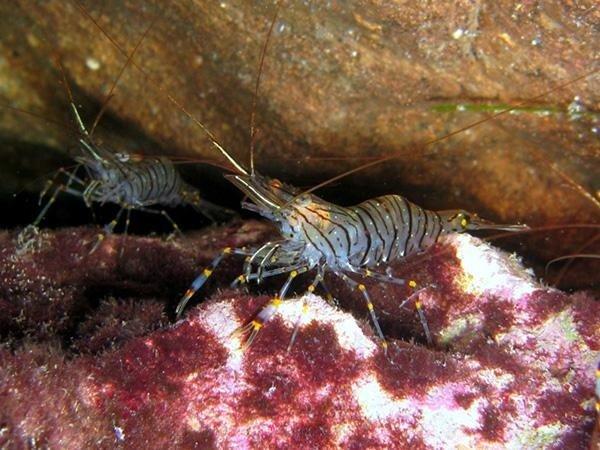} &
        \includegraphics[height=2cm]{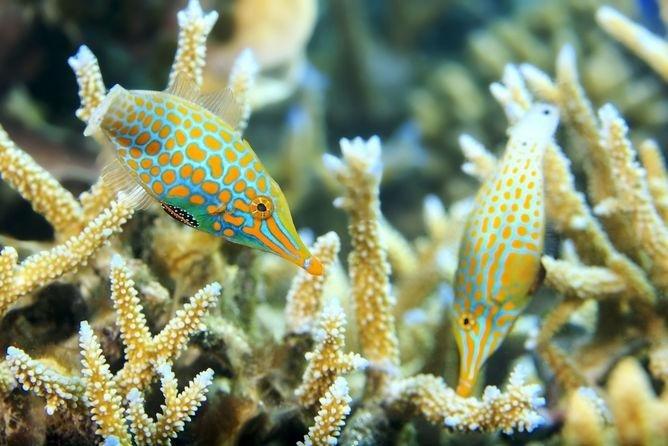} &
        \includegraphics[height=2cm]{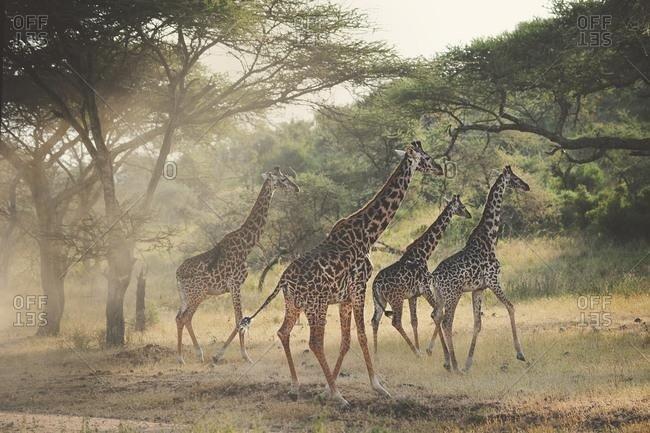}\\ 
        \raisebox{0cm}{\makebox[0pt][c]{\rotatebox{90}{\small Ground Truth}}} & 
        \includegraphics[height=2cm]{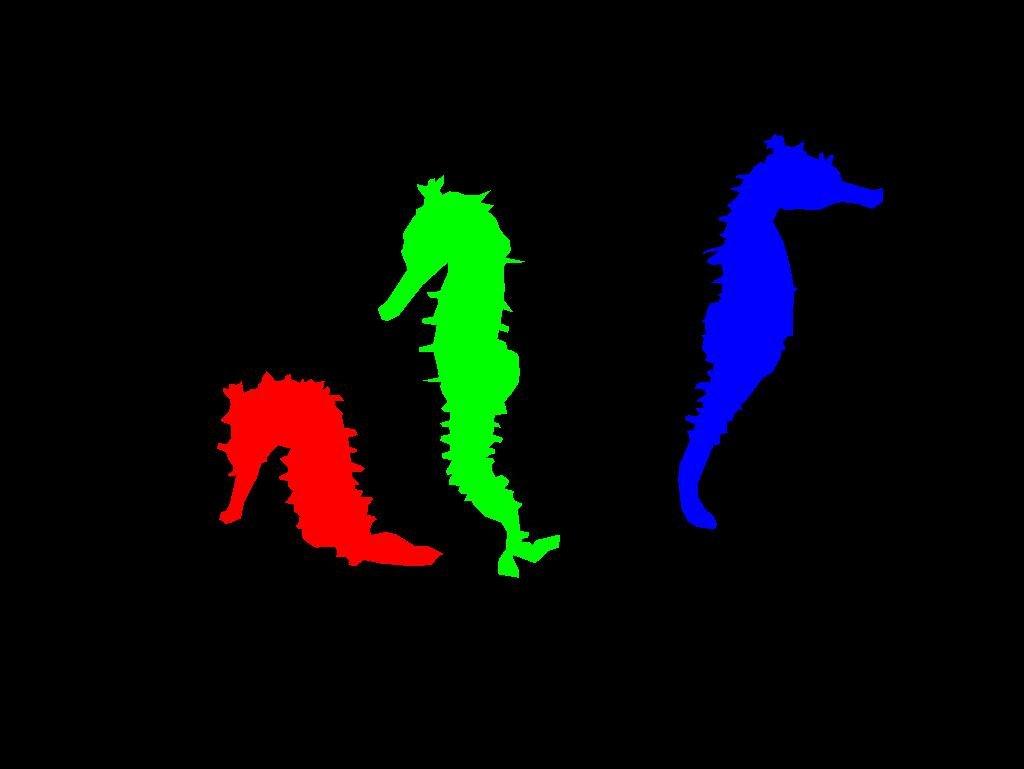} &
        \includegraphics[height=2cm]{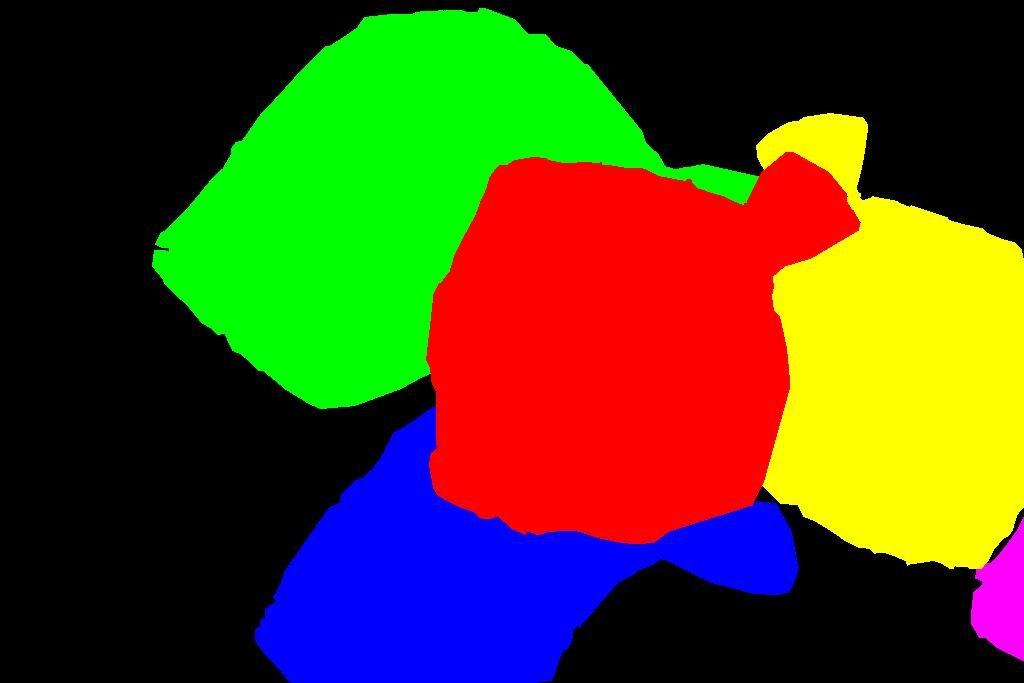} &
        \includegraphics[height=2cm]{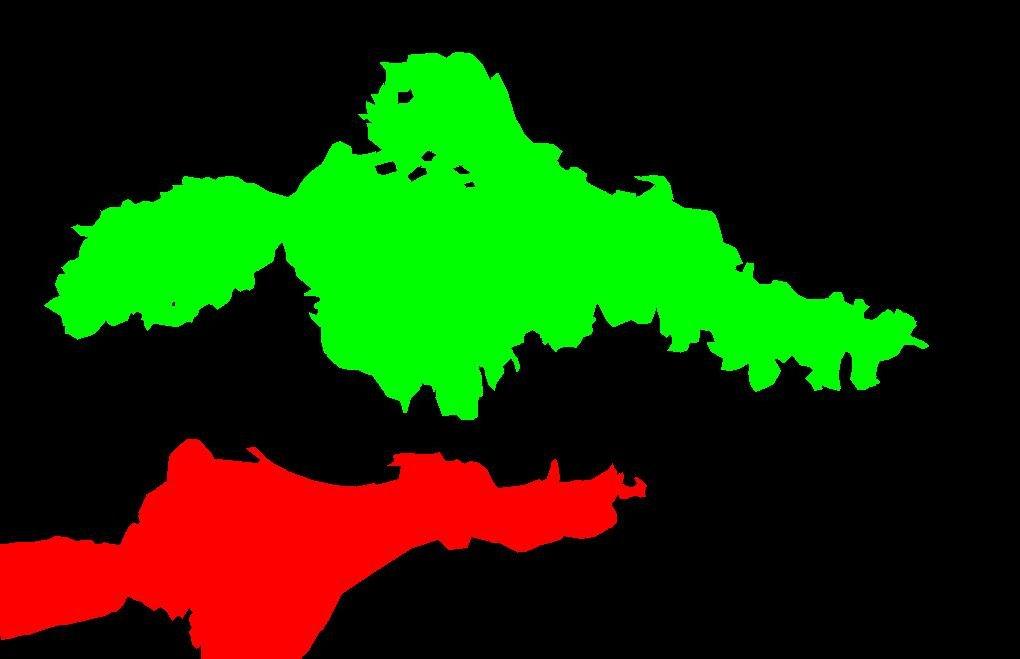} &
        \includegraphics[height=2cm]{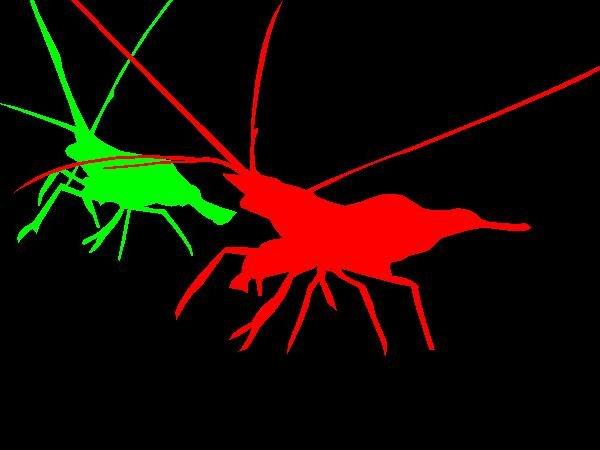} &
        \includegraphics[height=2cm]{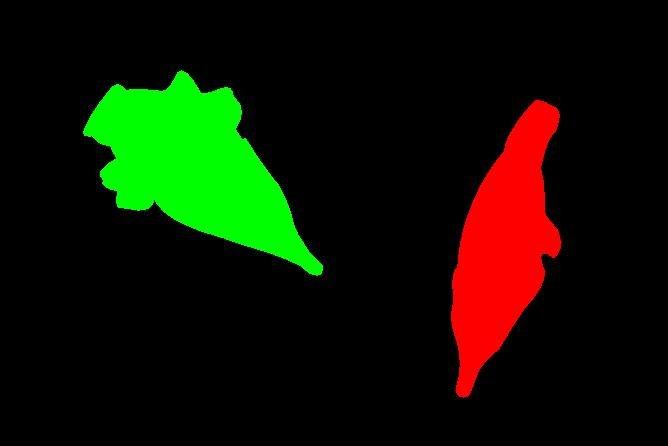} &
        \includegraphics[height=2cm]{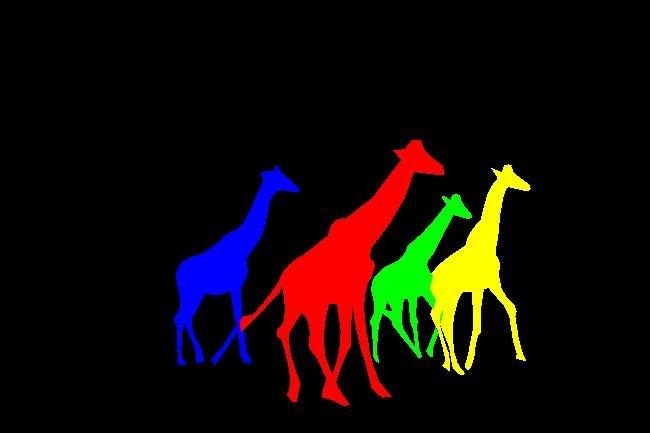} \\ 
        \raisebox{0cm}{\makebox[0pt][c]{\rotatebox{90}{\small Mask R-CNN}}} & 
        \includegraphics[height=2cm]{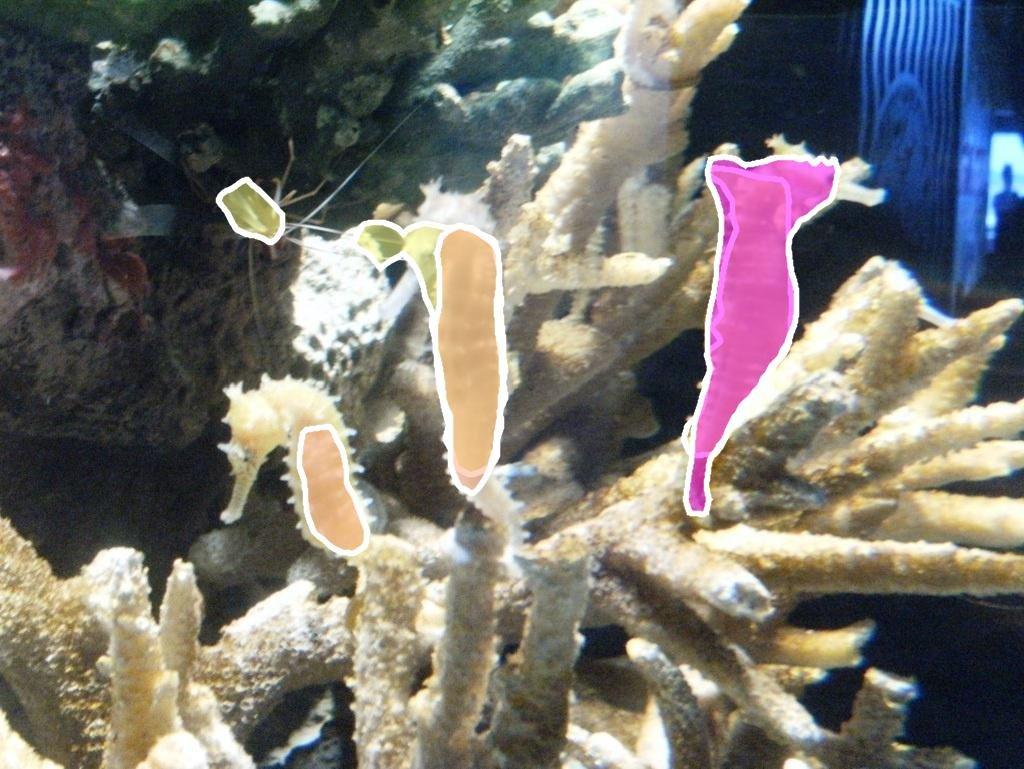} &
        \includegraphics[height=2cm]{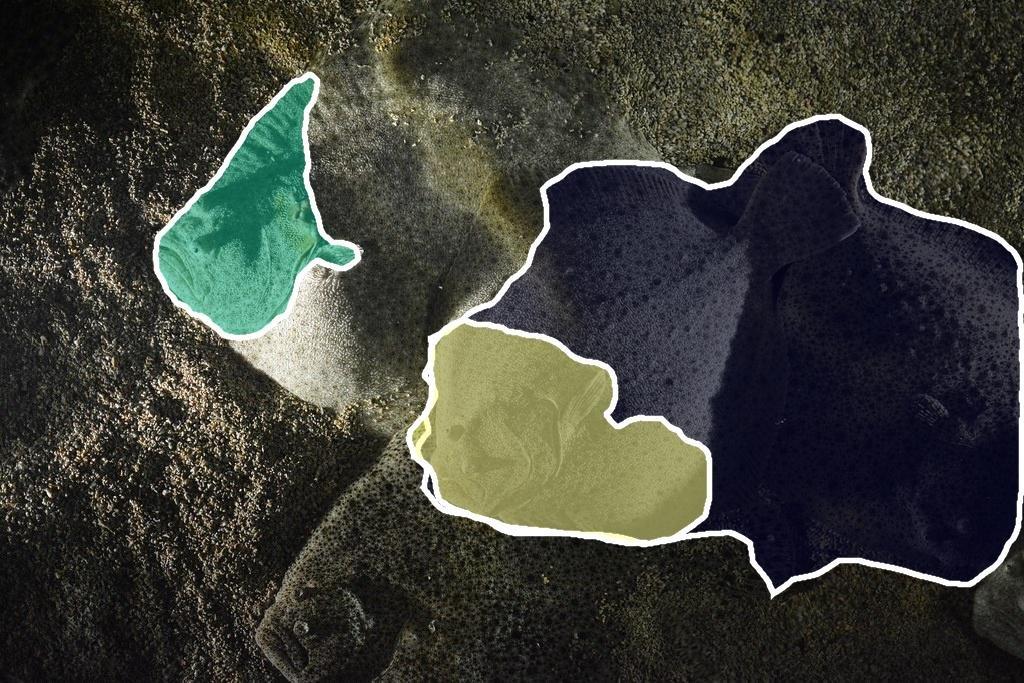} &
        \includegraphics[height=2cm]{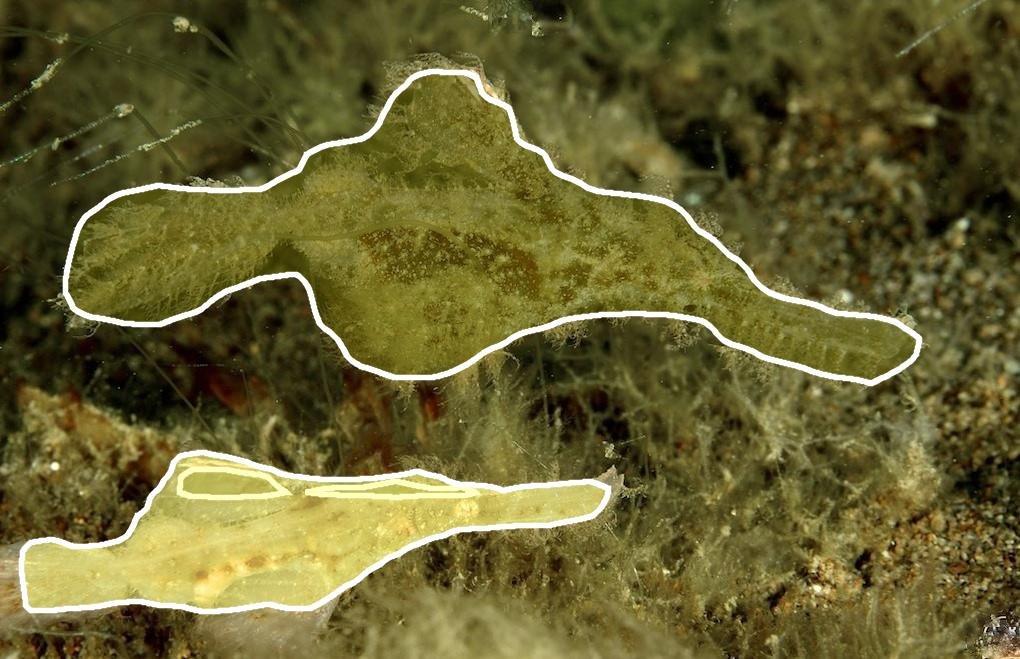} &
        \includegraphics[height=2cm]{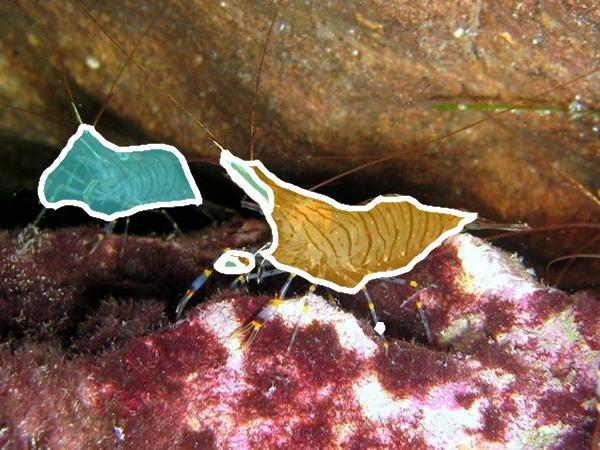} &
        \includegraphics[height=2cm]{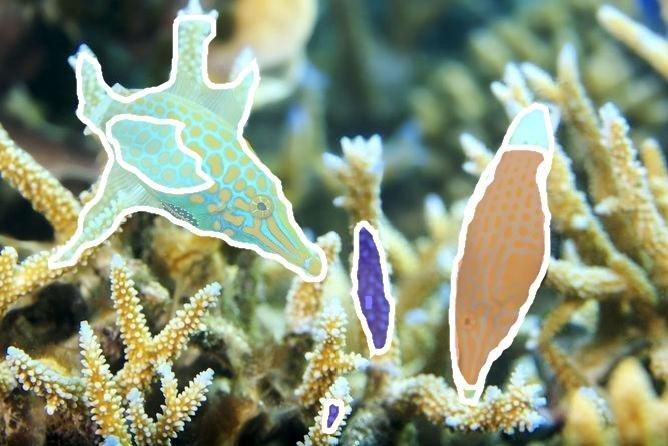} &
        \includegraphics[height=2cm]{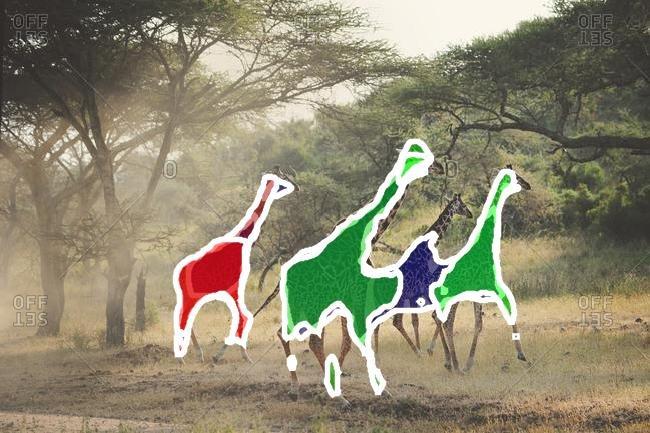} \\ 
        \raisebox{0.4cm}{\makebox[0pt][c]{\rotatebox{90}{\small SOLOv2}}} & 
        \includegraphics[height=2cm]{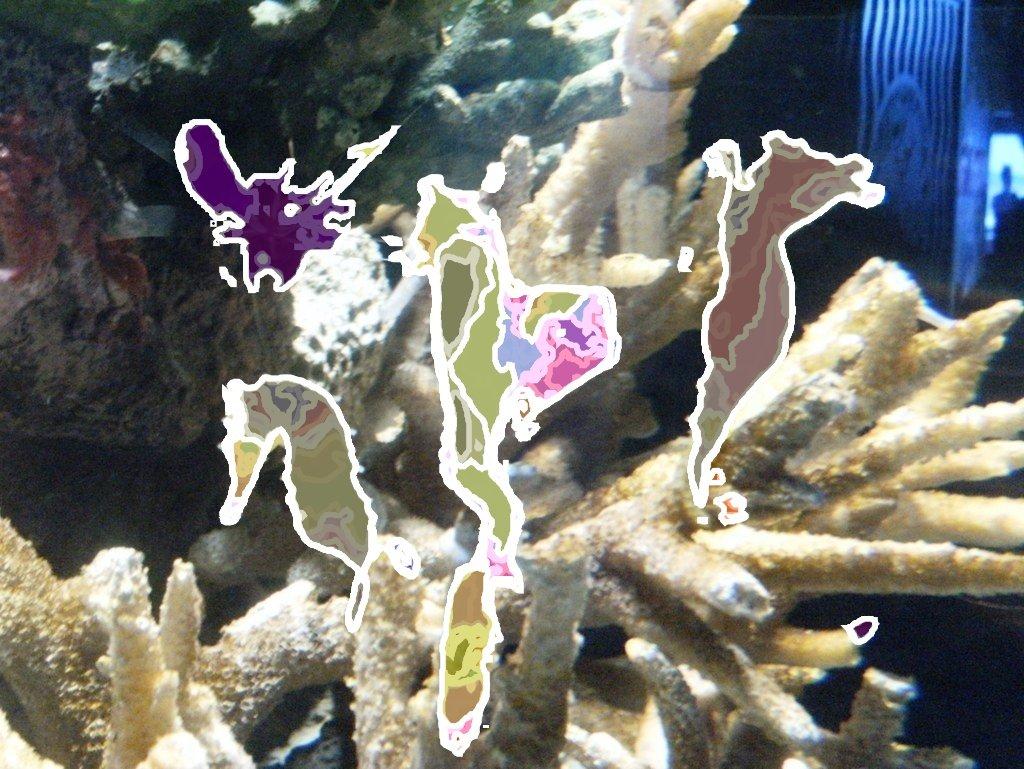} &
        \includegraphics[height=2cm]{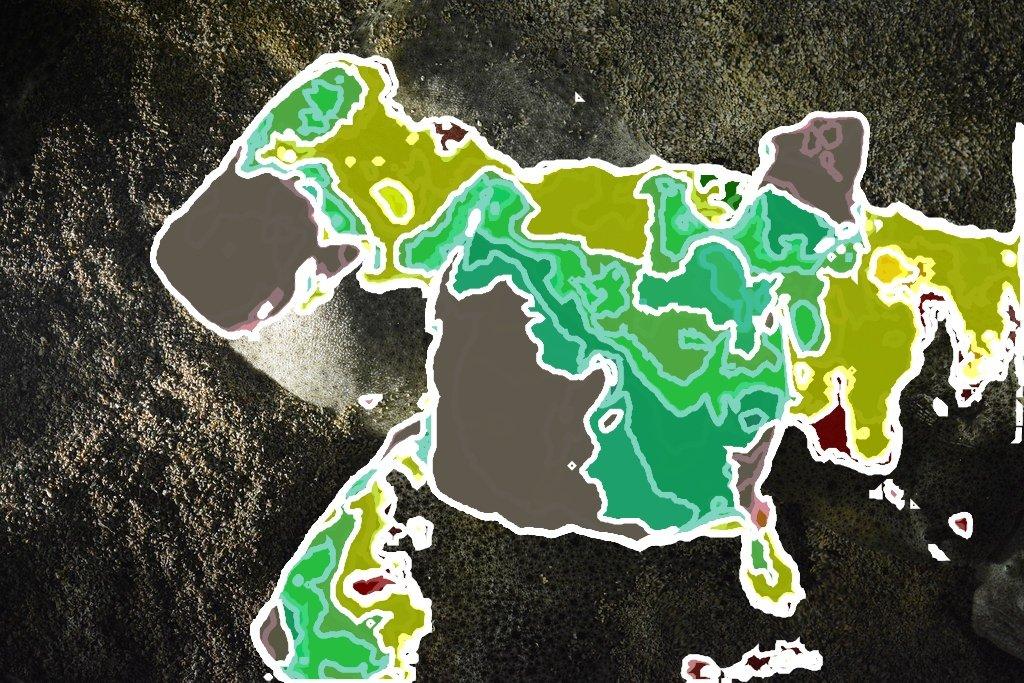} &
        \includegraphics[height=2cm]{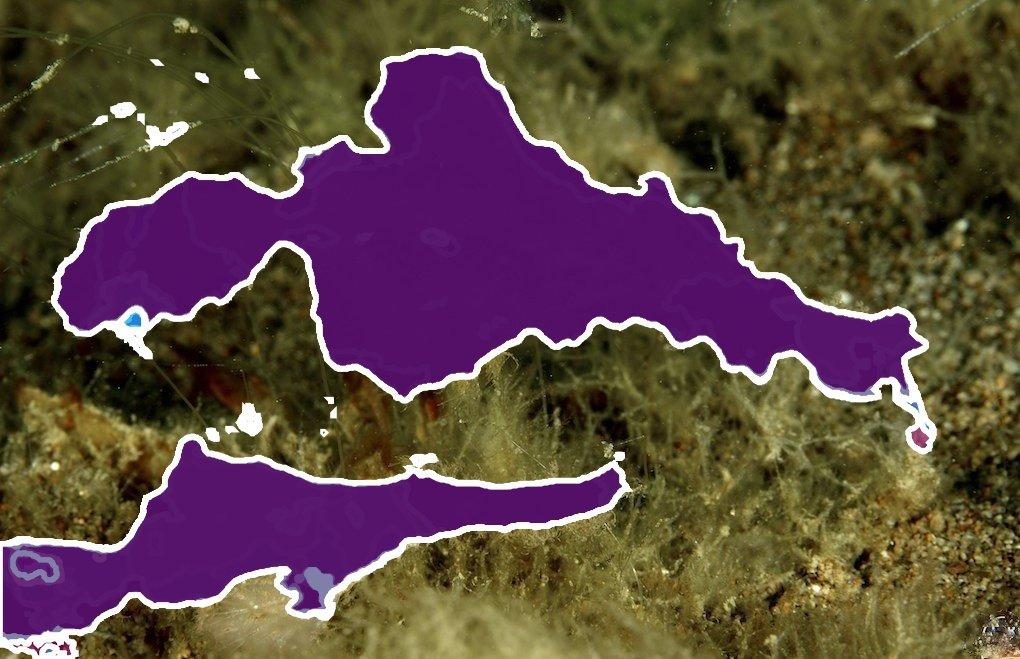} &
        \includegraphics[height=2cm]{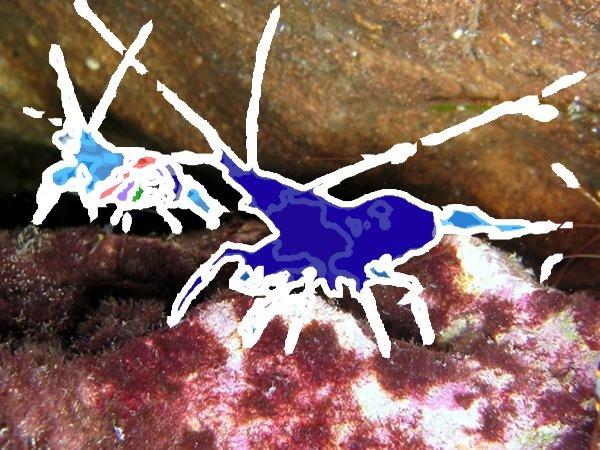} &
        \includegraphics[height=2cm]{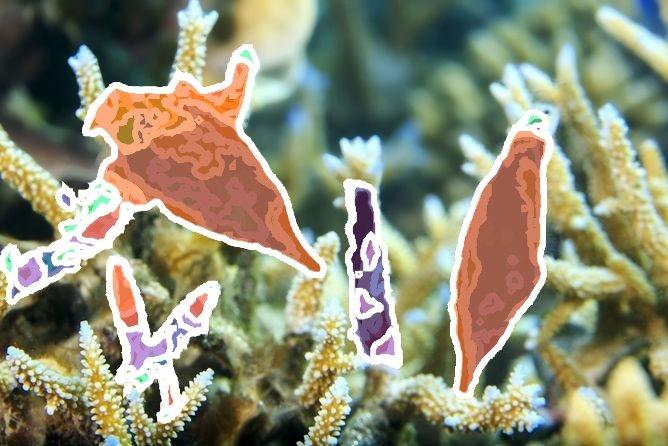} &
        \includegraphics[height=2cm]{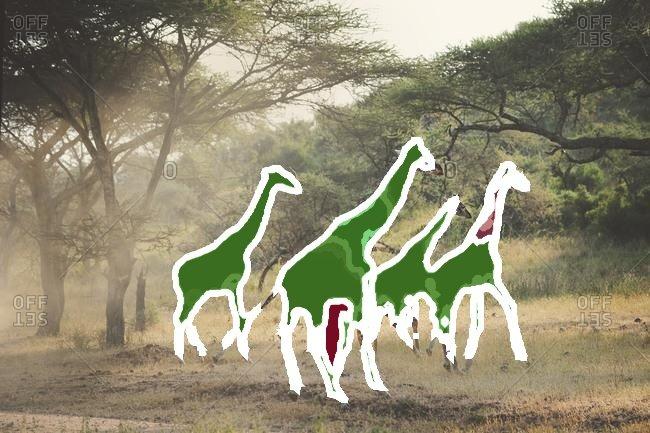} \\ 
        \raisebox{0.2cm}{\makebox[0pt][c]{\rotatebox{90}{\small OSFormer}}} & 
        \includegraphics[height=2cm]{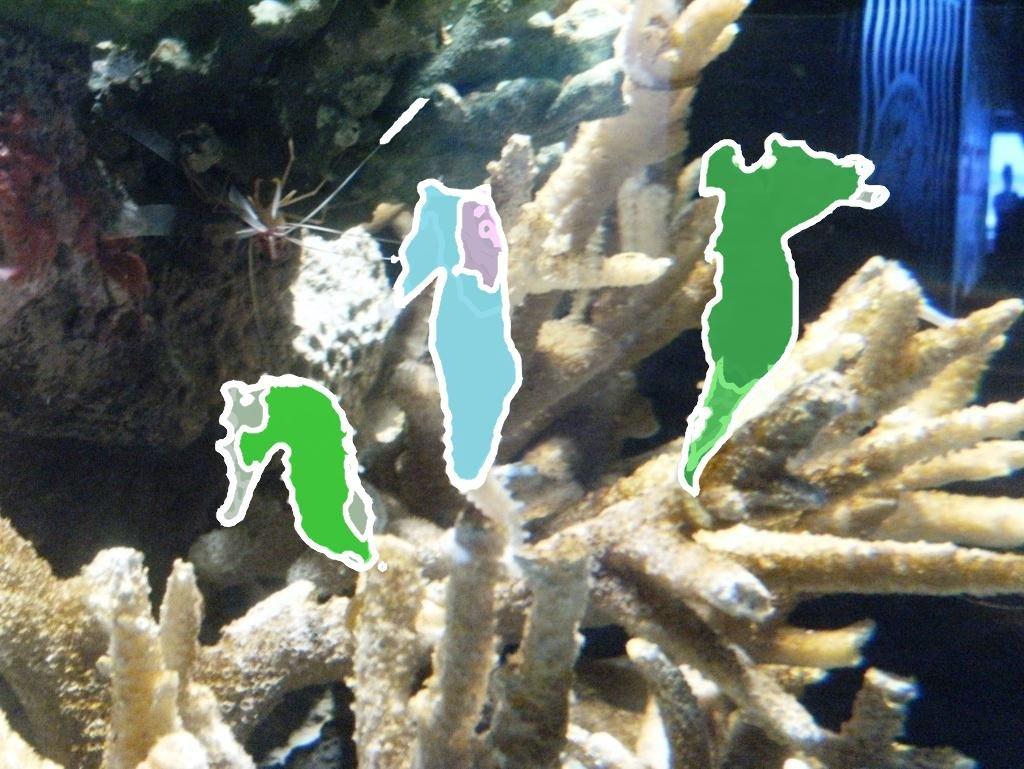} &
        \includegraphics[height=2cm]{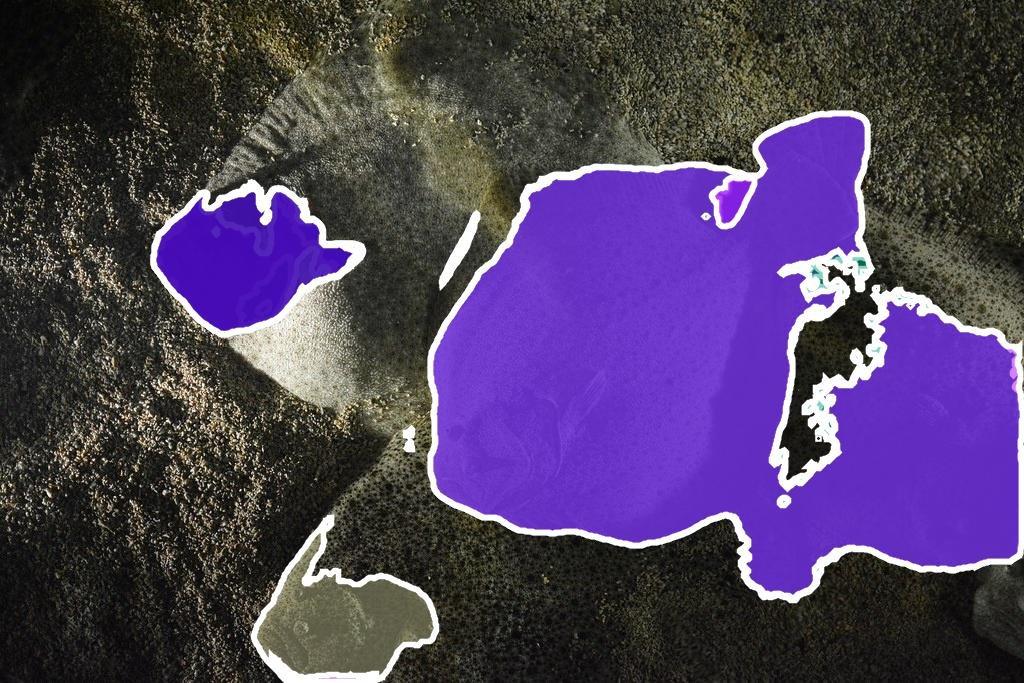} &
        \includegraphics[height=2cm]{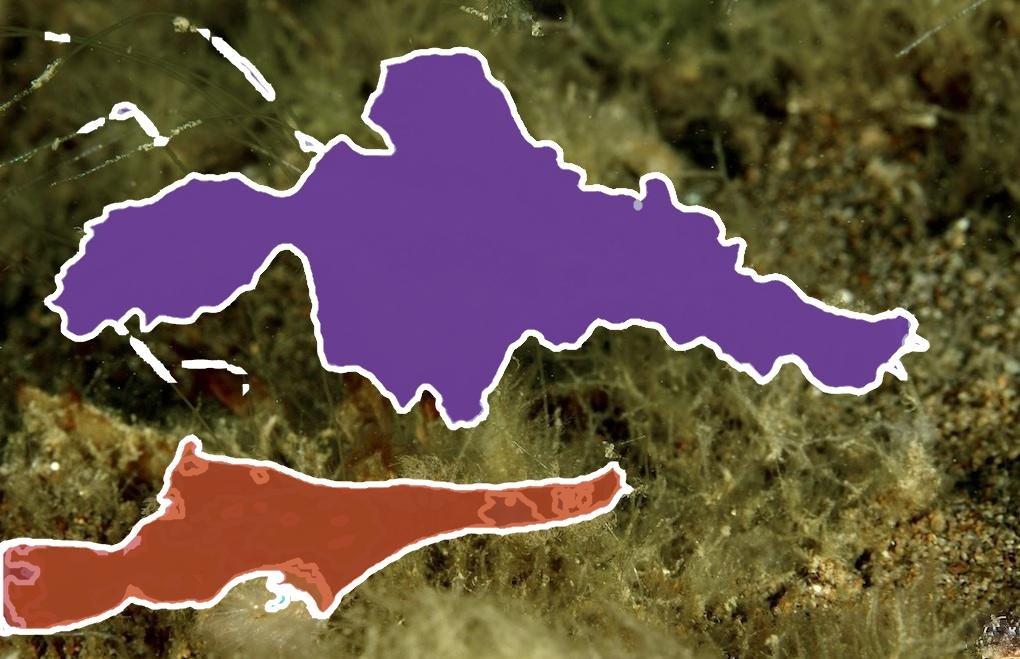} &
        \includegraphics[height=2cm]{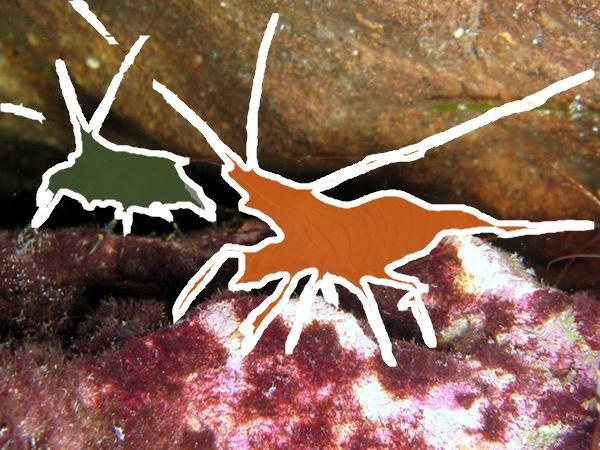} &
        \includegraphics[height=2cm]{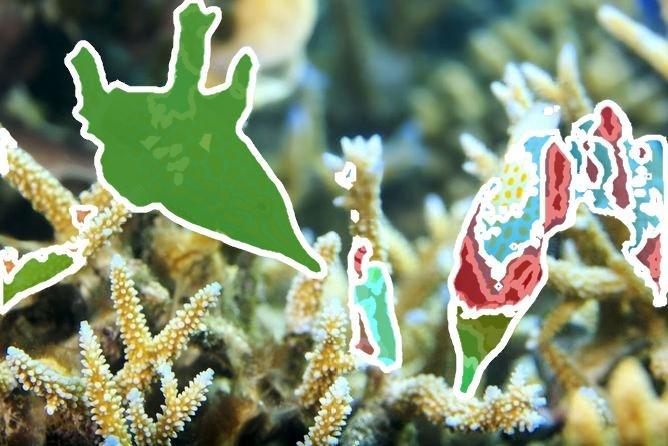} &
        \includegraphics[height=2cm]{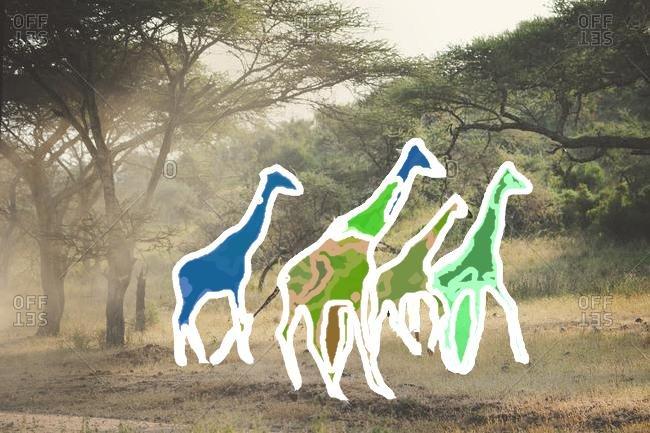} \\ 
        \raisebox{0.2cm}{\makebox[0pt][c]{\rotatebox{90}{\small SAM-Auto}}} & 
        \includegraphics[height=2cm]{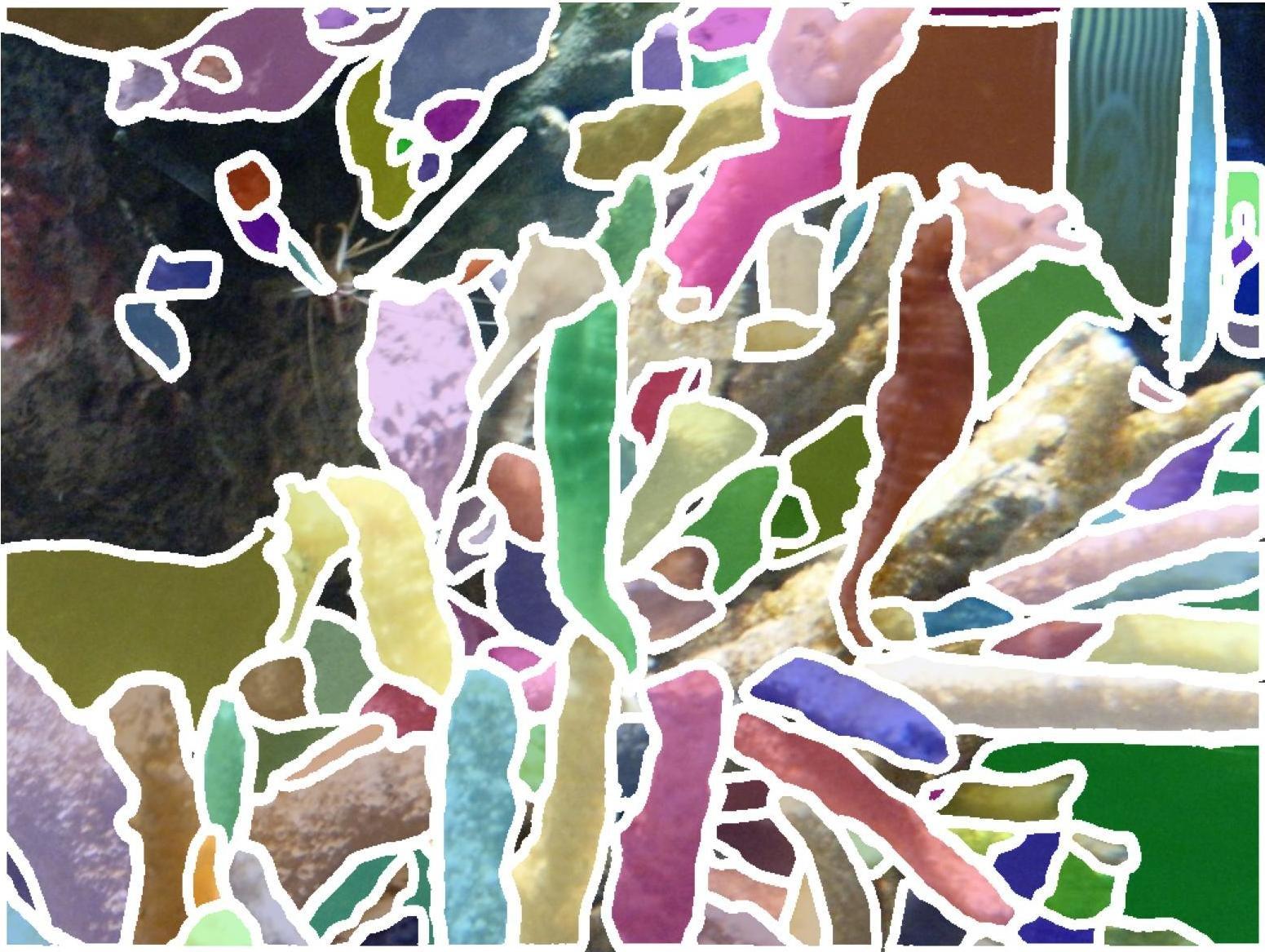} &
        \includegraphics[height=2cm]{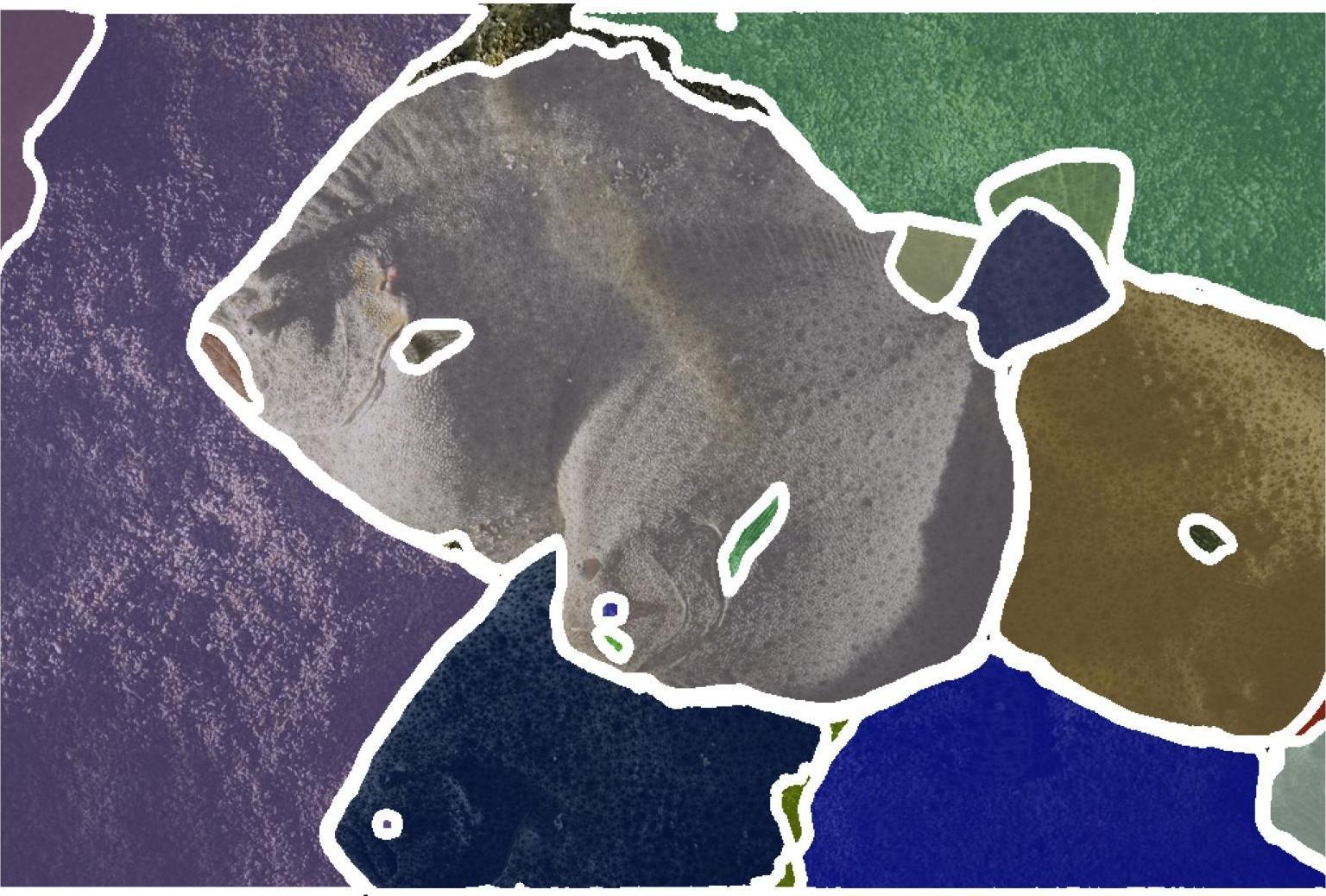} &
        \includegraphics[height=2cm]{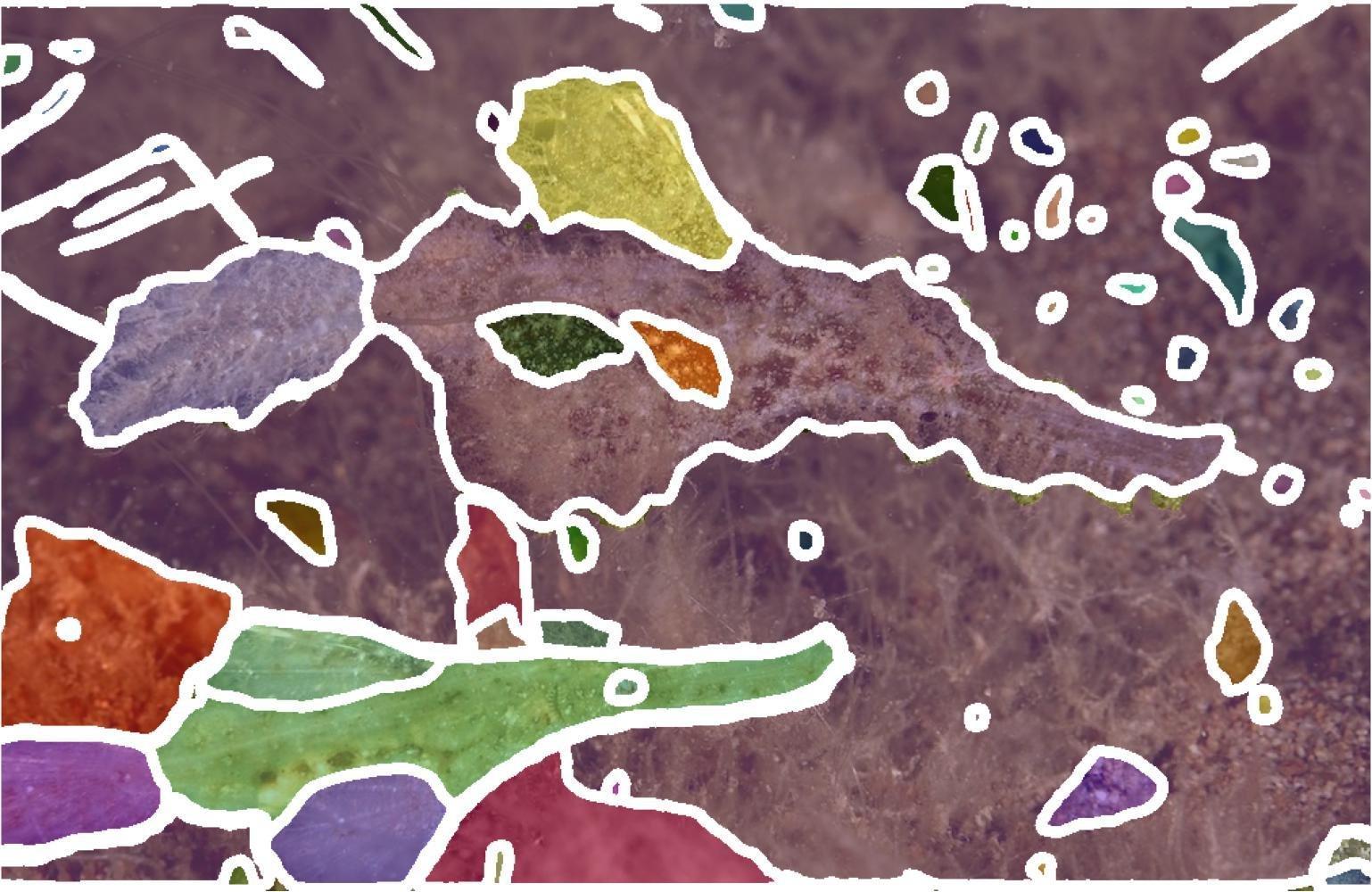} &
        \includegraphics[height=2cm]{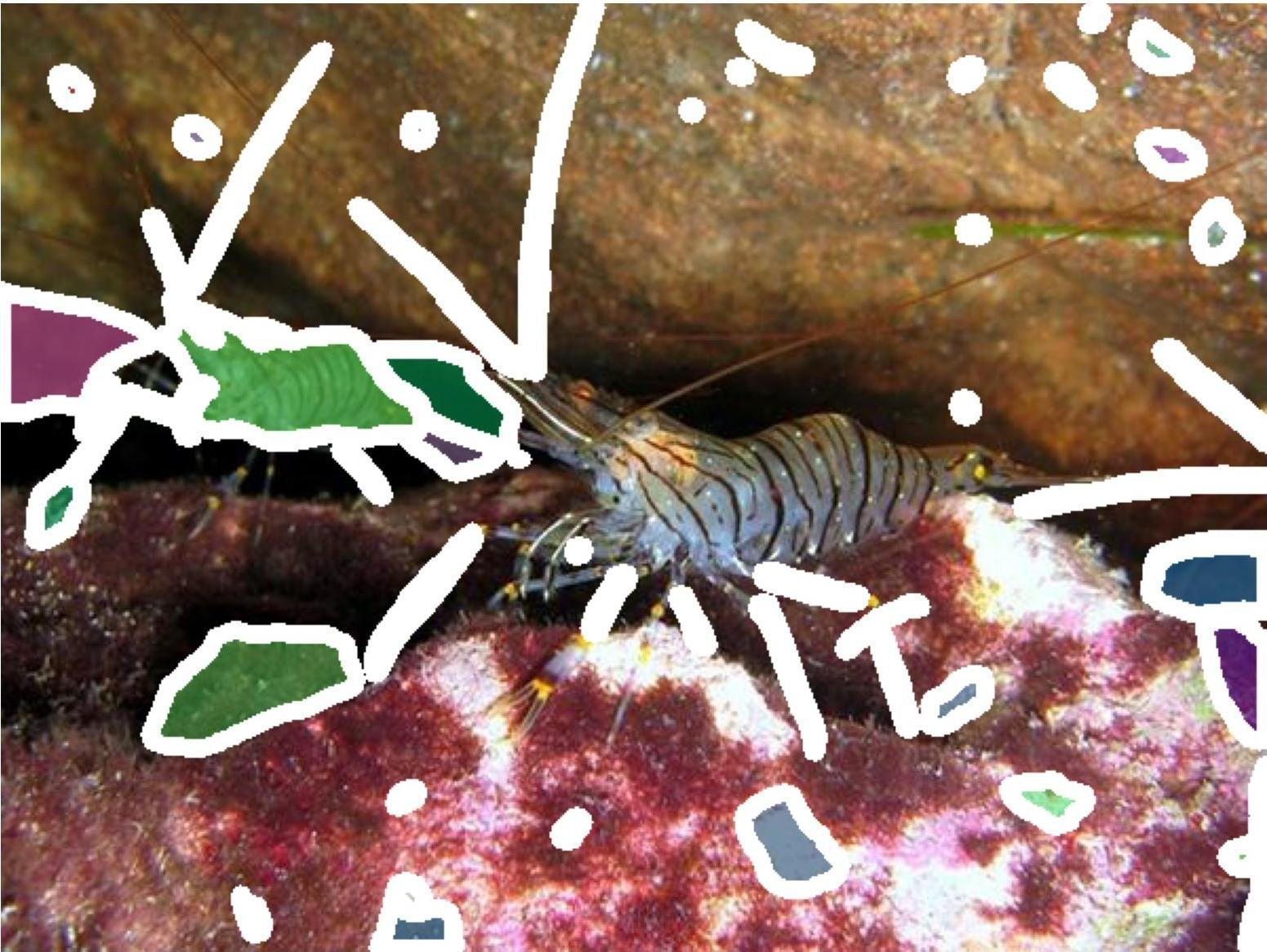} &
        \includegraphics[height=2cm]{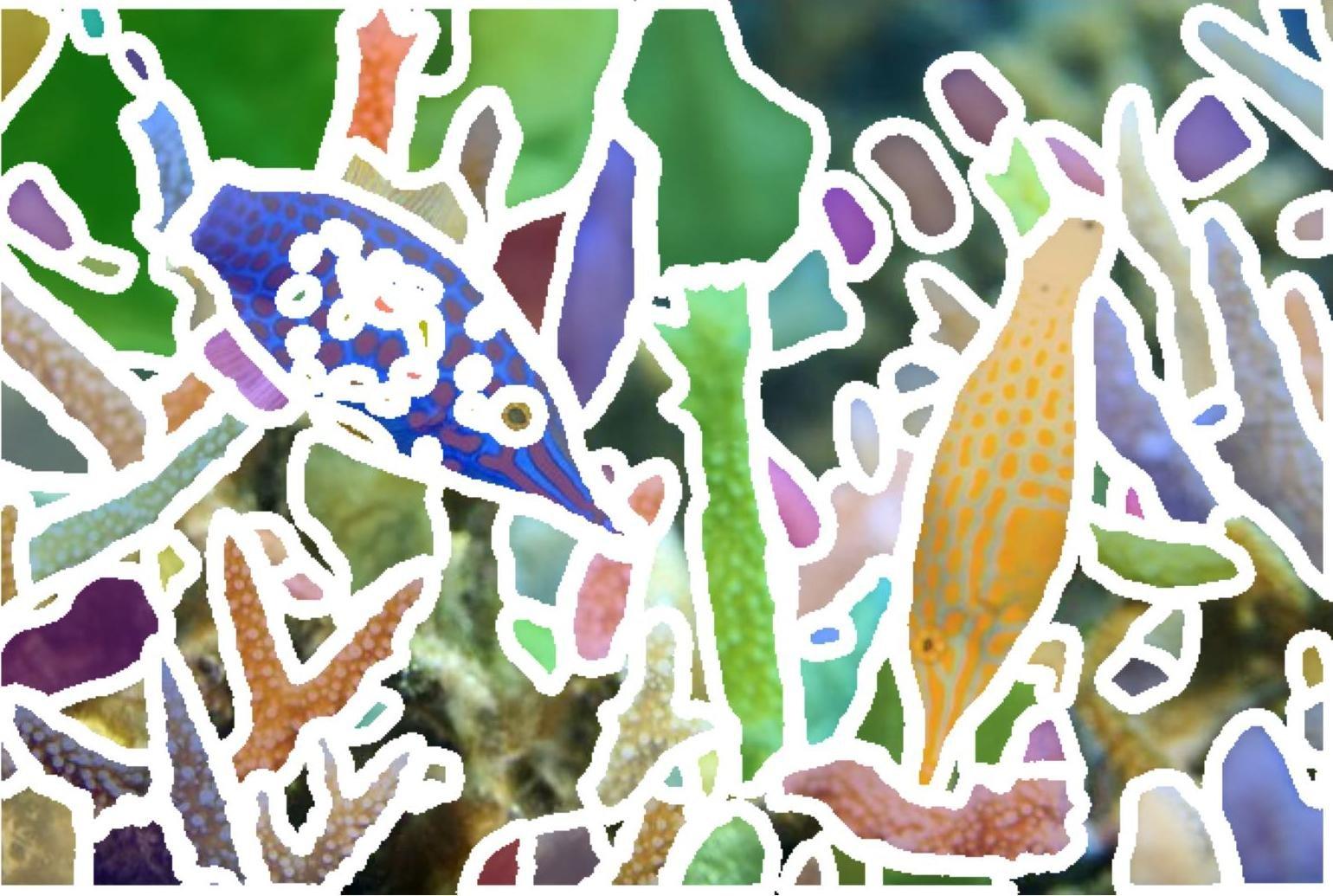} &
        \includegraphics[height=2cm]{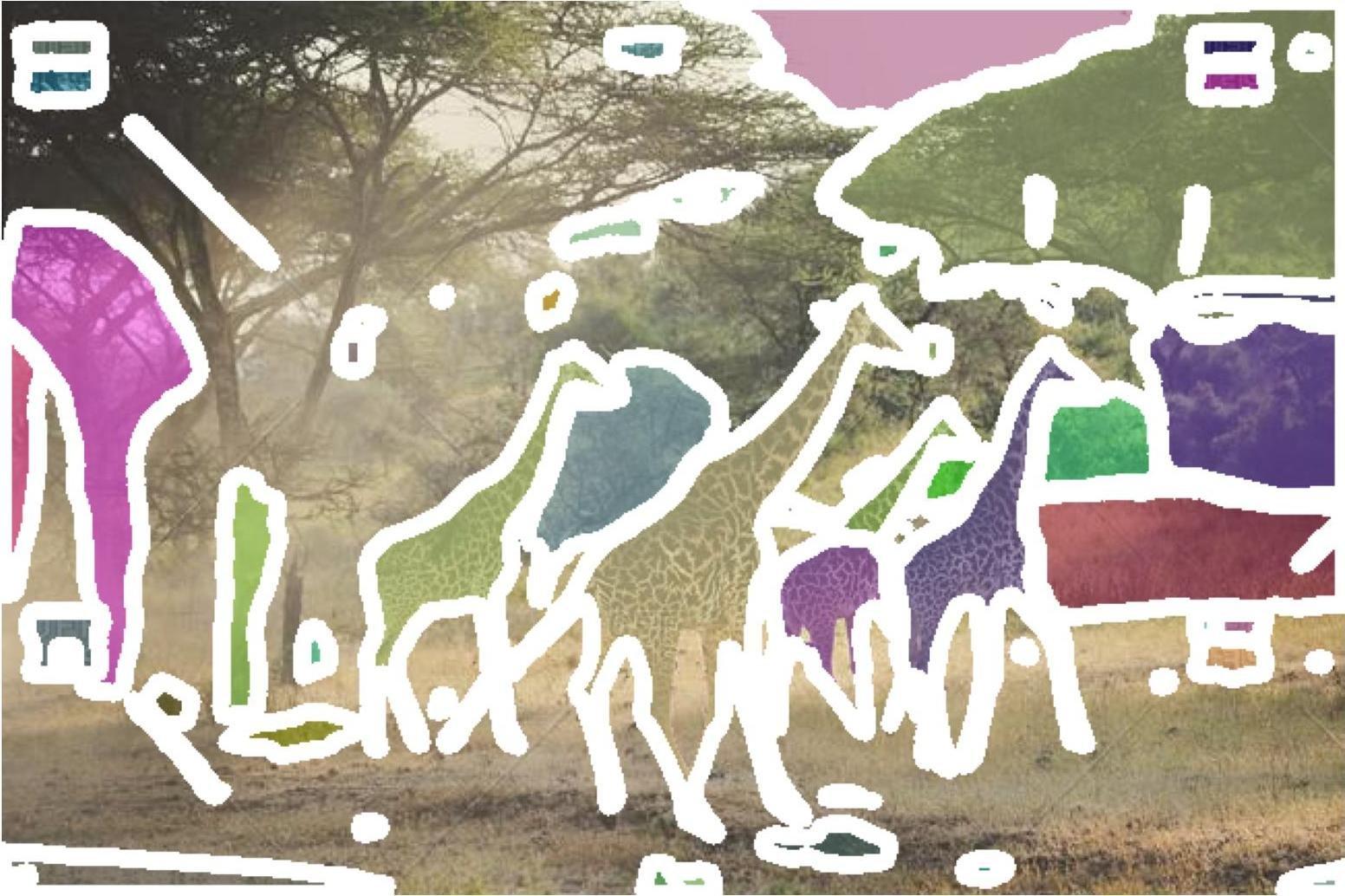} \\ 
        \raisebox{0.2cm}{\makebox[0pt][c]{\rotatebox{90}{\small SAM-Bbox}}} & 
        \includegraphics[height=2cm]{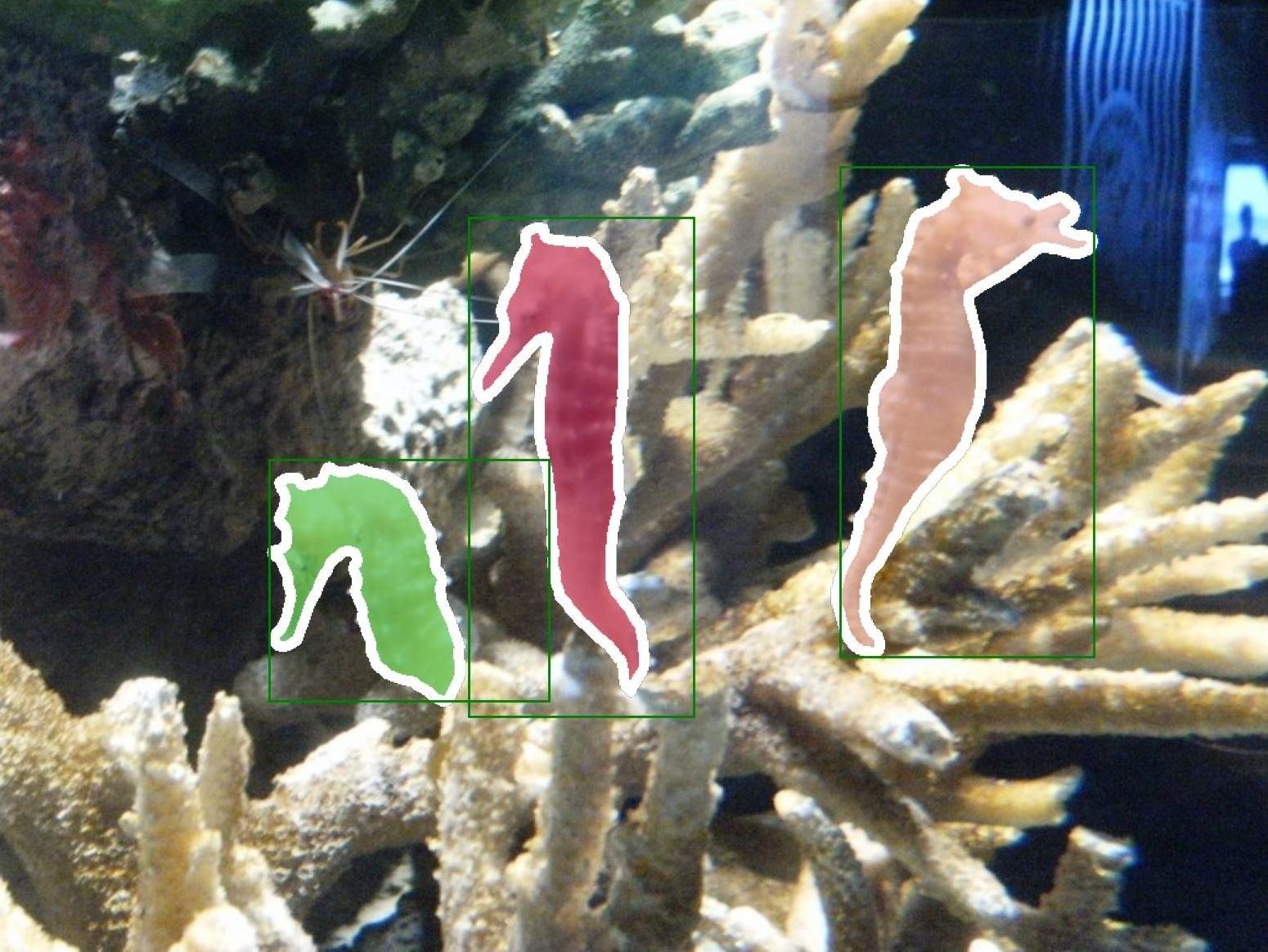} &
        \includegraphics[height=2cm]{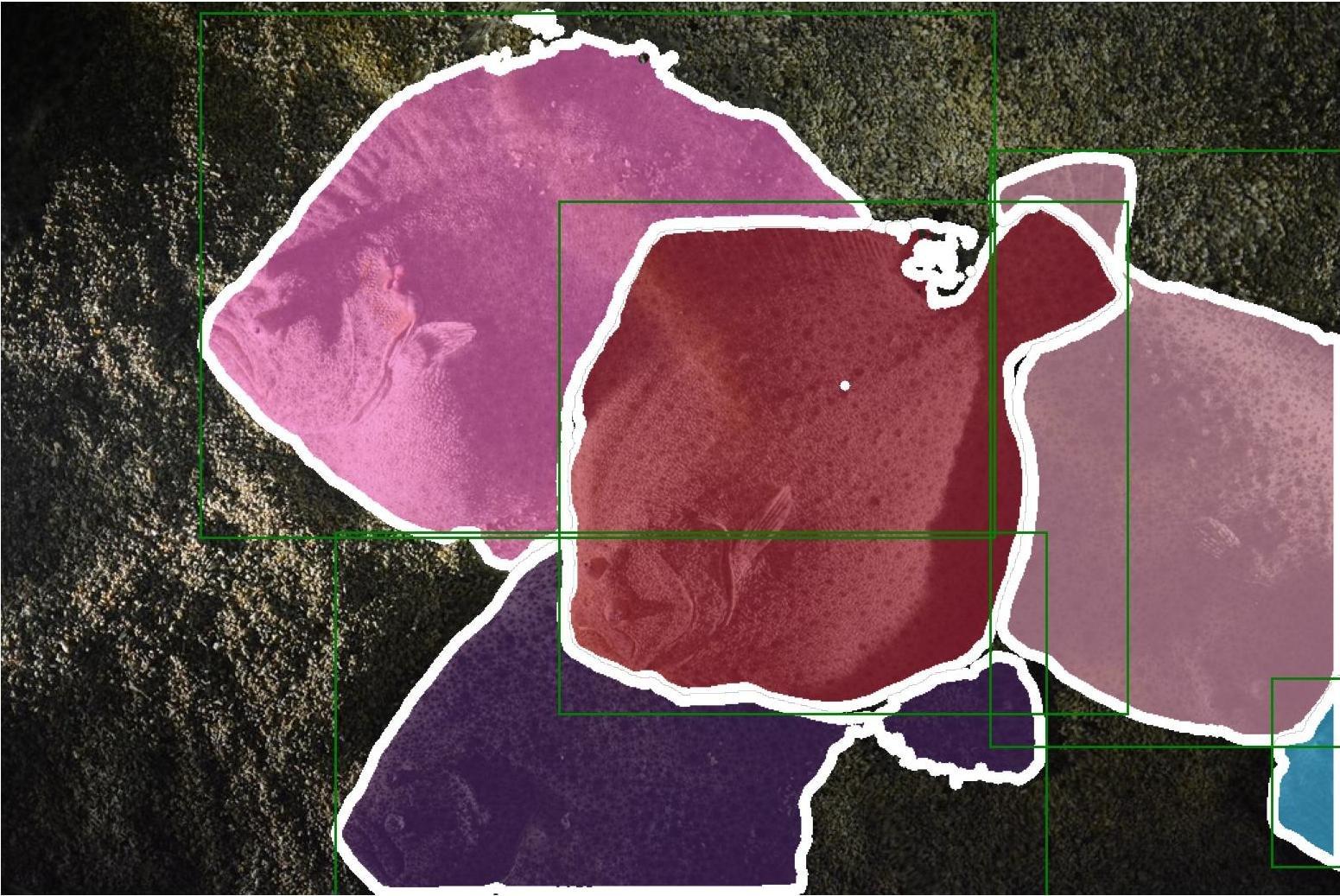} &
        \includegraphics[height=2cm]{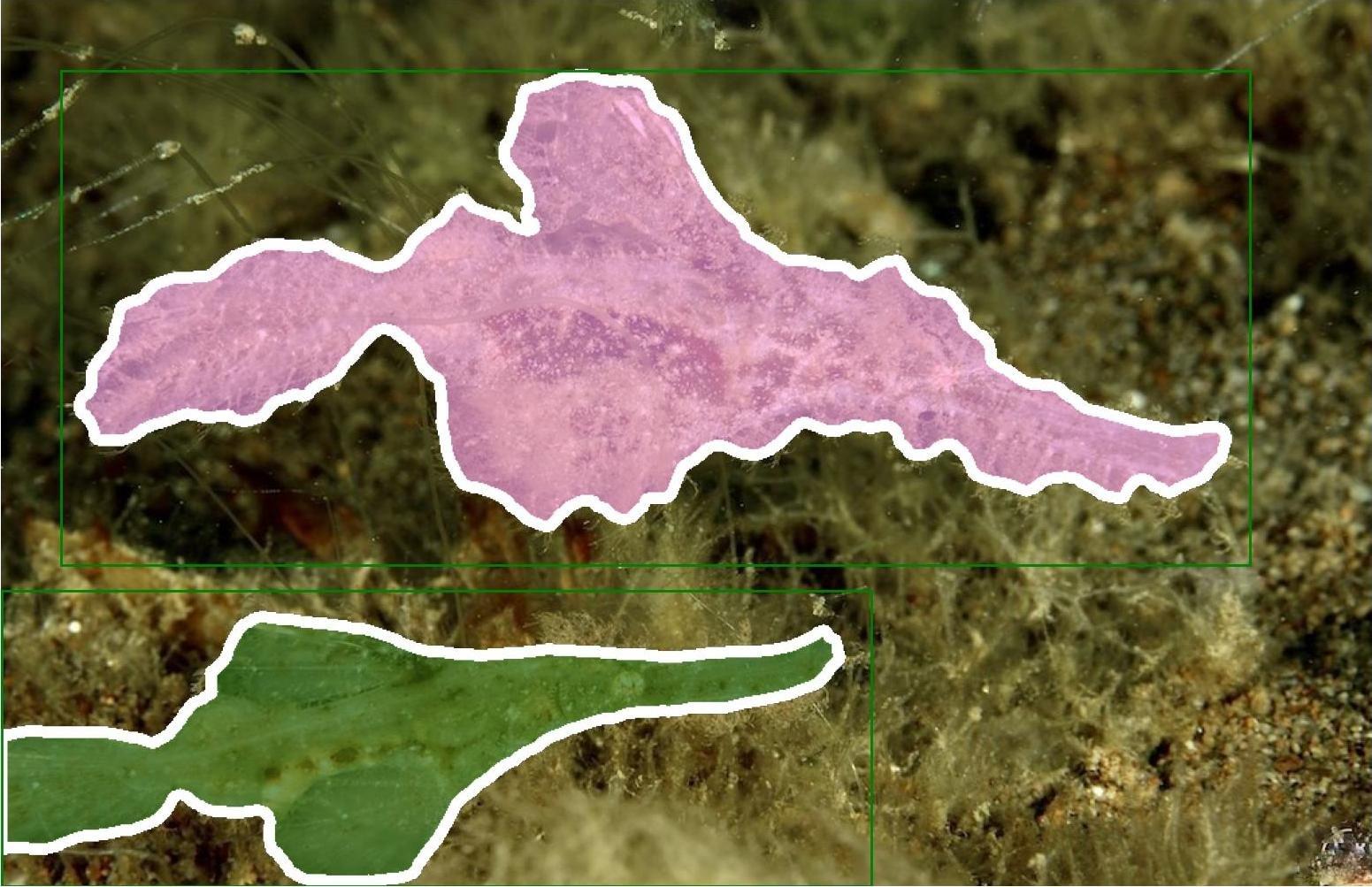} &
        \includegraphics[height=2cm]{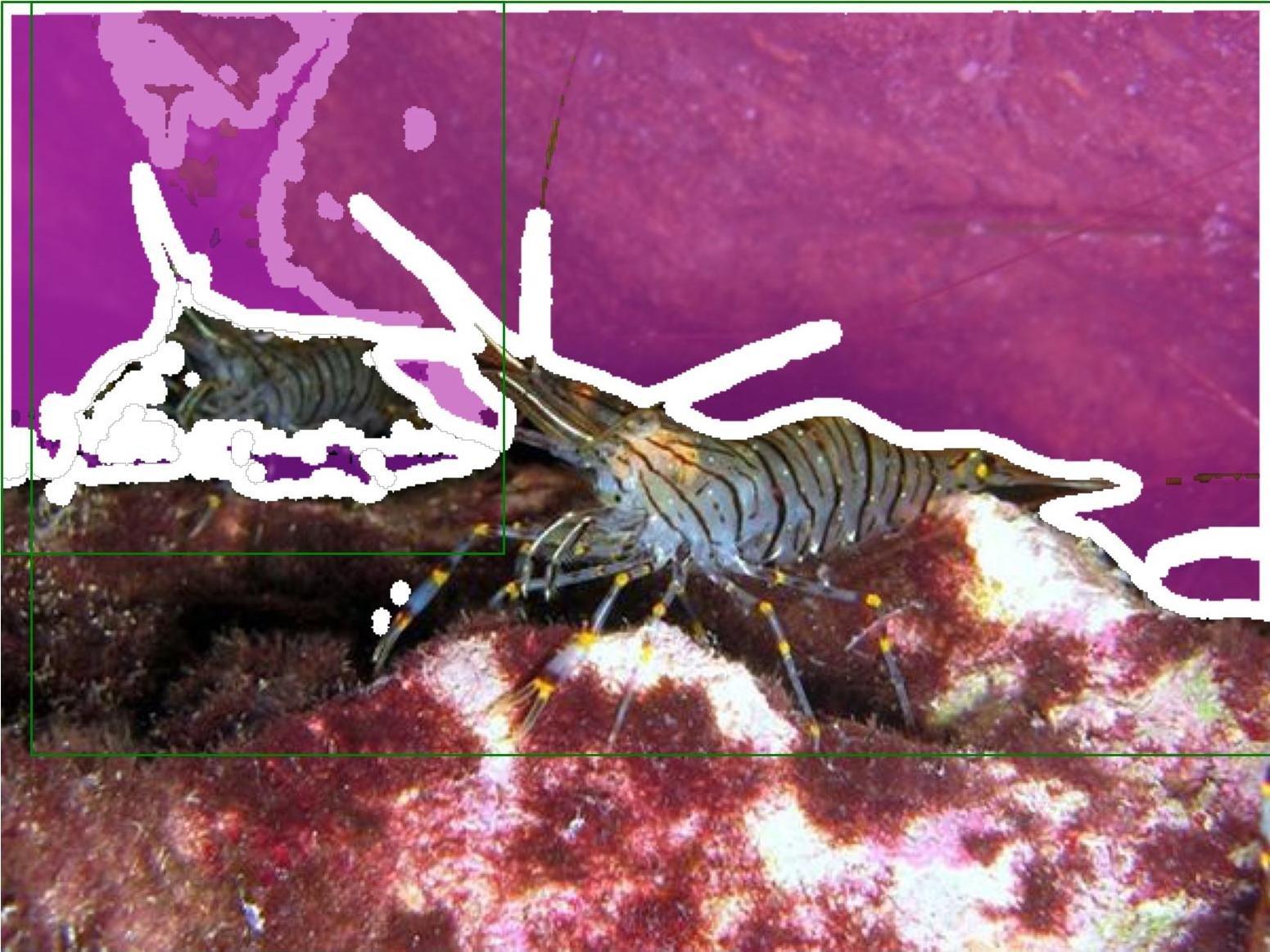} &
        \includegraphics[height=2cm]{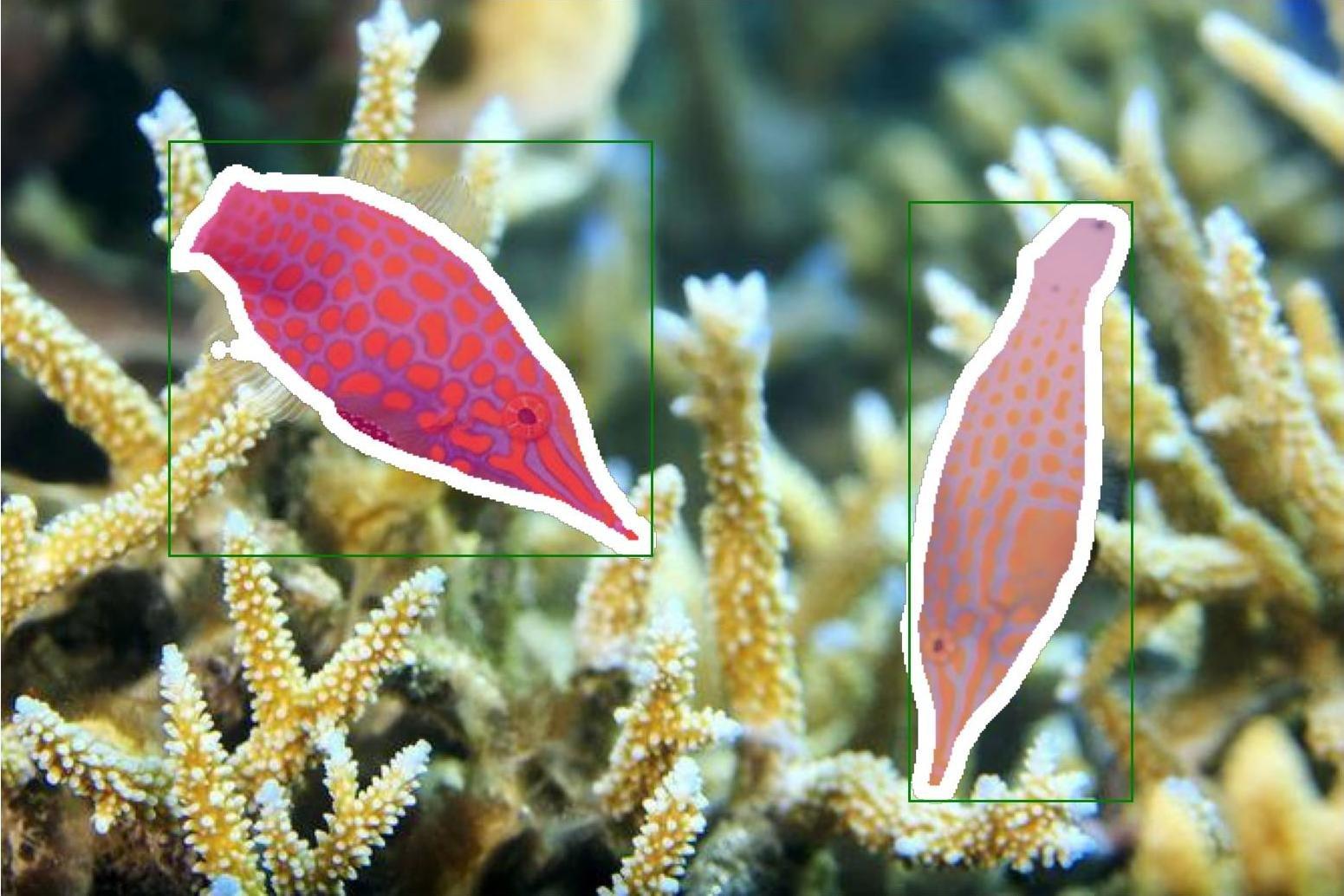} &
        \includegraphics[height=2cm]{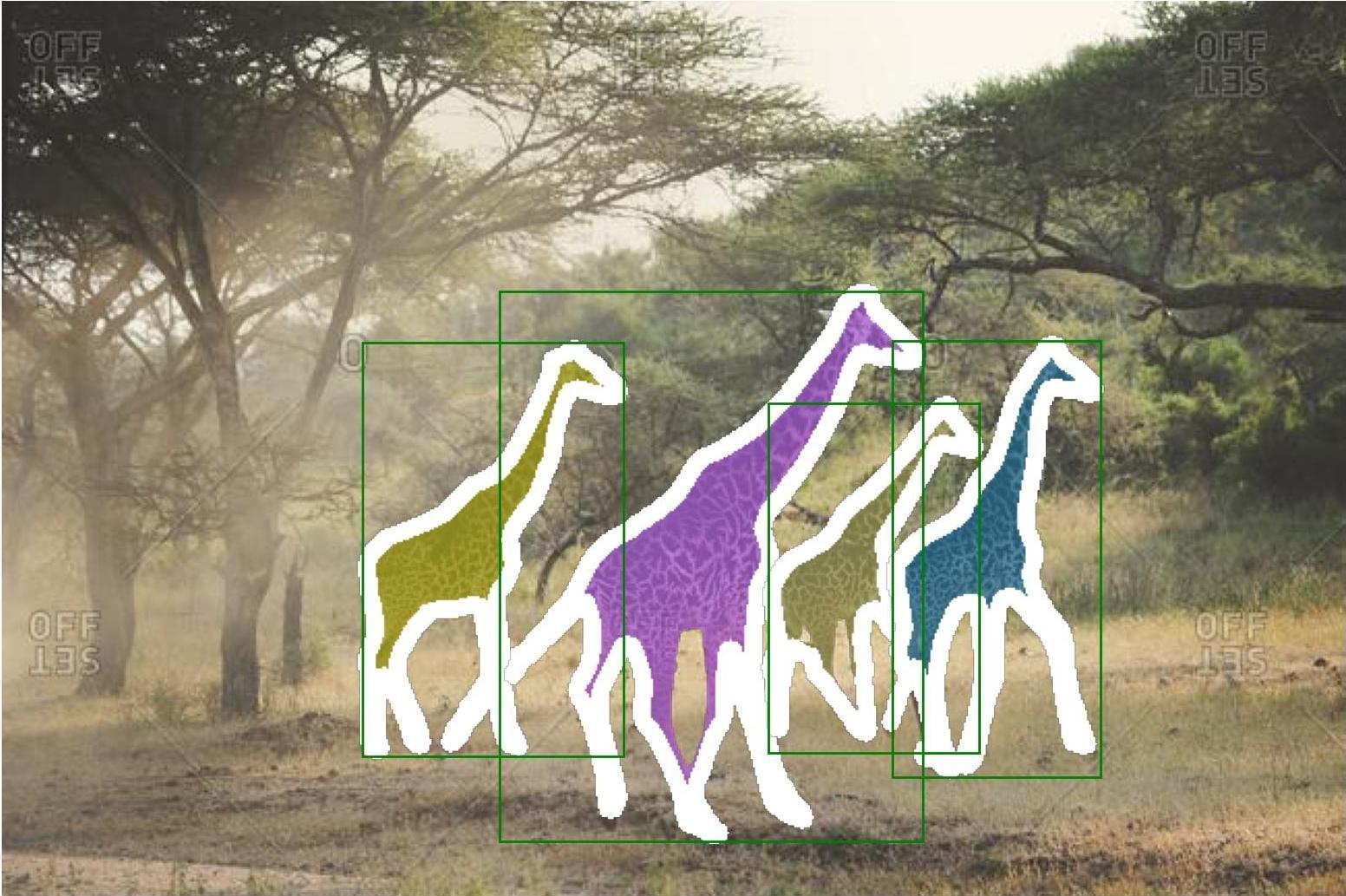} \\ 
        \raisebox{0.1cm}{\makebox[0pt][c]{\rotatebox{90}{\small \textbf{SAM2-Auto}}}} & 
        \includegraphics[height=2cm]{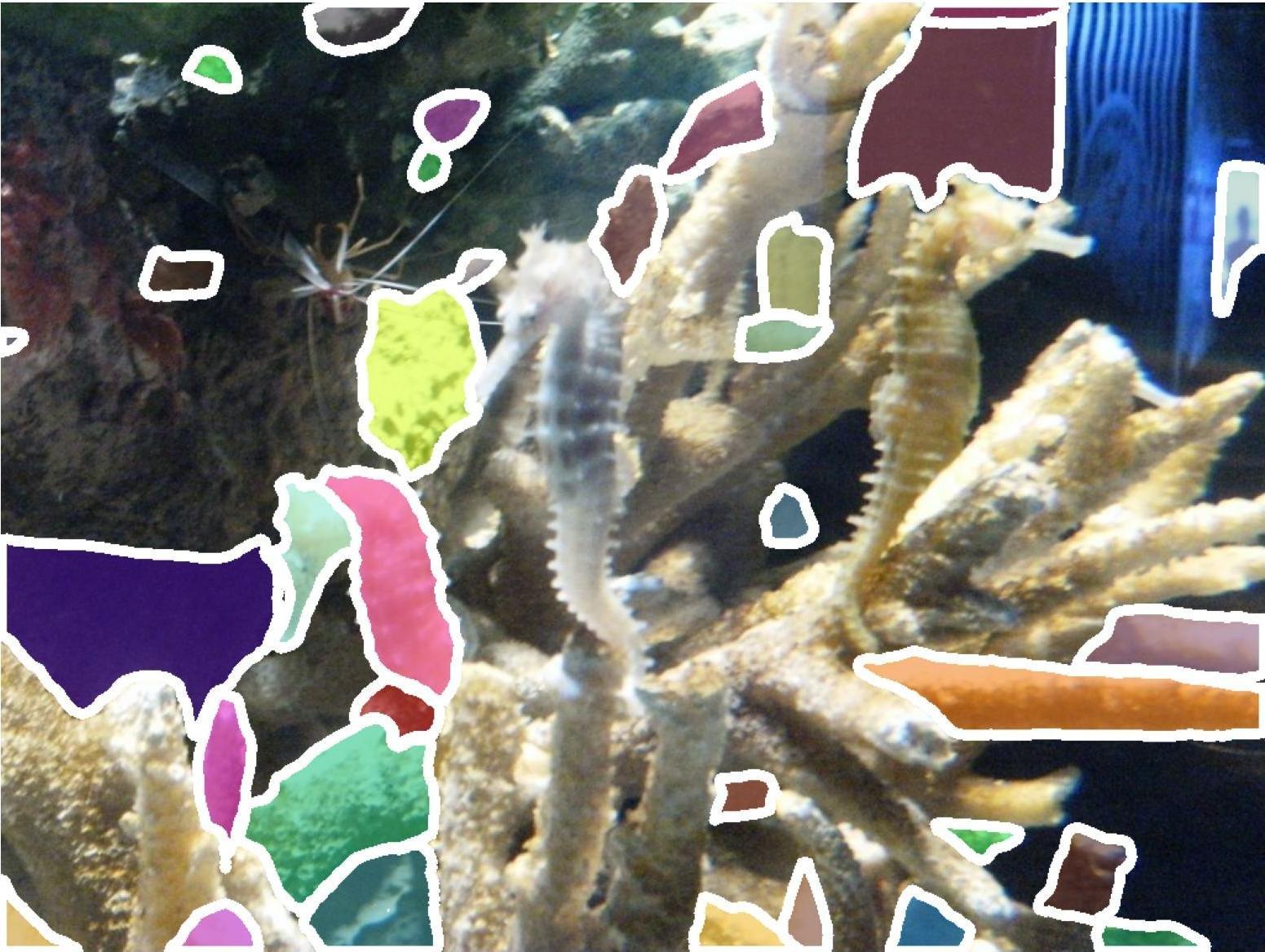} &
        \includegraphics[height=2cm]{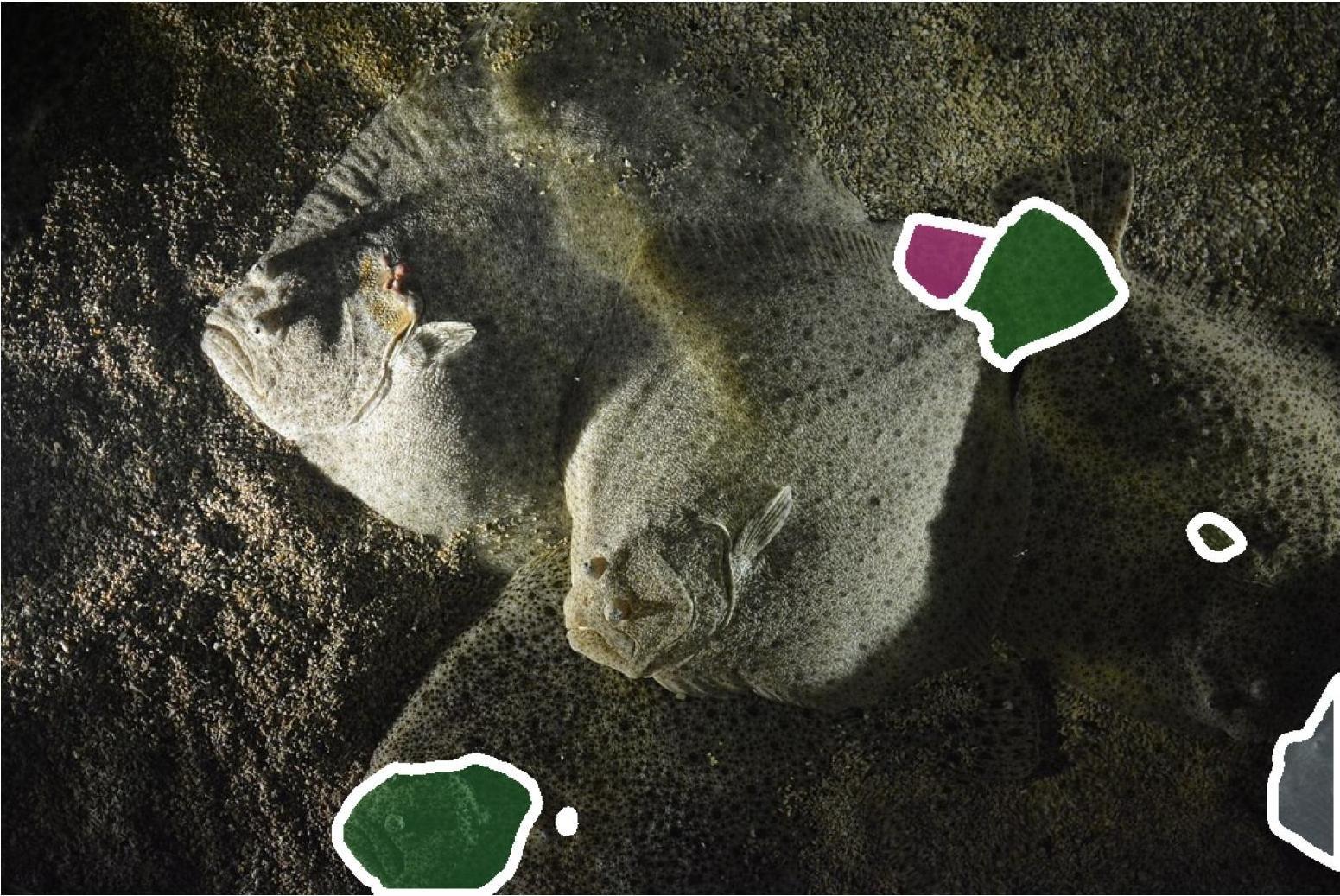} &
        \includegraphics[height=2cm]{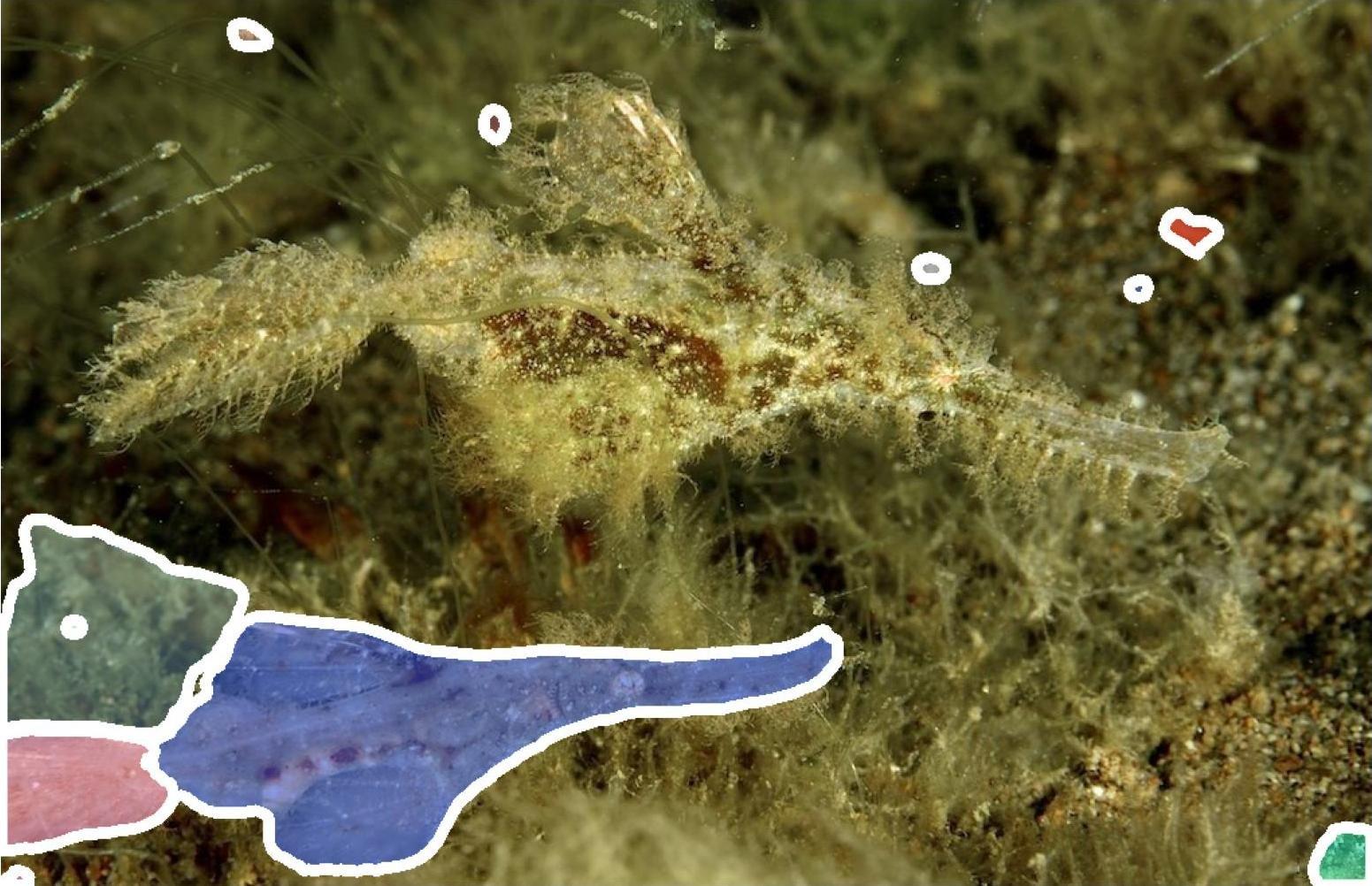} &
        \includegraphics[height=2cm]{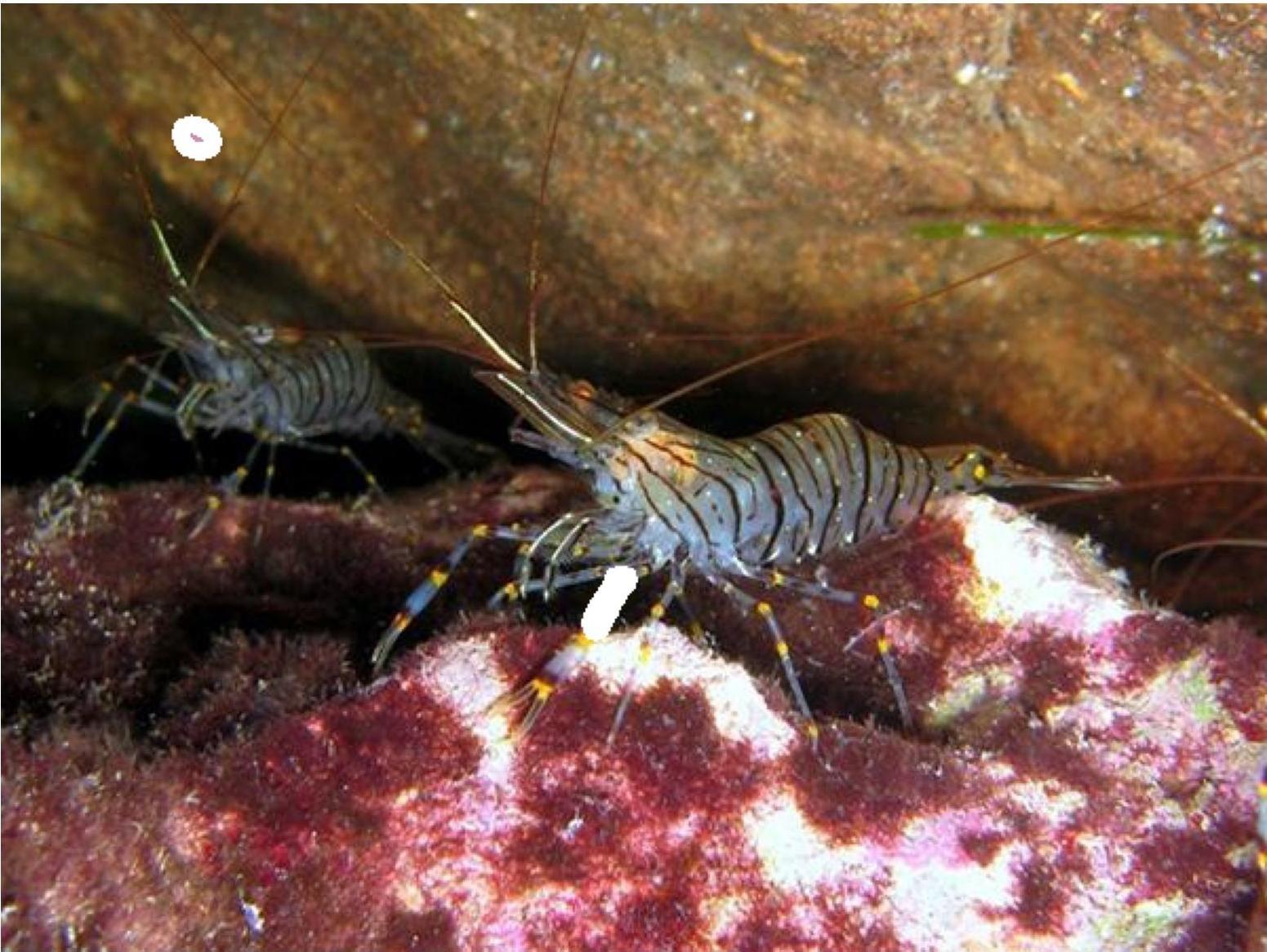} &
        \includegraphics[height=2cm]{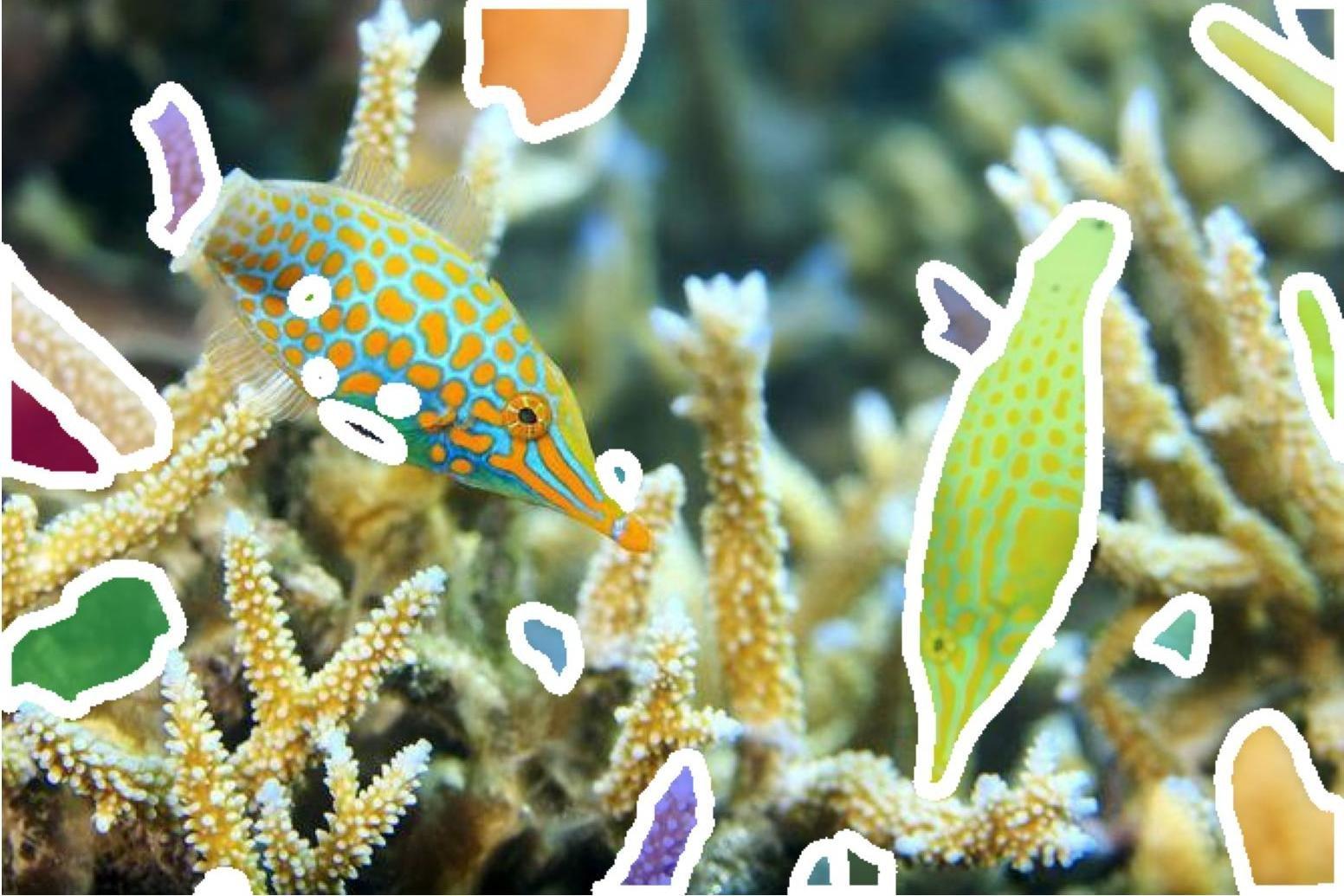} &
        \includegraphics[height=2cm]{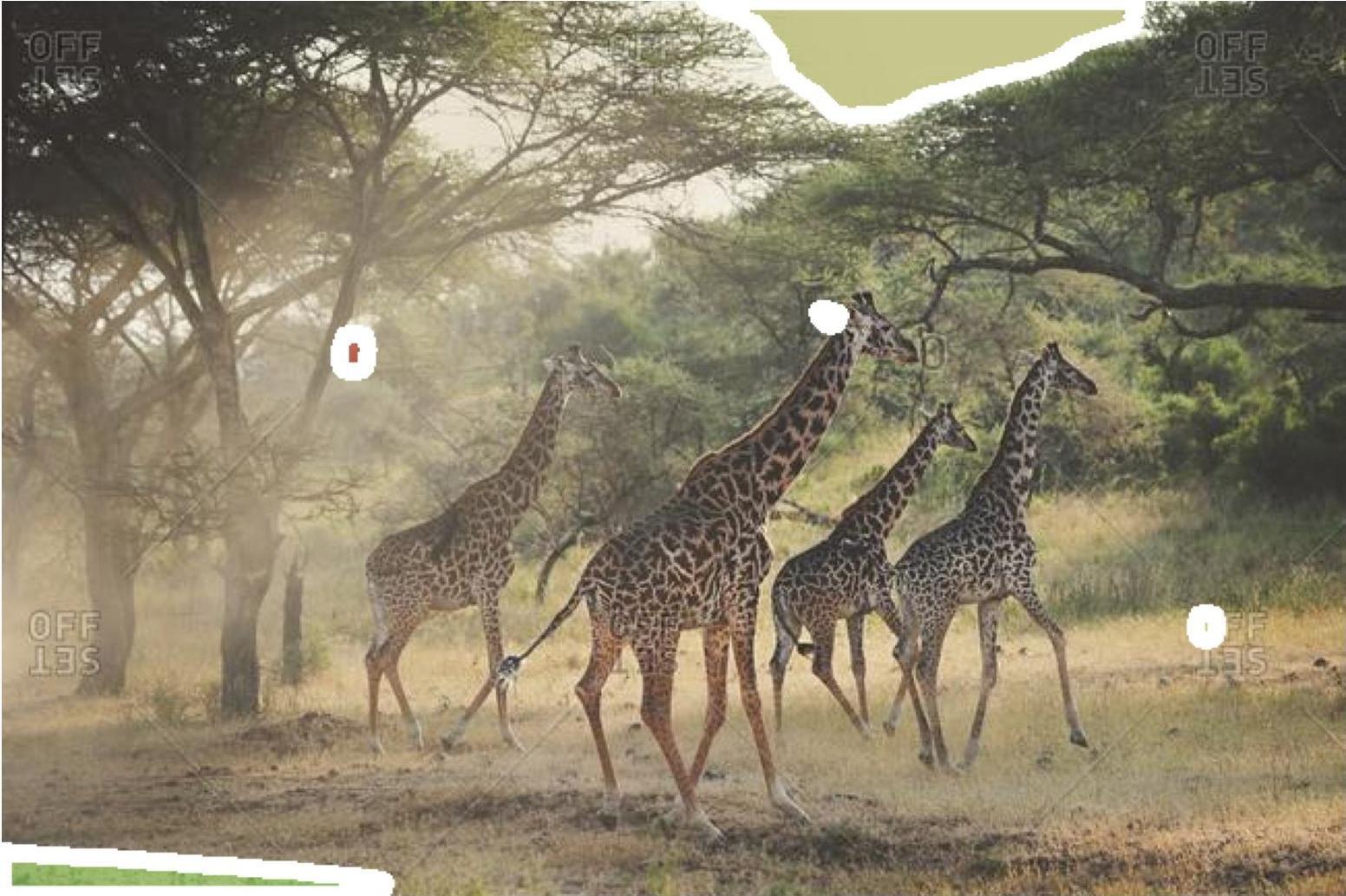} \\ 
        \raisebox{0.1cm}{\makebox[0pt][c]{\rotatebox{90}{\small \textbf{SAM2-Bbox}}}} & 
        \includegraphics[height=2cm]{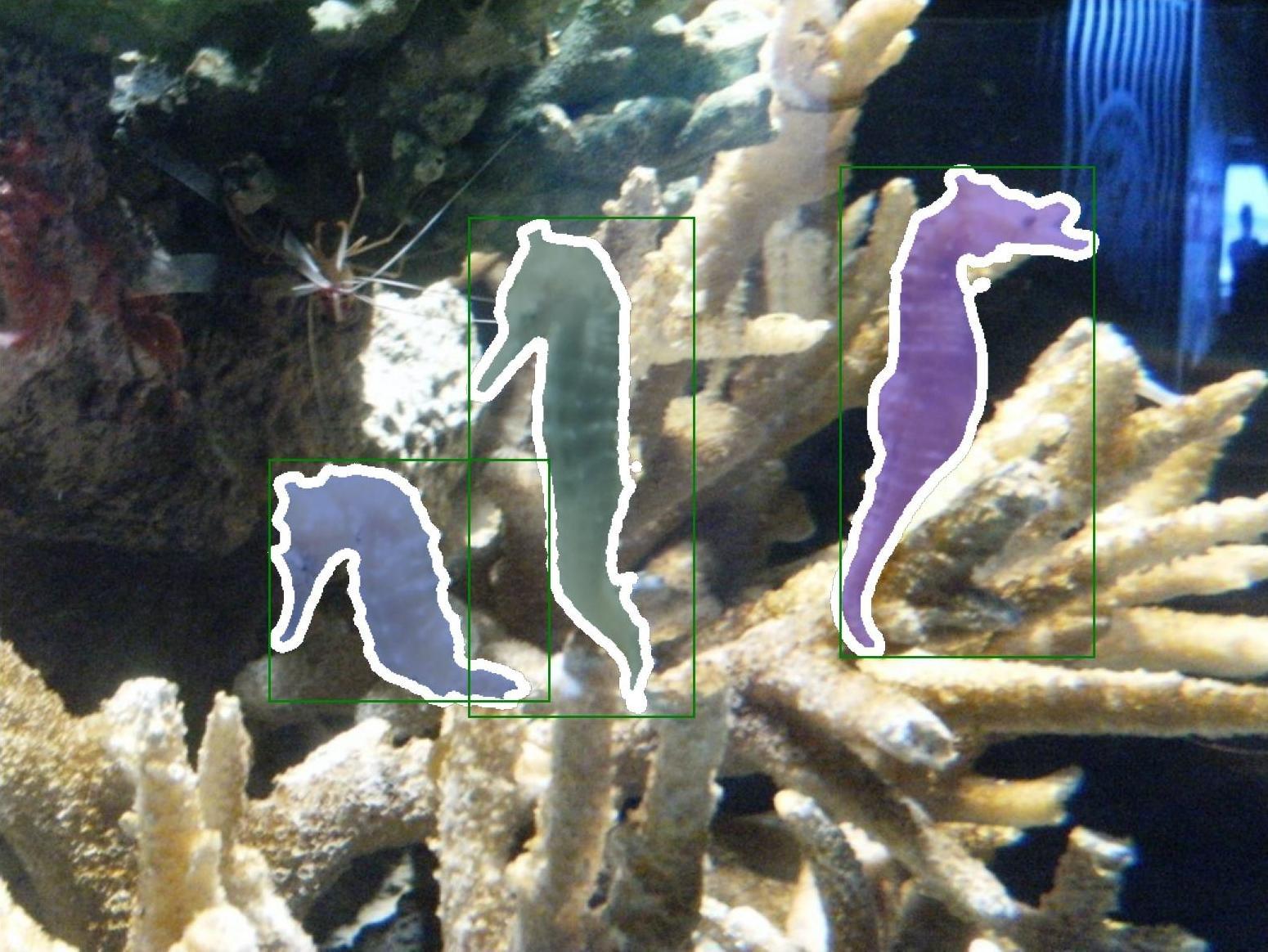} &
        \includegraphics[height=2cm]{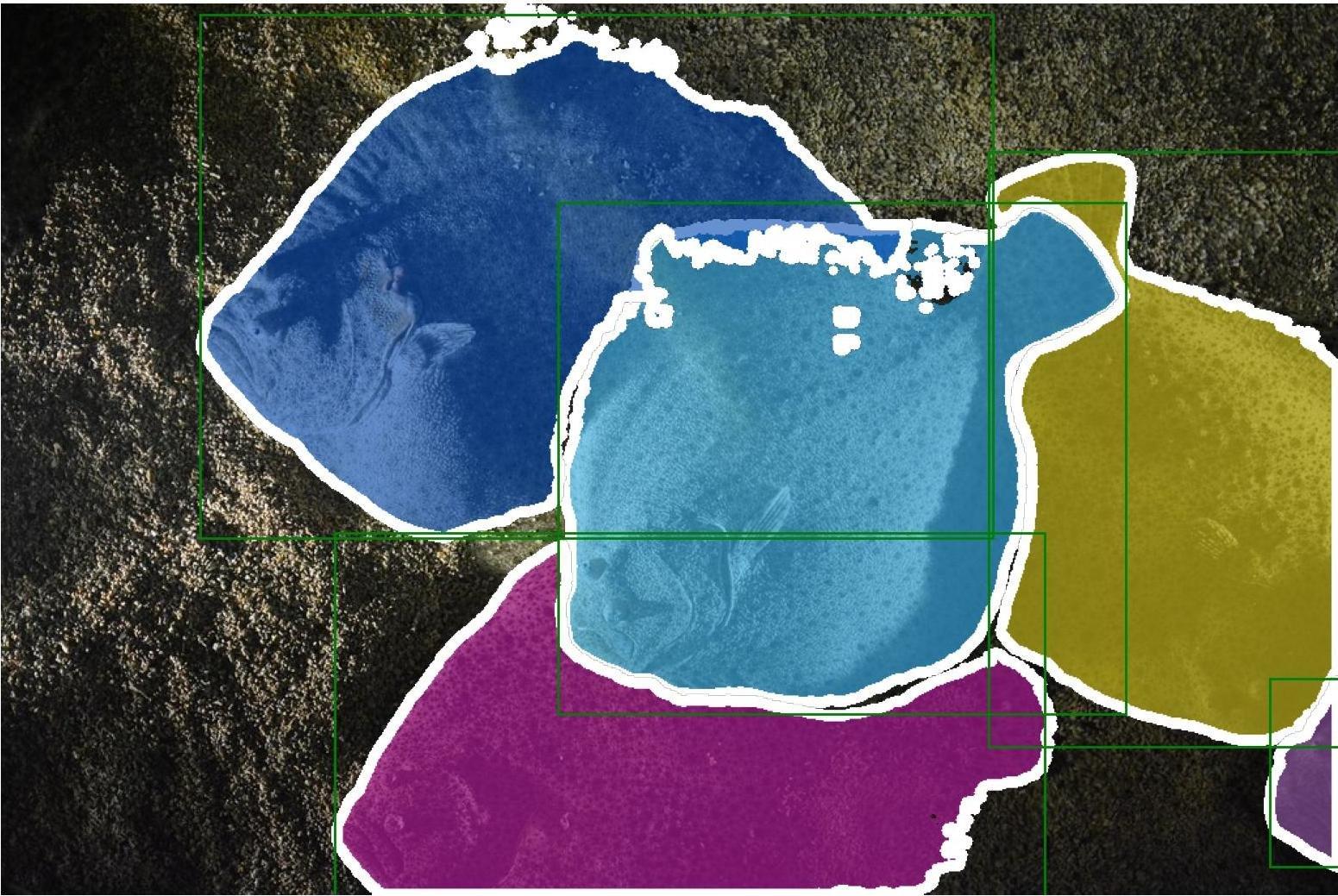} &
        \includegraphics[height=2cm]{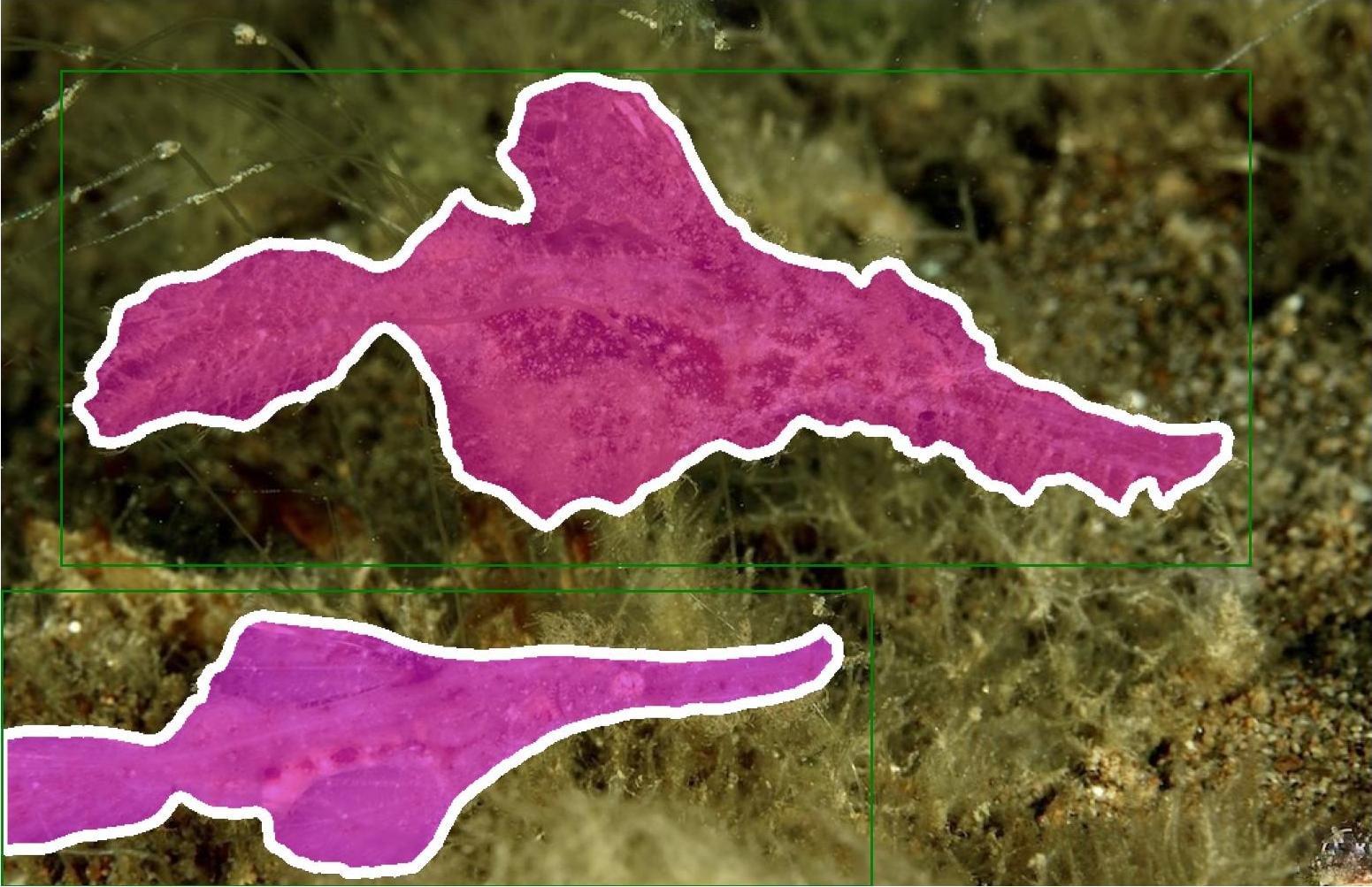} &
        \includegraphics[height=2cm]{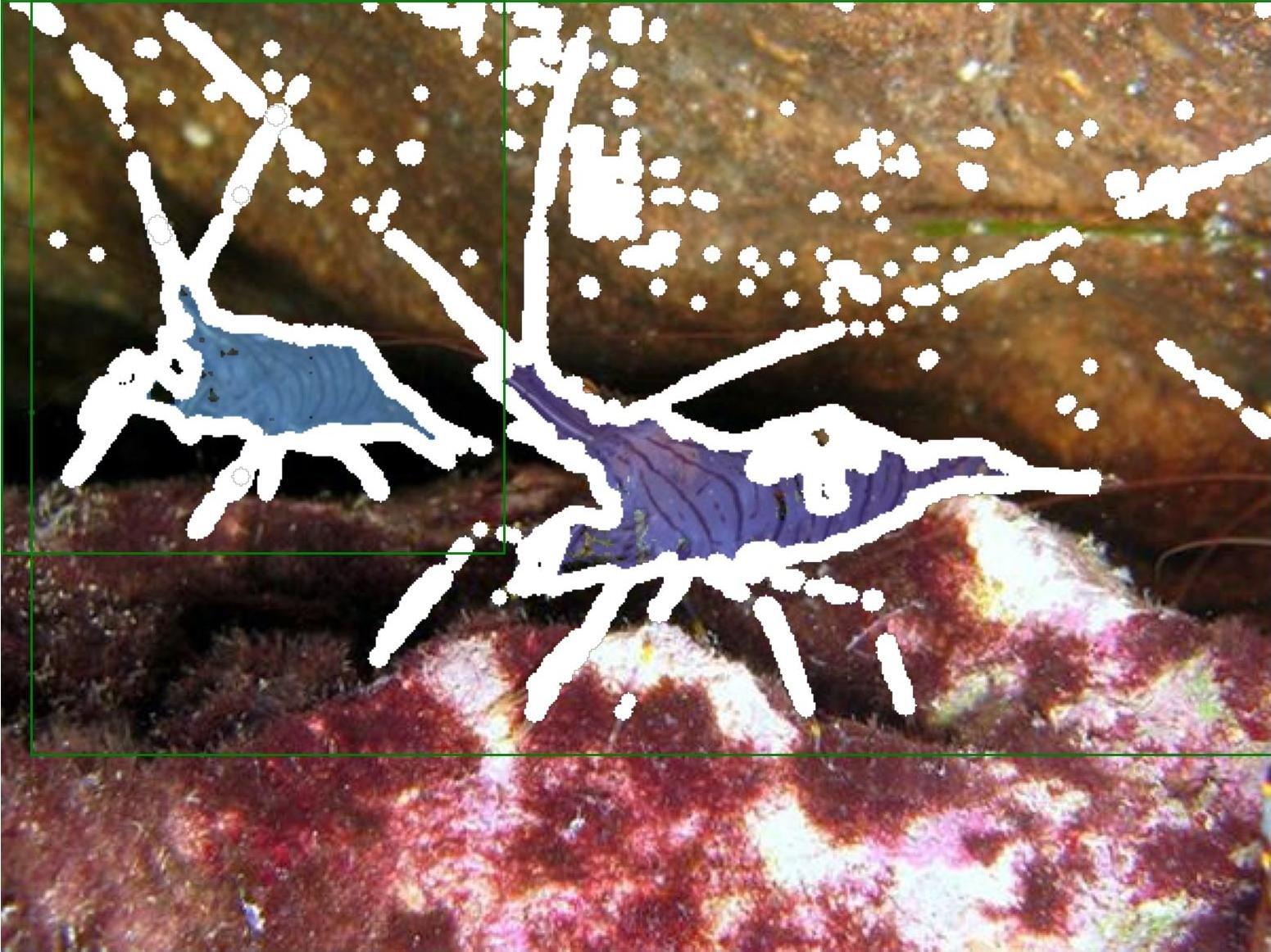} &
        \includegraphics[height=2cm]{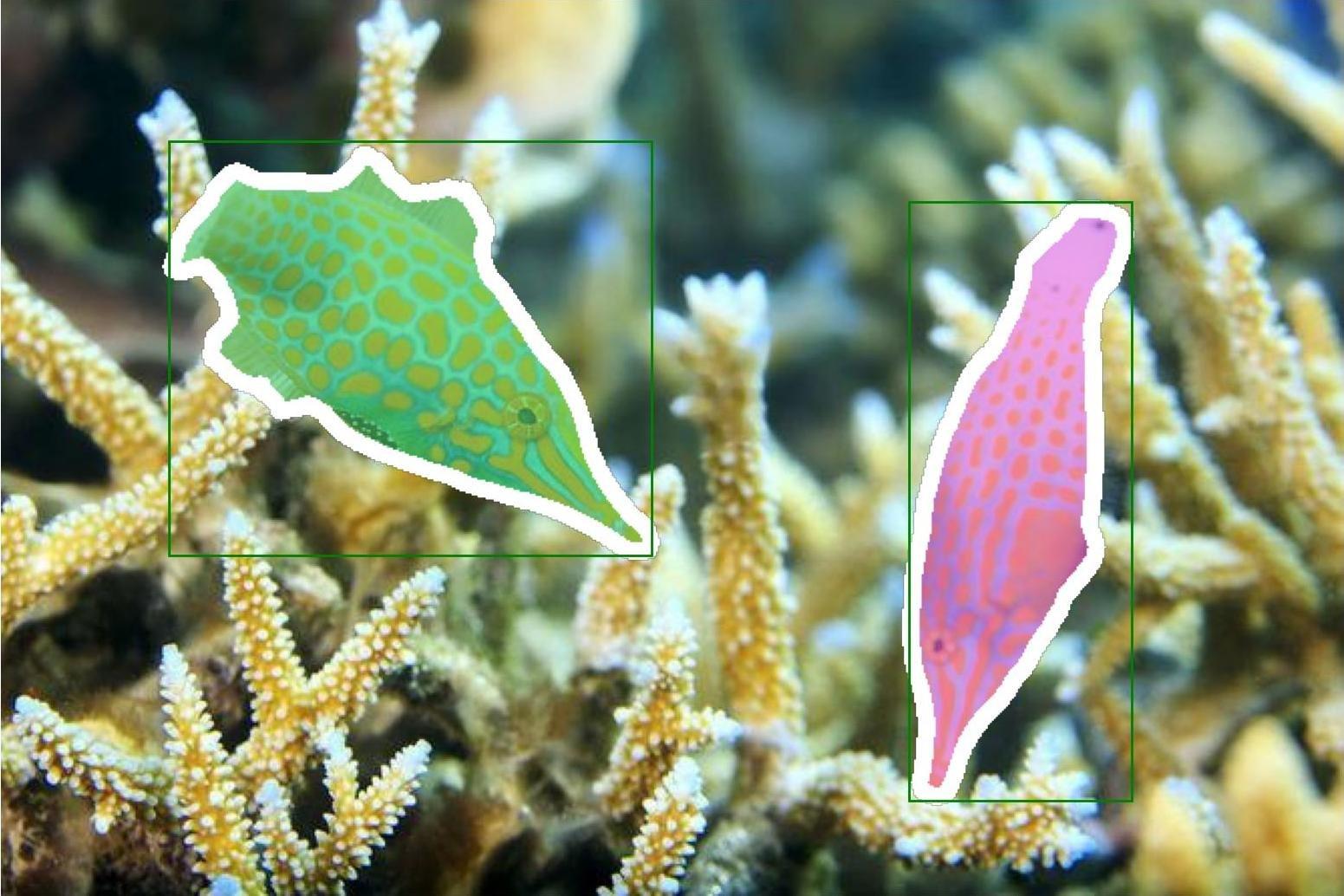} &
        \includegraphics[height=2cm]{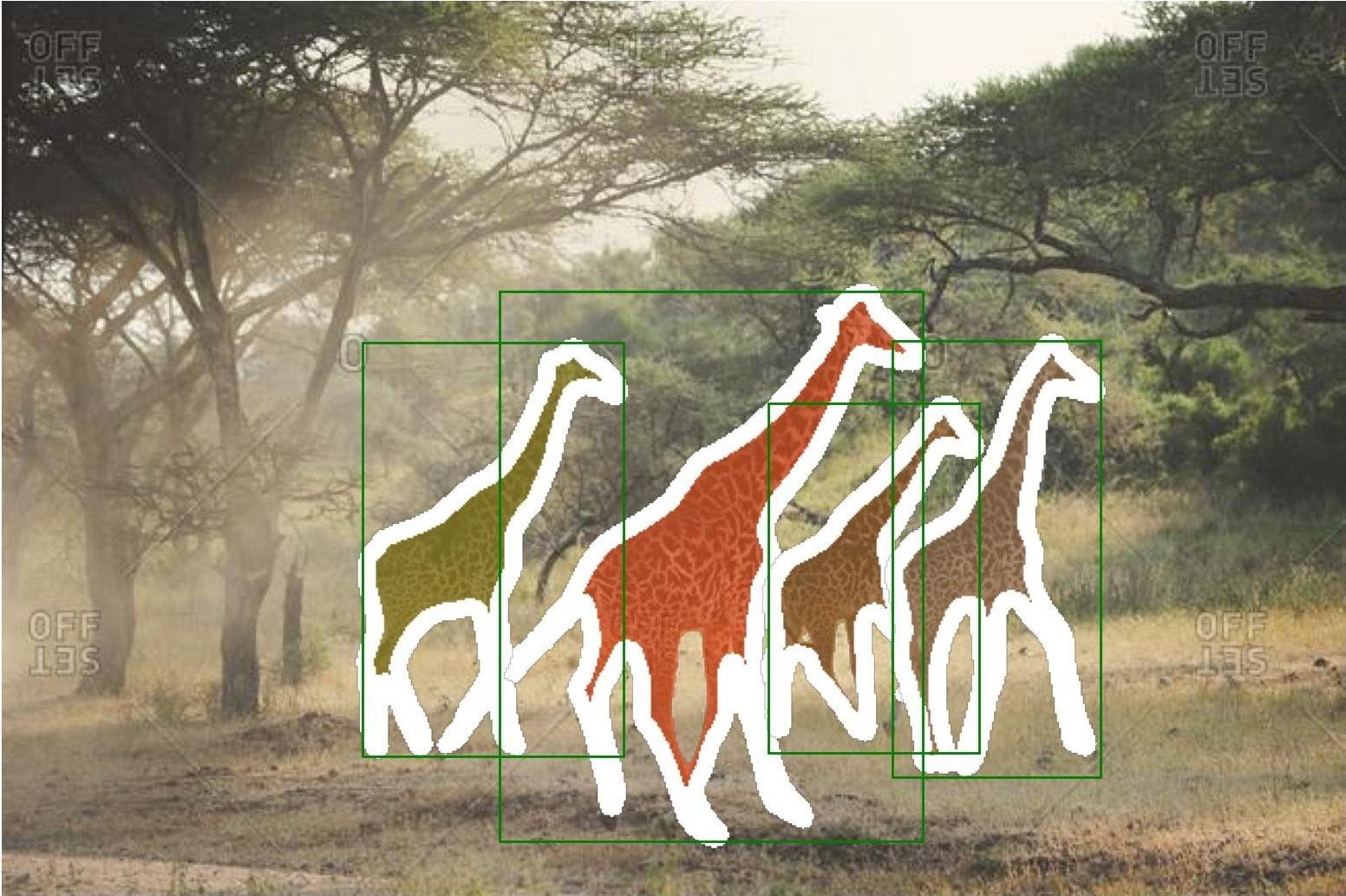} \\ 
    \end{tabular}
    \caption{Qualitative comparisons on COD10K and NC4K datasets for camouflaged instance segmentation.}
    \label{fig:CIS}
\end{figure*}

\subsubsection{Camouflaged Instance Segmentation}

\noindent\textbf{Quantitative Results.}
\tabref{table:CIS} shows the segmentation performance of SAM2 on camouflaged instances, which are more difficult to segment than salient instances in~\tabref{table:SIS}. 
In automatic mode, SAM2's performance is comparable to the unsupervised methods of task-specific algorithms and falls short of SAM, likely due to differences in parameter counts.
However, with box prompts, the performance of SAM2 improves dramatically. 
Specifically, on COD10K test set, the AP jumps from 10.6 to 68.8 with a large backbone, surpassing all other models. 
This suggests that the primary challenge of SAM2 in CIS is locating objects, but once the position is identified, it can produce precise segmentation.


\noindent\textbf{Qualitative Results.}
For qualitative analysis of the CIS task, as shown in~\figref{fig:CIS}, SAM-Auto can partially segment certain lightly concealed targets such as fish and giraffes, whereas SAM2-Auto has difficulty detecting camouflages. 
However, when provided with bounding box prompts, both SAM and SAM2 effectively segment camouflaged instances. Notably, SAM2 excels in capturing fine details, showcasing its strength in producing intricate features.

\begin{figure*}
    \centering
    \renewcommand{\arraystretch}{0.6}
    \begin{tabular}{@{}c@{\hskip 5pt}c@{\hskip 2pt}c@{\hskip 2pt}c@{\hskip 2pt}c@{\hskip 2pt}c@{\hskip 2pt}c@{}}
        \raisebox{0.5cm}{\makebox[0pt][c]{\rotatebox{90}{\small Image}}} & 
        \includegraphics[height=2cm]{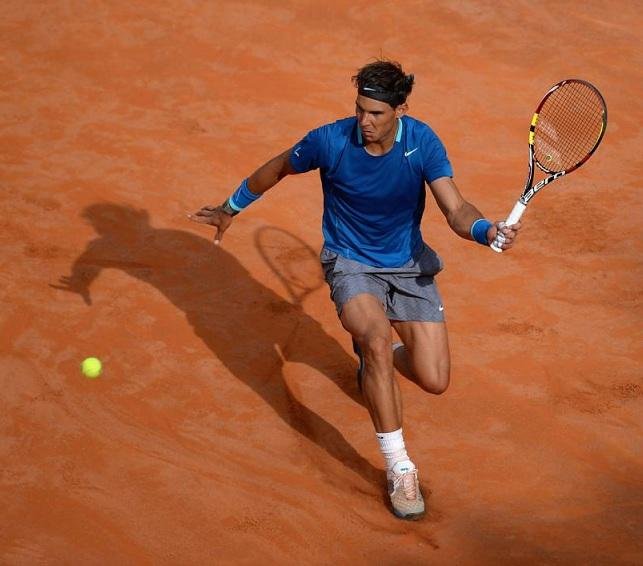} &
        \includegraphics[height=2cm]{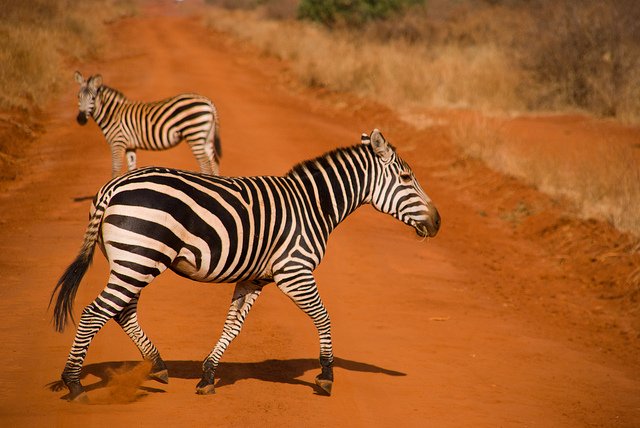} &
        \includegraphics[height=2cm]{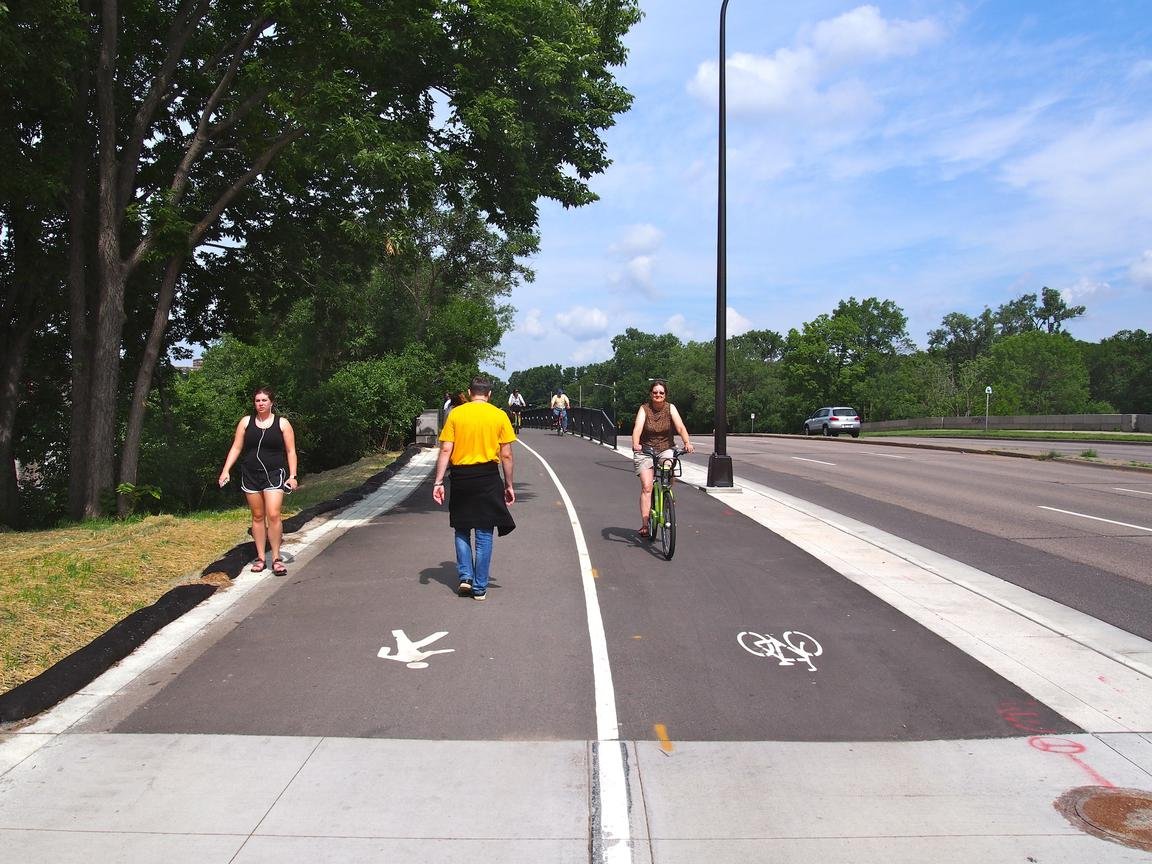} &
        \includegraphics[height=2cm]{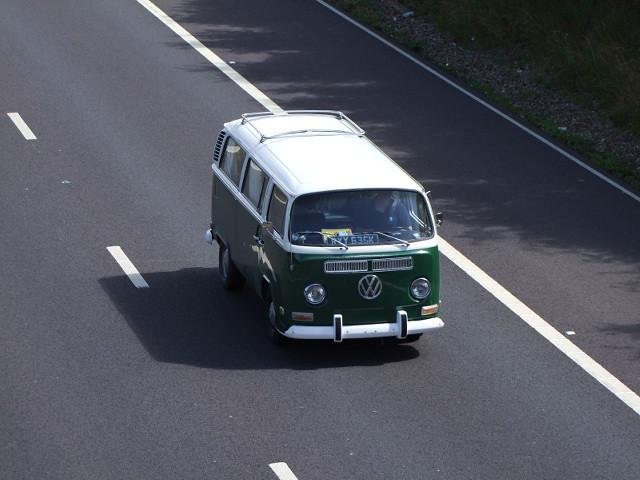} &
        \includegraphics[height=2cm]{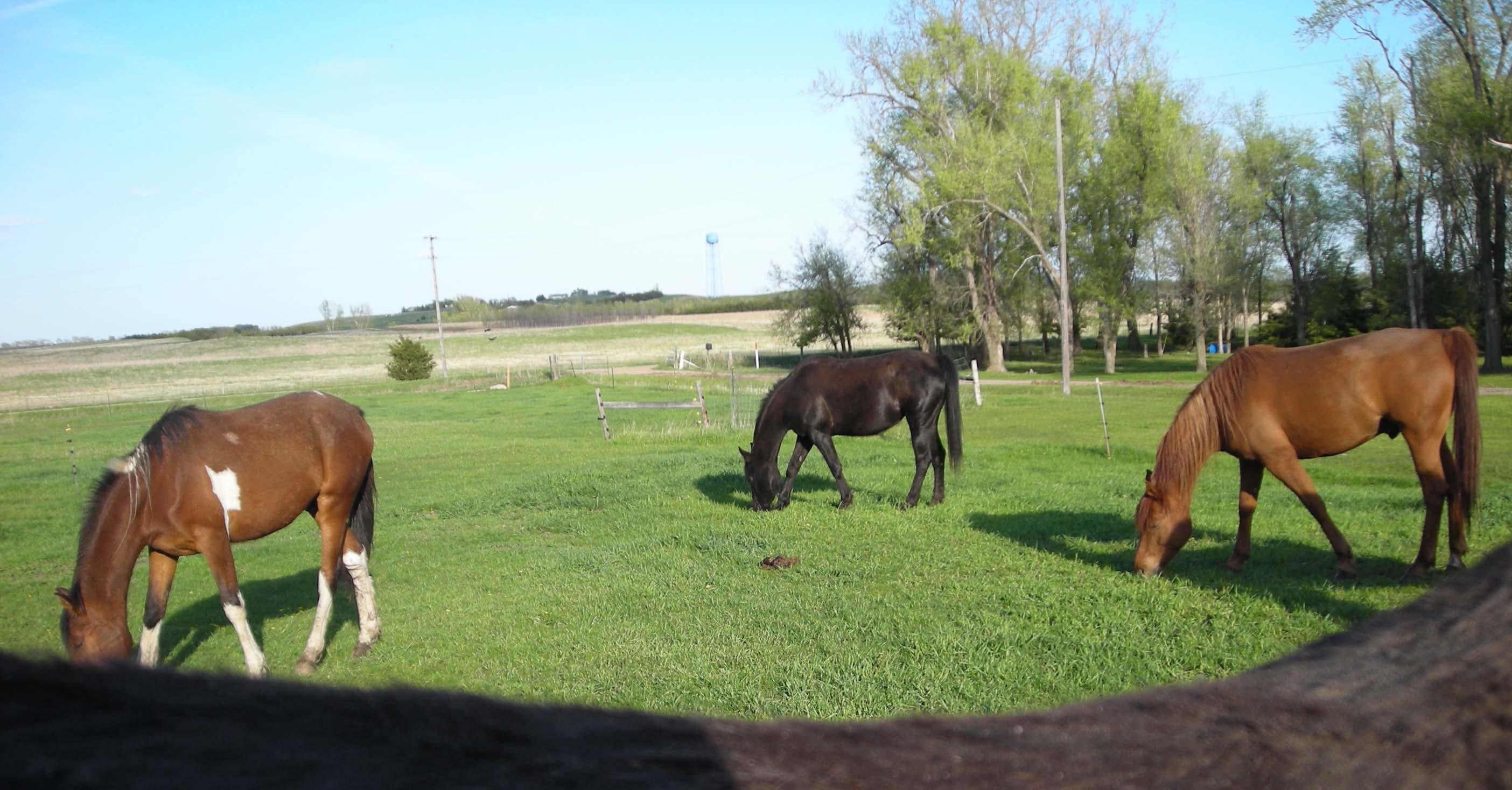} &
        \includegraphics[height=2cm]{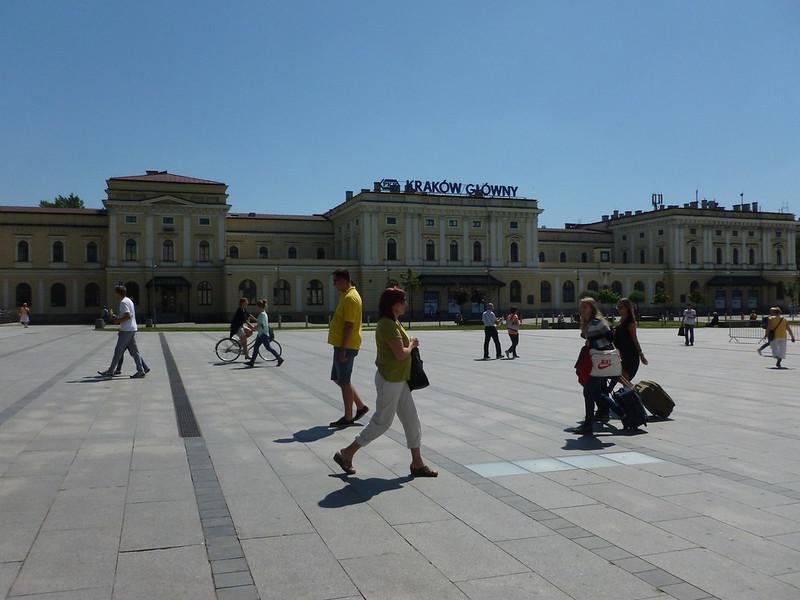}\\ 
        
        \raisebox{0cm}{\makebox[0pt][c]{\rotatebox{90}{\small Ground Truth}}} & 
        \includegraphics[height=2cm]{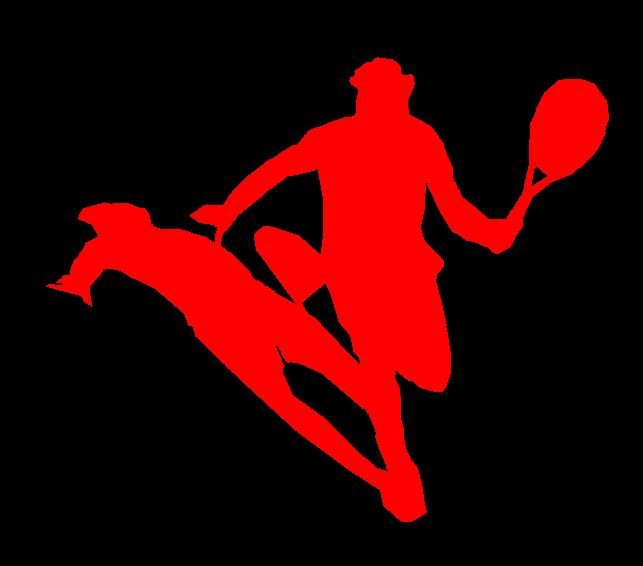} &
        \includegraphics[height=2cm]{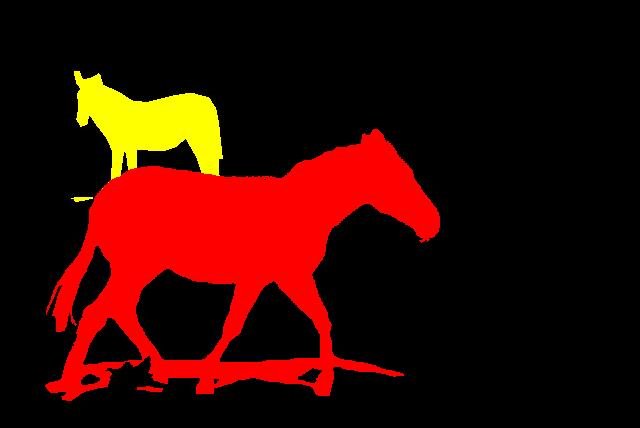} &
        \includegraphics[height=2cm]{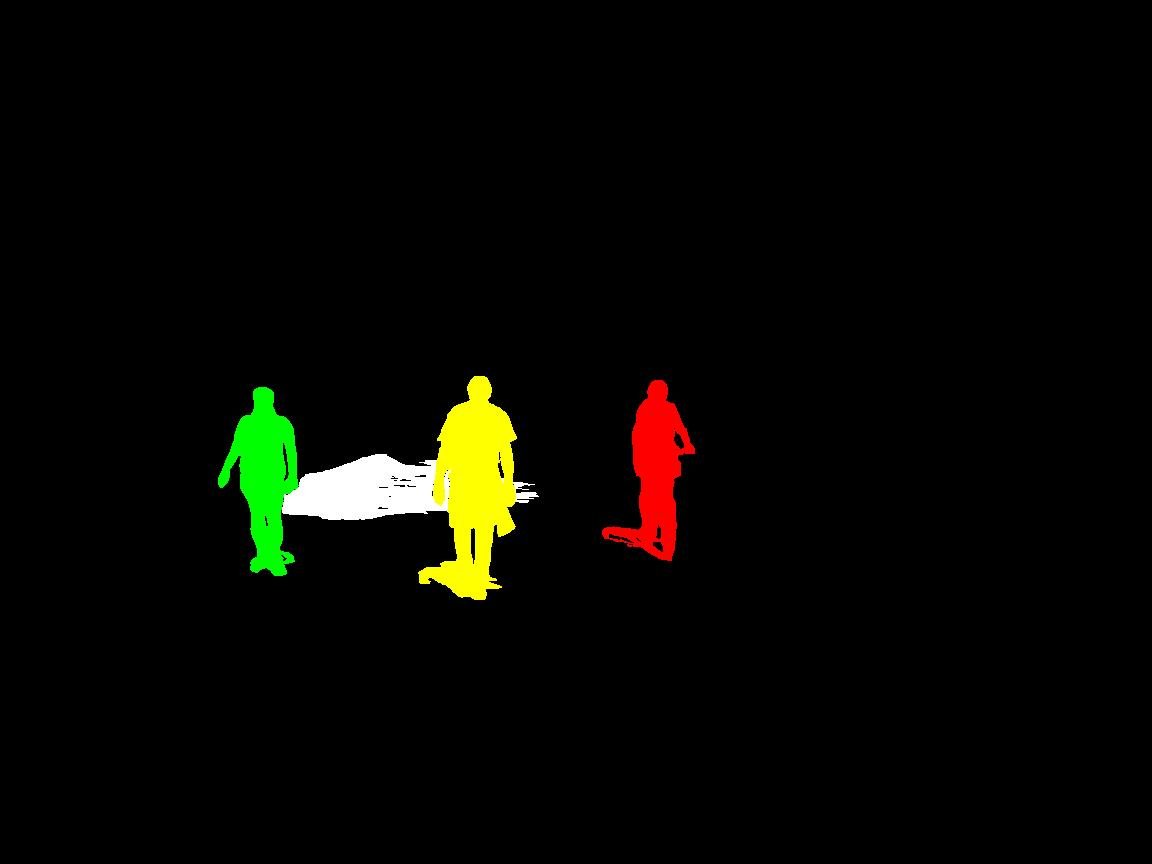} &
        \includegraphics[height=2cm]{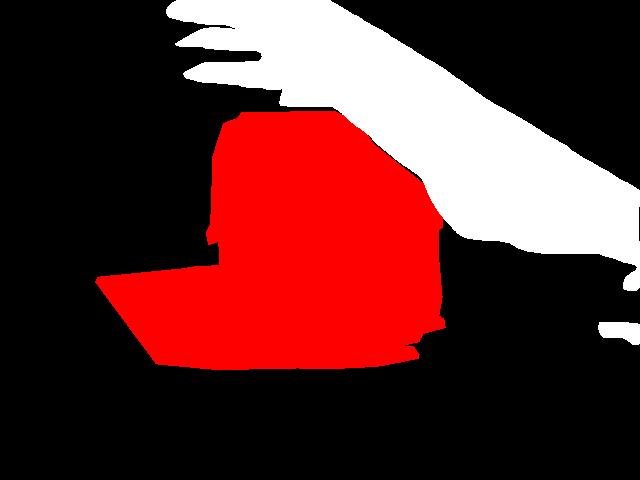} &
        \includegraphics[height=2cm]{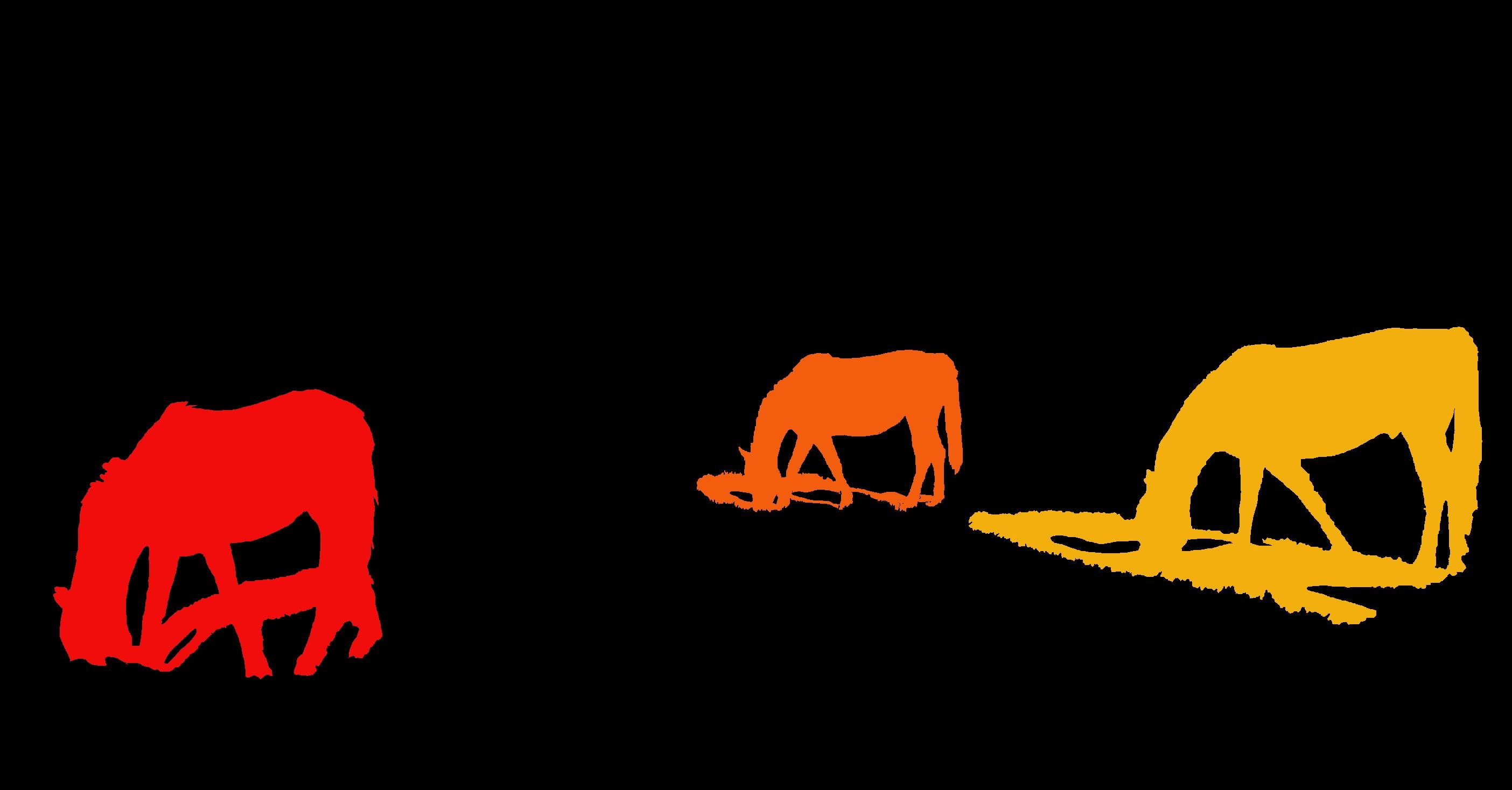} &
        \includegraphics[height=2cm]{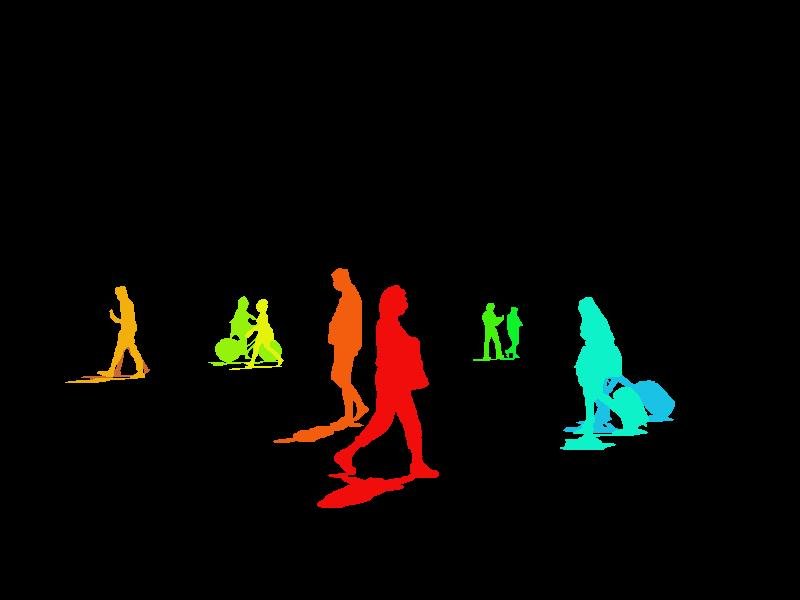}\\ 
        
        \raisebox{0.6cm}{\makebox[0pt][c]{\rotatebox{90}{\small SSIS}}} & 
        \includegraphics[height=2cm]{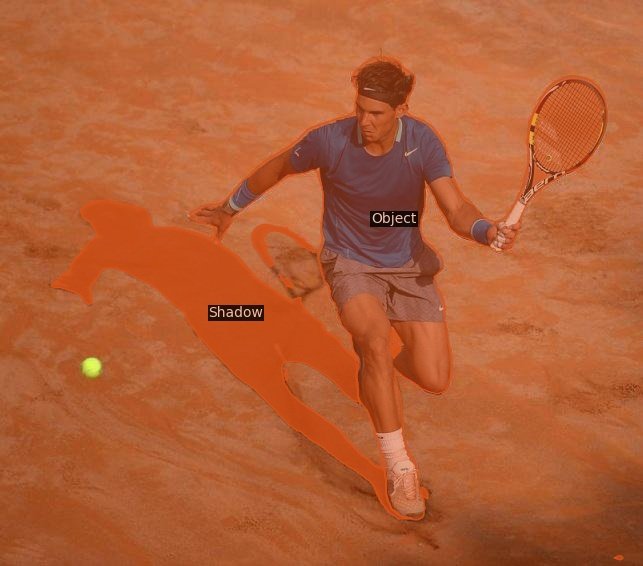} &
        \includegraphics[height=2cm]{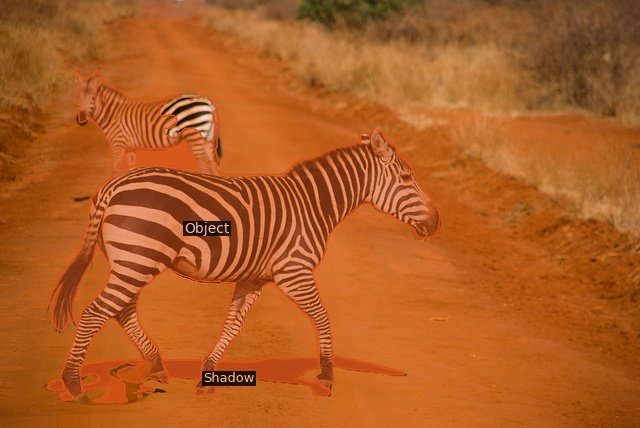} &
        \includegraphics[height=2cm]{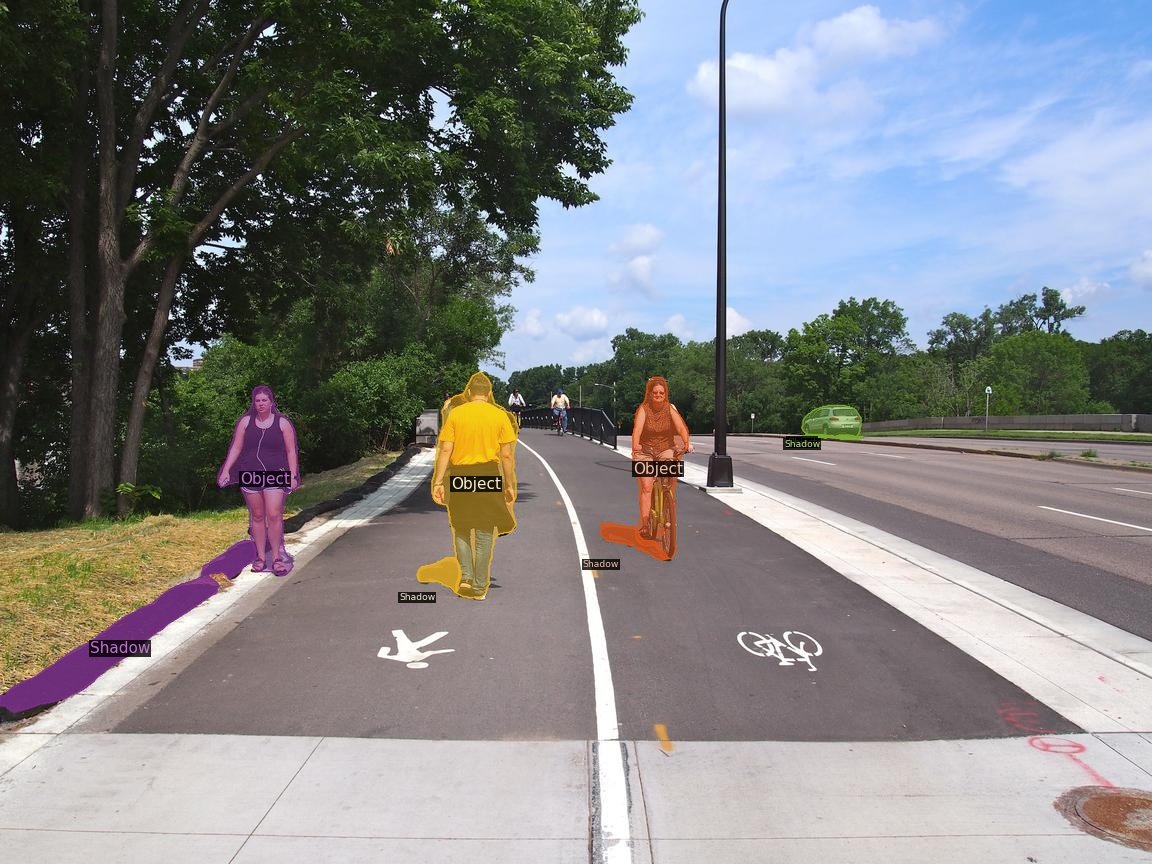} &
        \includegraphics[height=2cm]{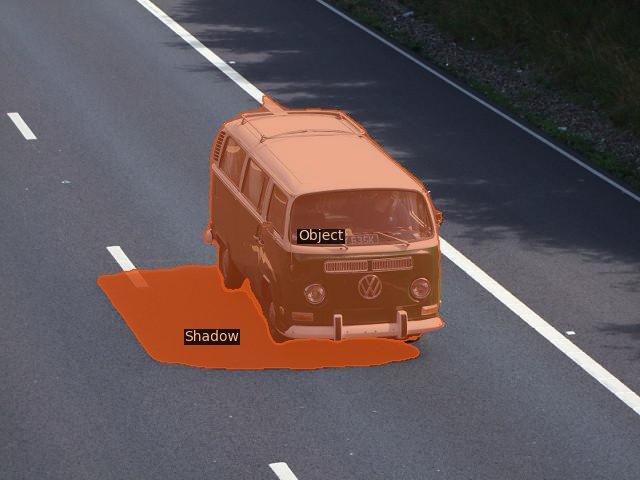} &
        \includegraphics[height=2cm]{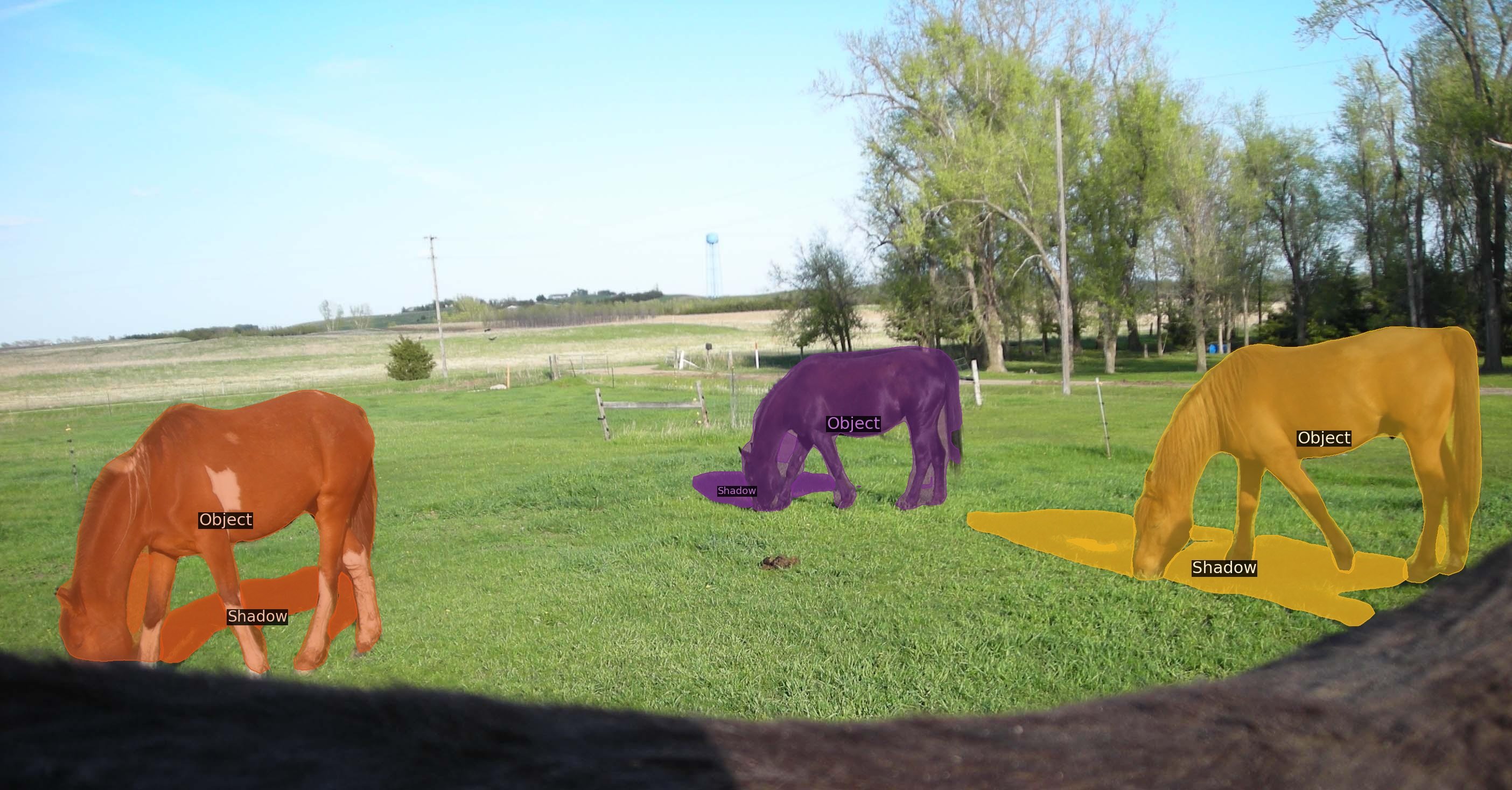} &
        \includegraphics[height=2cm]{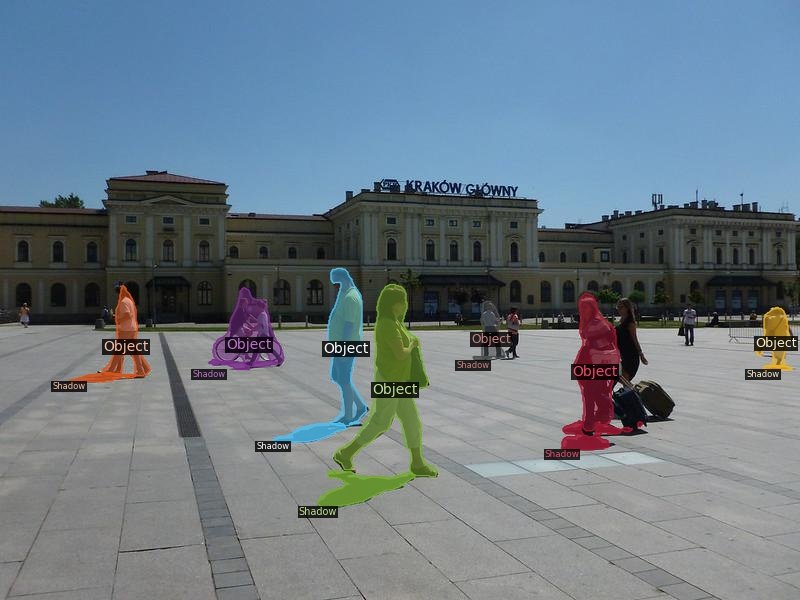}\\ 
        
        \raisebox{0.5cm}{\makebox[0pt][c]{\rotatebox{90}{\small SSISv2}}} & 
        \includegraphics[height=2cm]{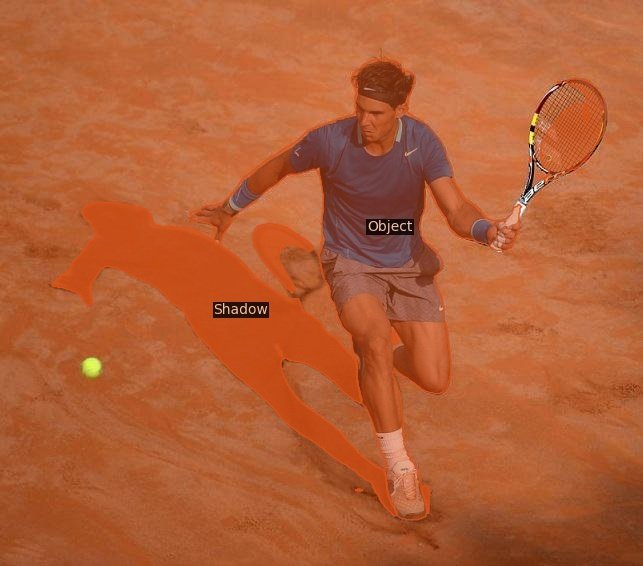} &
        \includegraphics[height=2cm]{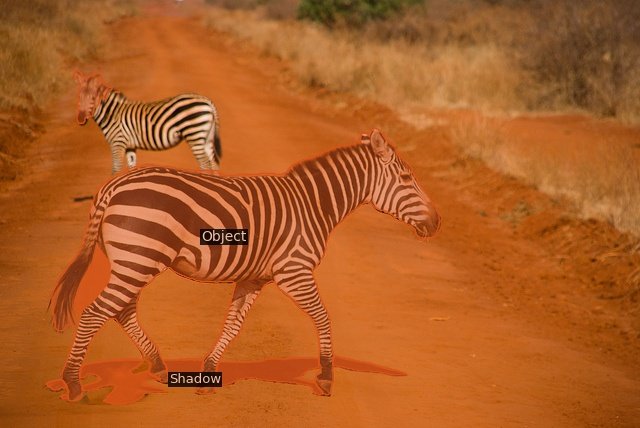} &
        \includegraphics[height=2cm]{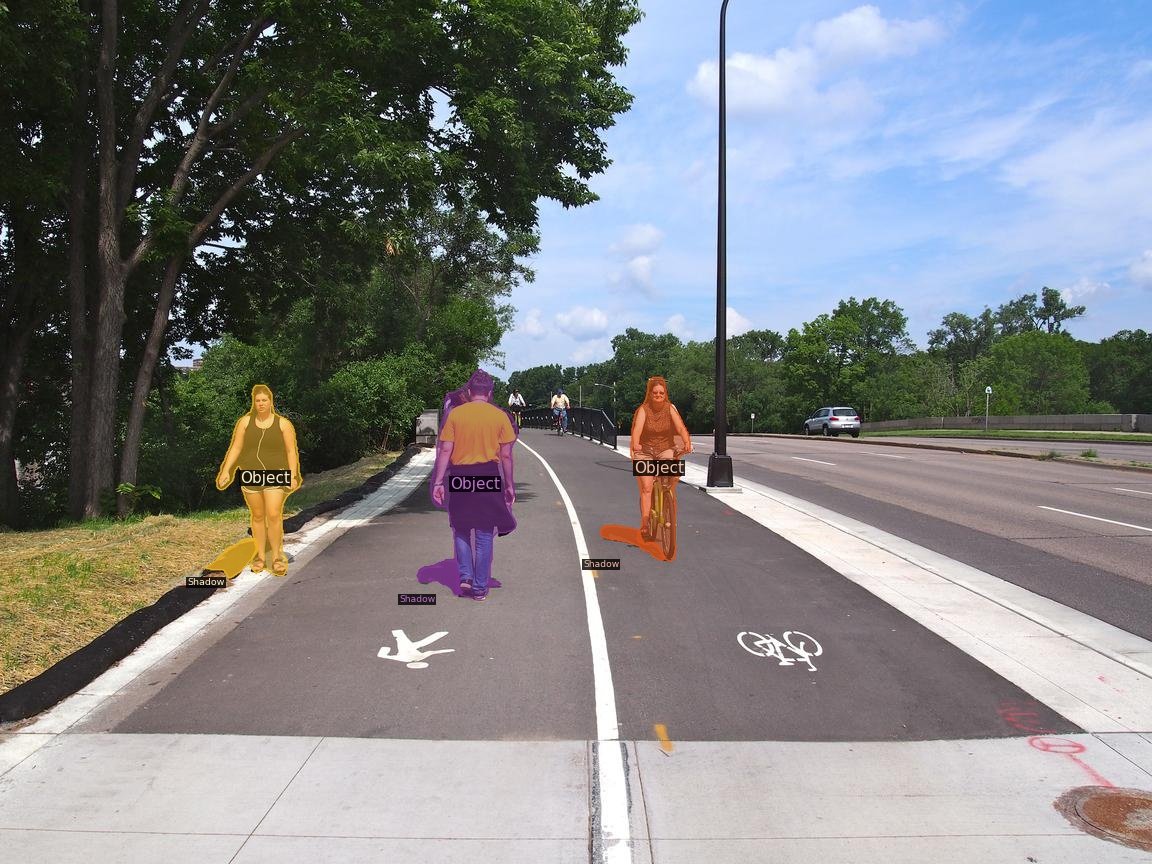} &
        \includegraphics[height=2cm]{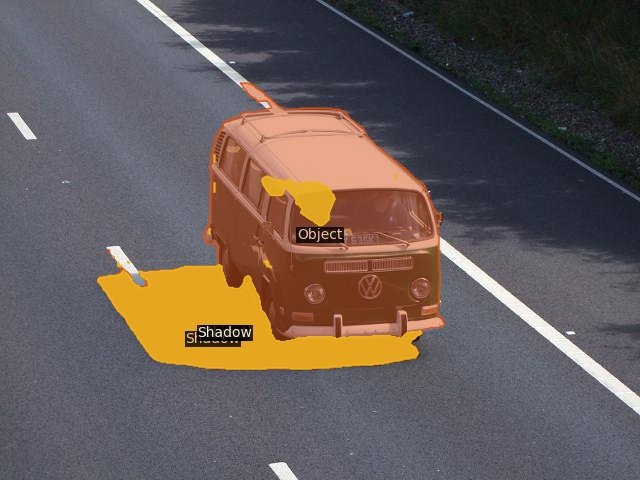} &
        \includegraphics[height=2cm]{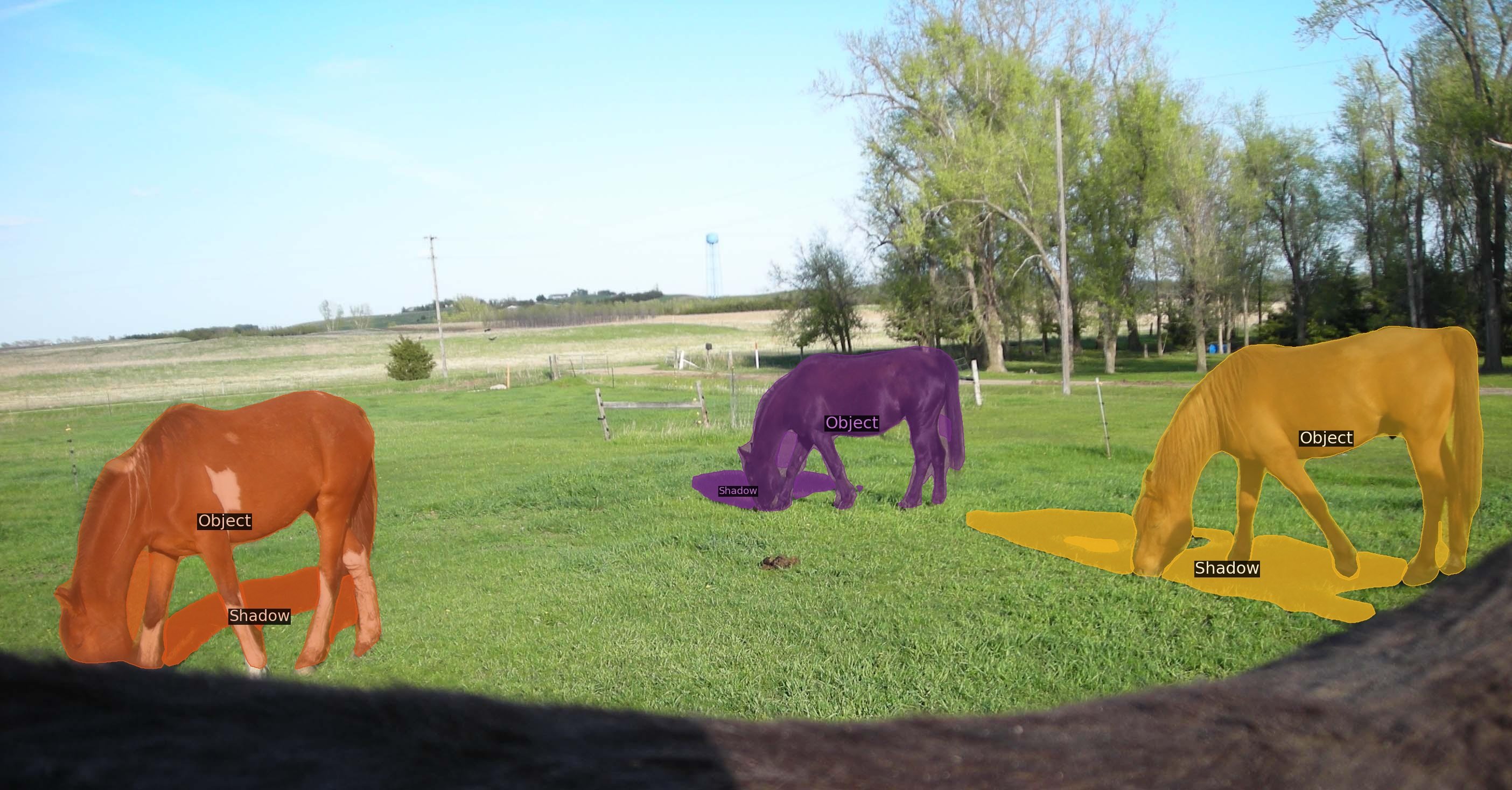} &
        \includegraphics[height=2cm]{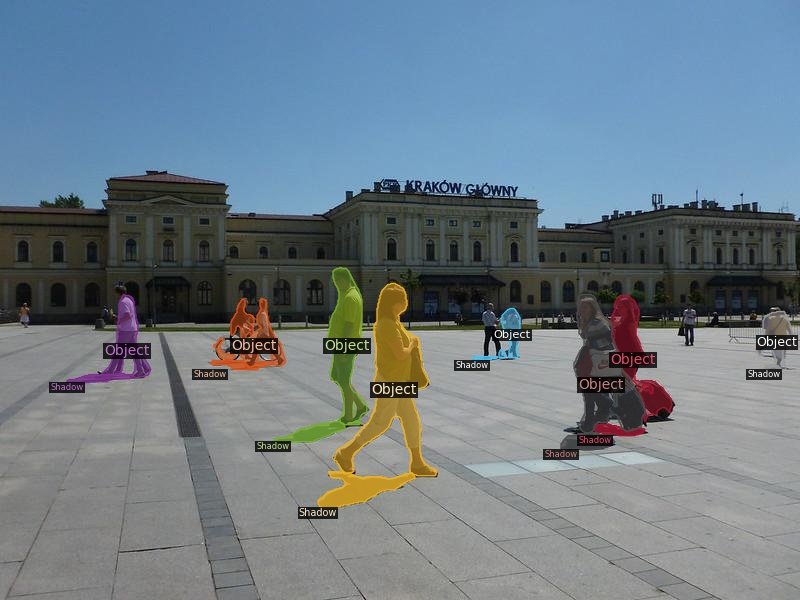}\\ 
        
        \raisebox{0.2cm}{\makebox[0pt][c]{\rotatebox{90}{\small SAM-Auto}}} & 
        \includegraphics[height=2cm]{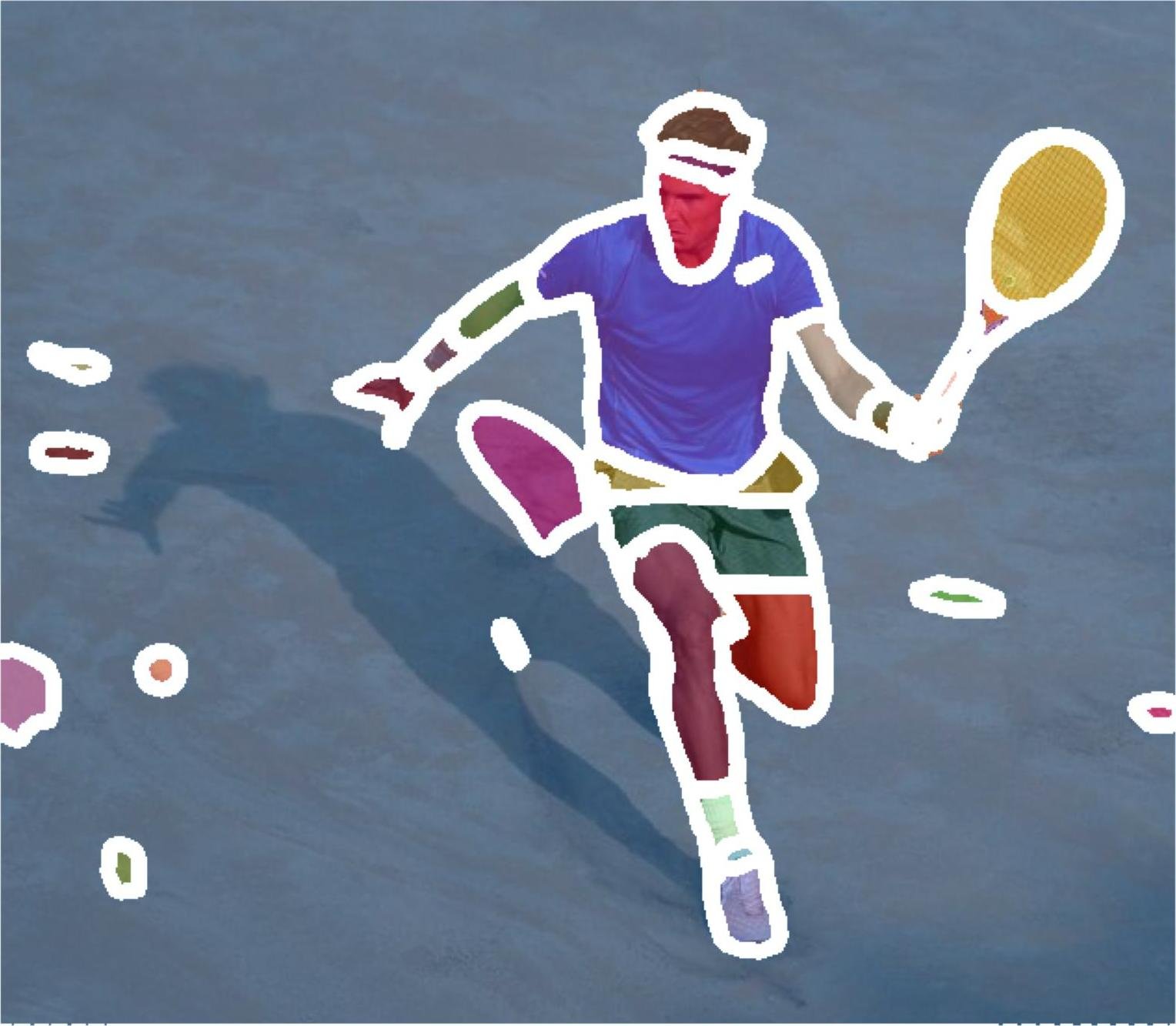} &
        \includegraphics[height=2cm]{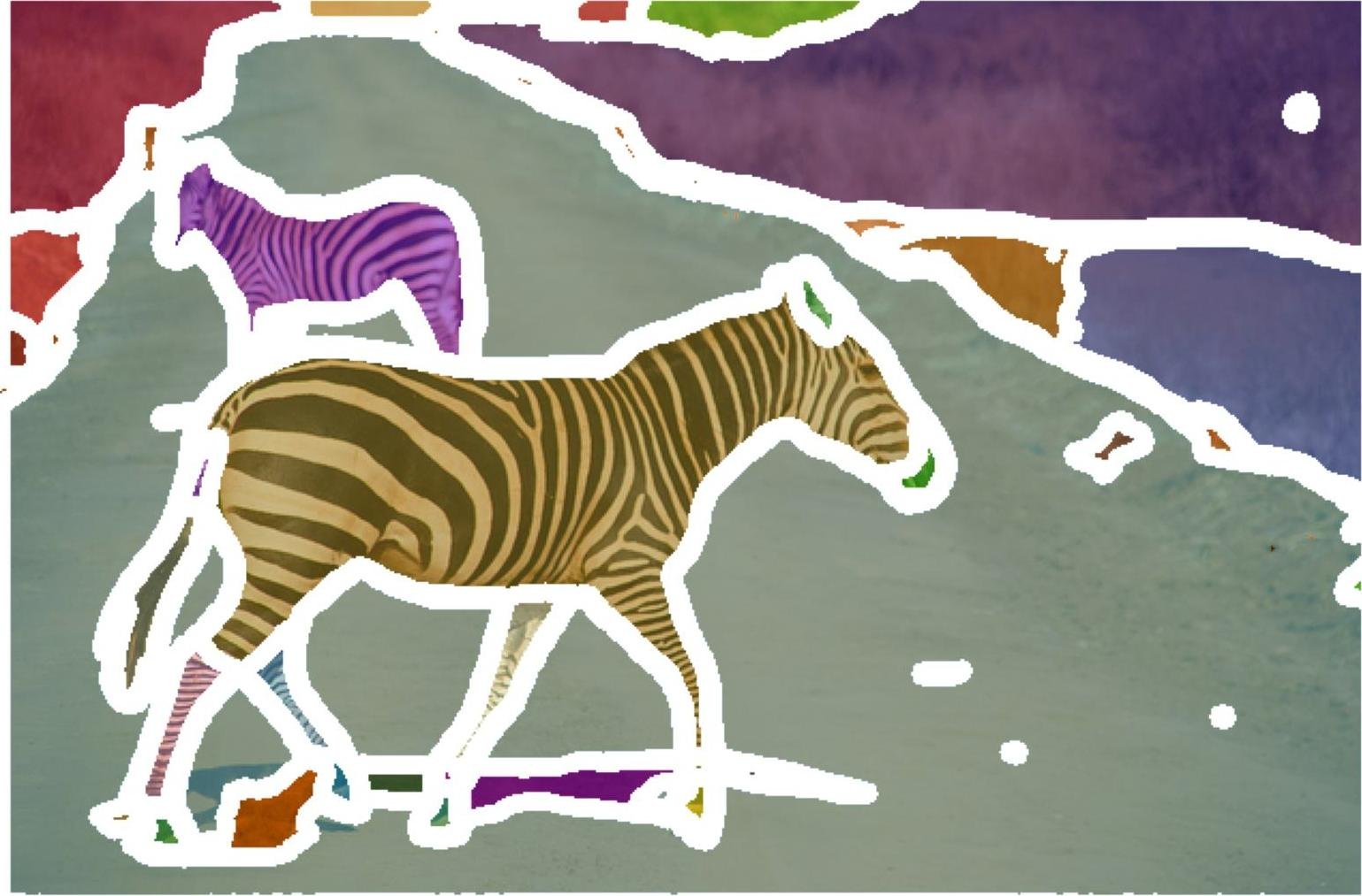} &
        \includegraphics[height=2cm]{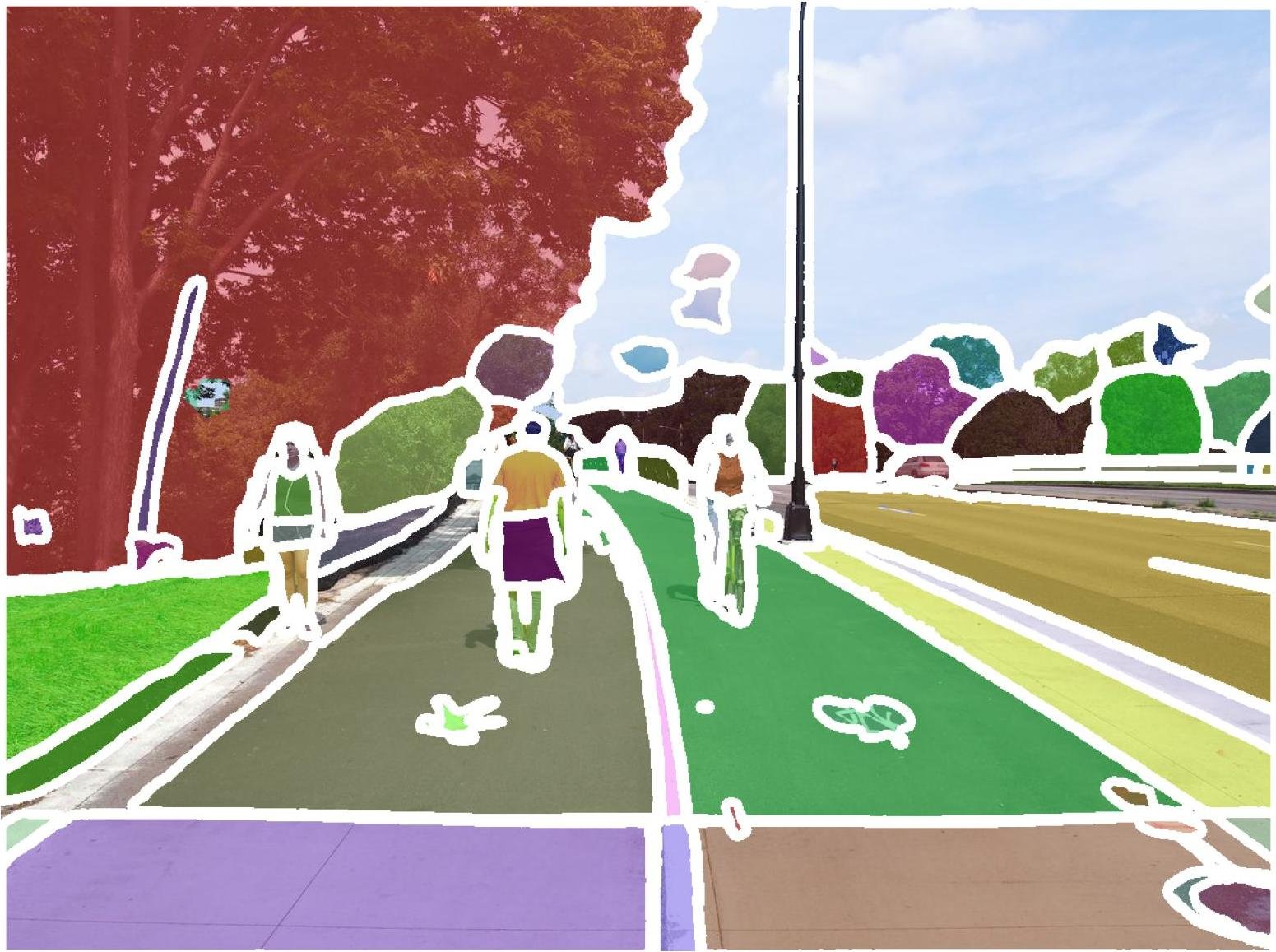} &
        \includegraphics[height=2cm]{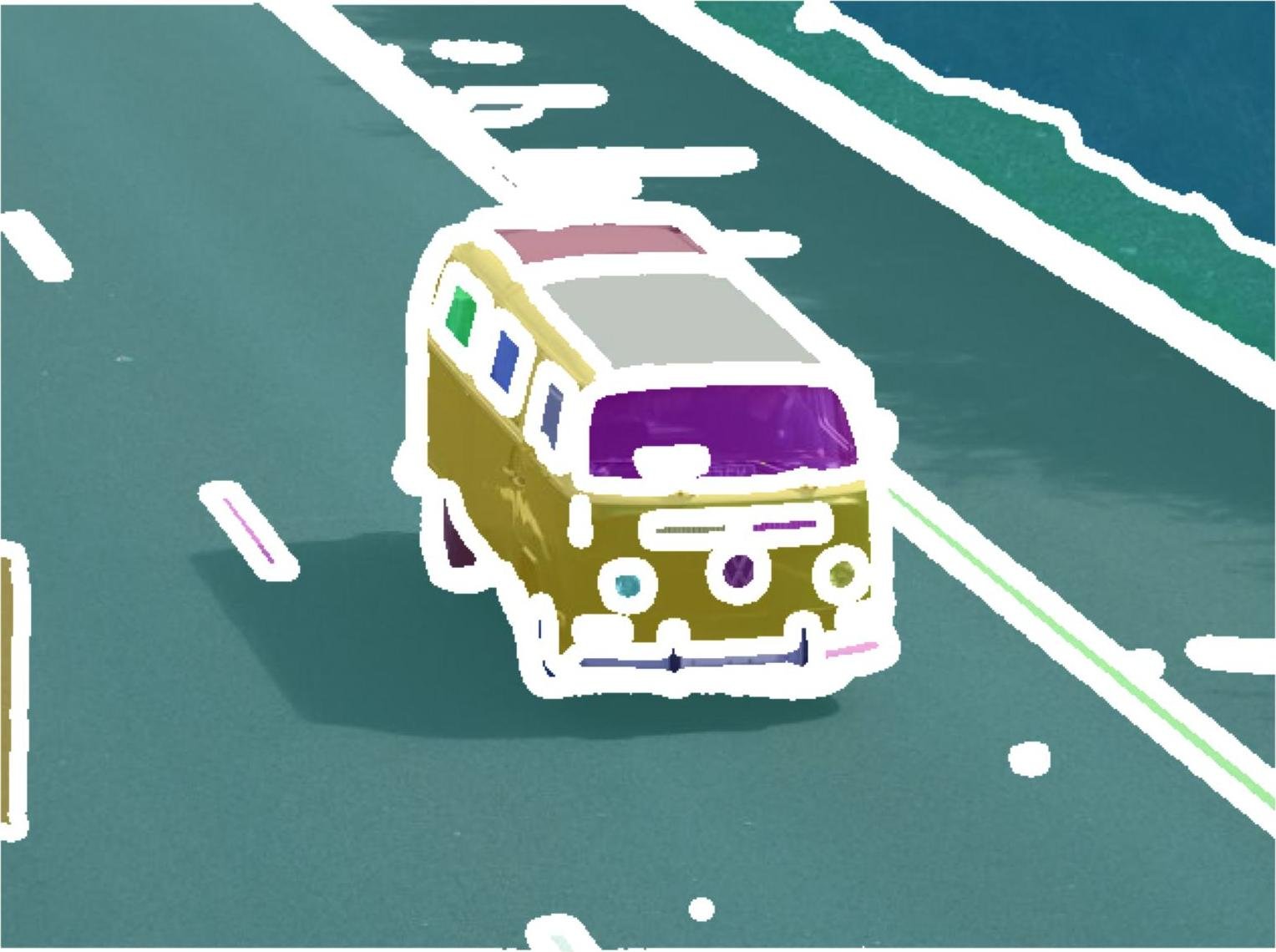} &
        \includegraphics[height=2cm]{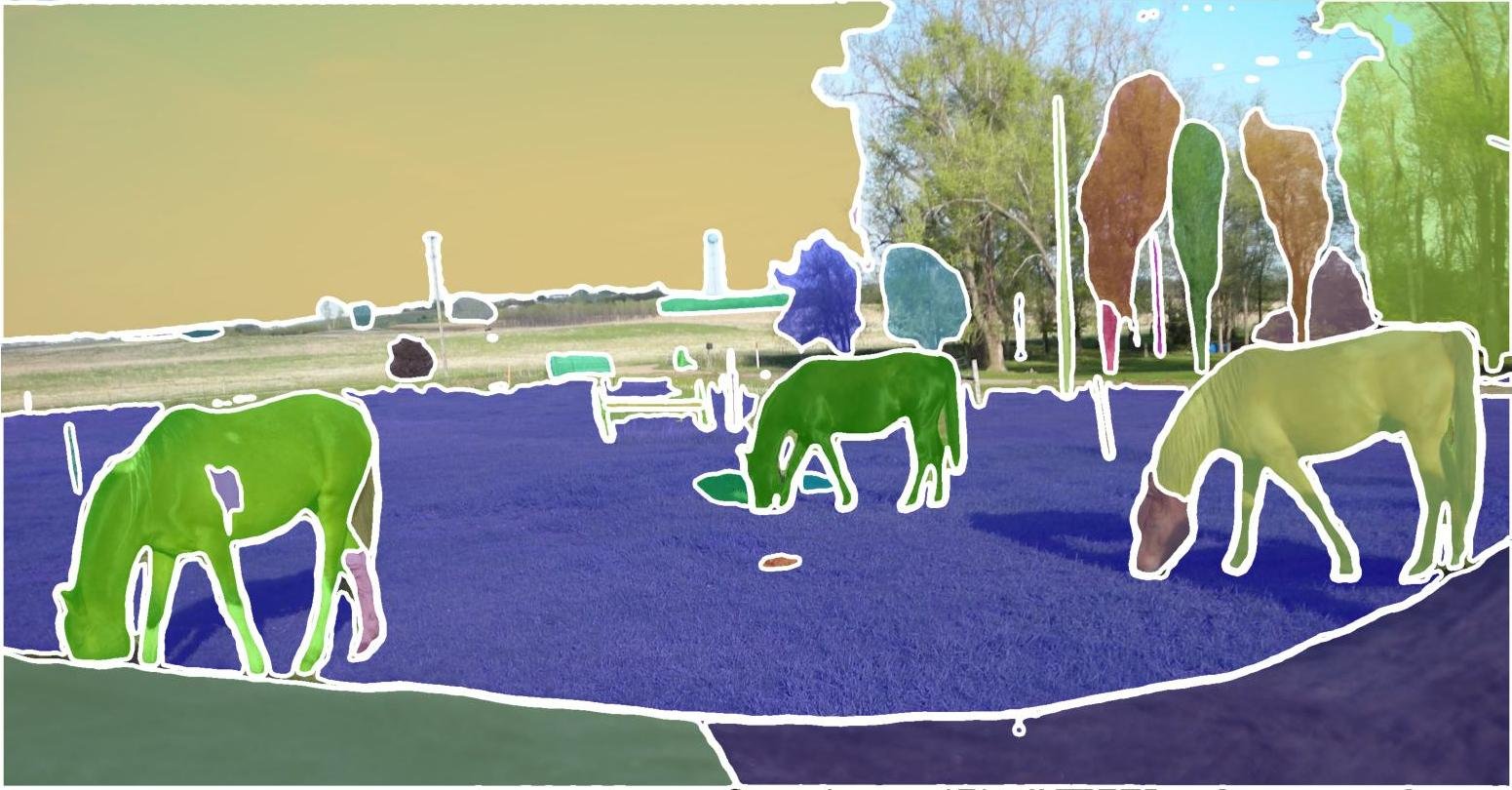} &
        \includegraphics[height=2cm]{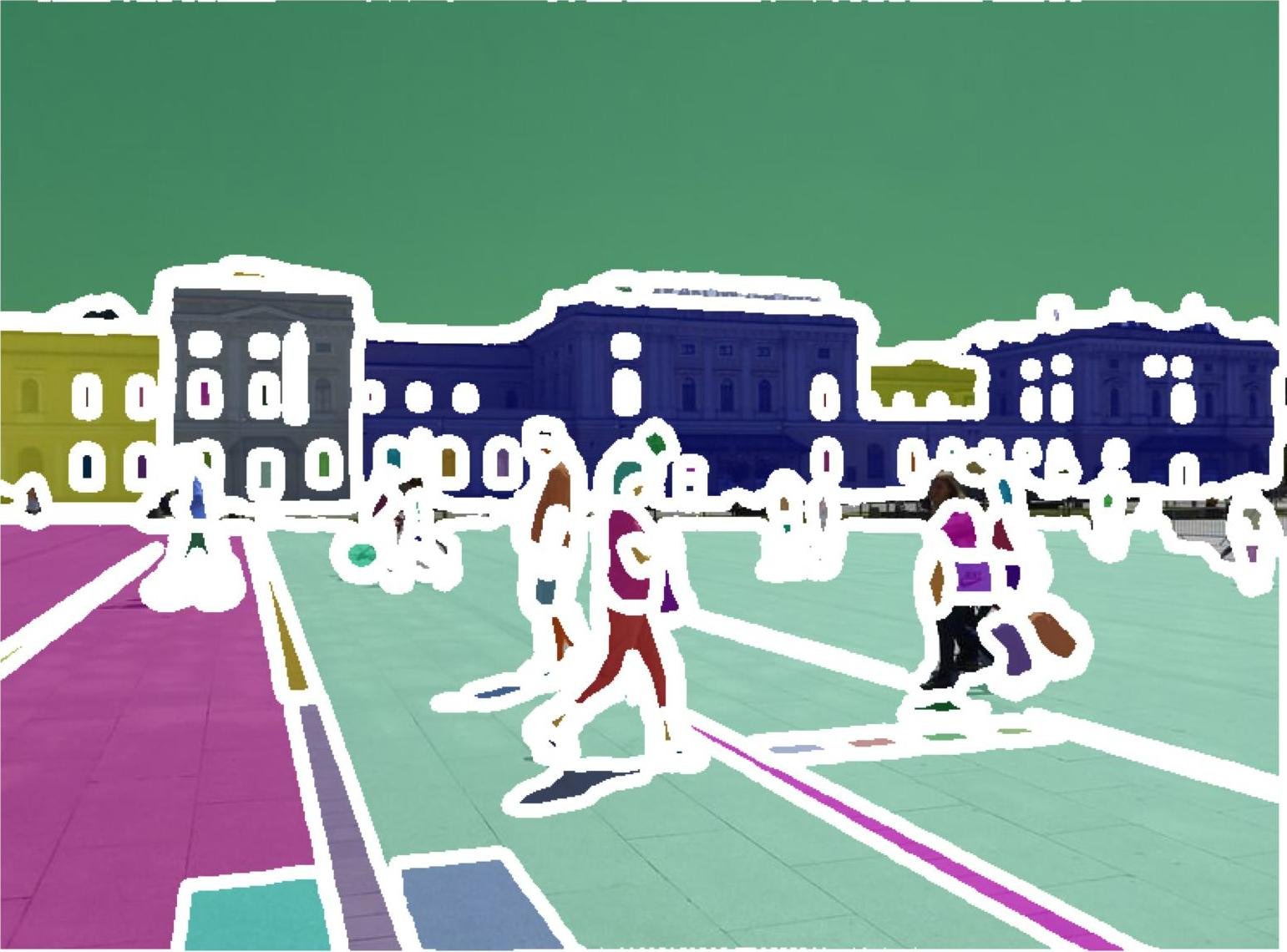}\\ 
        
        \raisebox{0.2cm}{\makebox[0pt][c]{\rotatebox{90}{\small SAM-Bbox}}} & 
        \includegraphics[height=2cm]{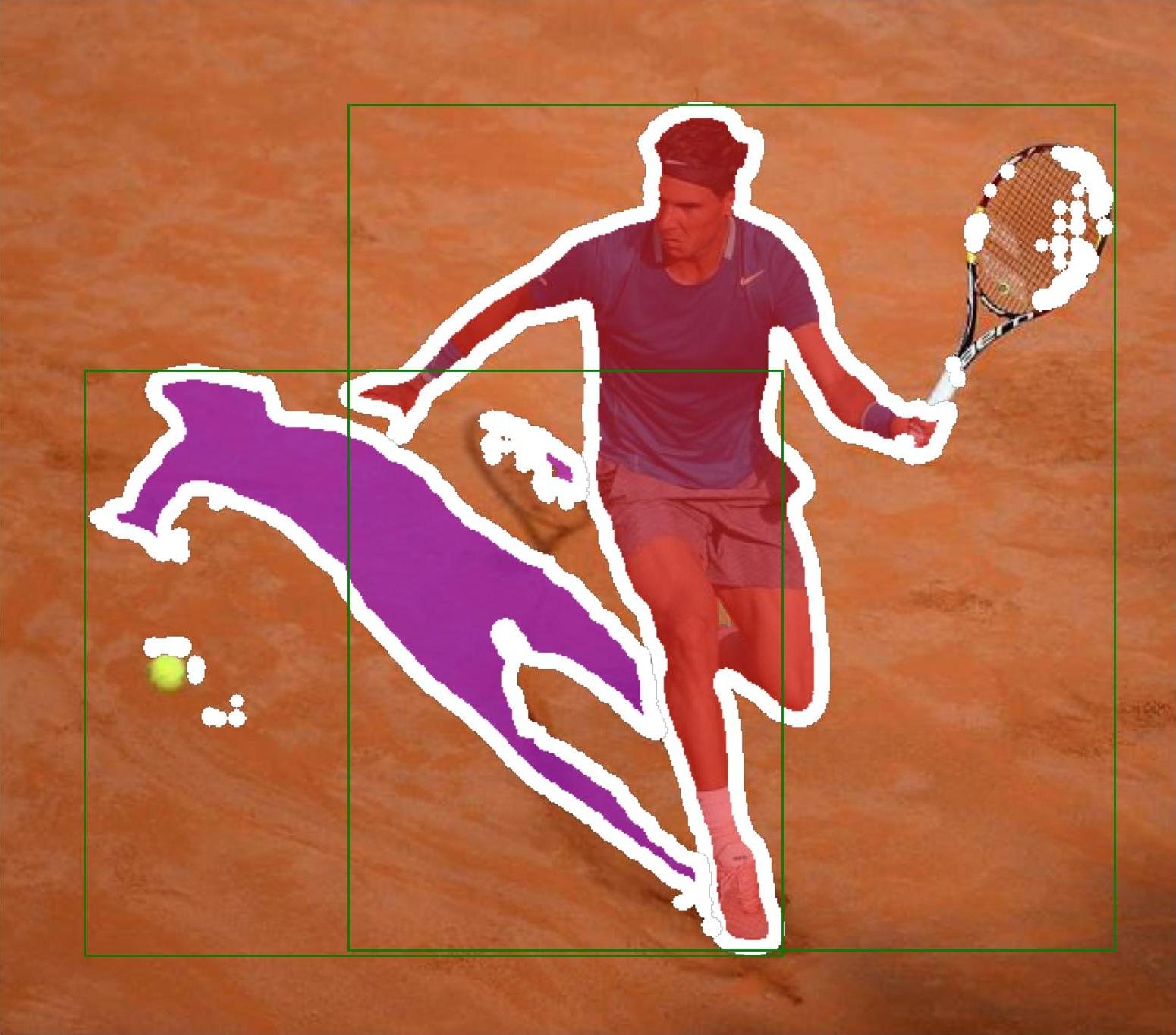} &
        \includegraphics[height=2cm]{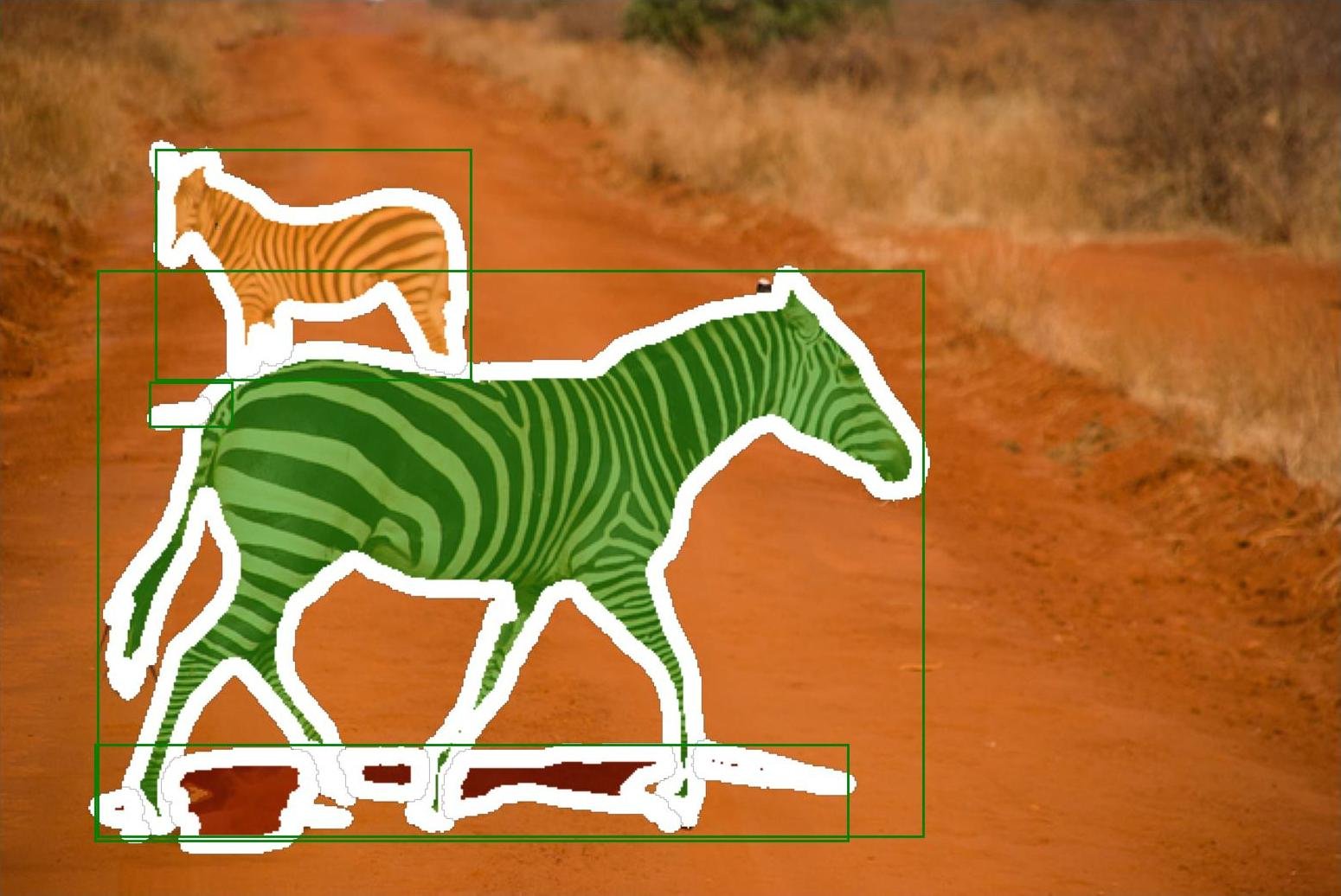} &
        \includegraphics[height=2cm]{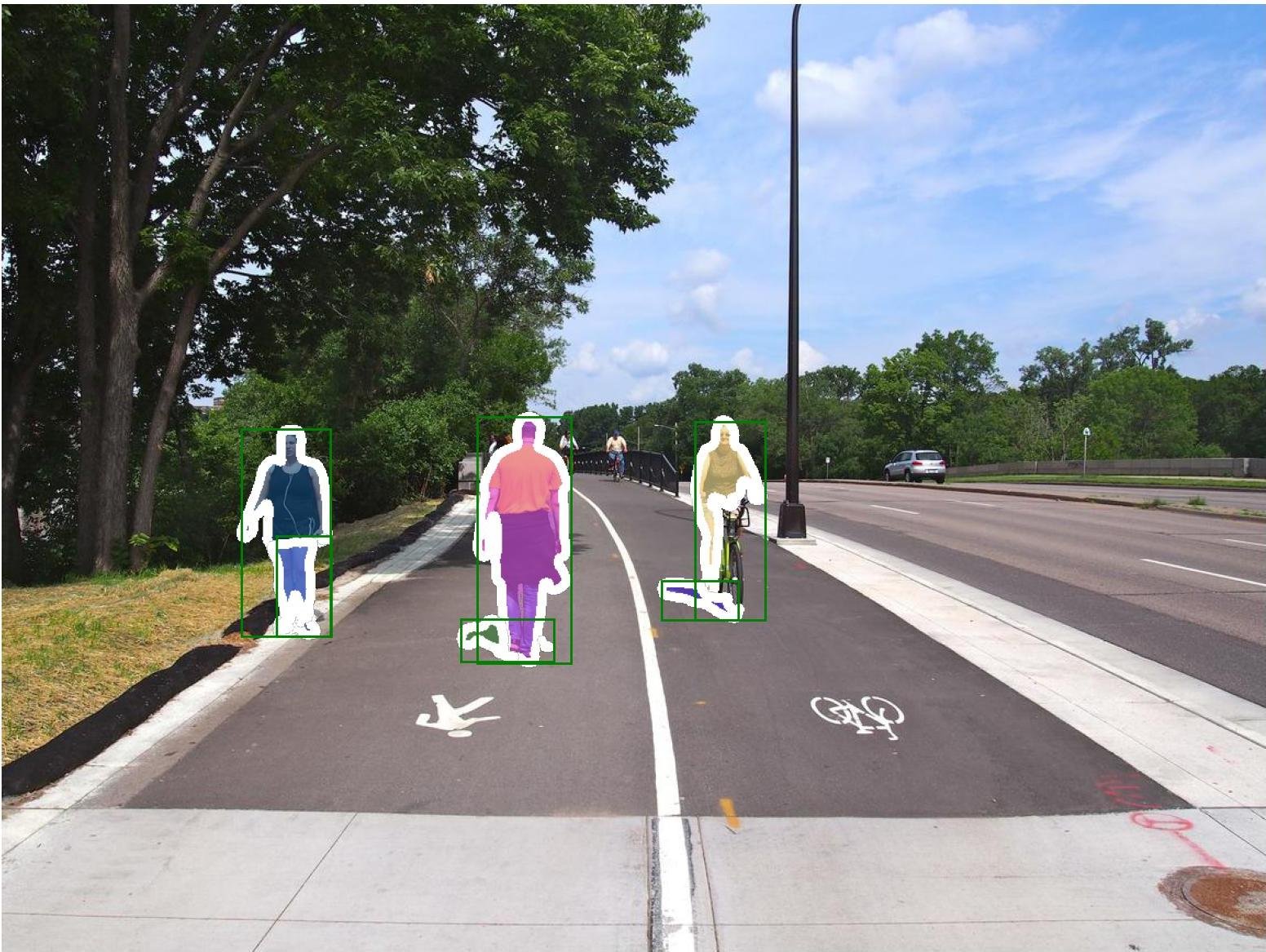} &
        \includegraphics[height=2cm]{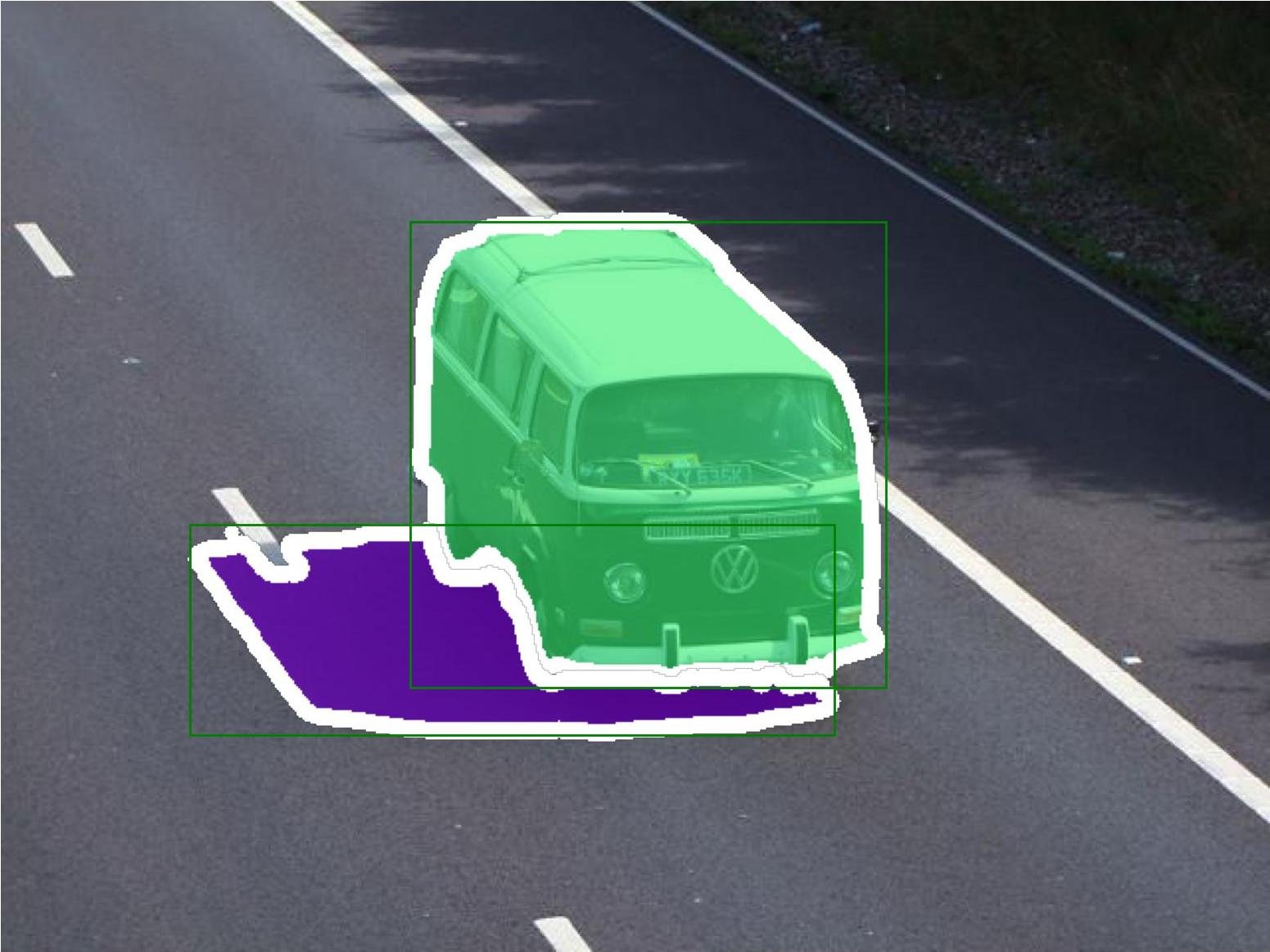} &
        \includegraphics[height=2cm]{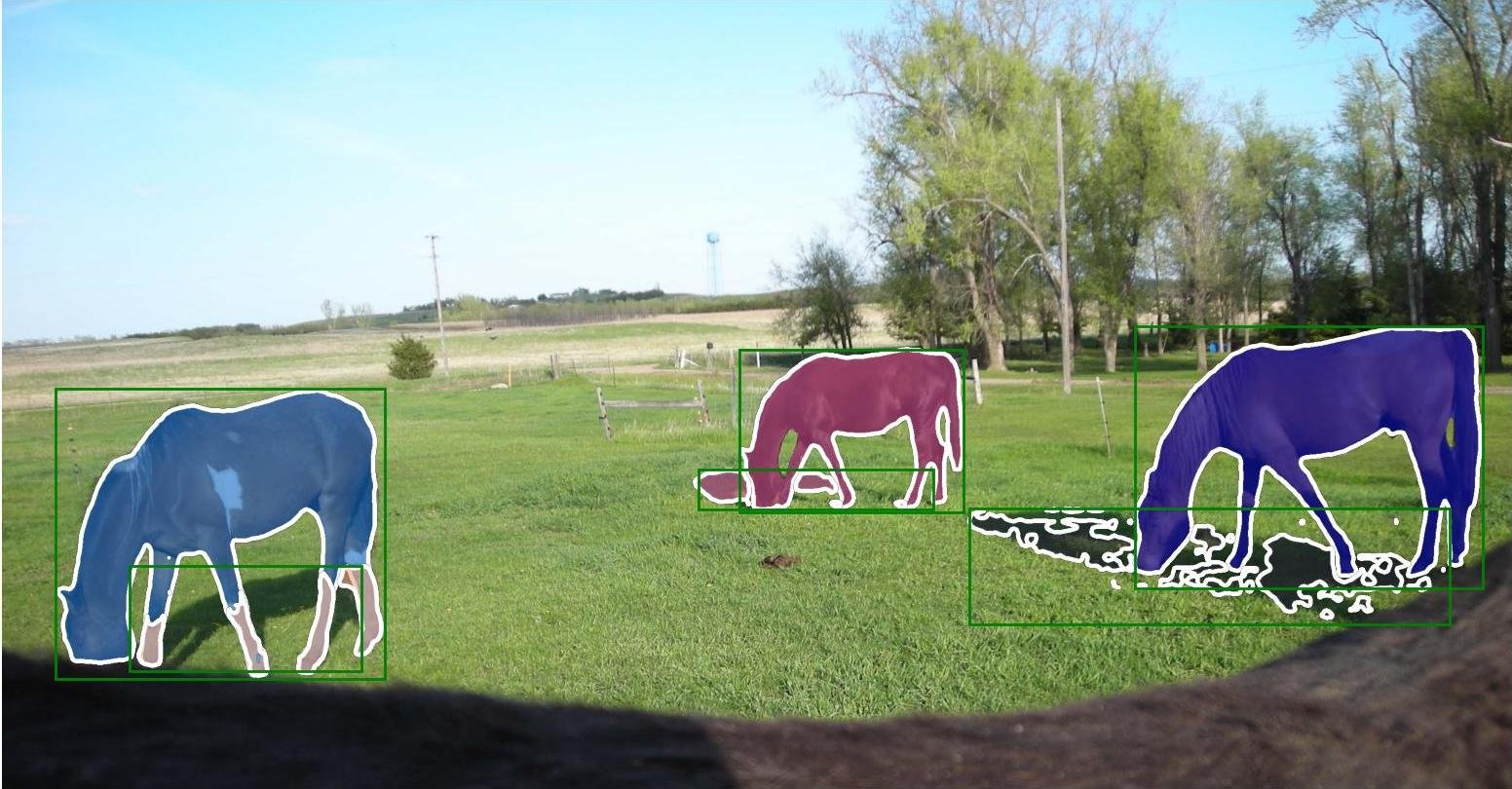} &
        \includegraphics[height=2cm]{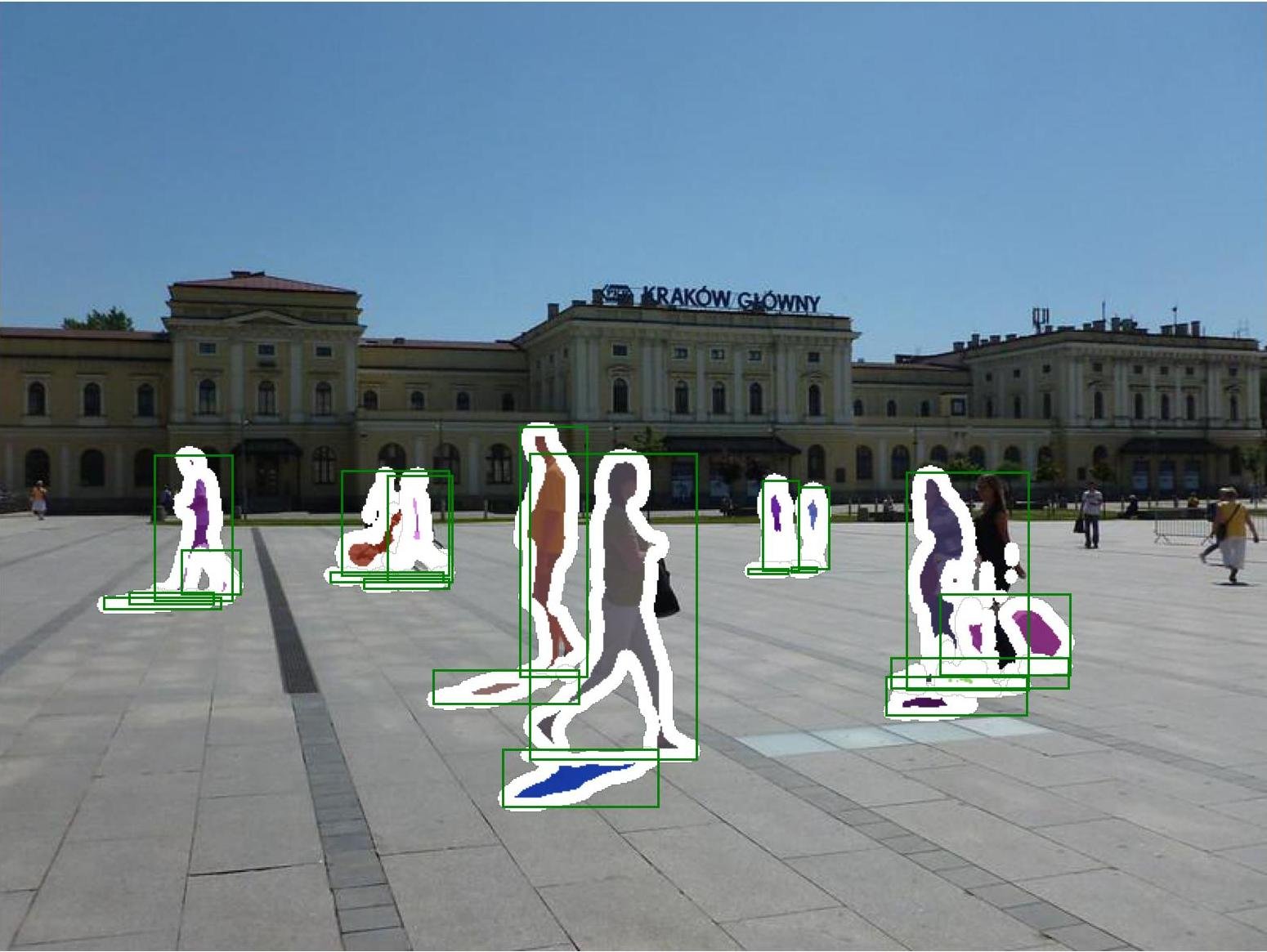}\\ 
        
        \raisebox{0.1cm}{\makebox[0pt][c]{\rotatebox{90}{\small \textbf{SAM2-Auto}}}} & 
        \includegraphics[height=2cm]{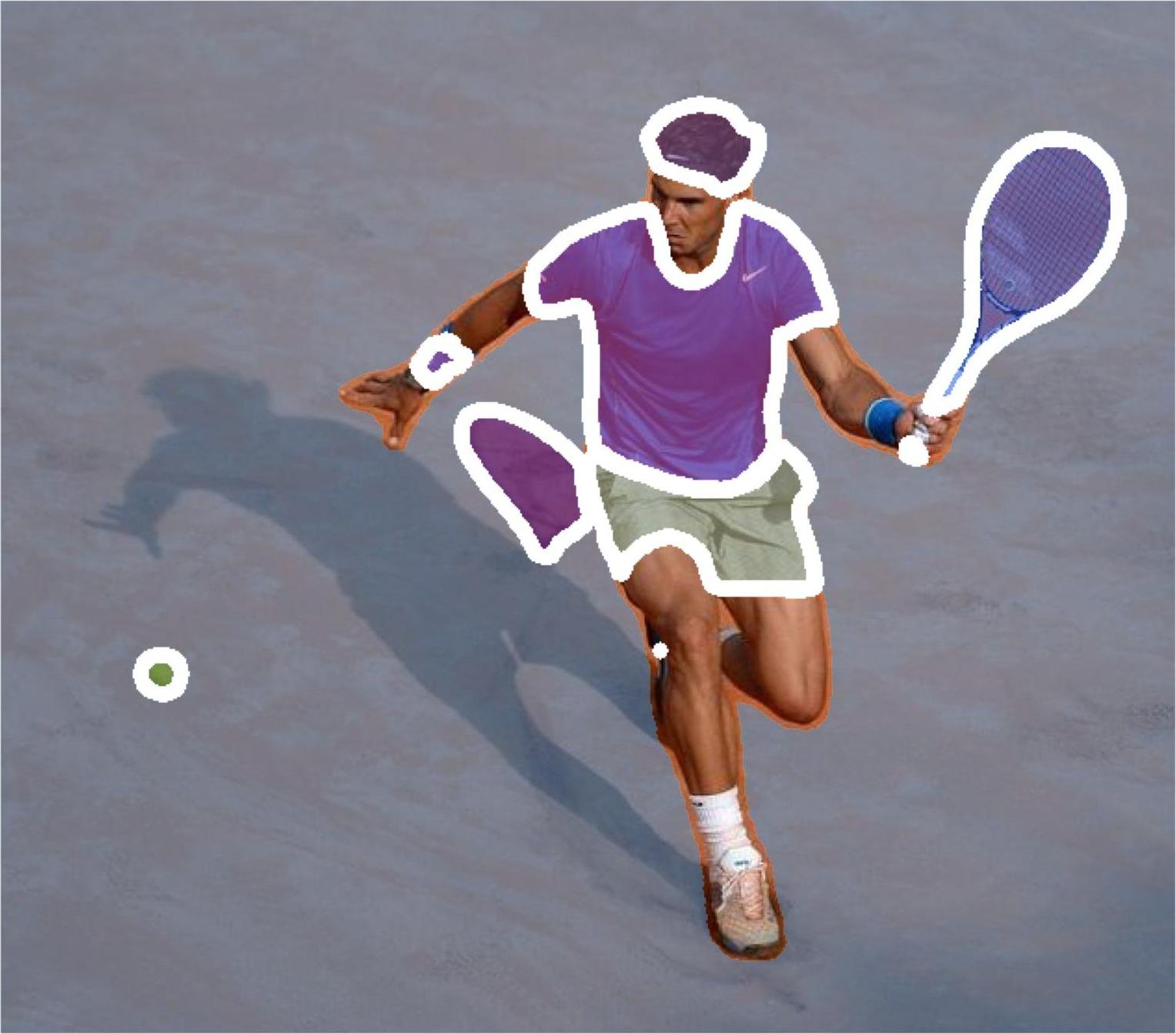} &
        \includegraphics[height=2cm]{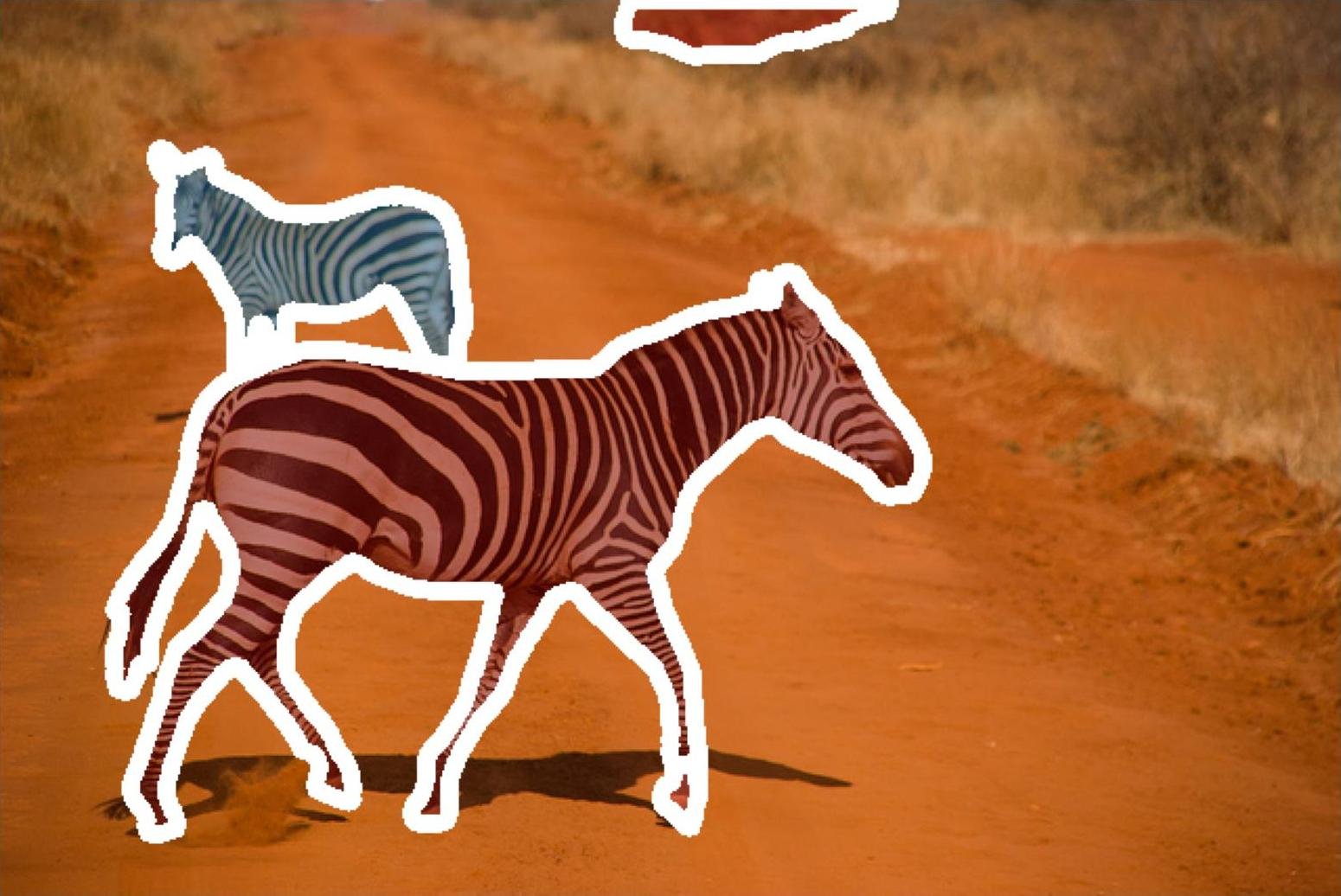} &
        \includegraphics[height=2cm]{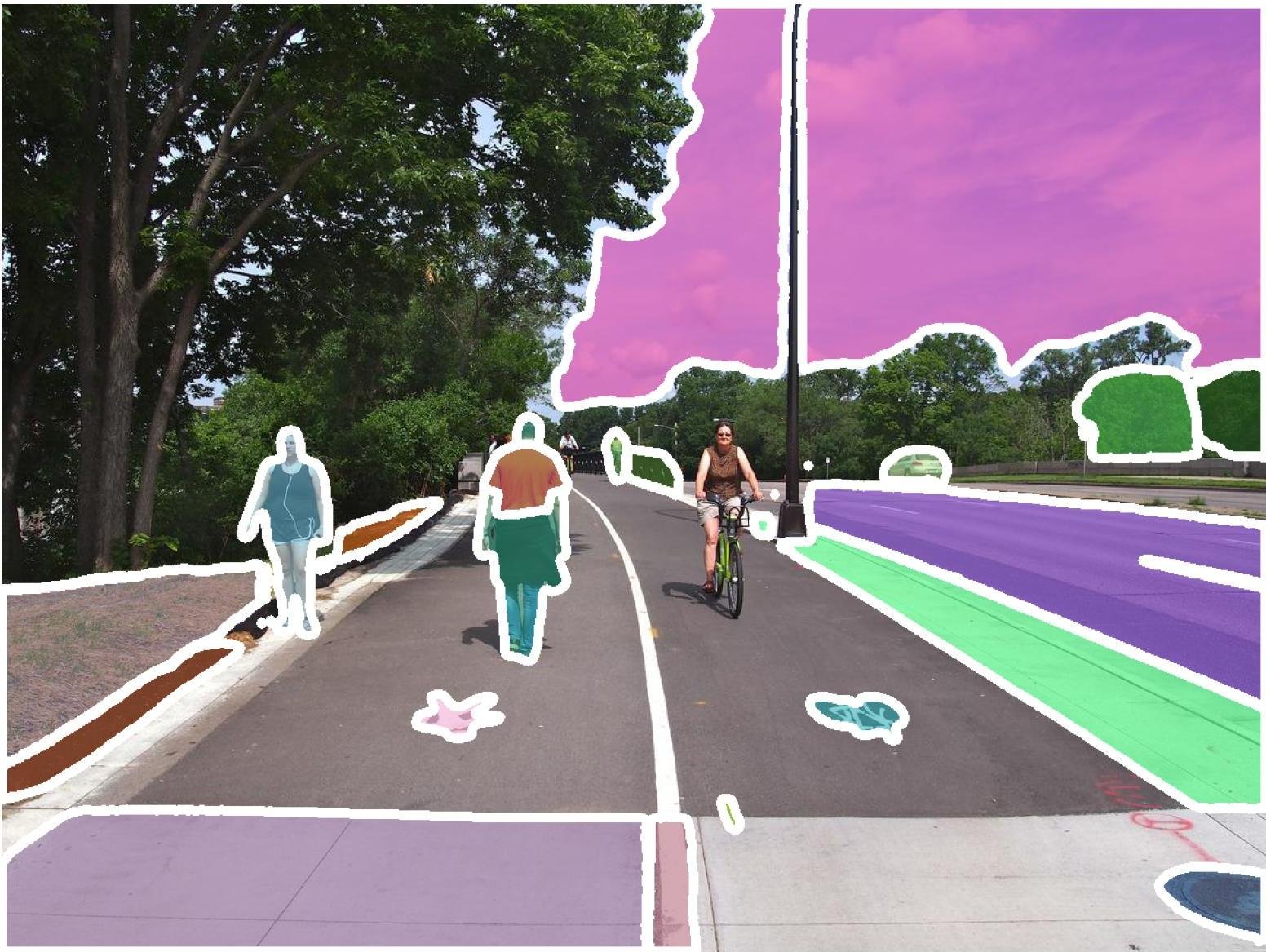} &
        \includegraphics[height=2cm]{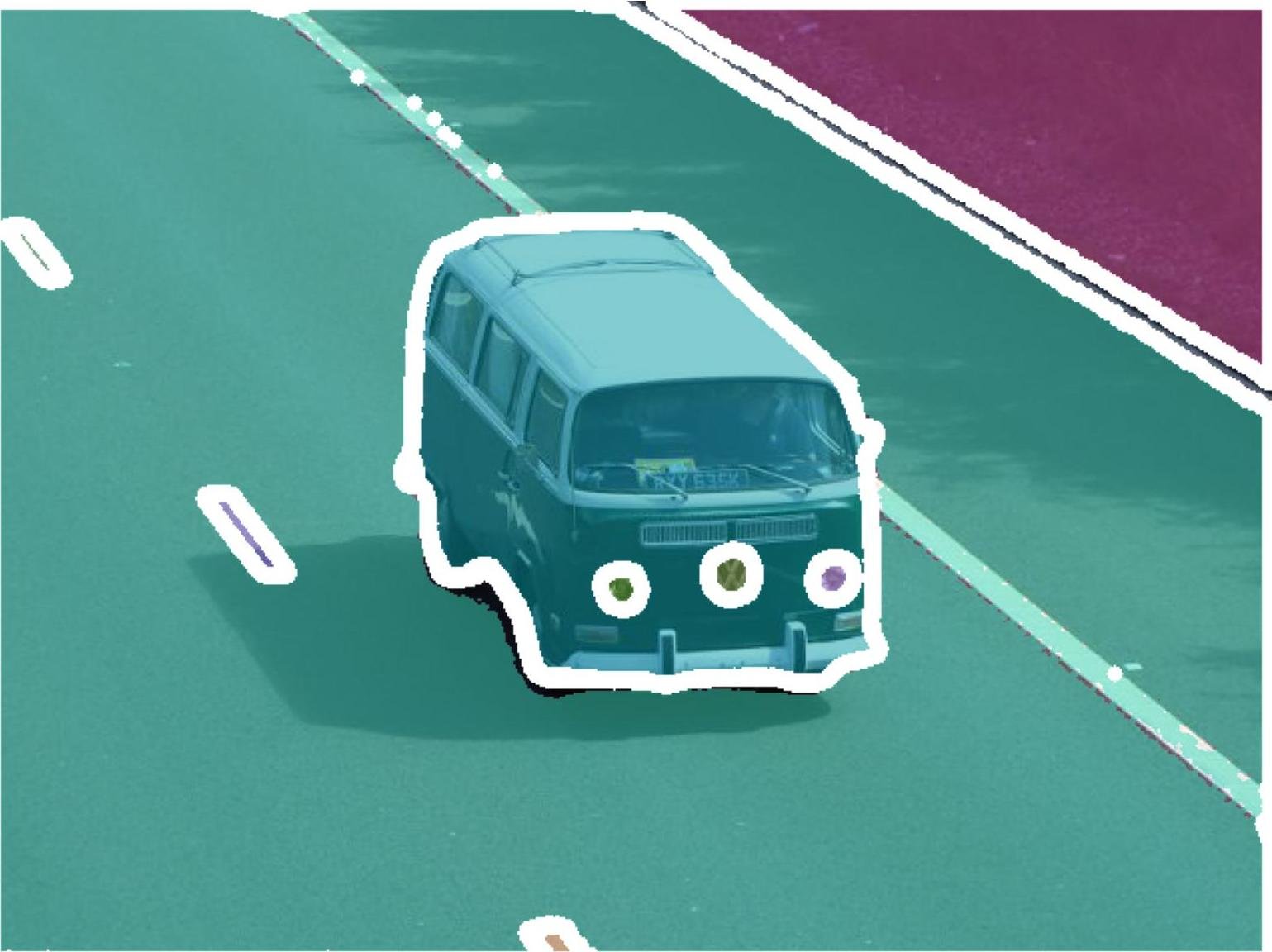} &
        \includegraphics[height=2cm]{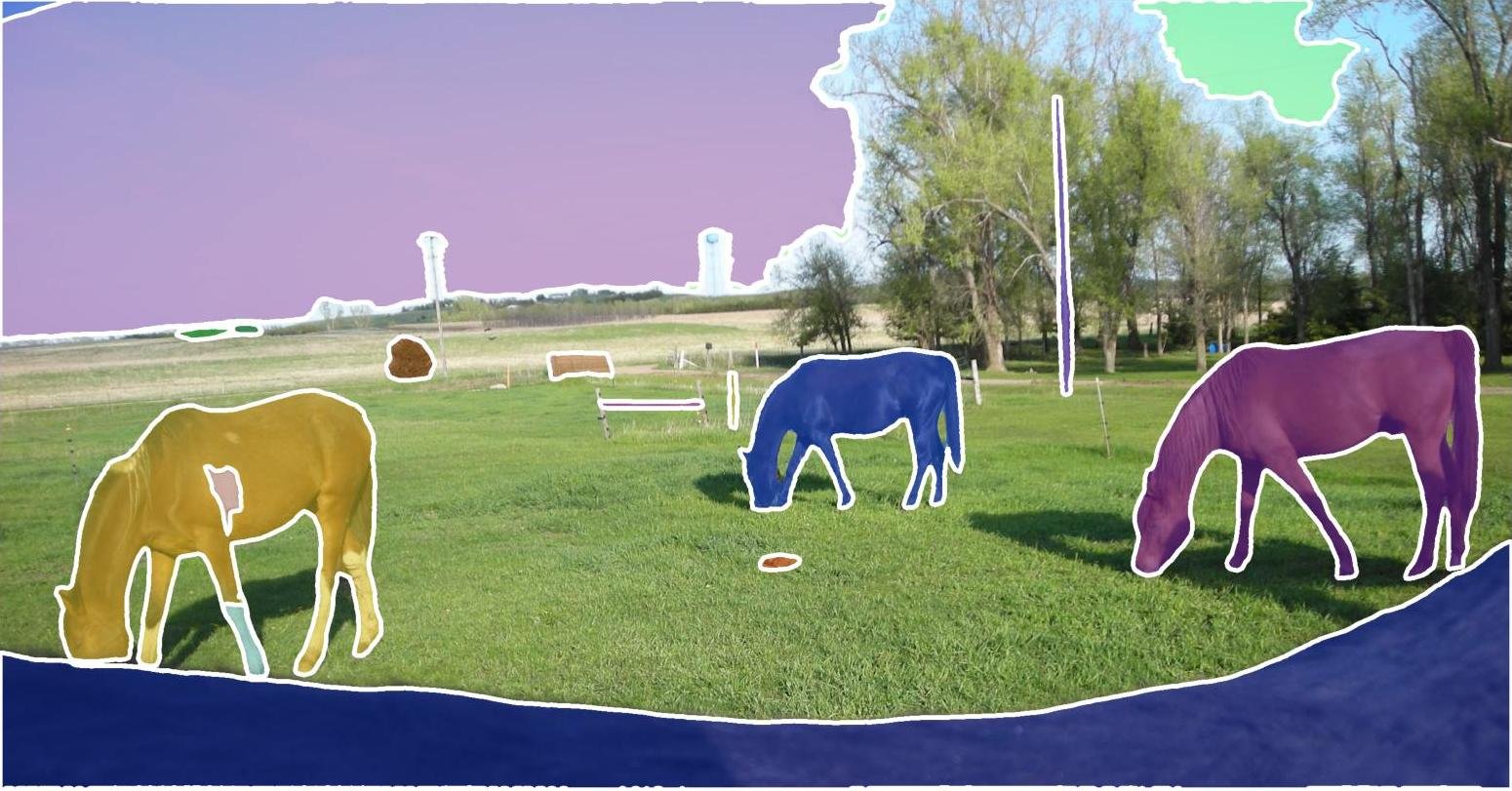} &
        \includegraphics[height=2cm]{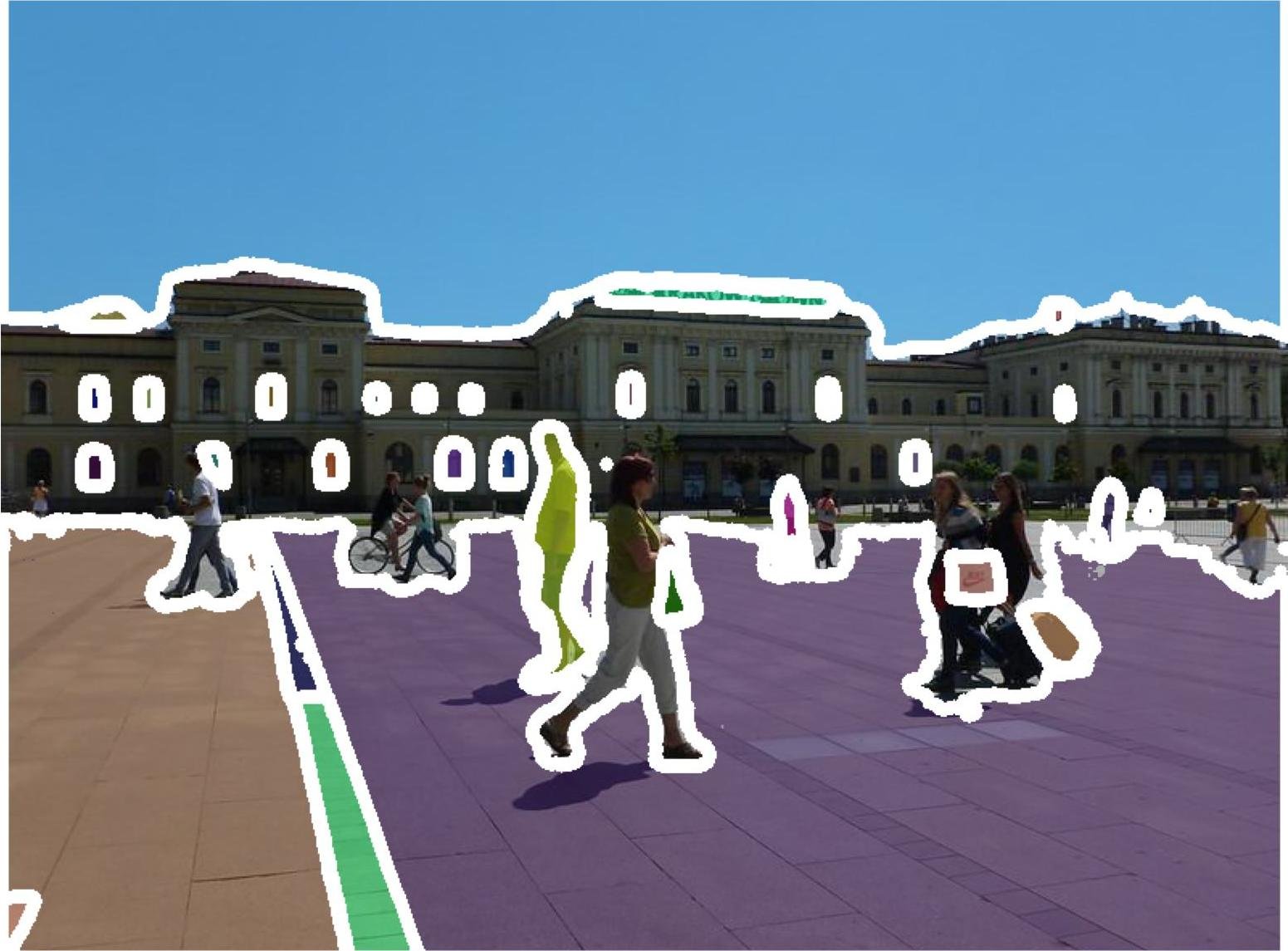}\\ 
        
        \raisebox{0.1cm}{\makebox[0pt][c]{\rotatebox{90}{\small \textbf{SAM2-Bbox}}}} & 
        \includegraphics[height=2cm]{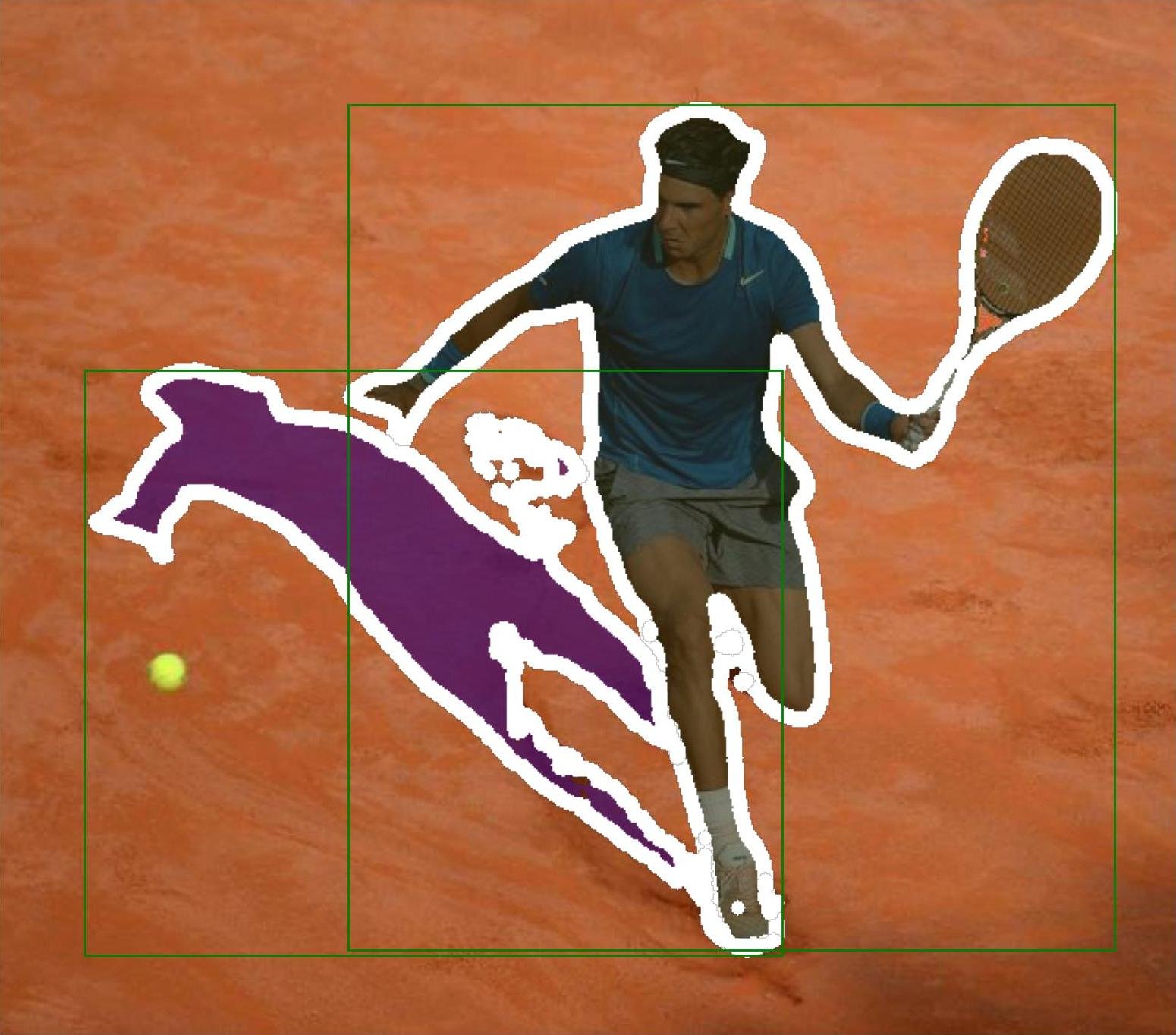} &
        \includegraphics[height=2cm]{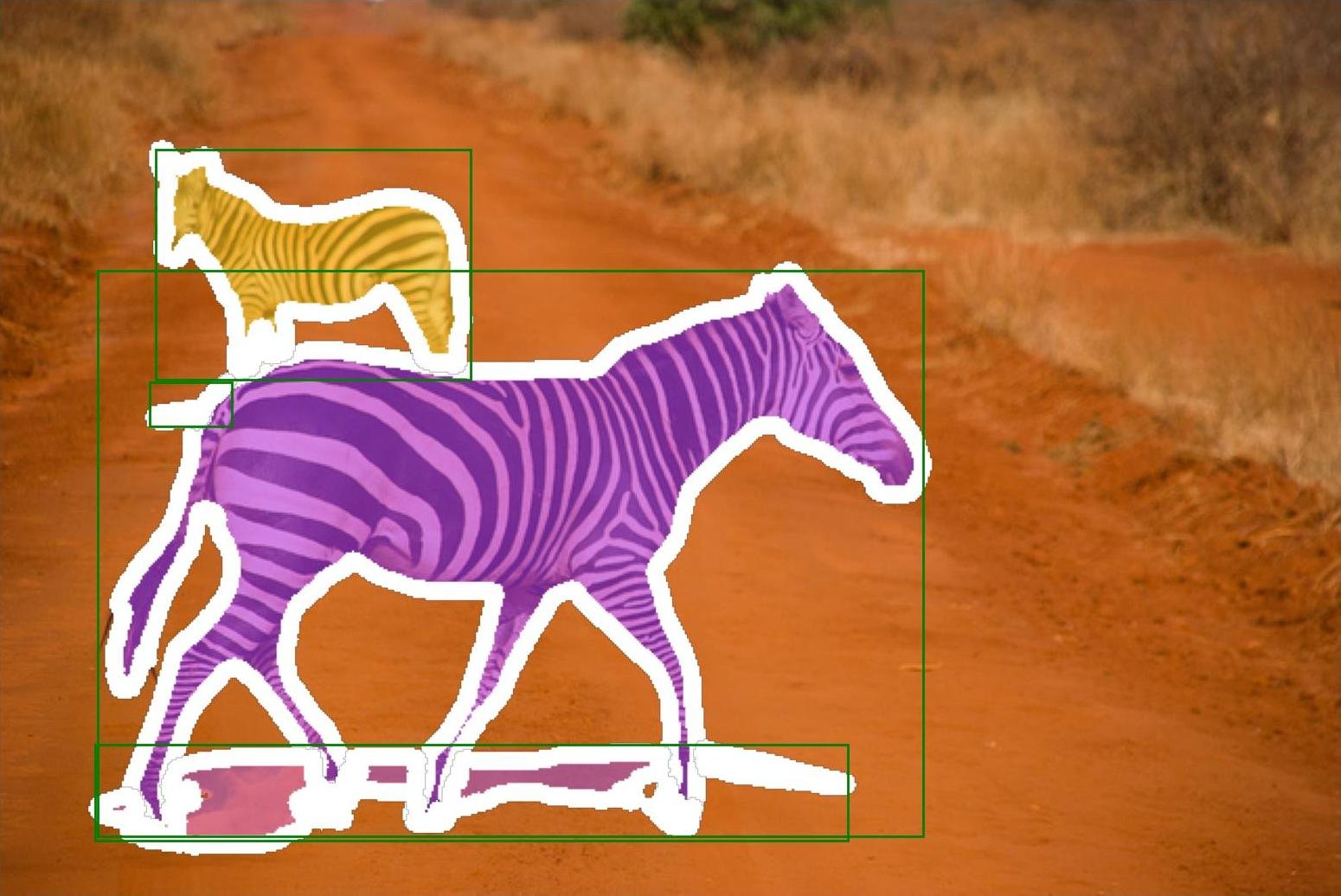} &
        \includegraphics[height=2cm]{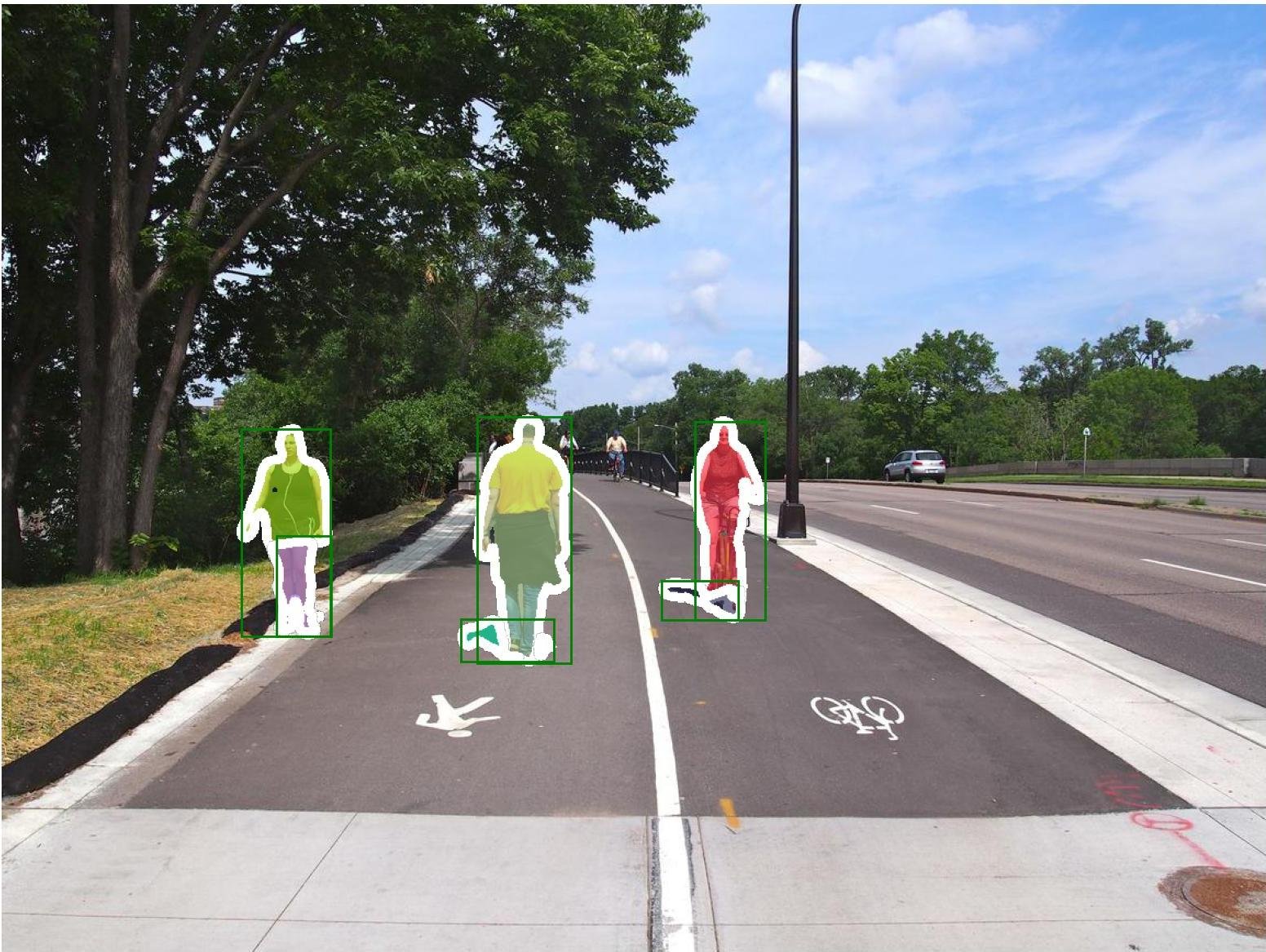} &
        \includegraphics[height=2cm]{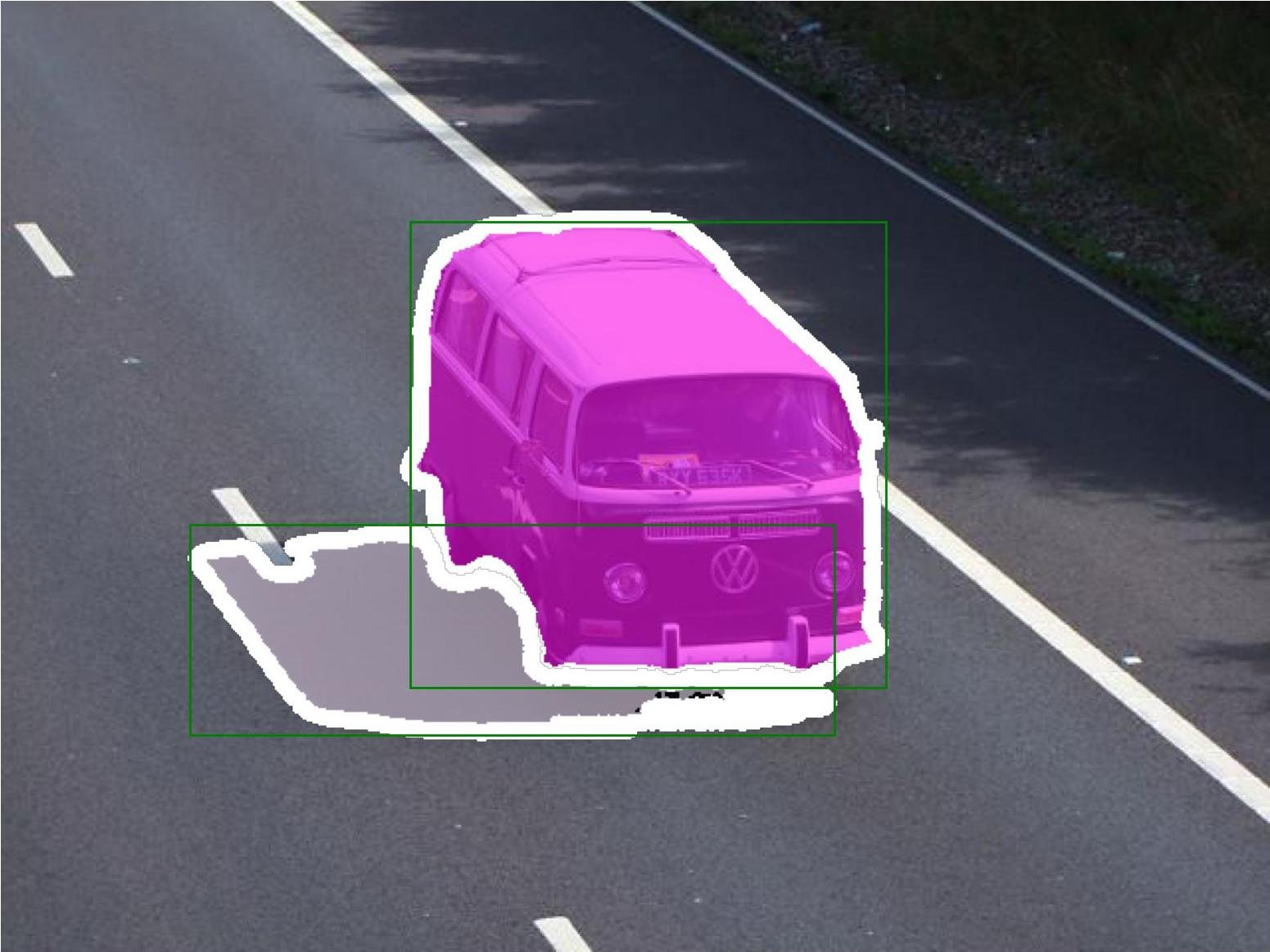} &
        \includegraphics[height=2cm]{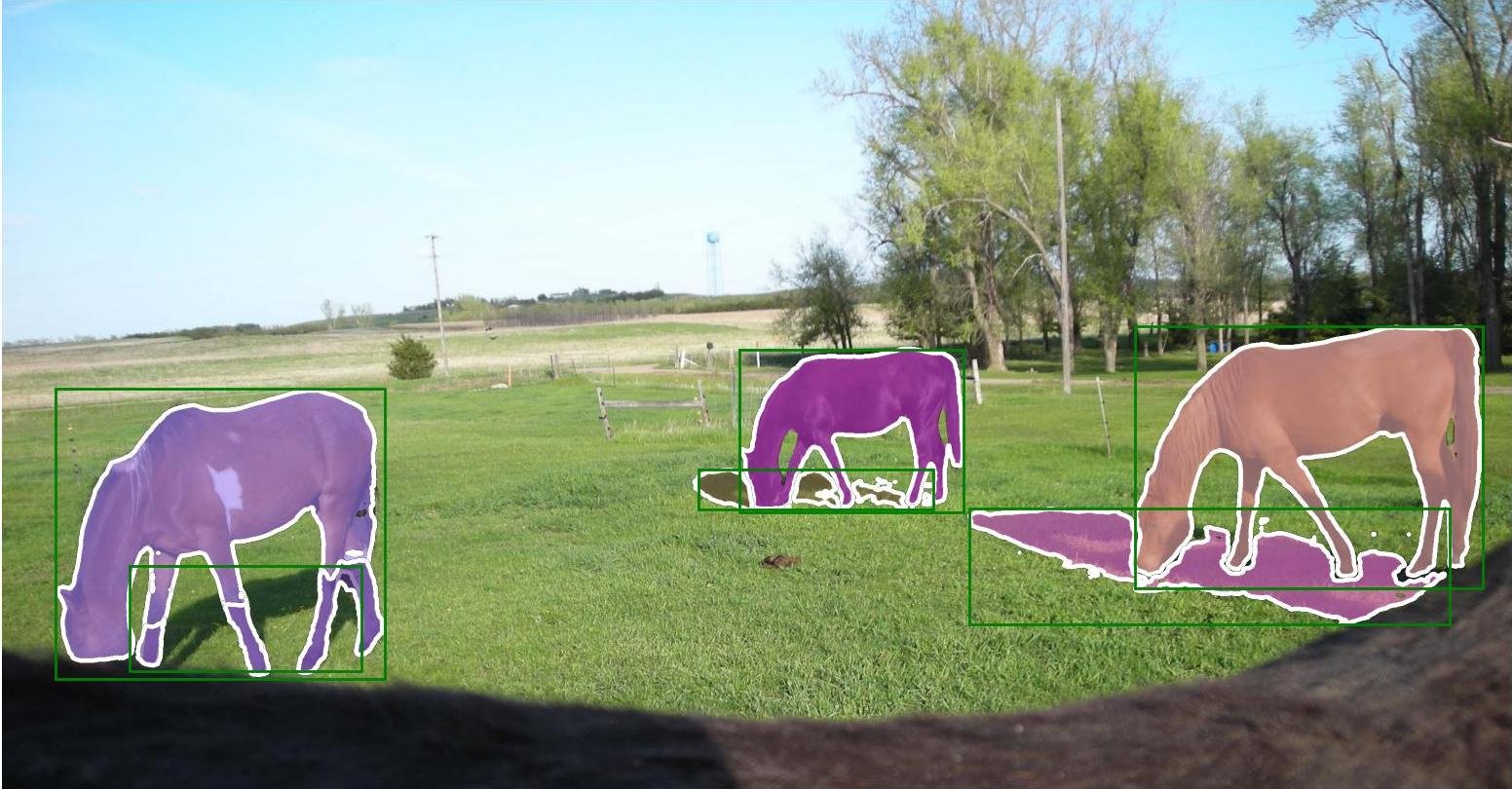} &
        \includegraphics[height=2cm]{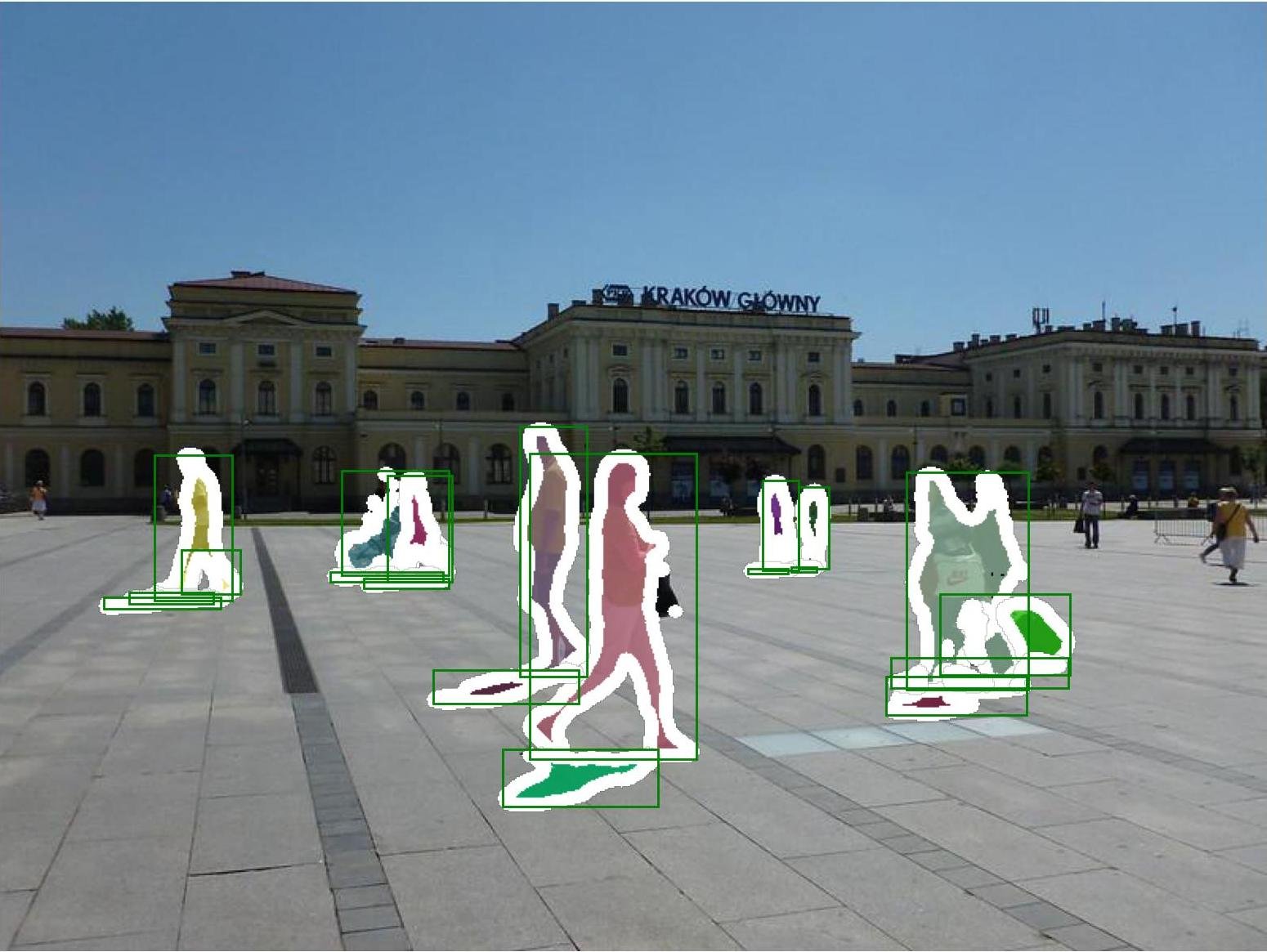}\\ 
    \end{tabular}
    \caption{Qualitative comparisons on SOBA-challenge and SOBA-test datasets for shadow instance detection.}
    \label{fig:shadow}
\end{figure*}



\begin{table*}[t!]
\begin{center}

\footnotesize
\renewcommand{\arraystretch}{1.2}
\renewcommand{\tabcolsep}{0.5mm}
\caption{\textbf{Quantitative comparisons of SAM2, SAM, and DIS methods on DIS5K, including DIS-VD, DIS-TE1, and DIS-TE2.} SAM-B, SAM-L, and SAM-H represent ViT-B, ViT-L, and ViT-H model types of SAM, respectively. SAM2-T, SAM2-B+, and SAM2-L represent Hiera-Tiny, Hiera-Base+, and Hiera-Large model types of SAM2, respectively. The best and second-best results are bolded and underlined respectively.}

\label{table:DIS1}

\begin{tabular}{r||r|cccccc|cccccc|cccccc}
\toprule
\multirow{2}{*}{Methods} & \multirow{2}{*}{Settings} 
& \multicolumn{6}{c|}{DIS-VD}   
& \multicolumn{6}{c|}{DIS-TE1} 
& \multicolumn{6}{c}{DIS-TE2} 
\\ 
& & $F_\beta^\text{max} \uparrow$ & $F_\beta^\omega \uparrow$ & $M \downarrow$ & $S_\alpha \uparrow$ & $E_{\phi}^\text{m} \uparrow$ & $H_{\gamma} \downarrow$ 
& $F_\beta^\text{max} \uparrow$ & $F_\beta^\omega \uparrow$ & $M \downarrow$ & $S_\alpha \uparrow$ & $E_{\phi}^\text{m} \uparrow$ & $H_{\gamma} \downarrow$ 
& $F_\beta^\text{max} \uparrow$ & $F_\beta^\omega \uparrow$ & $M \downarrow$ & $S_\alpha \uparrow$ & $E_{\phi}^\text{m} \uparrow$ & $H_{\gamma} \downarrow$ 
\\ \hline

IS-Net~\cite{qin2022highly}& Fully-supervised & .791 & .717 & .074 & .813 & .856 & 1116 & .74 & .662 & .074 & .787 & .820 & 149 & .799 & .728 & .070 & .823 & .858 & 340 \\

PF-DIS-R50~\cite{zhou2023dichotomous}&Fully-supervised & .823 & .763 & .062 & .843 & .891 & 1309 & .784 & .713 & .060 & .821 & .860 & 160 & .827 & .767 & .059 & .845 & .893 & 373 \\

UDUN-R50~\cite{pei2023unite}   &Fully-supervised & .823 & .763 & .059 & .838 & .892 & \underline{1097} & .784 & .720 & .059 & .817 & .864 & \underline{140} & .829 & .768 & .058 & .843 & .886 & \underline{325} \\
BiRefNet-S-B~\cite{zheng2024bilateral}  & Fully-supervised& \underline{.881} & \underline{.844} & \underline{.039} & \underline{.890} & \underline{.925} & \textbf{1029} & \underline{.857} & \underline{.819} & \underline{.038} & \underline{.884} & \textbf{.912} & \textbf{110} & \underline{.890} & \underline{.854} & \underline{.037} & \underline{.898} & \underline{.930} & \textbf{275} \\

MVANet-S-B~\cite{yu2024multi}  & Fully-supervised& \textbf{.914} & \textbf{.856} & \textbf{.036} & \textbf{.905} & \textbf{.938} &--&\textbf{.893} & \textbf{.823} & \textbf{.037} & \textbf{.879} & \underline{.911} &--&\textbf{.925} & \textbf{.874} & \textbf{.030} & \textbf{.915} & \textbf{.944}&- \\

\hline

SAM-B~\cite{kirillov2023segment} & Automatic & .215 & .132 & .258 & .398 & .392 & 1445 & .235 & .176 & .223 & .439 & .442 & 209 & .210 & .126 & .268 & .388 & .369 & 450 \\
SAM-L~\cite{kirillov2023segment} & Automatic  & .278 & .231 & .325 & .401 & .462 & 1402 & .365 & .311 & .268 & .481 & .531 & 224 & .286 & .227 & .331 & .397 & .441 & 464 \\
SAM-H~\cite{kirillov2023segment} & Automatic & .283 & .241 & .344 & .395 & .475 & 1417 & .402 & .352 & .261 & .505 & .555 & 223 & .283 & .228 & .349 & .386 & .449 & 471 \\
SAM-B~\cite{kirillov2023segment} & GT-Bbox & .671 & .623 & .150 & .681 & .774 & 1544 & .747 & .703 & .105 & .754 & .829 & 286 & .687 & .635 & .143 & .692 & .784 & 590 \\
SAM-L~\cite{kirillov2023segment}&  GT-Bbox & .739 & .698 & .117 & .739 & .817 & 1460 & .783 & .746 & .091 & .787 & .852 & 255 & .766 & .718 & .107 & .756 & .831 & 551 \\
SAM-H~\cite{kirillov2023segment}&  GT-Bbox  & .687 & .652 & .151 & .700 & .783 & 1468 & .755 & .721 & .106 & .766 & .833 & 244 & .708 & .666 & .141 & .713 & .791 & 543 \\
\hline
\rowcolor[RGB]{216,241,244}
\textbf{SAM2-T}~\cite{ravi2024sam} & Automatic & .306 & .209 & .169 & .471 & .407 & 1417 & .352 & .253 & .142 & .506 & .450 & 189 & .311 & .204 & .168 & .468 & .394 & 443 \\
\rowcolor[RGB]{216,241,244}
\textbf{SAM2-B+}~\cite{ravi2024sam} & Automatic & .428 & .311 & .156 & .515 & .477 & 1382 & .498 & .381 & .117 & .566 & .539 & 195 & .427 & .295 & .155 & .509 & .448 & 444 \\
\rowcolor[RGB]{216,241,244}
\textbf{SAM2-L}~\cite{ravi2024sam} & Automatic & .420 & .307 & .157 & .514 & .478 & 1385 & .494 & .382 & .117 & .570 & .550 & 196 & .442 & .310 & .147 & .518 & .464 & 444 \\
\hline
\rowcolor[RGB]{216,241,244}
\textbf{SAM2-T}~\cite{ravi2024sam} & GT-Bbox & .739 & .702 & .107 & .748 & .830 & 1646 & .791 & .756 & .0795 & .798 & .863 & 346 & .752 & .708 & .096 & .760 & .838 & 698 \\
\rowcolor[RGB]{216,241,244}
\textbf{SAM2-B+}~\cite{ravi2024sam} & GT-Bbox & .765 & .731 & .104 & .766 & .840 & 1560 & .834 & .805 & .069 & .829 & .888 & 313 & .775 & .734 & .102 & .770 & .842 & 642 \\
\rowcolor[RGB]{216,241,244}
\textbf{SAM2-L}~\cite{ravi2024sam} & GT-Bbox & .743 & .707 & .107 & .752 & .819 & 1533 & .828 & .796 & .0676 & .824 & .879 & 305 & .748 & .702 & .103 & .750 & .814 & 625 \\
\bottomrule
\end{tabular}
\end{center}
\end{table*}


\begin{table*}[t!]
\begin{center}

\footnotesize
\renewcommand{\arraystretch}{1.2}
\renewcommand{\tabcolsep}{0.5mm}
\caption{\textbf{Continued \tabref{table:DIS1}. Detailed comparisons of SAM2, SAM, and DIS methods on DIS5K test sets, including DIS-TE3, DIS-TE4 and overall DIS-TE.}}
\label{table:DIS2}

\begin{tabular}{r||r|cccccc|cccccc|cccccc}
\toprule
\multirow{2}{*}{Methods} & \multirow{2}{*}{Settings} 
& \multicolumn{6}{c|}{DIS-TE3}   
& \multicolumn{6}{c|}{DIS-TE4} 
& \multicolumn{6}{c}{Overall DIS-TE (1-4)} 
\\
& & $F_\beta^\text{max} \uparrow$ & $F_\beta^\omega \uparrow$ & $M \downarrow$ & $S_\alpha \uparrow$ & $E_{\phi}^\text{m} \uparrow$ & $H_{\gamma} \downarrow$ 
& $F_\beta^\text{max} \uparrow$ & $F_\beta^\omega \uparrow$ & $M \downarrow$ & $S_\alpha \uparrow$ & $E_{\phi}^\text{m} \uparrow$ & $H_{\gamma} \downarrow$ 
& $F_\beta^\text{max} \uparrow$ & $F_\beta^\omega \uparrow$ & $M \downarrow$ & $S_\alpha \uparrow$ & $E_{\phi}^\text{m} \uparrow$ & $H_{\gamma} \downarrow$ 
\\ \hline
IS-Net~\cite{qin2022highly} & Fully-supervised& .830 & .758 & .064 & .836 & .883 & 687 & .827 & .753 & .072 & .830 & .870 & 2888 & .799 & .726 & .070 & .819 & .858 & 1016 \\
PF-DIS-R50~\cite{zhou2023dichotomous}&Fully-supervised & .868 & .811 & .049 & .871 & .922 & 780 & .846 & .788 & .061 & .852 & .906 & 3347 & .831 & .770 & .047 & .847 & .895 & 1165 \\
UDUN-R50~\cite{pei2023unite}   &Fully-supervised& .865 & .809 & .050 & .865 & .917 & \underline{658} & .846 & .792 & .059 & .849 & .901 & \textbf{2785} & .831 & .772 & .057 & .844 & .892 & \underline{977} \\
BiRefNet-S-B~\cite{zheng2024bilateral}    &Fully-supervised& \underline{.919} & \underline{.886} & \textbf{.030} & \underline{.915} & \underline{.953} & \textbf{597} & \underline{.899} & \textbf{.860} & \textbf{.040} & \underline{.895} & \underline{.938} & \underline{2836} & \underline{.891} & \underline{.855} & \underline{.036} & \underline{.898} & \underline{.933} & \textbf{954} \\
MVANet-S-B~\cite{yu2024multi}  &Fully-supervised&\textbf{.936} & \textbf{.890} & \underline{.031} & \textbf{.920} & \textbf{.954} &--&\textbf{.911} & \underline{.857} & \underline{.041} & \textbf{.903} & \textbf{.944} &--&\textbf{.916} & \textbf{.861} & \textbf{.035} & \textbf{.904} & \textbf{.938}&- \\
\hline
SAM-B~\cite{kirillov2023segment}& Automatic & .220 & .120 & .270 & .386 & .373 & 890 & .233 & .118 & .298 & .366 & .395 & 3624 & .224 & .135 & .265 & .395 & .395 & 1293 \\
SAM-L~\cite{kirillov2023segment}& Automatic & .220 & .171 & .345 & .362 & .443 & 905 & .254 & .213 & .345 & .379 & .467 & 3528 & .281 & .230 & .322 & .404 & .471 & 1280 \\
SAM-H~\cite{kirillov2023segment}& Automatic & .235 & .190 & .351 & .368 & .453 & 905 & .272 & .233 & .337 & .394 & .491 & 3502 & .298 & .251 & .325 & .413 & .487 & 1275 \\
\hline
SAM-B~\cite{kirillov2023segment} & GT-Bbox & .624 & .573 & .171 & .647 & .745 & 1080 & .558 & .520 & .224 & .588 & .699 & 3667 & .654 & .608 & .161 & .670 & .764 & 1405 \\
SAM-L~\cite{kirillov2023segment}& GT-Bbox & .687 & .634 & .143 & .696 & .778 & 1021 & .613 & .576 & .191 & .639 & .734 & 3533 & .712 & .668 & .133 & .720 & .799 & 1340 \\
SAM-H~\cite{kirillov2023segment}& GT-Bbox & .629 & .583 & .176 & .654 & .748 & 997  & .576 & .545 & .218 & .611 & .707 & 3553 & .486 & .439 & .235 & .552 & .630 & 1325 \\
\hline

\rowcolor[RGB]{216,241,244}
\textbf{SAM2-T}~\cite{ravi2024sam} & Automatic & .308 & .203 & .169 & .470 & .391 & 877 & .268 & .179 & .192 & .445 & .382 & 3613 & .310 & .210 & .168 & .472 & .404 & 1280 \\
\rowcolor[RGB]{216,241,244}
\textbf{SAM2-B+}~\cite{ravi2024sam} & Automatic & .391 & .265 & .159 & .494 & .437 & 880 & .381 & .277 & .179 & .488 & .465 & 3509 & .424 & .305 & .153 & .514 & .472 & 1257 \\
\rowcolor[RGB]{216,241,244}
\textbf{SAM2-L}~\cite{ravi2024sam} & Automatic & .390 & .266 & .157 & .497 & .437 & 877 & .385 & .279 & .177 & .491 & .464 & 3521 & .428 & .309 & .150 & .519 & .479 & 1259 \\
\hline
\rowcolor[RGB]{216,241,244}
\textbf{SAM2-T}~\cite{ravi2024sam} & GT-Bbox & .698 & .653 & .126 & .715 & .807 & 1203 & .622 & .587 & .179 & .652 & .748 & 3766 & .716 & .676 & .120 & .731 & .814 & 1503 \\
\rowcolor[RGB]{216,241,244}
\textbf{SAM2-B+}~\cite{ravi2024sam} & GT-Bbox & .714 & .671 & .135 & .719 & .806 & 1169 & .633 & .601 & .188 & .657 & .741 & 3677 & .739 & .703 & .124 & .744 & .819 & 1450 \\
\rowcolor[RGB]{216,241,244}
\textbf{SAM2-L}~\cite{ravi2024sam} & GT-Bbox & .678 & .630 & .139 & .698 & .765 & 1127 & .603 & .569 & .187 & .639 & .719 & 3679 & .714 & .674 & .124 & .728 & .794 & 1434 \\
\bottomrule
\end{tabular}
\end{center}
\end{table*}

\begin{figure*}[!t]
    \centering
    \renewcommand{\arraystretch}{0.5}
    \begin{tabular}{@{}c@{\hskip 5pt}c@{}}
        \raisebox{0.5cm}{\makebox[0pt][c]{\rotatebox{90}{\small Image}}} & 
        \includegraphics[height=2cm,width=15cm]{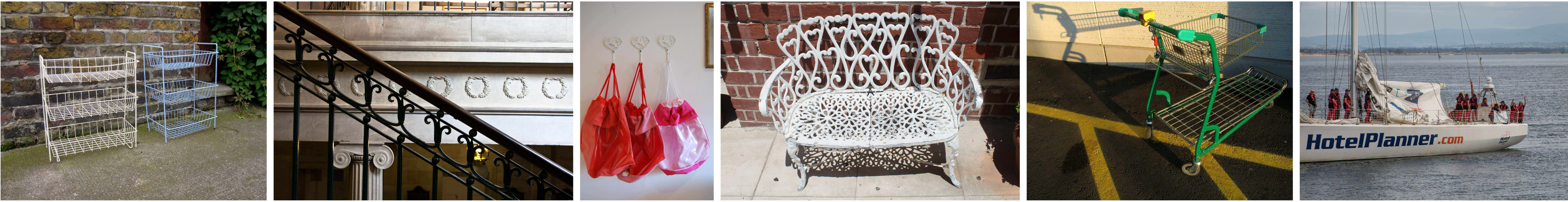} \\ [1pt]
        \raisebox{0cm}{\makebox[0pt][c]{\rotatebox{90}{\small Ground Truth}}} & 
        \includegraphics[height=2cm,width=15cm]{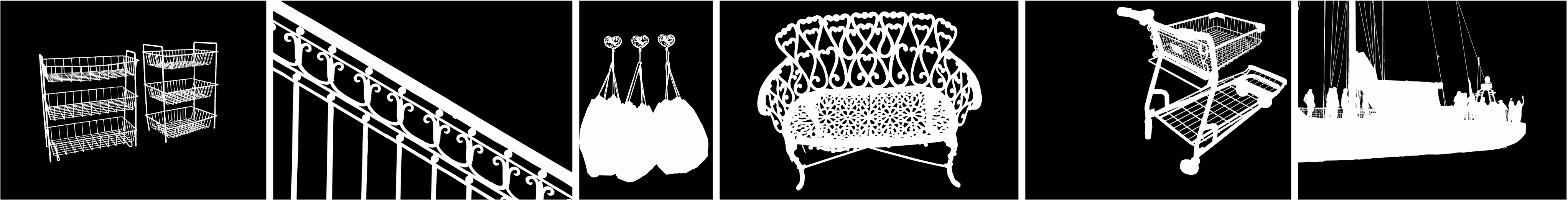} \\ [1pt]   
        \raisebox{0.5cm}{\makebox[0pt][c]{\rotatebox{90}{\small IS-Net}}} & 
        \includegraphics[height=2cm,width=15cm]{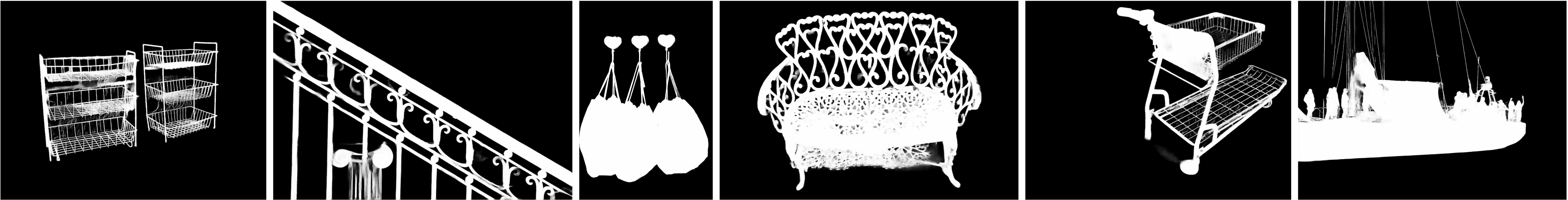} \\ [1pt]   
        \raisebox{0.5cm}{\makebox[0pt][c]{\rotatebox{90}{\small UDUN}}} & 
        \includegraphics[height=2cm,width=15cm]{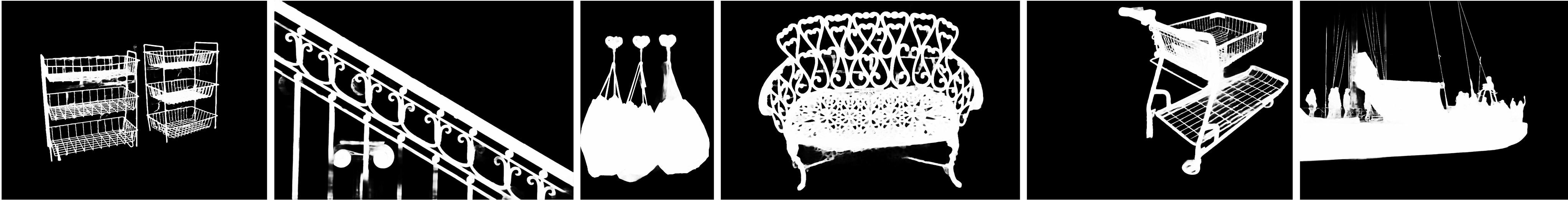} \\ [1pt]   
        \raisebox{0.4cm}{\makebox[0pt][c]{\rotatebox{90}{\small MVANet}}} & 
        \includegraphics[height=2cm,width=15cm]{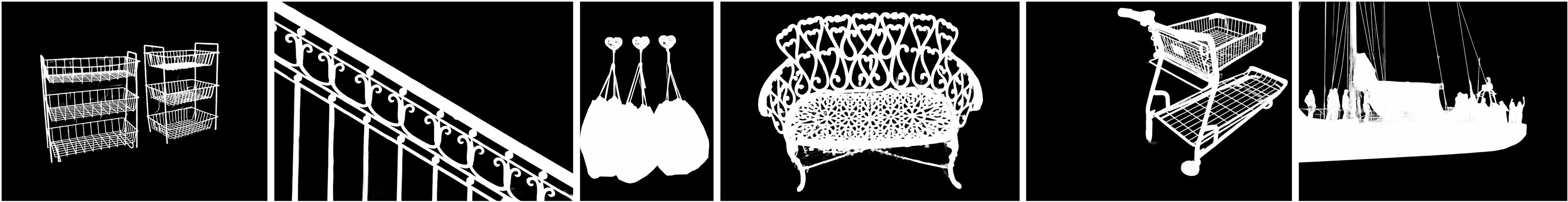} \\ [1pt]   
        \raisebox{0.2cm}{\makebox[0pt][c]{\rotatebox{90}{\small SAM-Auto}}} & 
        \includegraphics[height=2cm,width=15cm]{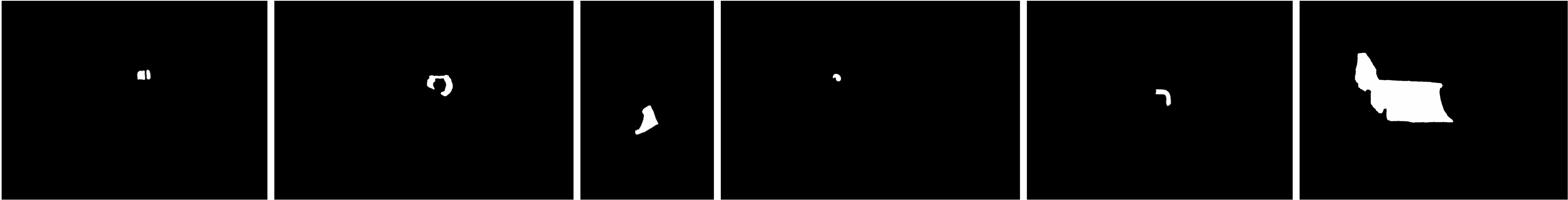} \\ [1pt]   
        \raisebox{0.2cm}{\makebox[0pt][c]{\rotatebox{90}{\small SAM-Bbox}}} & 
        \includegraphics[height=2cm,width=15cm]{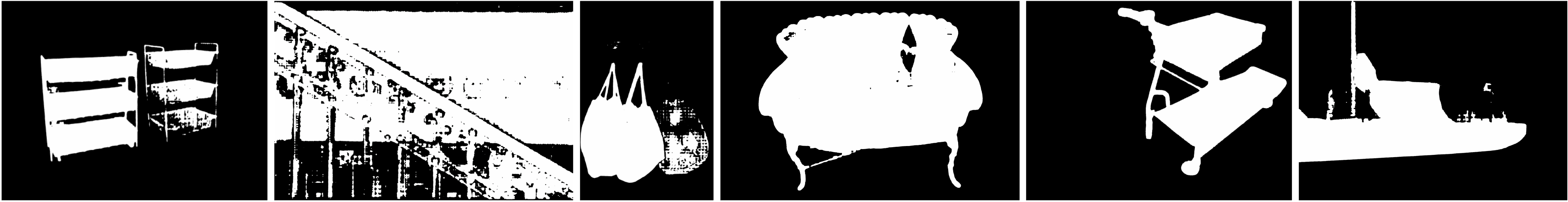} \\ [1pt]   
        \raisebox{0.1cm}{\makebox[0pt][c]{\rotatebox{90}{\small \textbf{SAM2-Auto}}}} & 
        \includegraphics[height=2cm,width=15cm]{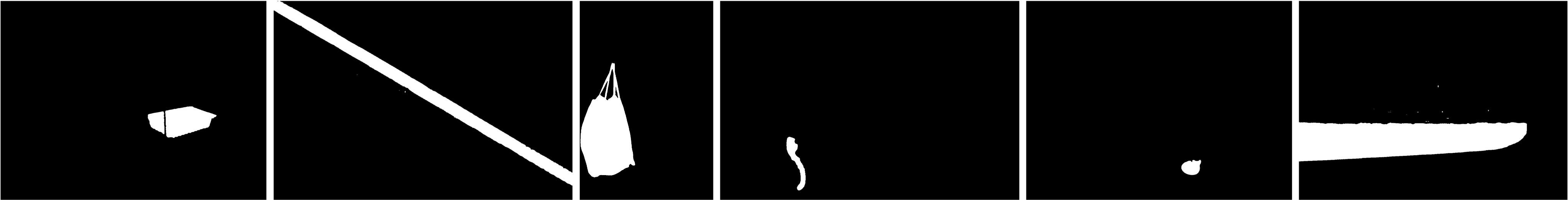} \\ [1pt]   
        \raisebox{0.1cm}{\makebox[0pt][c]{\rotatebox{90}{\small \textbf{SAM2-Bbox}}}} & 
        \includegraphics[height=2cm,width=15cm]{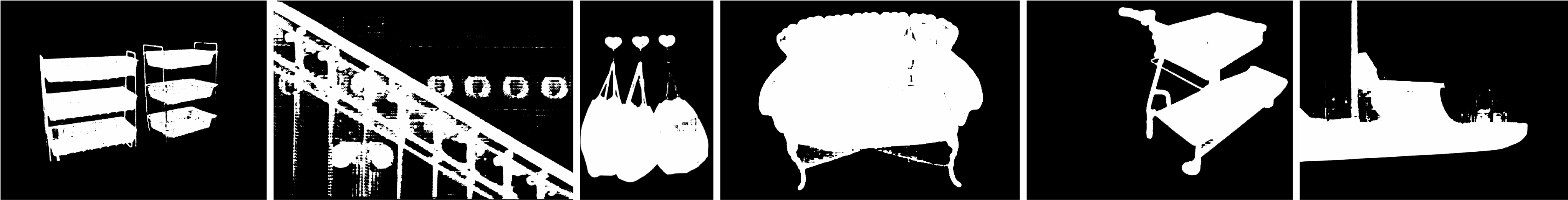} \\    
    \end{tabular}
    \caption{Qualitative comparisons on DIS5K dataset for dichotomous image segmentation.}
    \label{fig:DIS}
\end{figure*}

\subsubsection{Shadow Instance Segmentation}

\noindent\textbf{Quantitative Results.}
It is important to note that in shadow instance detection tasks, the matching degree between the shadow and the object needs to be measured. However, SAM2 lacks this functionality, so our comparison does not involve measuring this aspect. 
In our experiments, we treat instances and shadows as separate entities, rather than pairs of instances and corresponding shadows.
Based on the~\tabref{table:shadow}, SAM2 models perform exceptionally well in the GT-Bbox setting, with SAM2-T achieving AP scores of 51.9 on the SOBA-challenge and 58.9 on the SOBA-test, surpassing all SAM models and task-specific approaches. 
However, an interesting phenomenon is observed: using different backbones with SAM2 does not lead to significant performance differences. 
In fact, SAM2 with larger backbones has the potential to decreases segmentation performance, and the same phenomenon exists for SAM models. 
Switching to automatic mode results in a significant drop in performance for both SAM2 and SAM, but the changes in backbone parameter size do not greatly impact segmentation results. 
Therefore, to improve the performance of SAM2 on the SID task, it is not appropriate to simply increase the depth and parameter size of the model.

\noindent\textbf{Qualitative Results.}
As shown in~\figref{fig:shadow}, both SAM and SAM2 are effective in segmenting instances in automatic mode, but face challenges in accurately identifying shadows. 
This may be caused by the lack of instance-shadow IoU matching operations in SAM2.
When provided with box prompts, both models show significant improvement in shadow segmentation, with SAM2 having a slight edge in capturing shadow details. 
Despite these improvements, the overall quality of shadow segmentation by SAM2 still falls short compared to corresponding instances.

\subsubsection{Dichotomous Image Segmentation}
Dichotomous image segmentation focuses on identifying class-agnostic foreground objects in natural scenes. In automatic prediction mode, SAM generates multiple binary masks for each sample. To select the most suitable foreground mask, we use a maximum Intersection over Union (IoU) strategy, choosing the mask with the highest IoU score.
\noindent\textbf{Quantitative Results.}
\tabref{table:DIS1} and \tabref{table:DIS2} present the qualitative comparison results of SAM and SAM2 against task-specific methods. In the automatic setting, SAM2 models, particularly SAM2-T, show marked improvement over SAM models. 
Concretely, SAM2-T achieves an \(F_{\beta}^\text{max}\) of 0.306 on DIS-VD, compared to 0.215 for SAM-B. SAM2-B+ and SAM2-L further enhance performance, with SAM2-B+ reaching 0.428 in \(F_{\beta}^\text{max}\) on DIS-VD, surpassing all SAM variants. 
These improvements indicate better segmentation quality and alignment, as evidenced by higher \(S_\alpha\) and \(E_m\) values. In the GT-Bbox setting, SAM2 models demonstrate significant gains across all metrics, except for HCE. For instance, SAM2-B+ achieves \(F_{\beta}^\text{max}\) of 0.765 on DIS-VD, surpassing SAM-L's 0.739, although the HCE metric increases by 100 points. 
Generally, the HCE metric is more sensitive to the structural refinement of the segmentation map compared to traditional accuracy metrics like weighted F-measure, mean absolute error, and mean enhanced alignment measure. 
It indicates that SAM2 enhances the overall perception of the target, but it still struggles to identify the dominant area while accurately segmenting detailed object structures.
Overall, SAM2 models offer substantial improvements over SAM across the vast majority of metrics in all datasets, nearly approaching the performance of the fully supervised method IS-Net. 
However, the HCE scores illustrate that both SAM and SAM2 have limited potential in representing detailed structures.


\noindent\textbf{Qualitative Results.} 
Further analysis of the qualitative results, particularly in~\figref{fig:DIS}, reveals that both SAM and SAM2 encounter difficulties in identifying foreground objects, whether in automatic mode or with box prompts. 
Notably, when receiving a bounding box prompt, SAM can roughly outline the main body of objects, while SAM2 enhances this capability by improving segmentation completeness. 
For example, SAM can segment the body of a ship with the aid of a bounding box prompt (see the seventh column of~\figref{fig:DIS}), but it tends to miss smaller details such as the mast and thin lines. 
Thus, although SAM2 improves the accuracy for locating foreground objects in natural scenes, it still falls short in accurately capturing the full extent of dominant areas of targets and rendering intricate structural details.




\section{Discussion}
We conducted extensive quantitative and qualitative evaluations of SAM2 on various class-agnostic instance-level segmentation benchmarks.
In CIS, SIS, and SID tasks, SAM2 outperforms task-specific methods in the GT-Bbox mode. 
For the relatively straightforward SIS task, SAM2 attains an impressive AP score of 93.4 on SIP test set. 
For the more challenging CIS task, SAM2 reaches a significant AP score of 73.5 on the NC4K test set, far exceeding the performance of specific methods. 
In the SID task, SAM2 segments instances effectively but struggle with shadow matching. 
As observed in the qualitative comparison, SAM2 without the GT-Bbox is almost impossible to segment out shadows.
In the DIS task, SAM2 performs poorly, even in settings with box prompts it is not able to perform granular segmentation for complex structured objects.
Overall, SAM2 with GT-Bbox achieves better results for both salient and camouflaged objects, especially for salient instances. However, its ability to handle very delicate objects requires improvement.

In comparison with SAM, we found that SAM2 falls short of the performance of SAM in automatic mode across the CIS, SIS, and SID tasks. 
Interestingly, in GT-Bbox mode, SAM2 significantly outperforms SAM.
Besides, we observe that SAM2 models with larger backbones does not always enhance the performance, and even degrades it, which is especially noticeable in the automatic mode.
These findings provide valuable insights for future applications of SAM2 in instance-level segmentation tasks.

\section{Conclusion}
In this study, we evaluate the zero-shot performance of SAM2 in class-agnostic instance-level segmentation tasks across four scenarios: Salient Instance Segmentation (SIS), Camouflaged Instance Segmentation (CIS), Shadow Instance Detection (SID), and Dichotomous Image Segmentation (DIS). 
In the automatic setting, SAM2 underperforms compared to SAM and task-specific methods in the SIS, CIS, and SID tasks. 
When provided with bounding box prompts, especially in the DIS task, SAM2 demonstrates its capability to generate more refined masks. 
The experimental results demonstrate that SAM2 excels in class-agnostic instance-level segmentation when guided by prompts, as well as its potential capabilities in diverse scenarios. 
In future work, we aim to fine-tune SAM2 and develop adapters to boost its performance across various instance-level segmentation tasks.


{
\bibliographystyle{IEEEtran}
\bibliography{SAMv2}
}

\vfill

\end{document}